\documentclass[runningheads]{llncs}

\usepackage{eccv}

\usepackage{eccvabbrv}

\usepackage{graphicx}
\usepackage{booktabs}

\usepackage[accsupp]{axessibility}  

\newcommand{\YJ}[1]{\textbf{\color{blue}[YJ: #1]}}

\newcommand{\JH}[1]{\textbf{\color{teal}[JH: #1]}}

\newcommand{\NJ}[1]{\textbf{\color{magenta}[NJ: #1]}}

\newcommand{\SH}[1]{\textbf{\color{violet}[SH: #1]}}

\newcommand{\TODO}[1]{\textbf{\color{red}[TODO: #1]}}

\renewcommand{\YJ}[1]{}
\renewcommand{\JH}[1]{}
\renewcommand{\SH}[1]{}
\renewcommand{\NJ}[1]{}
\renewcommand{\TODO}[1]{}

\usepackage{microtype}

\renewcommand{\subsubsection}[1]{\vspace{0.8em}\noindent\textbf{#1}}

\usepackage{lipsum}
\usepackage{bbm}
\usepackage{multirow}
\usepackage{makecell}
\usepackage[table]{xcolor}

\usepackage{enumitem}

\usepackage{kotex}

\usepackage{overpic}

\newcommand{\firstc}[1]{\colorbox{BestColor}{#1}}
\newcommand{\secondc}[1]{\colorbox{SecondColor}{#1}}
\newcommand{\thirdc}[1]{\colorbox{ThirdColor}{#1}}

\definecolor{BestColor}{rgb}{1, 0.7, 0.7}
\definecolor{SecondColor}{rgb}{1, 0.85, 0.7} 
\definecolor{ThirdColor}{rgb}{1, 1, 0.7}

\newcommand{\first}[1]{\cellcolor{BestColor}#1}
\newcommand{\second}[1]{\cellcolor{SecondColor}#1}
\newcommand{\third}[1]{\cellcolor{ThirdColor}#1}

\newcommand{\real}{\mathbbm{R}}

\newcommand{\ourmethod}{\textit{LoGoColor}\xspace}

\usepackage{hyperref}

\usepackage{orcidlink}

\begin{document}

\title{LoGoColor: Local-Global 3D Colorization \\ for 360\textdegree~Scenes} 

\titlerunning{LoGoColor: Local-Global 3D Colorization for 360\textdegree~Scenes}


\author{Yeonjin Chang\inst{1} \and
Juhwan Cho\inst{1} \and
Seunghyeon Seo\inst{1} \and \\
Wonsik Shin\inst{1} \and
Nojun Kwak\inst{1}}

\authorrunning{Y.~Chang et al.}

\institute{Seoul National University \\
\email{\{yjean8315,hj99cho,zzzlssh,wonsikshin,nojunk\}@snu.ac.kr}}

\maketitle

\begin{abstract}

Single-channel 3D reconstruction is widely used in fields such as robotics and medical imaging. 
While these methods are good at reconstructing 3D geometry, their outputs are typically uncolored 3D models, making 3D colorization necessary for visualization.
Recent 3D colorization studies address this problem by distilling 2D image colorization models. 
However, these approaches suffer from an inherent inconsistency of 2D image models. 
This results in colors being averaged during training, leading to monotonous and oversimplified results, particularly in complex 360$^{\circ}$ scenes.
In contrast, we aim to preserve color diversity by generating a new set of consistently colorized training views, thereby suppressing the averaging process.
Nevertheless, mitigating the averaging process introduces a new challenge: ensuring strict multi-view consistency across these colorized views.
To achieve this, we propose \ourmethod, a pipeline designed to preserve color diversity by eliminating this guidance-averaging process with a `Local-Global' approach: we partition the scene into subscenes and explicitly tackle both inter-subscene and intra-subscene consistency using a fine-tuned multi-view diffusion model. 
We demonstrate our method achieves quantitatively and qualitatively more consistent and plausible 3D colorization on complex 360$^{\circ}$ scenes than existing methods.
Project page is available at \url{https://yeonjin-chang.github.io/LoGoColor/}.
\end{abstract}
\begin{figure}
    \centering
    \includegraphics[width=0.99\textwidth]{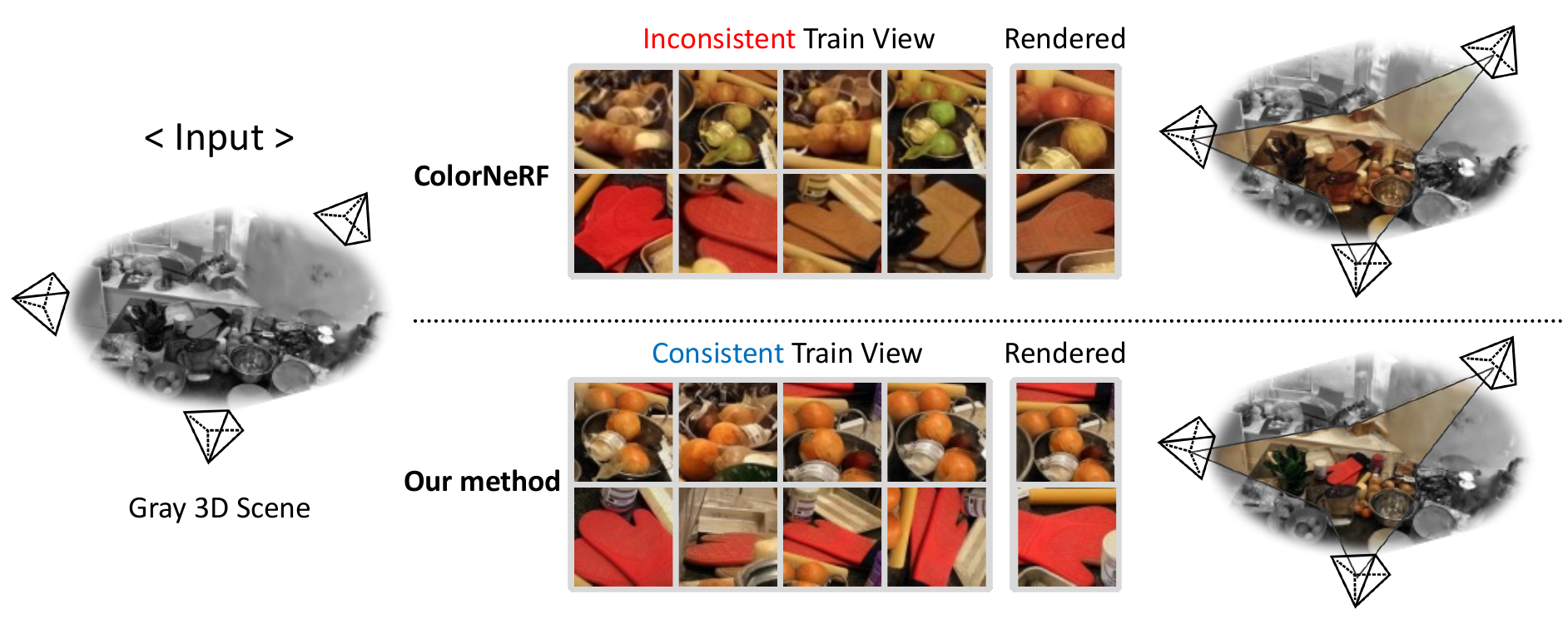}
    \vspace{-5pt}
    \caption{
    We propose \ourmethod to achieve color-rich 3D colorization by minimizing guidance from image models. This avoids the guidance-averaging of prior works, which relied on inconsistent image model outputs and led to monotonous results. To do so, we explicitly handle consistency with a Local-Global approach, ensuring both intra- and inter-subscene consistency.
    }
\label{fig:teaser}
\end{figure}

\section{Introduction}
\label{sec:intro}

Recent advancements in 3D reconstruction, driven by Neural Radiance Fields (NeRF)~\cite{nerf} and 3D Gaussian Splatting (3DGS)~\cite{3dgs}, have enabled high-fidelity novel view synthesis.
These breakthroughs have spurred a wide range of subsequent research, including dynamic scene representation~\cite{gafni2021dynamic, lin2024gaussianflow}, efficient training~\cite{plenoxels, hu2022efficientnerf}, scene editing~\cite{yuan2022nerfediting, gaussctrl2024}, and 3D scene generation~\cite{poole2022dreamfusion,zhang2024text2nerf}, increasing their applicability in VR/AR.
One line of subsequent research has focused on multi-view reconstruction from single-channel images, with prior works focusing on optimizing 3D models from thermal or (near-)infrared images, or reconstructing 3D volumes from X-ray images in the medical domain \cite{ye2024thermalnerf,liu2025thermalgs,li2021nirpolar,cai2024saxnerf}. 
However, they primarily focus on generating 3D geometry for specific applications, such as robot manipulation \cite{singh2023robotnerf} or medical assistance \cite{cai2024xgaussian}. 
While effective for these specific tasks, the resulting single-channel models lack the rich visual data of full-color RGB models.
This limits their versatility for general-purpose applications like VR/AR and their compatibility with standard 3D pipelines.
To bridge this gap and make these single-channel 3D reconstructions truly versatile, a robust 3D colorization step is essential.

3D colorization task, however, presents a crucial challenge beyond the vividness and plausibility required in 2D image colorization: ensuring \textbf{color consistency across views}. 
Existing 3D colorization methods achieve this by leveraging image model outputs, either by iteratively updating the 3D representation with local patch outputs~\cite{colornerf}, or by training from pre-generated ones~\cite{chromadistill}.
However, these approaches essentially average the outputs of an image colorization model, either across iterations or training views. 
This reliance on averaging, while effective at achieving consistency, implicitly assumes a restricted color distribution. 
This assumption breaks down in 360-degree real-world scenes, which are often composed of many distinct objects and complex geometric regions.
Consequently, this approach fails to robustly colorize these intricate areas, resulting in muted and oversimplified colors as shown in \cref{fig:teaser}.

In this work, we rethink 3D consistency to robustly colorize complex 360-degree scenes.
To preserve their natural color diversity, we minimize independent guidance from image colorization models and eliminate the guidance-averaging process. 
Instead, we design our pipeline to generate a new set of consistently colorized training views, with minimal reliance on the image model knowledge.
While this approach bypasses the averaging issue, it shifts the challenge to how we generate these training views to be \textit{consistent}.
To address this, we propose \ourmethod, a Local-Global approach designed to achieve consistency without resorting to averaging.
We partition the scene into subscenes and explicitly tackle both inter-subscene (global) and intra-subscene (local) consistency, using a fine-tuned multi-view diffusion model to learn and ensure these relationships.

Specifically, our pipeline begins by reconstructing a geometry-only 3D model from the single-channel inputs. 
Leveraging the training view cameras and the learned geometry, we then apply our \textit{View-based Subscene Decomposition} method to partition the 3D scene into subscenes that maximize coverage while minimizing overlap.
For each subscene, we select a representative base view and colorize it with an image colorization model.
We then calibrate these base views with a multi-view diffusion model that enforces global consistency among subscenes.
The calibrated base views serve as color references for the subsequent colorization of all training views, yielding globally and locally consistent colorized images.

We demonstrate through experiments that our method yields 3D models with diverse and consistent color from single-channel images. 
We compare our approach with existing works~\cite{colornerf, chromadistill}, as well as \cite{genn2n} for consistent 3D editing and \cite{colormnet}, an inherently 3D-aware video colorization method.
Both qualitative and quantitative results show that our model achieves superior color diversity and consistency.
To quantify this improvement, we report the normalized Colorfulness (\emph{nColorfulness}), which measures Colorfulness~\cite{hasler2003measuring} after removing the overall image tint. 
This metric demonstrates that our model produces more diverse colors by avoiding the averaging effect.
Notably, even when we adapt the baselines to the 3DGS framework to enhance their geometric reconstruction, their inherent averaging process inevitably leads to monotonous colorization. 
In contrast, our method successfully preserves the rich color diversity of the scene.
We also demonstrate the effectiveness of our Local-Global approach through ablation studies. 
Finally, we reconstruct colorized 3D models from Near-infrared (NIR) multi-view images, demonstrating that our method applies robustly to single-channel image modalities.

\section{Related Work}
\label{sec:related}

\subsubsection{3D reconstruction.}
Reconstructing 3D scenes from 2D observations is a long-standing problem in computer vision, and recent years have seen significant progress through neural representations.
Neural Radiance Fields (NeRF)~\cite{nerf} have emerged as a dominant framework by learning volumetric radiance fields via differentiable volume rendering.
Since its introduction, NeRF has been extended in various directions: few-shot reconstruction~\cite{yu2021pixelnerf, seo2023flipnerf, seo2023mixnerf, yang2023freenerf}, scene generalization~\cite{yu2021pixelnerf, rematas2021sharf, chen2021mvsnerf}, dynamic and unbounded scenes~\cite{gafni2021dynamic, mipnerf360}, and faster optimization~\cite{yu2021plenoctrees, muller2022instant}.
More recently, 3D Gaussian Splatting (3DGS)~\cite{3dgs} has gained attention as a real-time alternative to NeRF, representing scenes with rasterized Gaussian primitives instead of volumetric fields.
Its efficiency has sparked numerous follow-ups on anti-aliasing~\cite{yan2024multi, yu2024mip, liang2024analytic}, 3D content generation~\cite{tang2024dreamgaussian, yuan2024gavatar, zou2024triplane, lin2025diffsplat}, dynamic scenes~\cite{yang2024deformable, wu20244d, yan20244d}, and so on.
In our work, we shift focus to single-channel 3D reconstructions, where geometry is recovered without color.
Unlike the above methods that jointly learn color and shape, our method addresses the open challenge of colorizing geometry-only 3D scenes in a globally and locally consistent manner.

\subsubsection{Single channel 3D reconstruction.}
Single-channel 3D reconstruction is motivated by the limitations of RGB and depth sensors in challenging conditions, such as low-light or highly reflective environments. 
While standard modalities often degrade in these settings, thermal and NIR signals remain robust, enabling geometry inference directly from intensity variations.
Early studies combined thermal or IR cues with photogrammetry to jointly recover geometry and temperature~\cite{cabrelles2009,iwaszczuk2011}, or employed thermal-only silhouette intersection for volumetric recovery~\cite{chen2015reconstruction}.
Similar attempts using NIR polarization~\cite{li2021nirpolar} demonstrate single-channel cues can convey geometric information.
Recently, radiance-based representations such as NeRF and 3DGS have extended these ideas.
Thermal-NeRF~\cite{ye2024thermalnerf} and ThermalGS~\cite{liu2025thermalgs} reconstruct scene geometry and thermal emission directly from infrared input.
However, these neural single-channel approaches lack realistic color, leading to less fidelity than multi-modal reconstructions.
Our work addresses this limitation by coupling single-channel 3D representations with \textit{colorization}, producing plausible 3D colorized scenes under low-light, nighttime, or non-visible conditions.

\subsubsection{Image colorization.}
Colorizing grayscale images into plausible RGB counterparts is a fundamental challenge in computer vision.  
Early CNN-based approaches map luminance to chrominance~\cite{zhang2016colorful,iizuka2016let}, and later transformers capture broader spatial dependencies~\cite{kumar2021colorization}.  
Adversarial methods further enhance realism by training generative models to produce sharper, vivid colors~\cite{nazeri2018image,zhang2022bigcolor}.
Diffusion models~\cite{ho2020ddpm} have recently reframed colorization as an image-to-image translation task~\cite{img2img},  
offering greater color fidelity and diversity~\cite{img2imgturbo}.  
However, as these methods operate in a single-view setting, they fail to ensure cross-view color consistency.
Parallel studies translate single-channel modalities, such as thermal into visible spectra.  
Early cross-modal mappings~\cite{berg2018generating} learn direct transformations between thermal and RGB domains, while later GAN- and diffusion-based approaches~\cite{luo2022pearlgan,nair2023t2vddpm}  
generate visually plausible single-view results but still lack multi-view color coherence. 
In contrast, our method colorizes multi-view single-channel inputs (grayscale, thermal, or NIR) to reconstruct consistent and realistic 3D color scenes, providing a unified framework for multi-view colorization in 3D.

\subsubsection{3D colorization.}
3D colorization has been extensively studied for point clouds acquired from LiDAR or Time-of-Flight cameras, which inherently lack color information~\cite{liu2019pccn, cao2018point, shinohara2021point2color,gao2023scene}. 
Recently, there have been attempts to colorize 3D reconstructions derived from grayscale images.
Notable works include ColorNeRF~\cite{colornerf}, which integrates colorization into NeRF by training on image patches processed by an image colorization model, and ChromaDistill~\cite{chromadistill}, which distills chromatic information from pre-generated 2D color views into a Plenoxel representation. 
However, these methods achieve multi-view consistency primarily by averaging the outputs of 2D image models.
To address this limitation, a concurrent work, Color3D~\cite{color3d}, fine-tunes a personalized colorizer on a single reference view to propagate consistent colors.
Unlike relying on a single view, we propose a Local-Global approach that mitigates this averaging effect by generating a set of consistently colorized training views, thereby preserving the scene color diversity.
\begin{figure}[tb]
    \centering
    \includegraphics[width=0.99\textwidth]{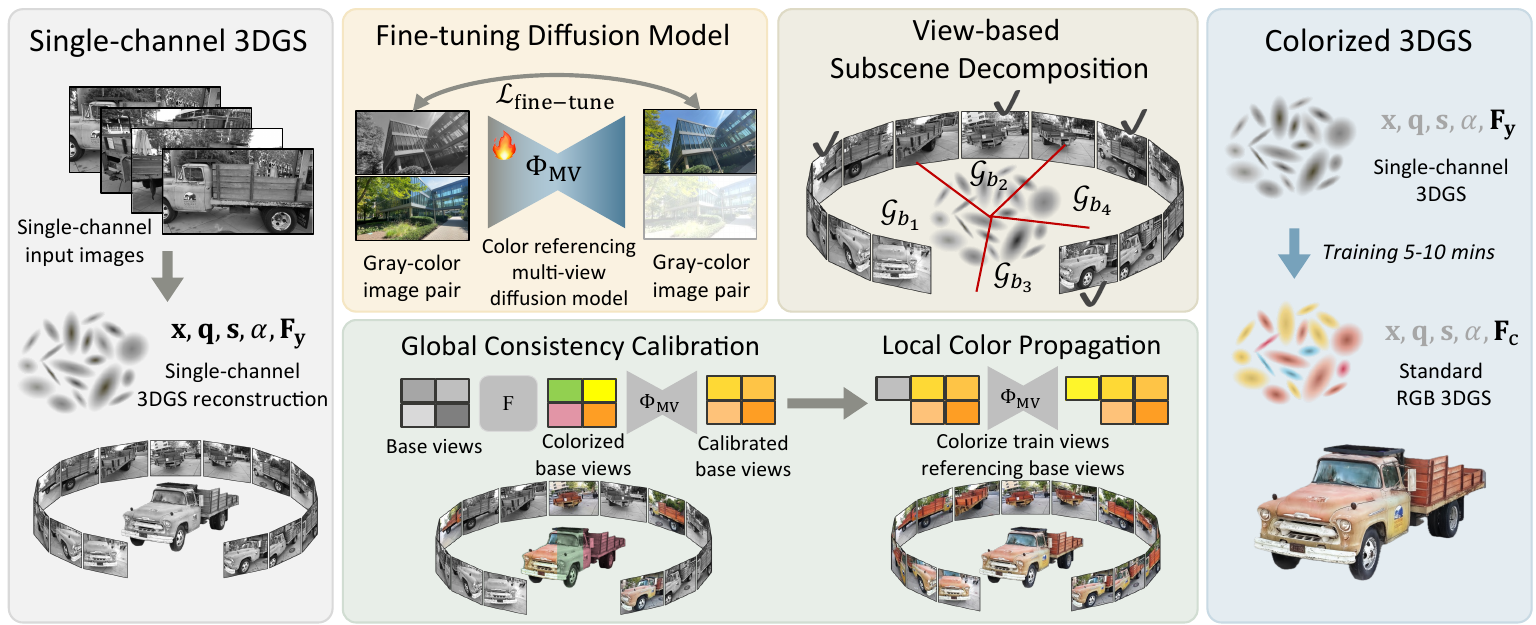}
    \caption{
        {\bf Overview of \ourmethod~--} We first reconstruct single-channel 3D Gaussians from multi-view grayscale images to recover scene geometry.
        Using this geometry, we decompose the scene into subscenes and select their corresponding base views.
        In parallel, we fine-tune a multi-view diffusion model to transfer color from reference views.
        We then calibrate global consistency among the base views and propagate color across all training views, ultimately producing a fully colorized 3D Gaussian model.
    }
    \label{fig:overview}
\end{figure}

\section{Method}
\label{sec:method}
We illustrate our proposed approach in~\cref{fig:overview}.
We first optimize single-channel 3D Gaussians from the given single-channel multi-view images (\cref{sec:gray3d}).
Given the scene geometry and training view cameras, we divide the 3D scene into subscenes that have minimal overlap and whose union covers most of the full scene (\cref{sec:subscene}).
For the subsequent Local-Global 3D colorization, we first fine-tune a multi-view diffusion model that colorizes an input image using the color from a reference image (\cref{sec:finetune}). 
Leveraging this model, we colorize the training views to ensure both inter- and intra-subscene consistency.
We begin by colorizing one base view from each 3D subscene using an image colorization model. 
Since these base views are colorized independently, they are globally inconsistent. 
To resolve this and achieve inter-subscene consistency, we pass them through the fine-tuned multi-view diffusion model to create a globally consistent reference set (\cref{sec:global}).
Next, to ensure intra-subscene consistency, we again use this fine-tuned model to colorize the remaining training views, referencing the base views (\cref{sec:local}).
Finally, these consistent views are used as pseudo-ground truth to optimize the color components of the 3D Gaussian model.

\subsection{Single-channel 3D Reconstruction}
\label{sec:gray3d}
We first aim to reconstruct the scene geometry from given single-channel multi-view images $\mathcal{I}_g = \{I_g^1, I_g^2, \dots, I_g^T\}$, where $I_g^*\in\real^{1 \times H \times W }$ and $T$ is the total number of views.
We represent the 3D geometry using a set of single-channel 3D Gaussian primitives, modified from 3D Gaussian Splatting (3DGS)~\cite{3dgs}.
Standard 3DGS defines each Gaussian primitive with position $\mathbf{x}\in\real^3$, rotation $\mathbf{q}\in\real^4$, scaling $\mathbf{s}\in\real^3$, opacity $\alpha\in\real$, 
and spherical harmonic coefficients $\mathbf{F}_c \in\real^{3 \times h}$ where $h$ is the number of coefficients used to compute its view-dependent color $\mathbf{c} \in\real^3$.
Since our input $\mathcal{I}_g$ is single-channel, we adjust this representation by replacing the color component $\mathbf{F}_c$ with single-channel luminance coefficients $\mathbf{F}_y \in\real^{1 \times h}$, used to compute view-dependent luminance $y\in\real$,
while retaining the core geometric parameters ($\mathbf{x}, \mathbf{q}, \mathbf{s}, \alpha$).
We consequently modify the rendering process to compute an $\alpha$-blended luminance $\mathbf{\hat{Y}}$ for each pixel $p$:
\begin{equation}
  \mathbf{\hat{Y}}(p) = \sum_{i}^{N} y_i \alpha'_i \prod_{j}^{i-1} (1 - \alpha'_j),
\label{eq:blending}
\end{equation}
where $N$ is the number of Gaussian primitives contributing to the pixel $p$ and $\alpha'$ is the projected 2D opacity at pixel $p$. 
We optimize all parameters $\mathbf{x}, \mathbf{q}, \mathbf{s}, \alpha, \mathbf{F}_y$ by minimizing a combined $\mathcal{L}_1$ and D-SSIM loss between the rendered luminance $\mathbf{\hat{Y}}$ and the ground-truth image $I_g$:
\begin{align}
    \mathcal{L}_\text{geom} = (1 - \lambda_\text{s})\mathcal{L}_1(\mathbf{\hat{Y}}, I_g) + \lambda_\text{s} \mathcal{L}_\text{D-SSIM}(\mathbf{\hat{Y}}, I_g).
\label{eq:loss}
\end{align}
This process reconstructs a precise geometric foundation for our subsequent ``Local-Global'' 3D colorization pipeline.

\begin{figure}[tb]
  \centering
    \includegraphics[width=0.99\linewidth]{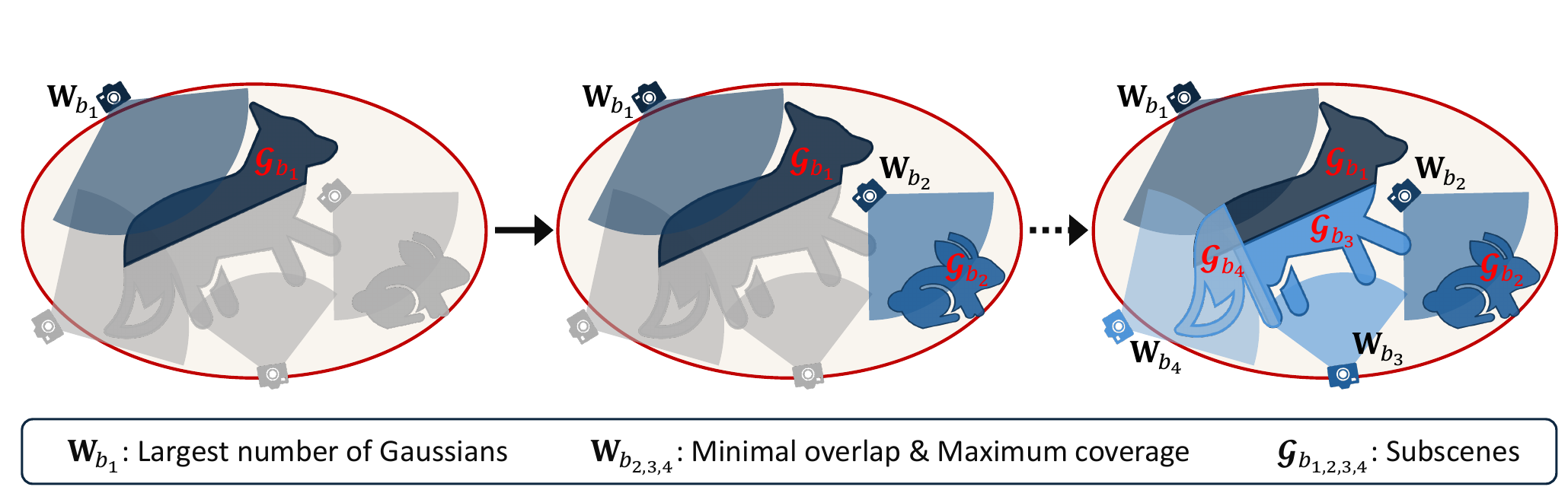}
    
   \caption{\textbf{Our View-based Subscene Decomposition --} Starting from the base view $\mathbf{W}_{b_1}$ that observes the largest number of Gaussians, we use a greedy algorithm to iteratively select subsequent base views that maximize coverage while minimizing overlap.
   }
   \label{fig:subscene}
\end{figure}

\subsection{View-based Subscene Decomposition}
\label{sec:subscene}

We decompose the 360-degree scene into multiple subscenes, based on the 3D Gaussian geometry reconstructed in \cref{sec:gray3d} and the training camera views. 
We design this decomposition to leverage the prior knowledge of an image colorization model for intricate parts of the complex scene while tackling inter- and intra-subscene consistency.
The objective of the decomposition is twofold: 1) ensure maximum coverage across the scene, and 2) find $K$ subscenes that have minimal overlap to prevent potential inconsistencies from multiple reference colors.
Directly addressing these conditions on the myriad of Gaussian primitives is complex, as it involves solving a large-scale partitioning problem on millions of points.
Instead, we reframe the problem as partitioning the scene from the perspective of camera poses that represent these Gaussian subsets.
Specifically, we associate each camera pose $\mathbf{W}_{t}$ with its set of visible Gaussian splats ($\mathcal{G}_t$)---the Gaussians used when rendering the image for that camera.
We thus partition the scene by selecting $K$ base views (camera poses) whose corresponding Gaussian sets best satisfy the conditions of maximum coverage and minimal overlap.

We find these base view cameras using a greedy algorithm.
First, we select the first base view $\mathbf{W}_{b_1}$ as the camera that observes the largest number of Gaussians. 
We initialize the covered region as $\mathcal{G}_\text{covered} = \mathcal{G}_{b_1}$.
Subsequently, for $k=2$ to $K$, we iteratively select the next base view $\mathbf{W}_{b_k}$ as the camera $\mathbf{W}_t\in \mathcal{W}$ that maximizes the ratio of newly covered Gaussians to the overlapping Gaussians:
\begin{equation}
    \mathbf{W}_{b_k} = \arg\max_{\mathbf{W}_t \in \mathcal{W}} \frac{|\mathcal{G}_t \setminus \mathcal{G}_\text{covered}|}{|\mathcal{G}_t \cap \mathcal{G}_\text{covered}| + 1}
\end{equation}
where adding 1 to the denominator prevents division by zero.
After selecting $\mathbf{W}_{b_k}$, we update the set of covered Gaussians: $\mathcal{G}_\text{covered} \leftarrow \mathcal{G}_\text{covered} \cup \mathcal{G}_{b_k}$.
This process yields $K$ base views and their corresponding subscenes that approximate our coverage and overlap goals.
See \cref{fig:subscene} for a visualization.
Note that this greedy selection minimizes overlap but does not eliminate it.
However, as we describe in \cref{sec:global}, any potential inconsistencies arising from this remaining overlap are resolved when we later use these $K$ base views to enforce global (inter-subscene) consistency.

\subsection{Multi-view Colorizing Model Fine-tuning}
\label{sec:finetune}

Our Local-Global approach requires a model capable of handling both inter-subscene and intra-subscene consistency while preserving the 3D structure from \cref{sec:gray3d}. 
Diffusion models have proven highly effective for image-to-image translation tasks, including colorization~\cite{img2img}, making them a strong candidate for our pipeline.
However, unlike standard image-to-image models that process images independently, our approach requires a model that can also ensure color consistency across multiple views of the scene.
To achieve this, we adopt the diffusion model SD-Turbo~\cite{sauer2024adversarial,sdturbo} as a base, and fine-tune it following the image-to-image training scheme of {pix2pix-Turbo}~\cite{img2imgturbo}.
Crucially, to enable the required multi-view referencing, we integrate a reference mixing layer from {DIFIX3D+}~\cite{difix3d}, which applies self-attention to the reference image $I_{ref}$ to guide the colorization of the single-channel input $I_g$.
We denote this multi-view model as $\Phi_\text{MV}$.

$\Phi_\text{MV}$ is fine-tuned to generate $\hat{I}_c\in\real^{3 \times H \times W }$ by applying the color of its reference image $I_{ref}$ to the input $I_g$ while preserving structure. 
To provide explicit guidance for color generation, we compute the loss in the LAB color space, which is more perceptually uniform than the standard RGB color space \cite{zhang2016colorful}. 
We convert both the model output $\hat{I}_c$ and the ground-truth $I_c$ to LAB space.  
The loss function is then composed of an $\mathcal{L}_{L_1}$ loss in LAB space, as well as Gram and LPIPS losses following~\cite{difix3d}:
\begin{equation}
    \mathcal{L}_\text{fine-tune} = \mathcal{L}_{L_1}(\hat{I}_c, I_c) + \lambda_\text{LPIPS} \mathcal{L}_\text{LPIPS}(\hat{I}_c, I_c) + \lambda_\text{Gram} \mathcal{L}_\text{Gram}(\hat{I}_c, I_c).
\end{equation}
By combining this objective with the reference-mixing layer architecture, $\Phi_\text{MV}$ learns to colorize in a multi-view consistent manner while preserving image structure.
This model plays a key role in \cref{sec:global} and \cref{sec:local}.

\subsection{Global Consistency Calibration}
\label{sec:global}
Using the $K$ base views selected in \cref{sec:subscene}, we generate inter-subscene consistent colorized base views, which will be used as references in \cref{sec:local}.
To achieve this, we first pass the $K$ base views $I_g^k$ through an image colorization model $\mathcal{F}$ to produce initial colorized base views $I'^{k}_{c}$:
\begin{equation}
    I'^{k}_{c} = \mathcal{F}(I_g^k) \quad \text{for } k = 1 \dots K.
\end{equation}
However, since the image colorization model $\mathcal{F}$ processes each input $I_g^k$ independently, it does not guarantee inter-subscene consistency among the resulting $\{I'^{k}_{c}\}$. To resolve this issue, we introduce a global consistency calibration step that iteratively refines each view using our multi-view diffusion model, $\Phi_\text{MV}$.

Specifically, for each $k \in [1, K]$, we generate a consistent view $I_c^k$ by combining the initial view $I'^{k}_{c}$ with a globally-calibrated version. 
This calibrated version is generated by passing the grayscale $k$-th view into $\Phi_\text{MV}$, which simultaneously references the color from all other $K-1$ views, $\{I'^{j}_{c}\}_{j \neq k}$. 
This entire calibration process for a single view $k$ is defined as:
\begin{equation}
    I_c^k = \frac{1}{2} \left( I'^{k}_{c} + \Phi_\text{MV}\left( \text{Grayscale}(I'^{k}_{c}), \{I'^{j}_{c}\}_{j \neq k} \right) \right).
\end{equation}
This process is repeated for all $K$ views to create the final consistent set $\{I_c^k\}_{k=1}^K$. 
Unlike $\mathcal{F}$, which processes views independently, this mechanism aggregates the color information from all $K$ image model outputs, thereby resolving potential inconsistencies.
These resulting inter-subscene consistent base views are then used as references to enforce intra-subscene consistency in \cref{sec:local}.

\subsection{Local Color Propagation}
\label{sec:local}

Using the inter-subscene consistent reference set obtained in \cref{sec:global}, we colorize all $T$ training views to ensure intra-subscene consistency.
To do this, we use $\Phi_\text{MV}$ to colorize each training view $I_g^t$. 
We pass $I_g^t$ as the structure input while providing the entire set of $K$ calibrated base views $\{I^{k}_{c}\}$ as the color reference:
\begin{equation}
    \hat{I}^t_c = \Phi_\text{MV}\left(I_g^t, \{I^{k}_{c}\}_{k=1}^K\right).
\end{equation}
The resulting image $\hat{I}_c^t$ thus achieves comprehensive consistency: 
intra-subscene consistency is achieved via the reference-mixing layer to query the relevant colors from the $K$ base views, while inter-subscene consistency is inherited from the globally-calibrated reference set.
By repeating this process for all $T$ views, we obtain a set of fully consistent training views, $\hat{\mathcal{I}}_c = \{\hat{I}_c^t\}_{t=1}^T$, for colorizing the single-channel 3D model.
Lastly, we use $\hat{\mathcal{I}}_c$ as pseudo-color ground-truth to colorize the single-channel 3D Gaussian model obtained in \cref{sec:gray3d}.
We freeze the geometry parameters ($\mathbf{x}, \mathbf{q}, \mathbf{s}, \alpha$) and extend the optimized single-channel luminance coefficients $\mathbf{F}_y \in\real^{1 \times h}$ with new, learnable color coefficients $\mathbf{F}_c \in\real^{3 \times h}$. 
We optimize only this new color component $\mathbf{F}_c$.
This optimization of the colors of Gaussian primitives yields the final colorized 3D model.

\begin{figure}[p]
    \centering
    \newcommand{\imgw}{0.2\textwidth}
    \newcommand{\imgh}{0.13\textwidth}
    
    \setlength{\tabcolsep}{0.5pt}
    \resizebox{0.99\linewidth}{!}{
    \begin{tabular}{@{}ccccccc@{}}
        & {Input Image} & {GenN2N} & {ColorMNet} & {ColorNeRF} & {ChromaDistill} & {Our Method} \\
        
        \raisebox{12pt}{\rotatebox{90}{TnT-Train}} &
        \includegraphics[width=\imgw]{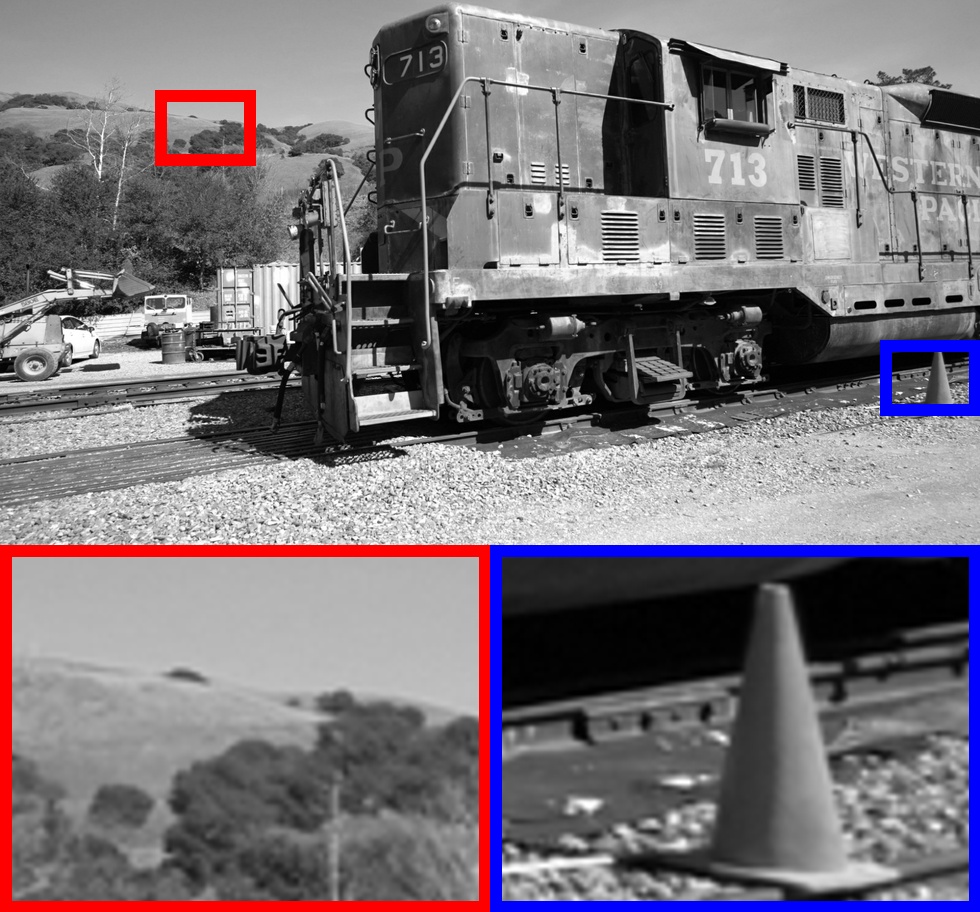} &
        \includegraphics[width=\imgw]{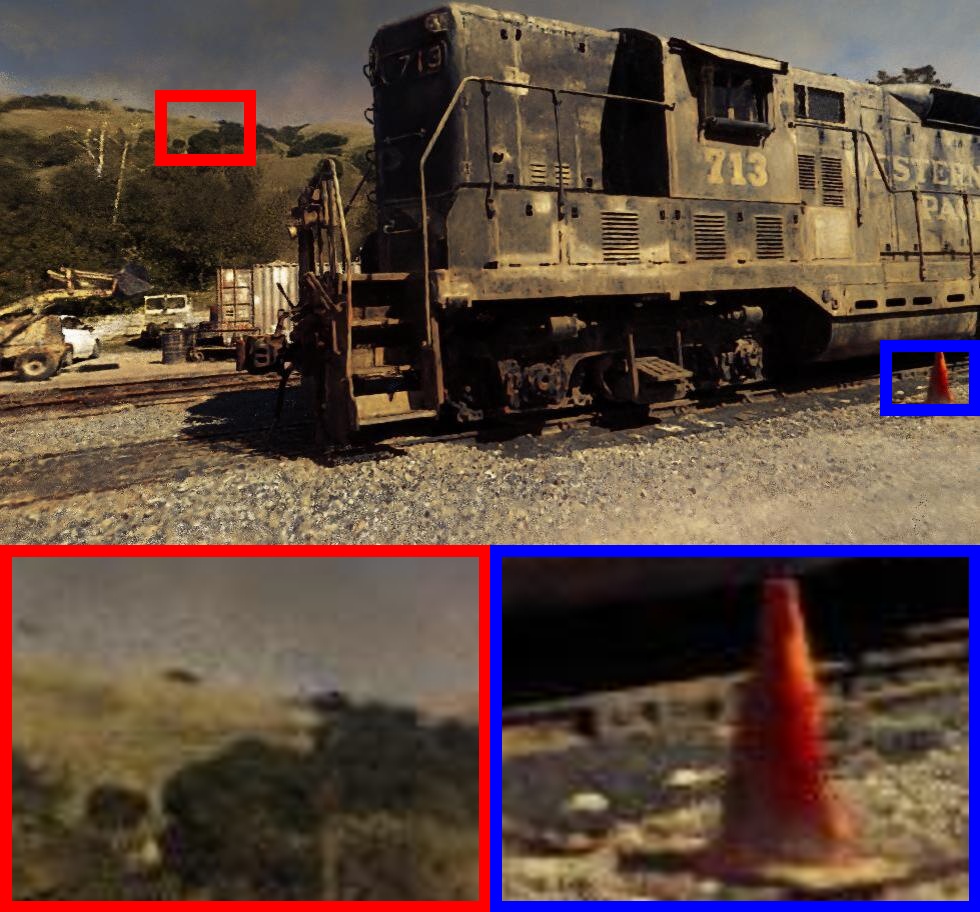} &
        \includegraphics[width=\imgw]{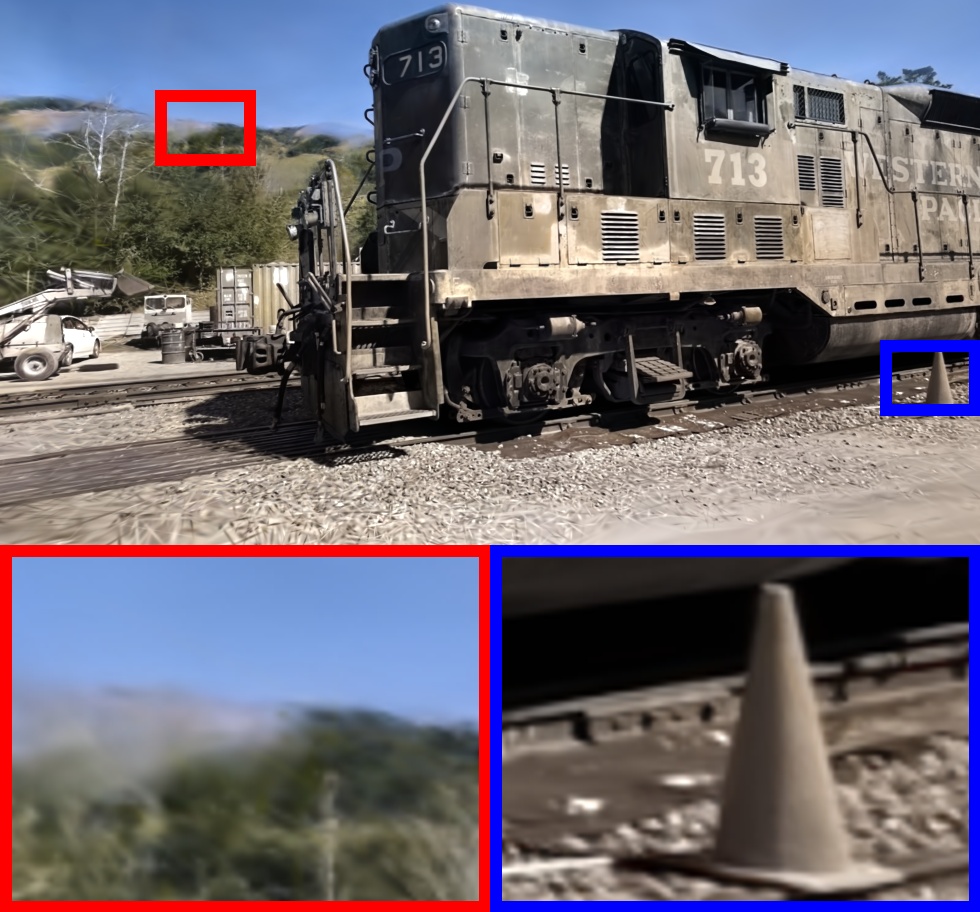} &
        \includegraphics[width=\imgw]{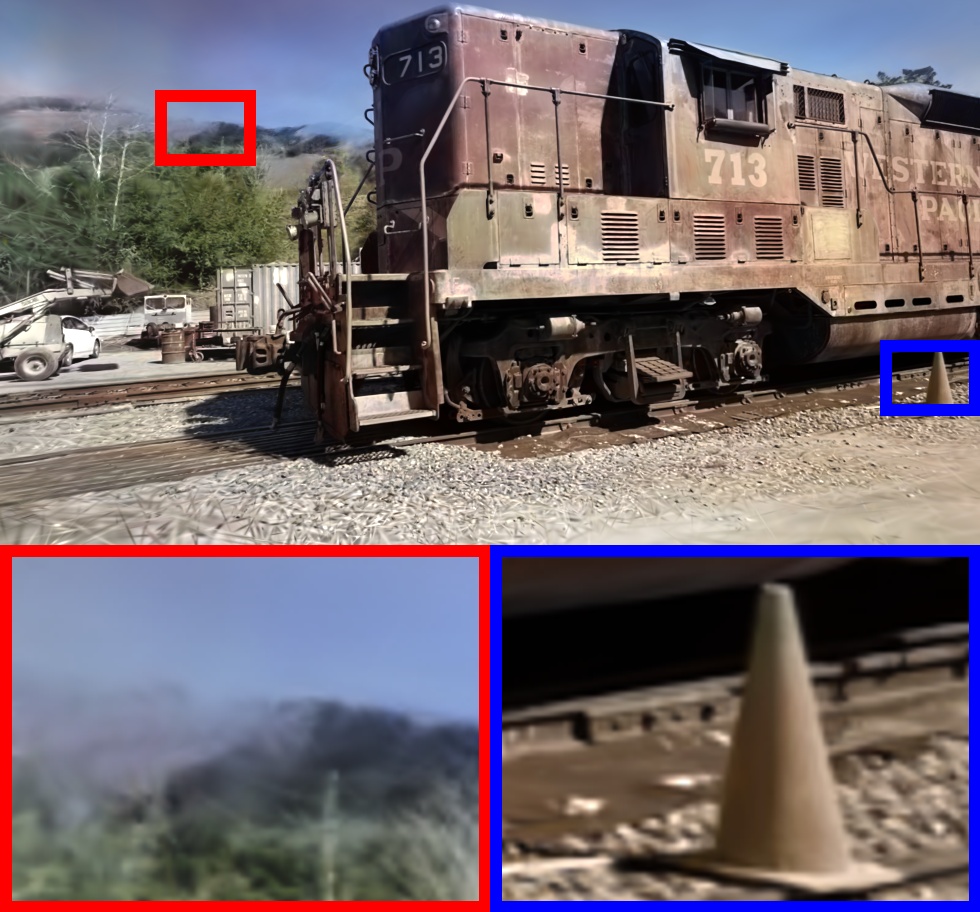} &
        \includegraphics[width=\imgw]{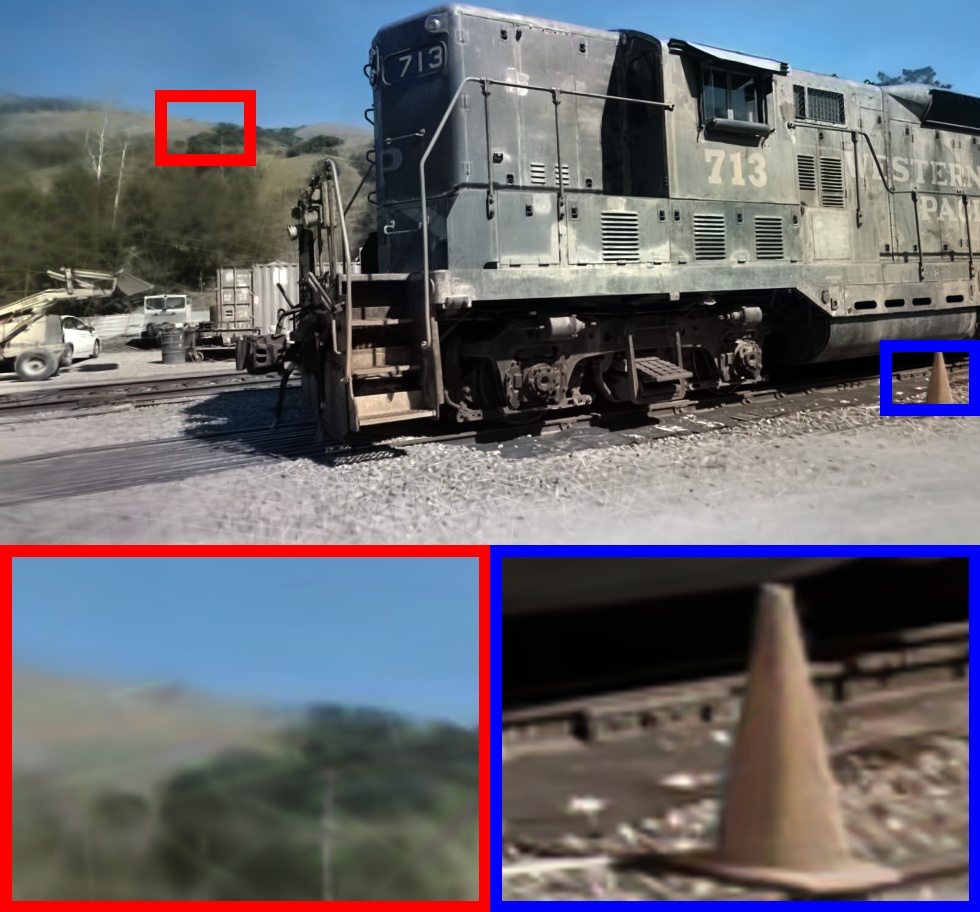} &
        \includegraphics[width=\imgw]{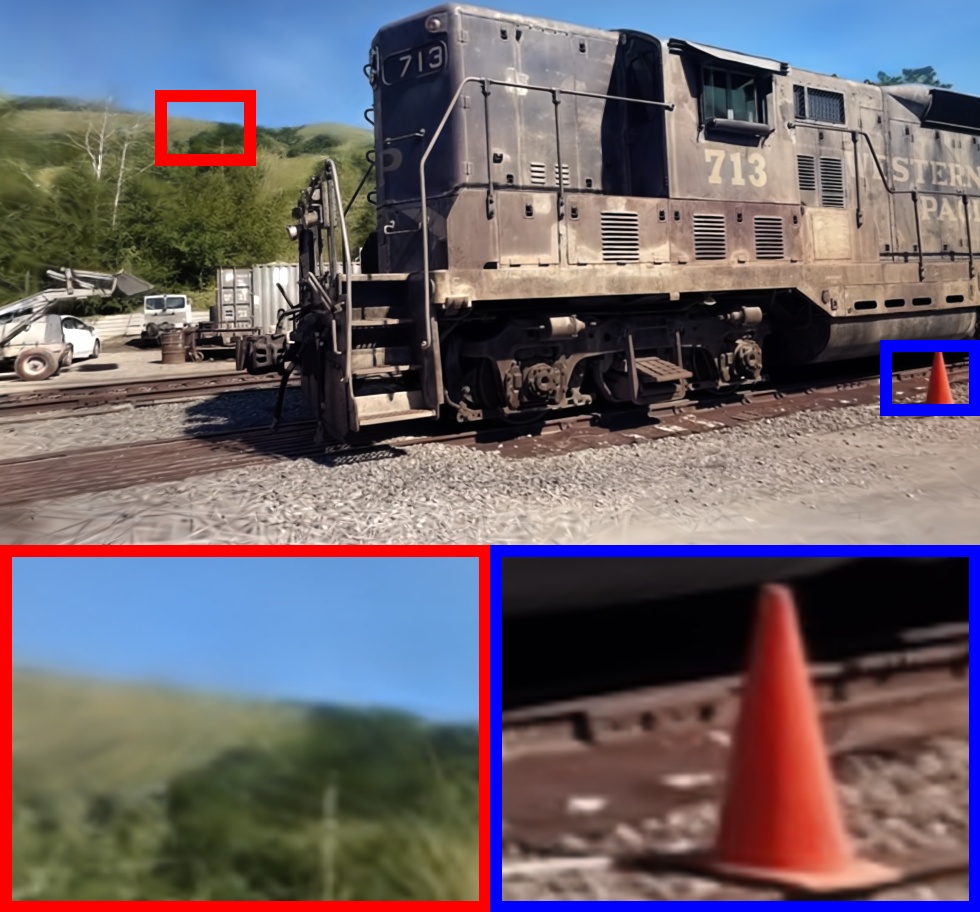} \\

        \raisebox{12pt}{\rotatebox{90}{TnT-Truck}} &
        \includegraphics[width=\imgw]{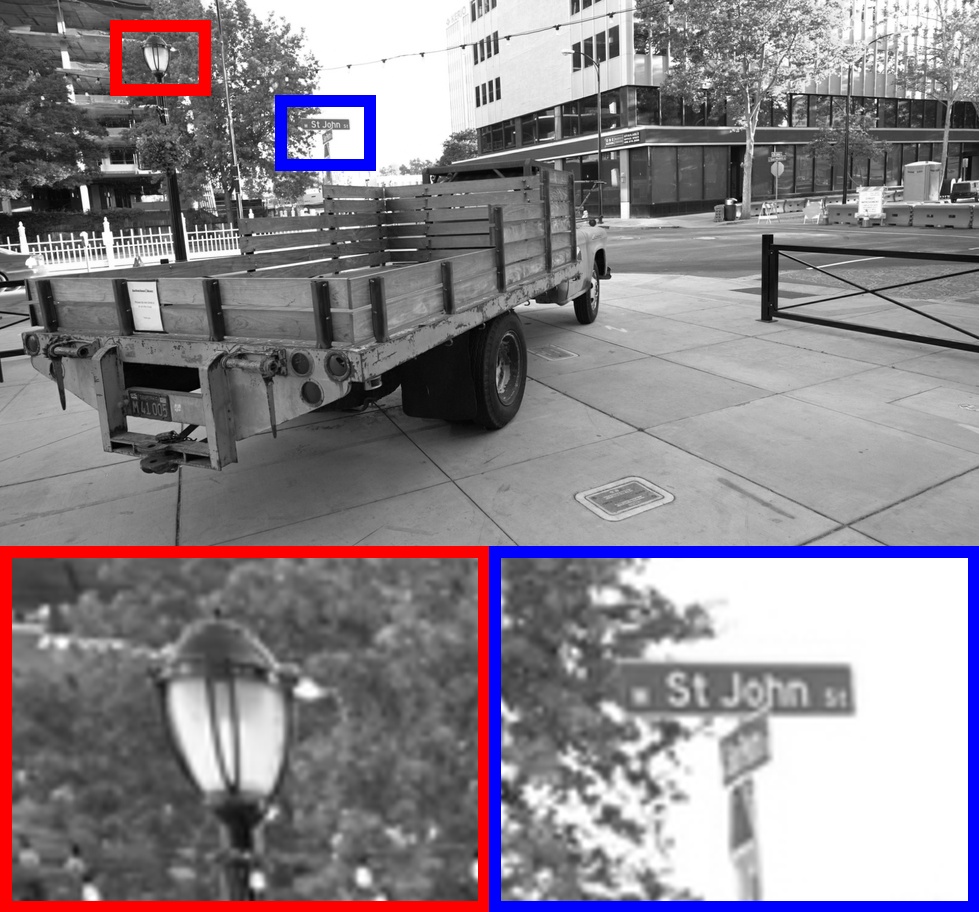} &
        \includegraphics[width=\imgw]{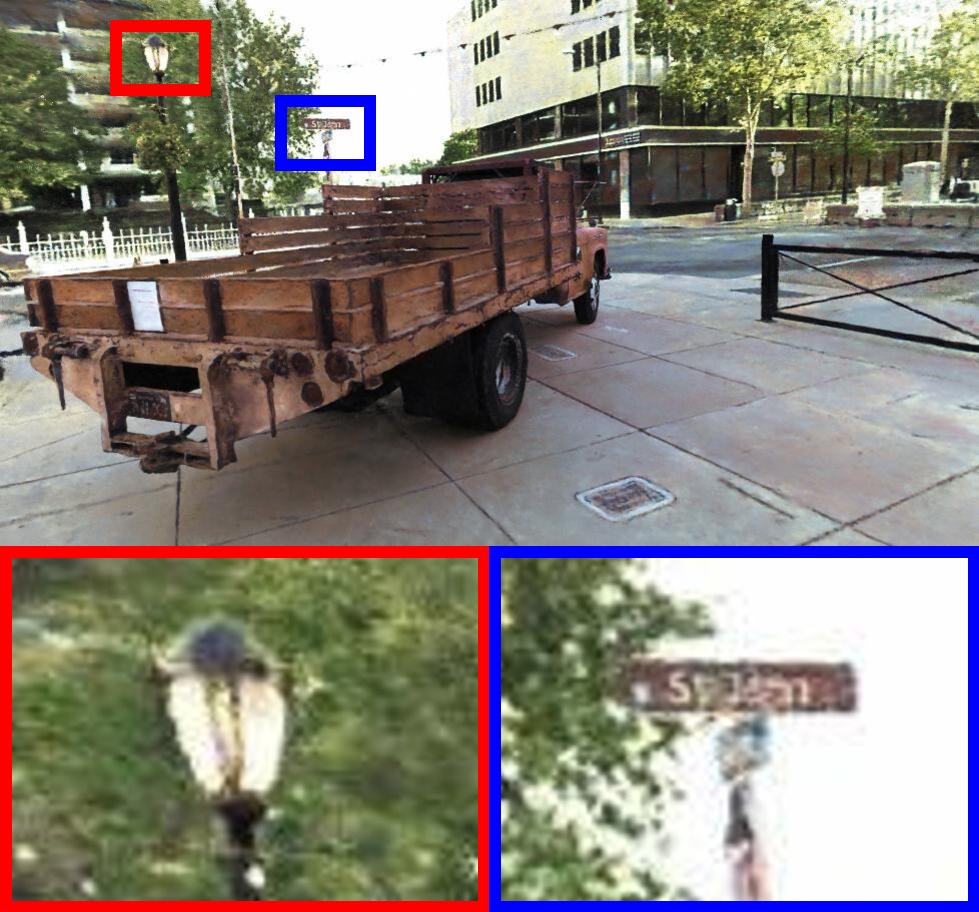} &
        \includegraphics[width=\imgw]{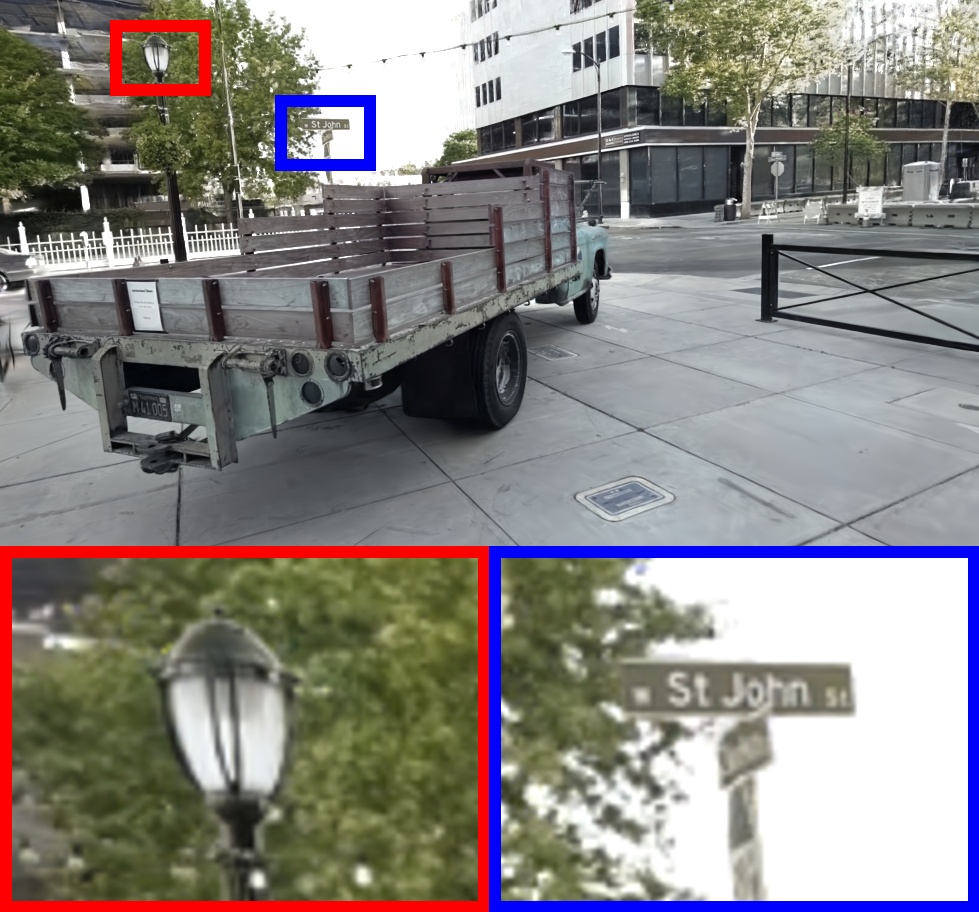} &
        \includegraphics[width=\imgw]{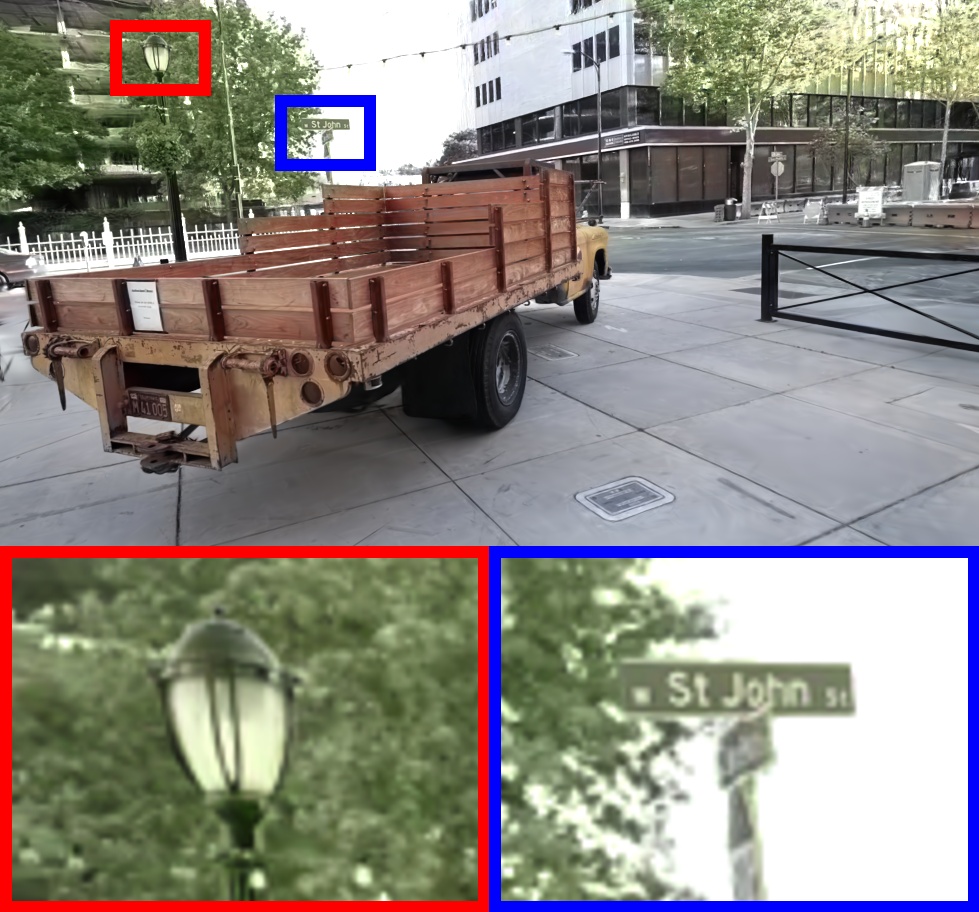} &
        \includegraphics[width=\imgw]{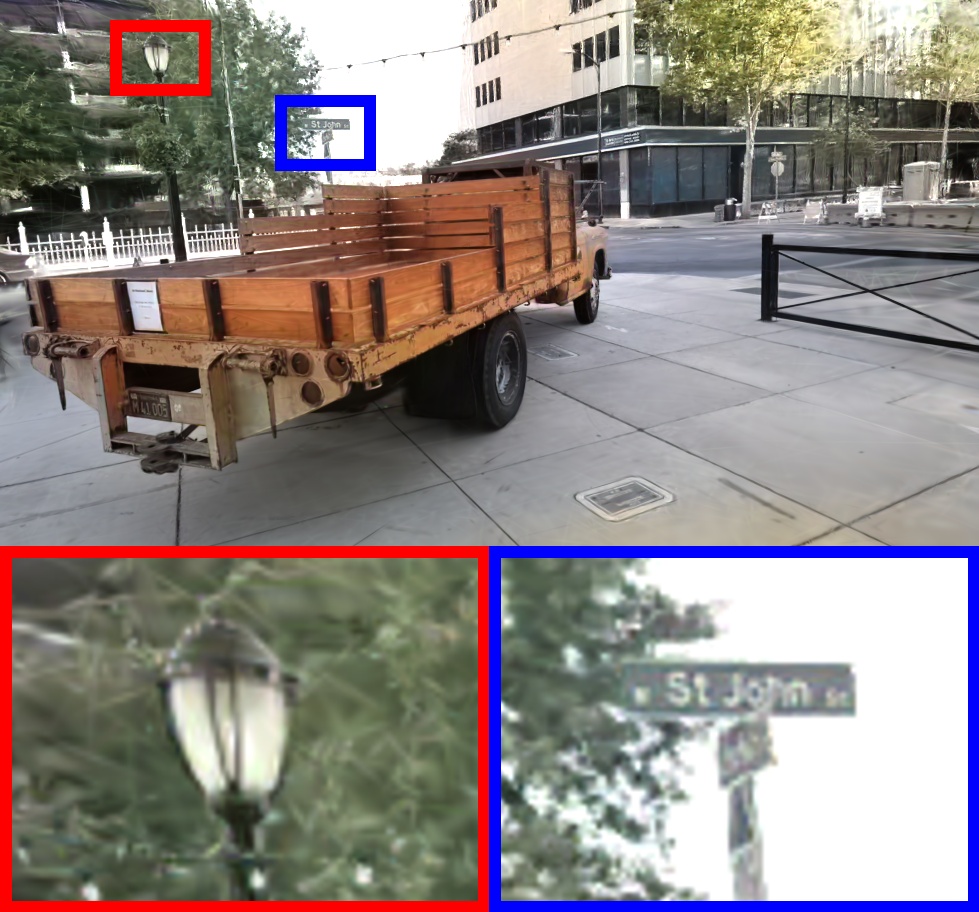} &
        \includegraphics[width=\imgw]{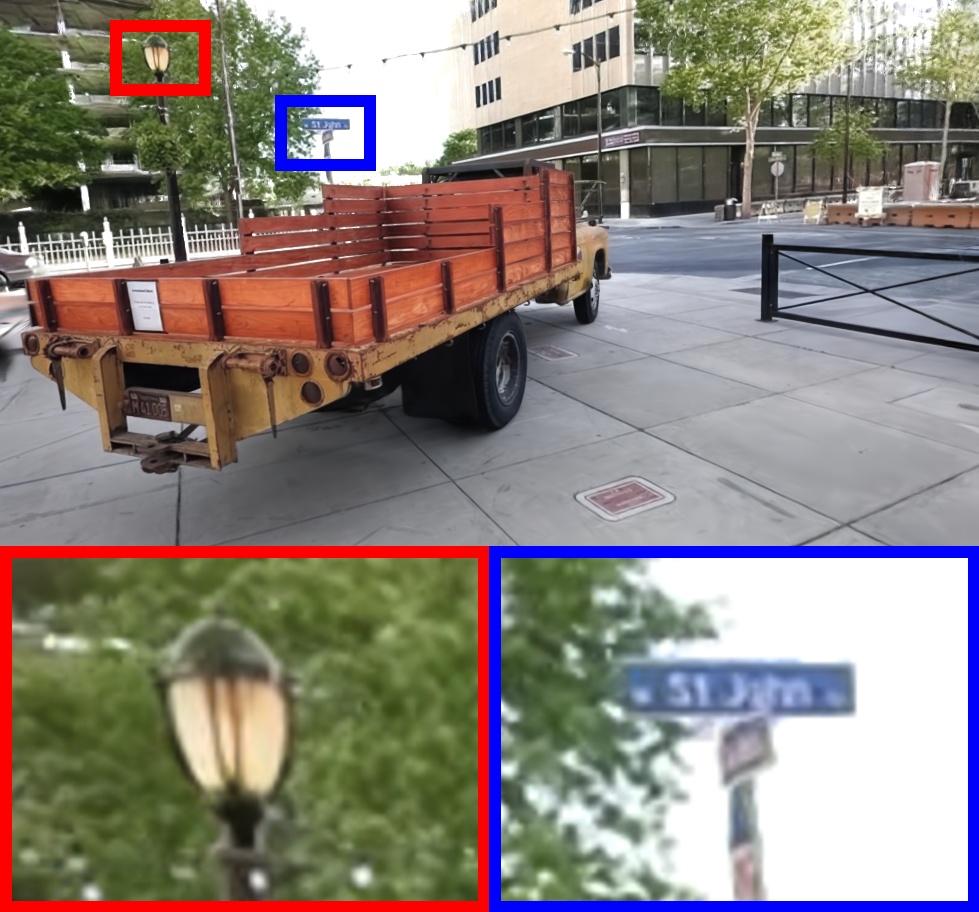} \\
        
        \raisebox{15pt}{\rotatebox{90}{360-Bonsai}} &
        \includegraphics[width=\imgw]{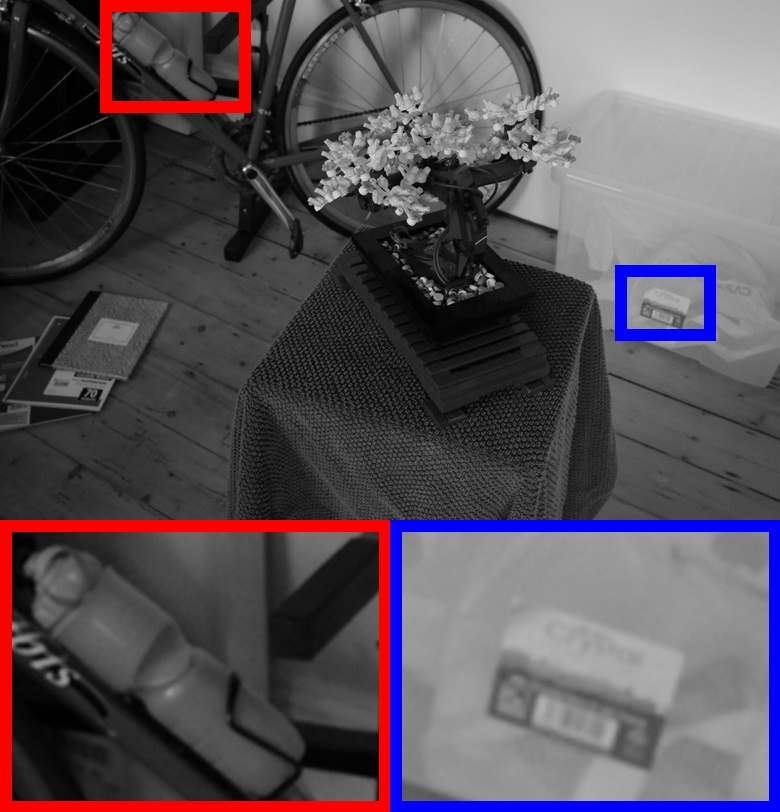} &
        \includegraphics[width=\imgw]{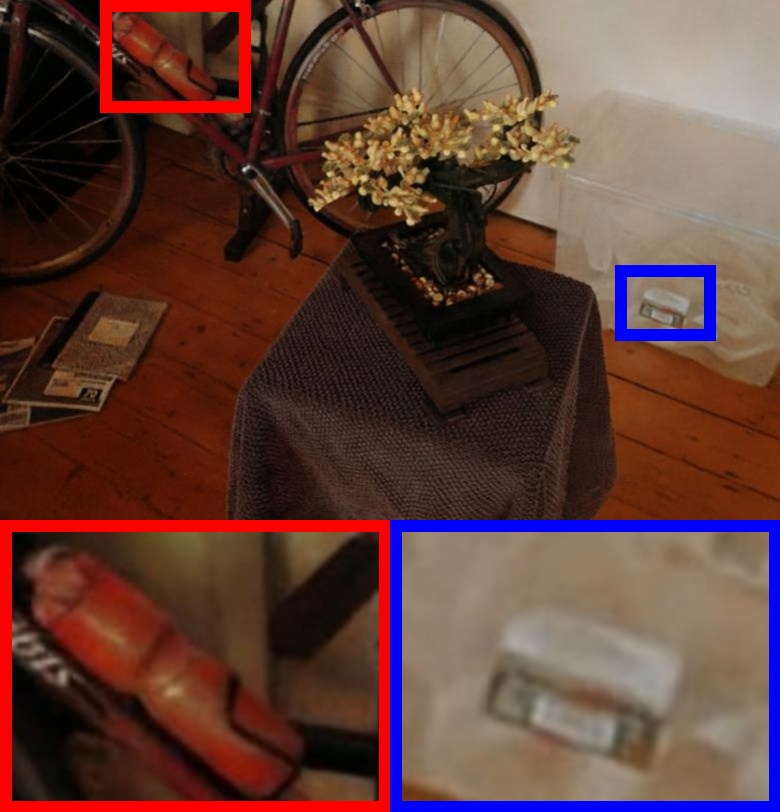} &
        \includegraphics[width=\imgw]{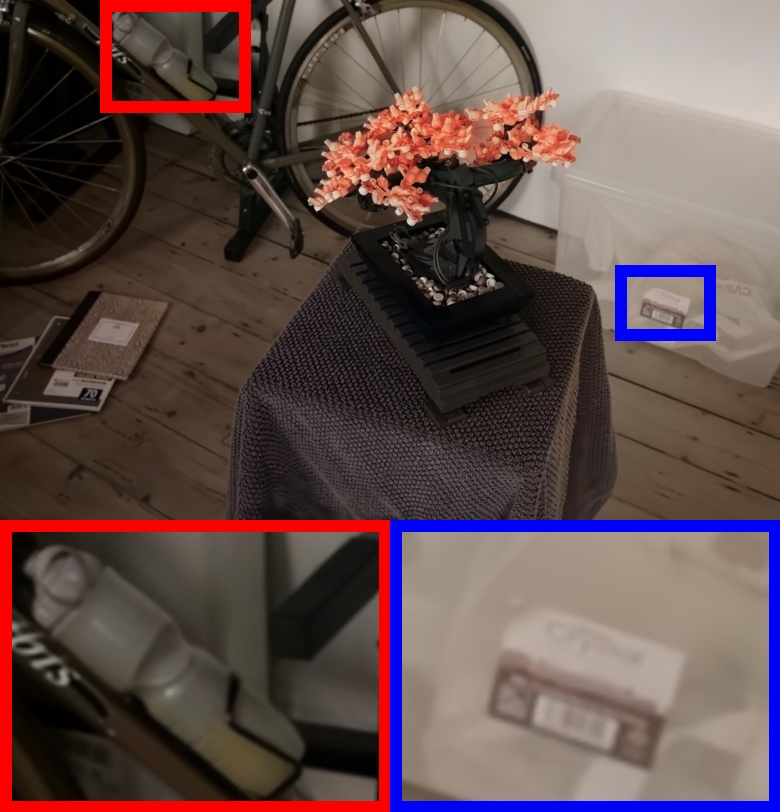} &
        \includegraphics[width=\imgw]{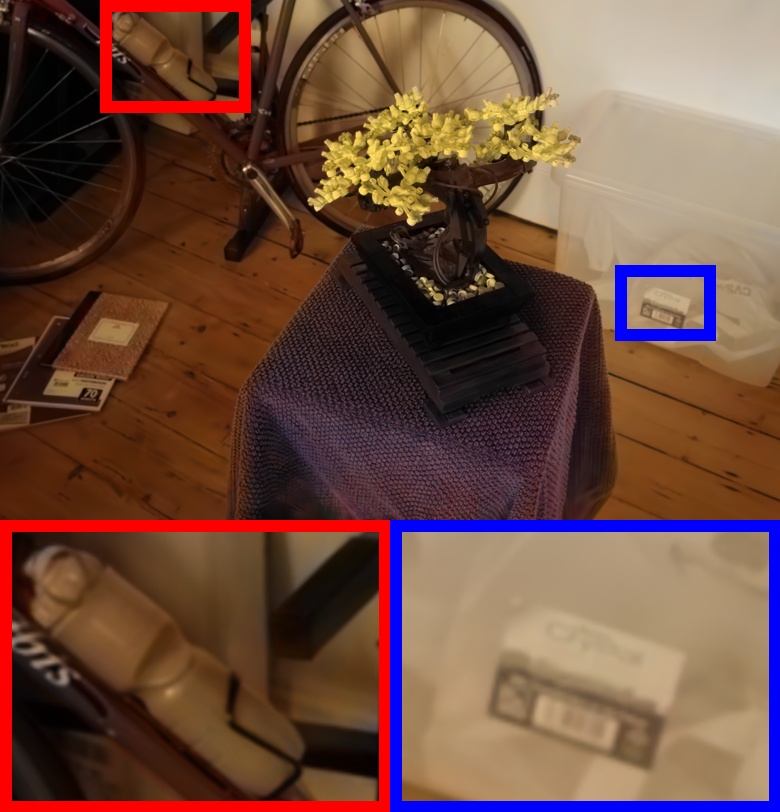} &
        \includegraphics[width=\imgw]{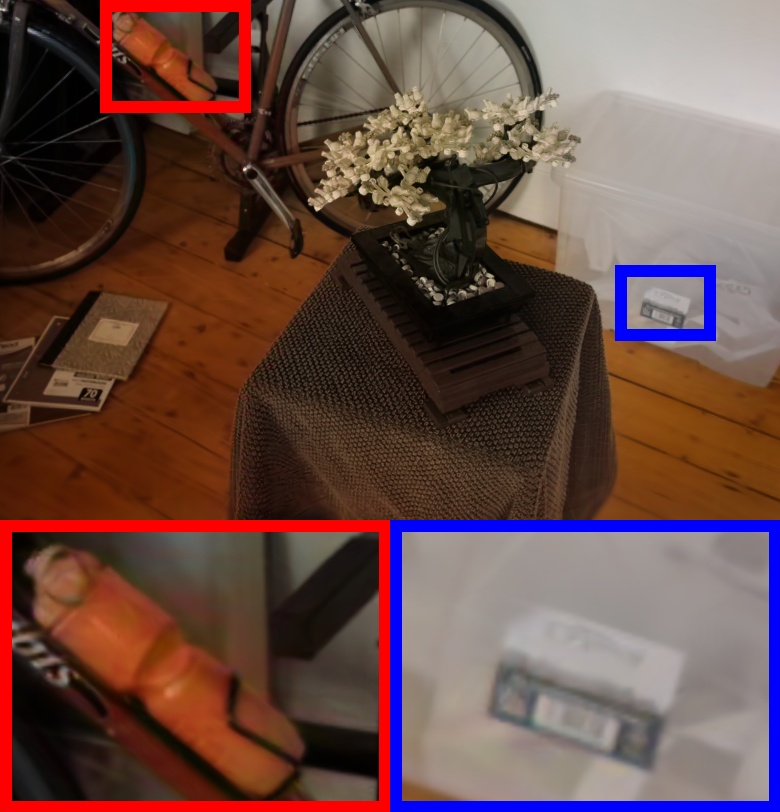} &
        \includegraphics[width=\imgw]{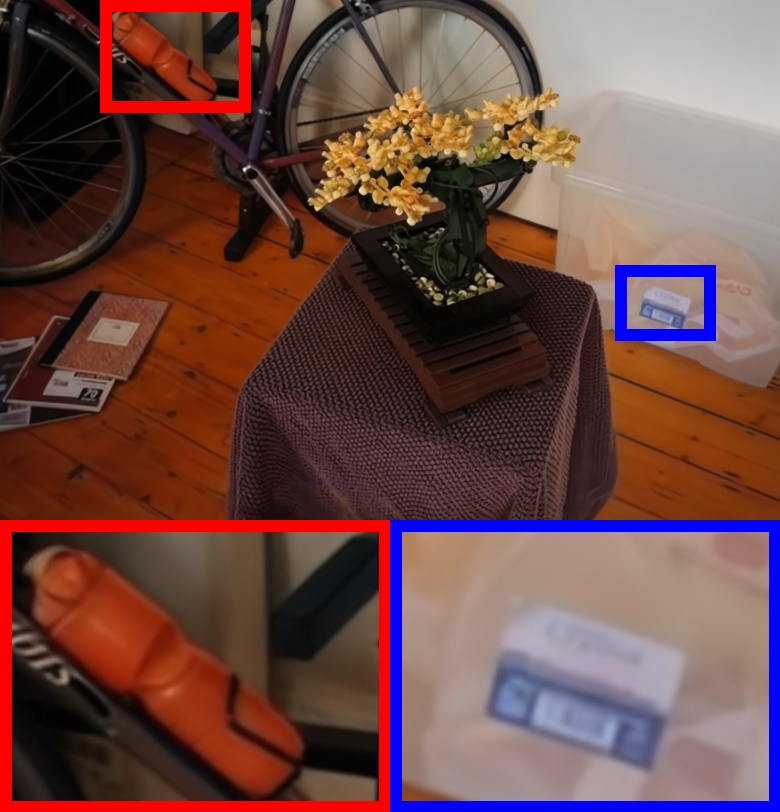} \\
        
        \raisebox{15pt}{\rotatebox{90}{360-Counter}} &
        \includegraphics[width=\imgw]{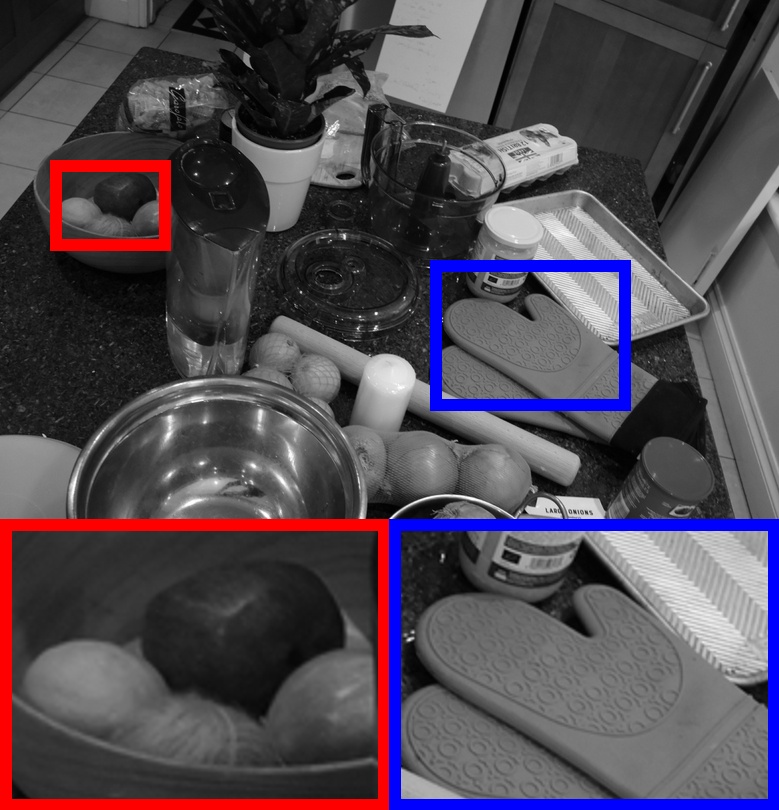} &
        \includegraphics[width=\imgw]{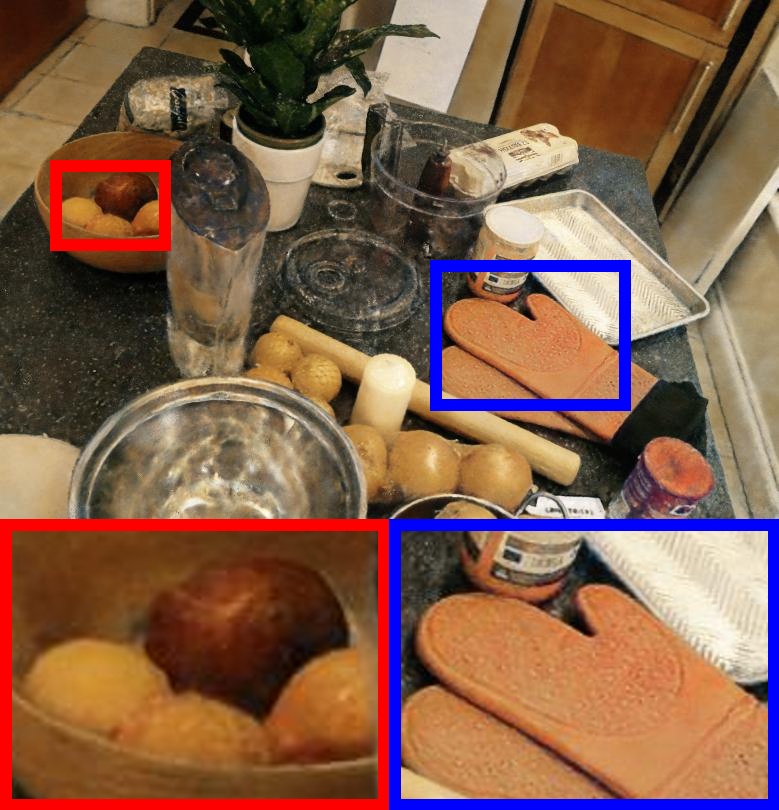} &
        \includegraphics[width=\imgw]{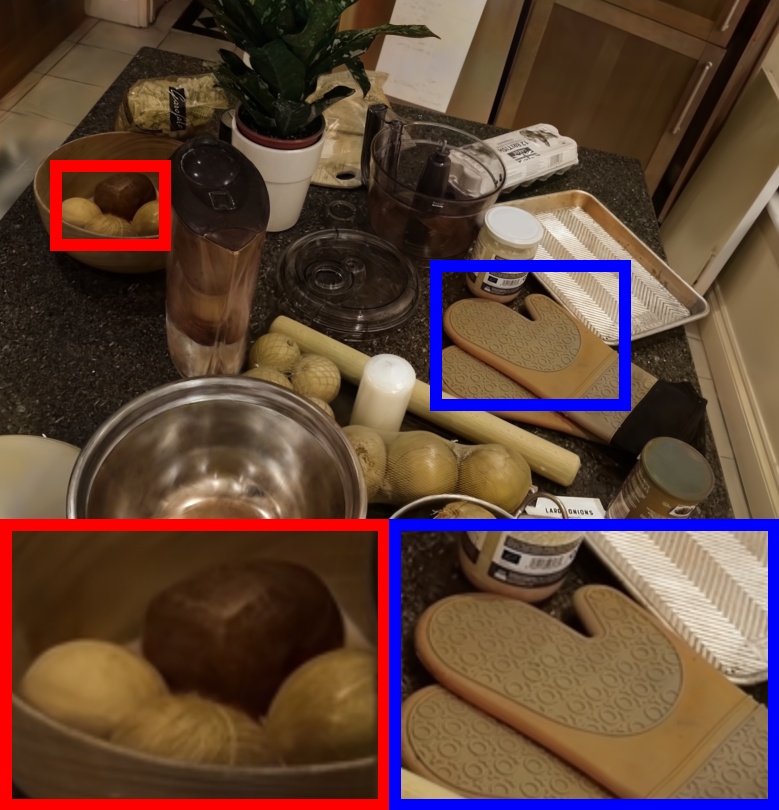} &
        \includegraphics[width=\imgw]{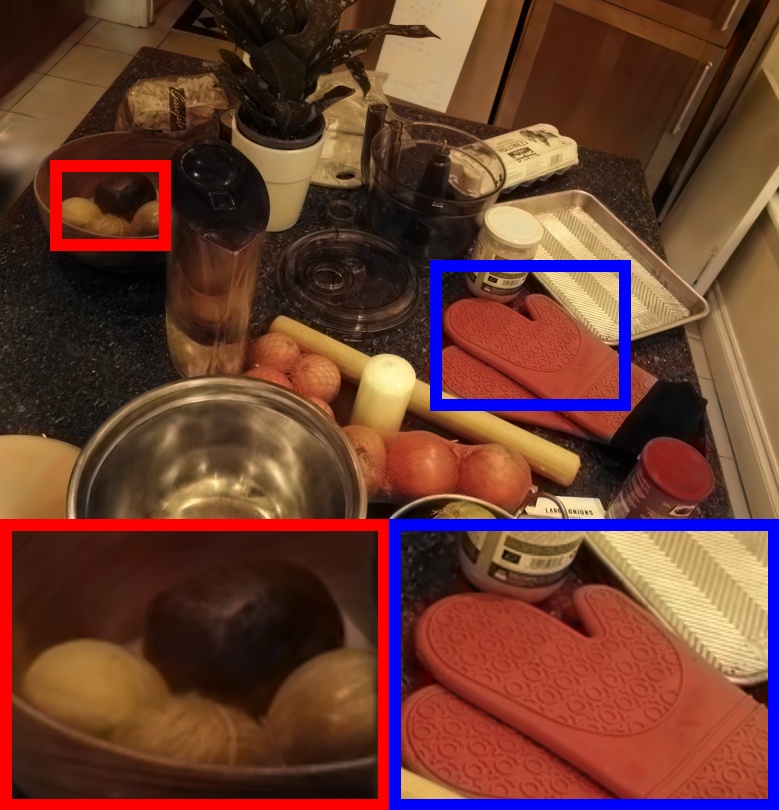} &
        \includegraphics[width=\imgw]{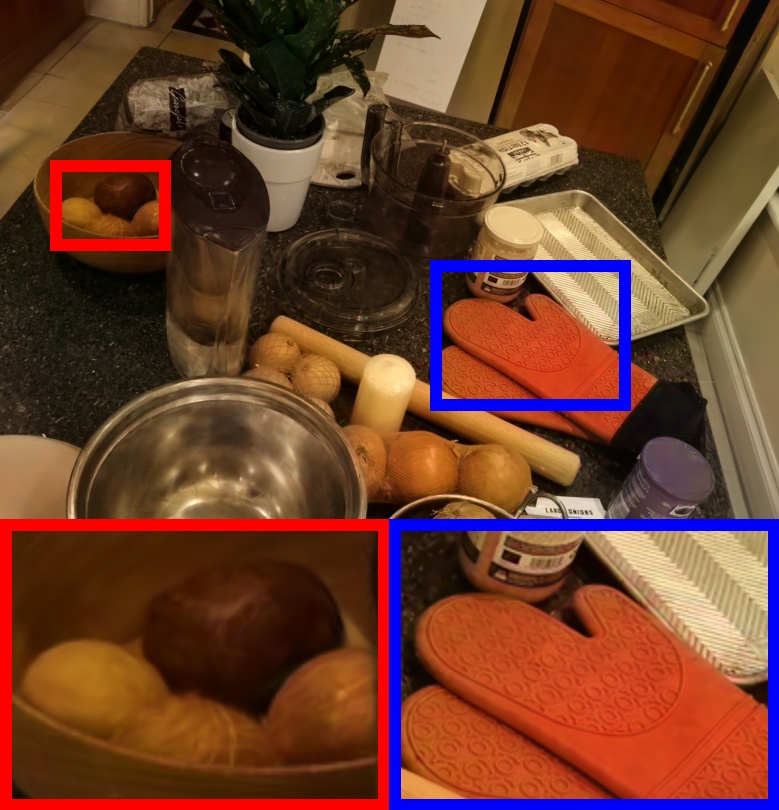} &
        \includegraphics[width=\imgw]{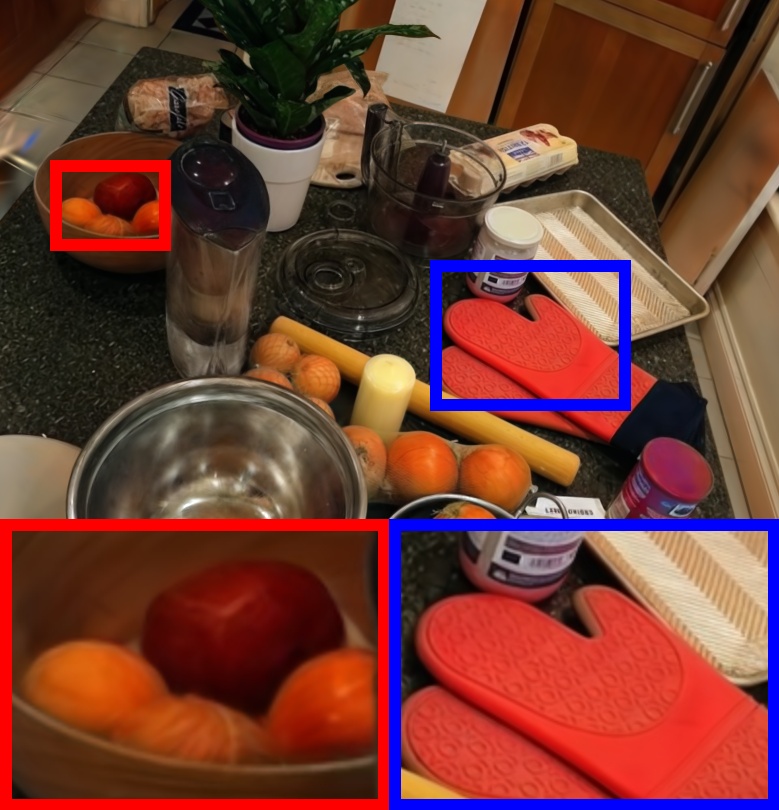} \\
        
        \raisebox{15pt}{\rotatebox{90}{360-Garden}} &
        \includegraphics[width=\imgw]{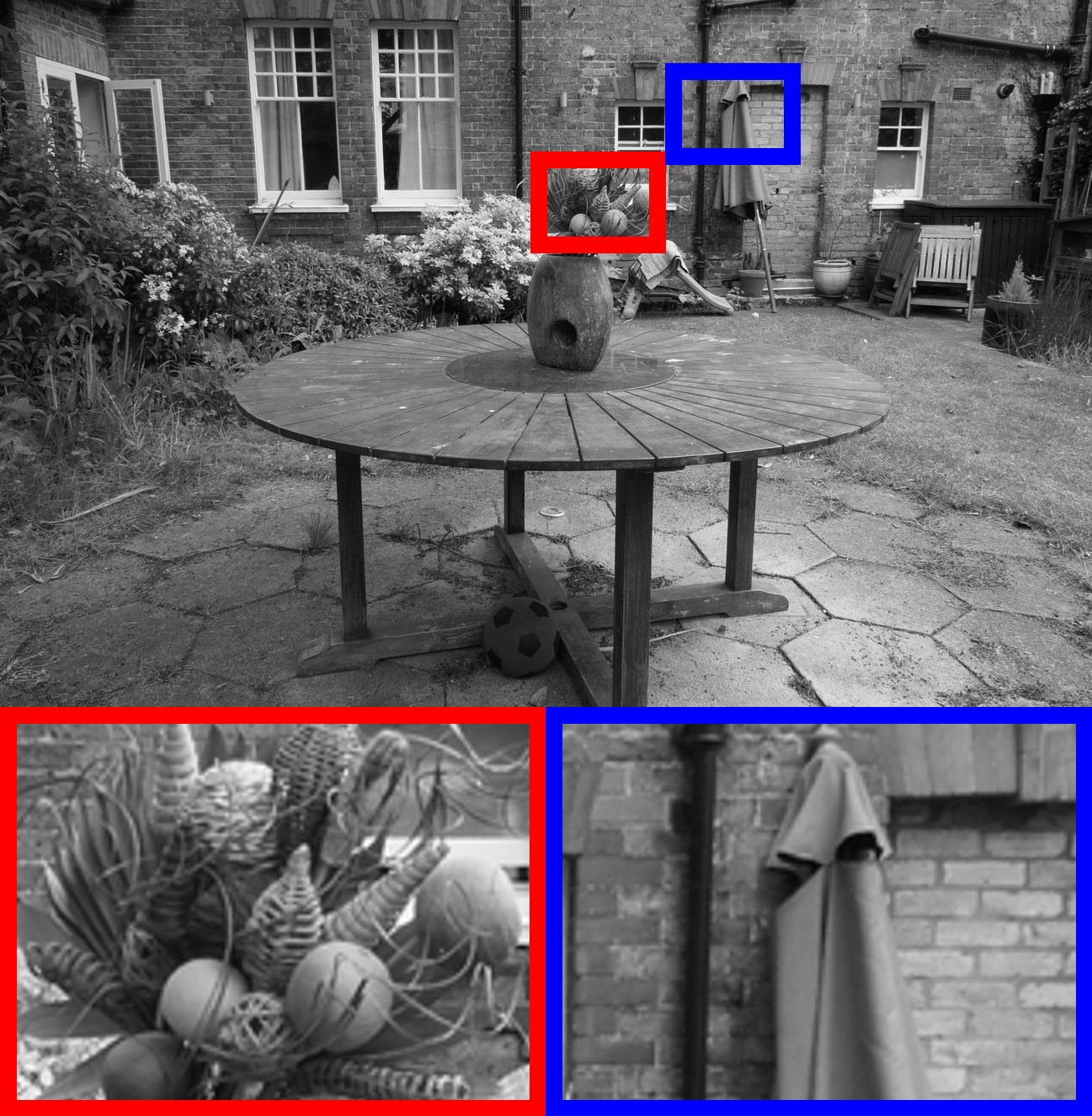} &
        \includegraphics[width=\imgw]{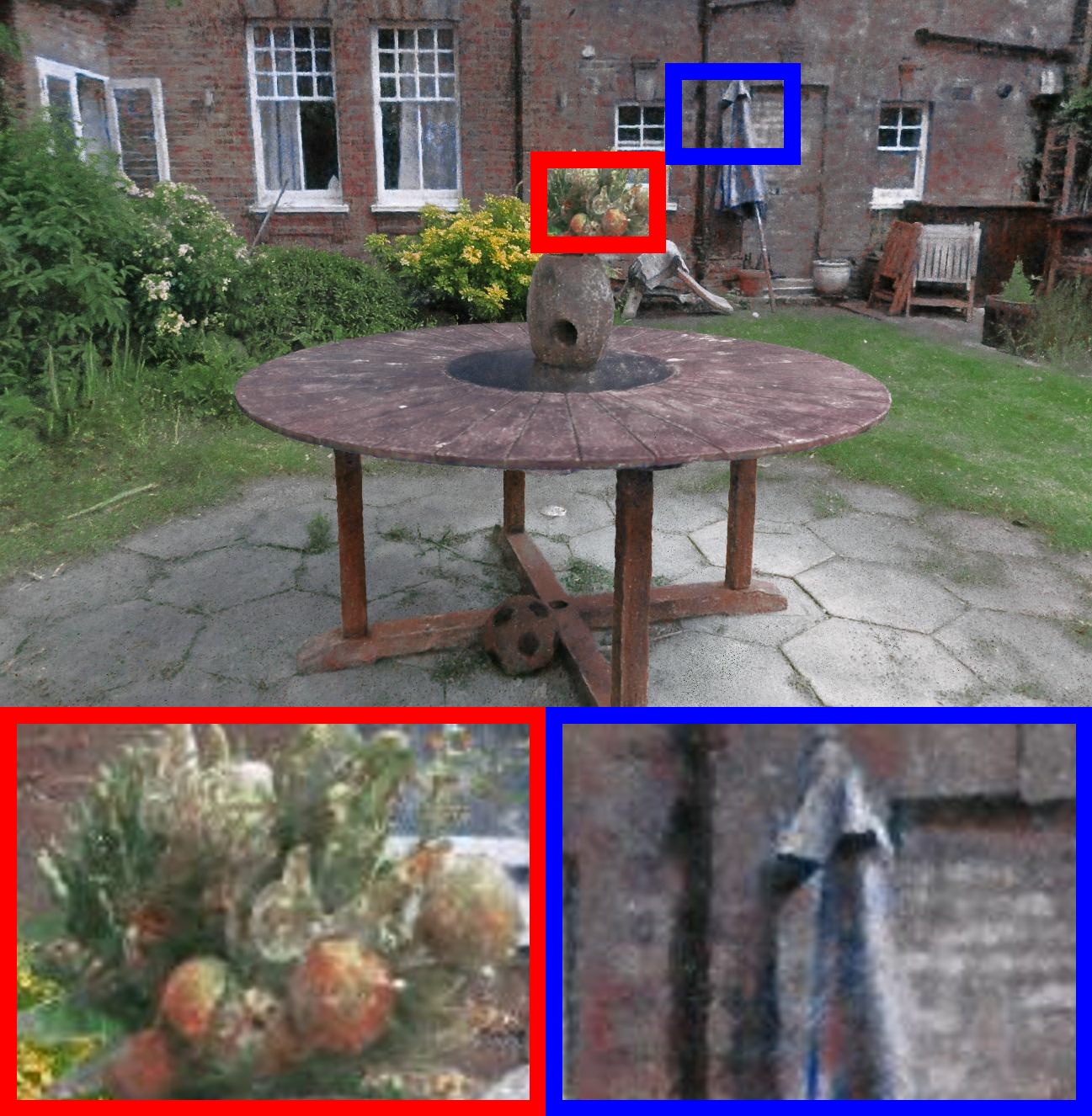} &
        \includegraphics[width=\imgw]{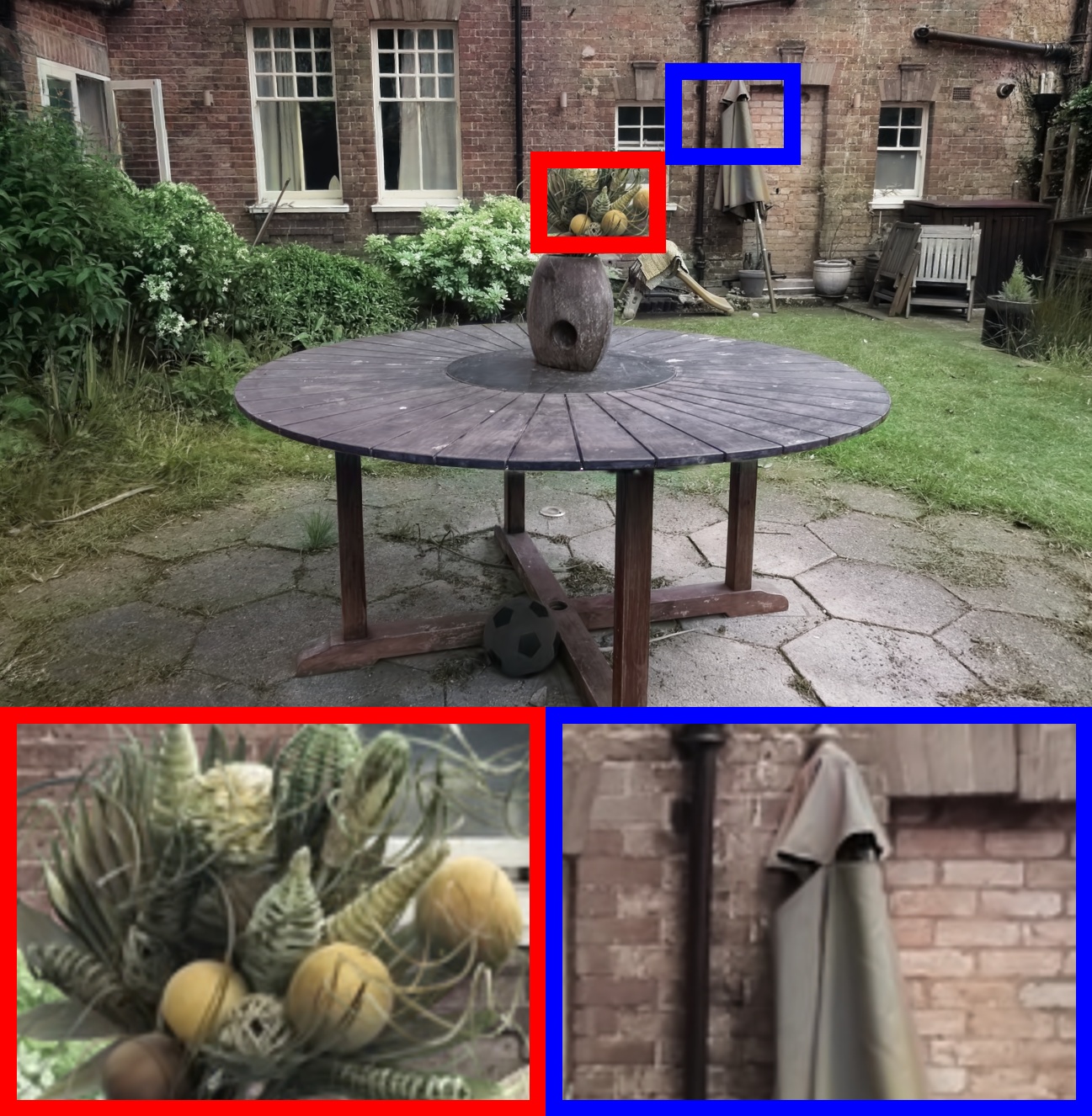} &
        \includegraphics[width=\imgw]{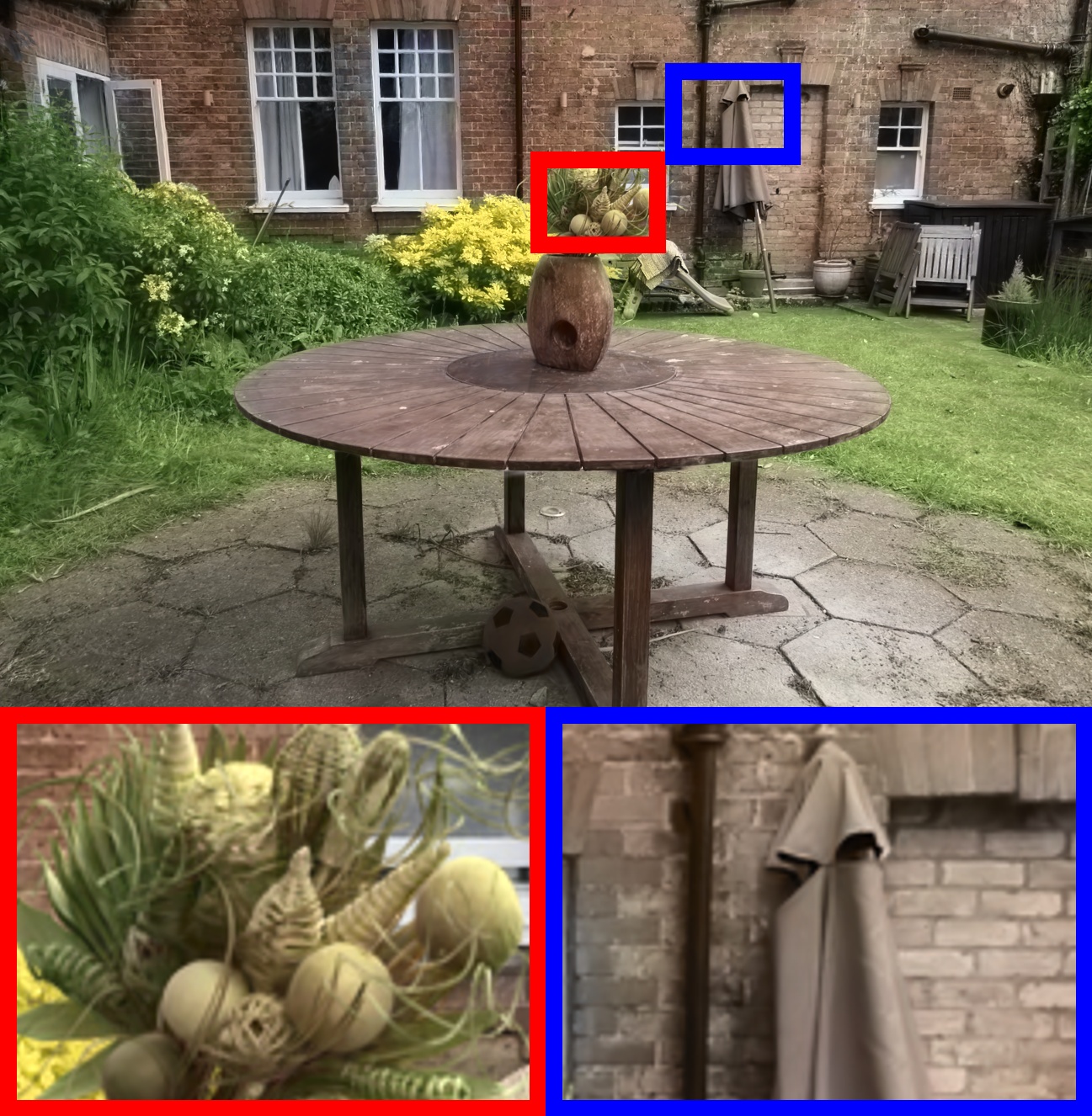} &
        \includegraphics[width=\imgw]{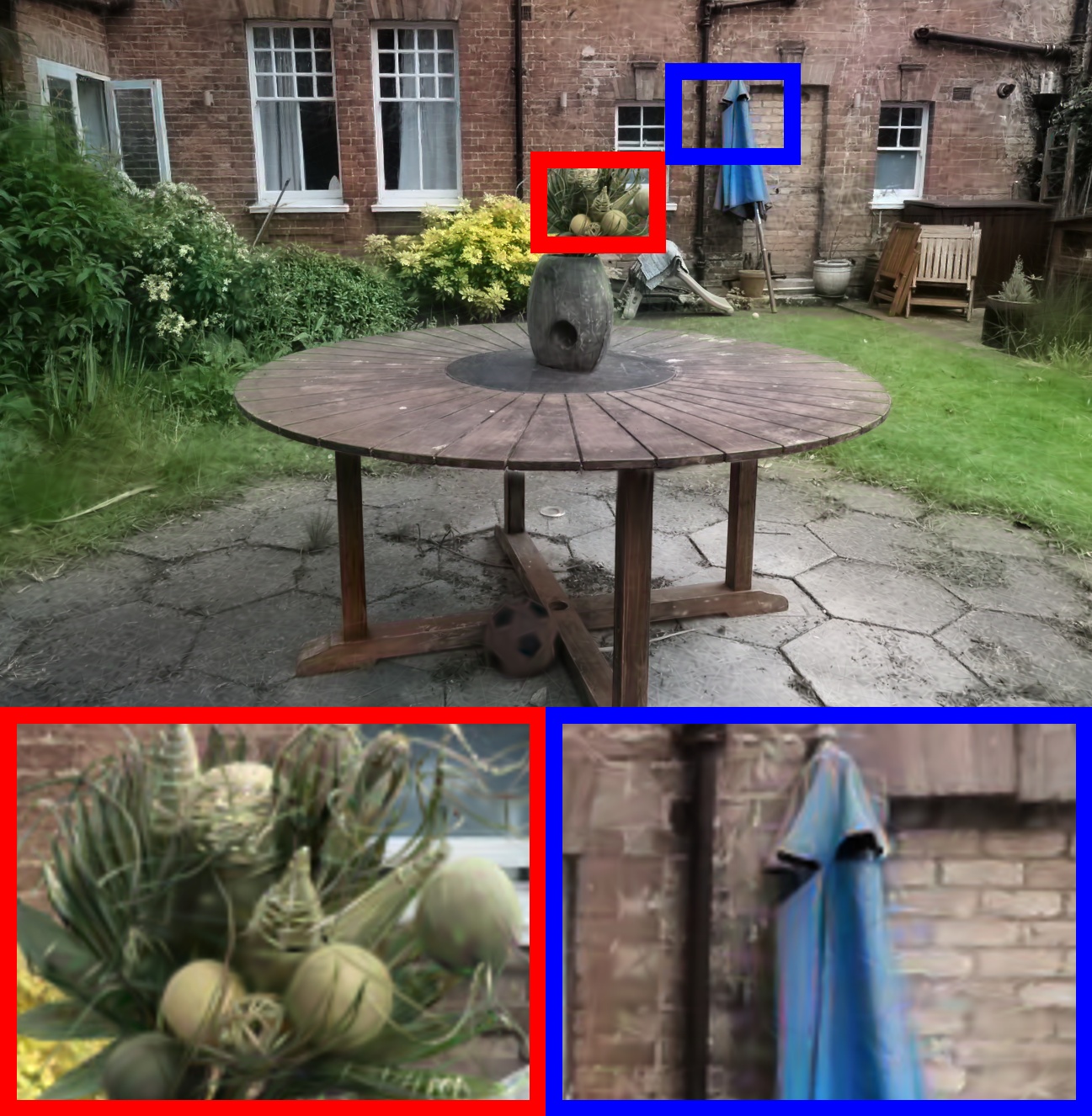} &
        \includegraphics[width=\imgw]{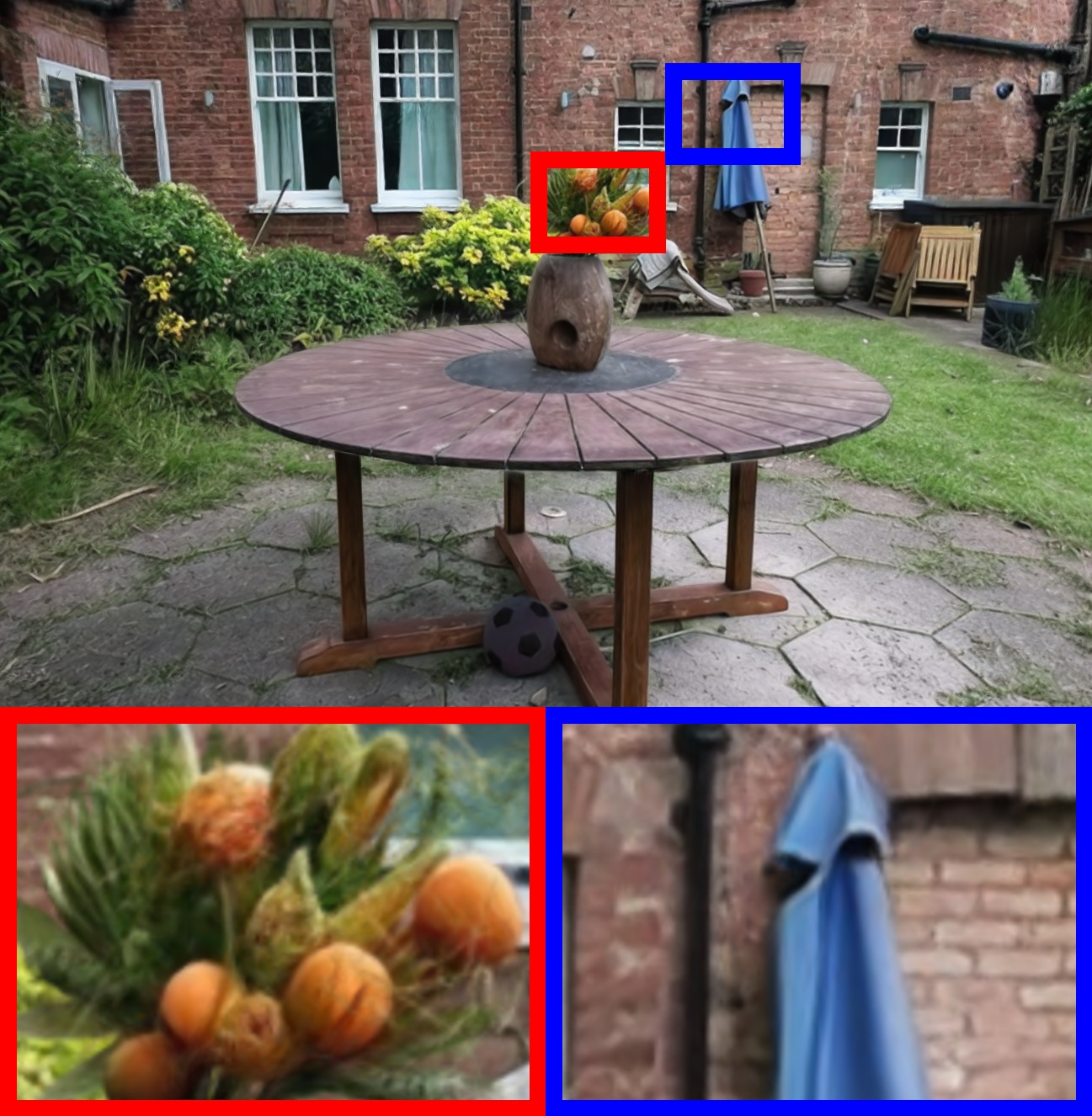} \\

        \raisebox{3pt}{\rotatebox{90}{DL3DV-Scene1}} &
        \includegraphics[width=\imgw]{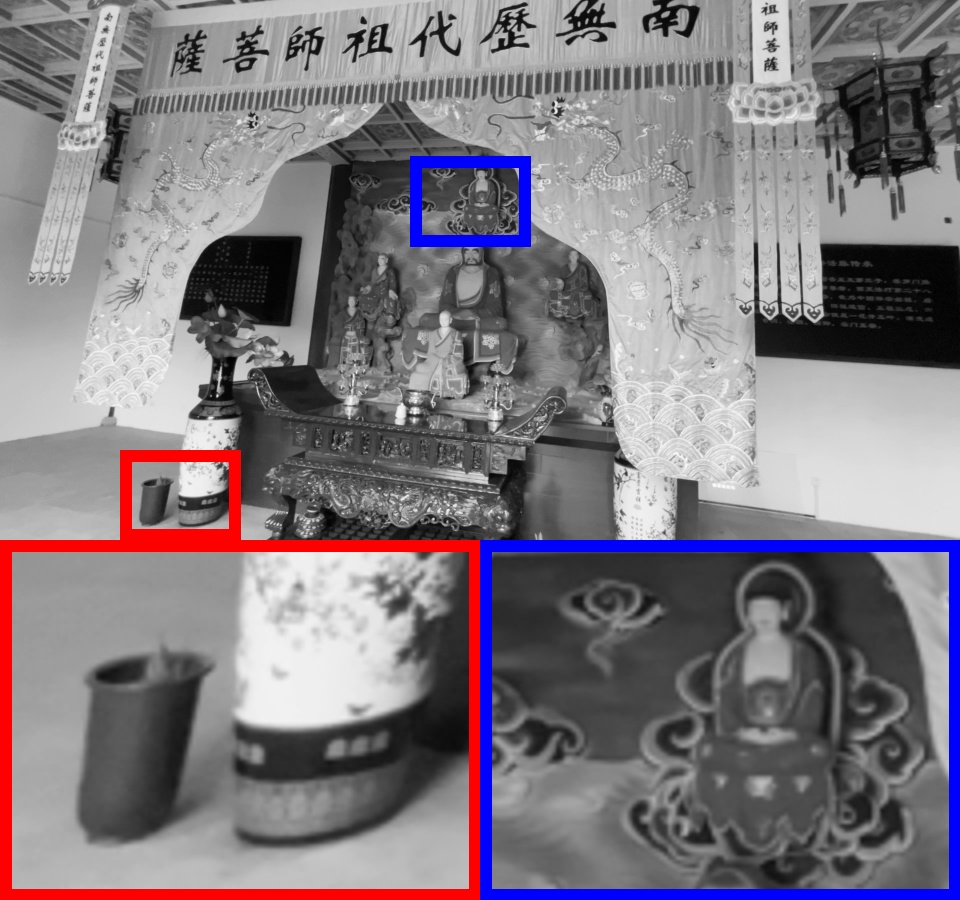} &
        \includegraphics[width=\imgw]{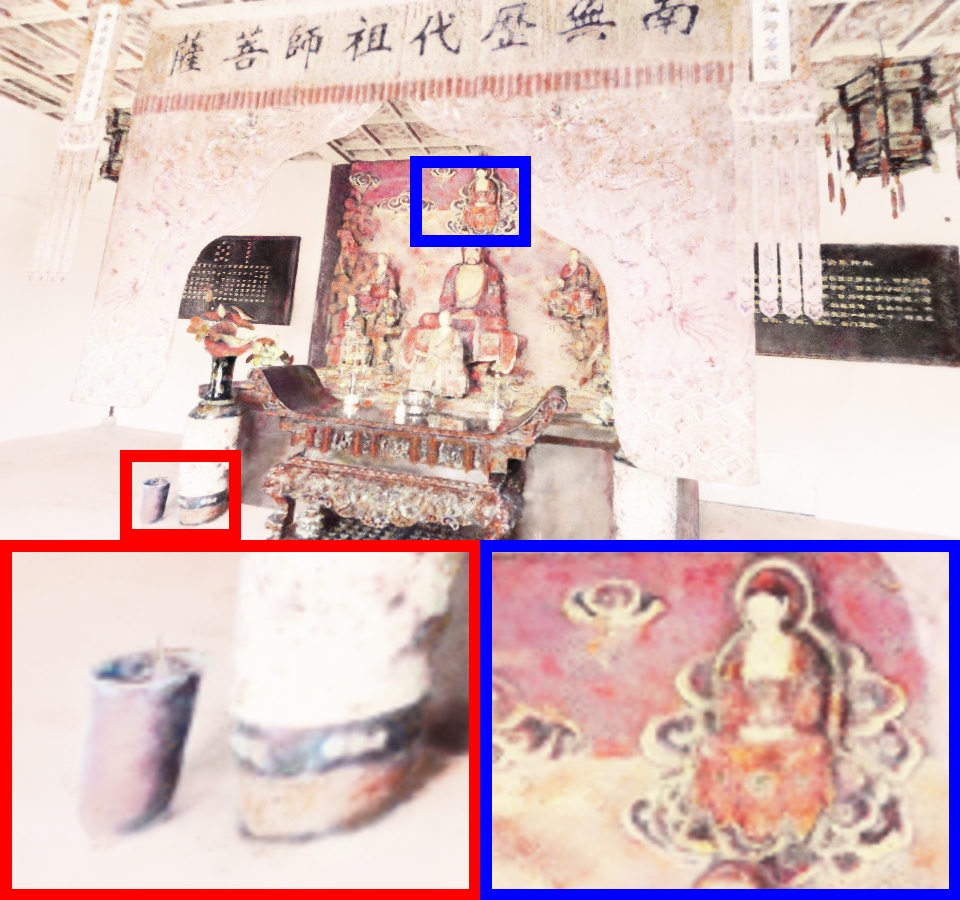} &
        \includegraphics[width=\imgw]{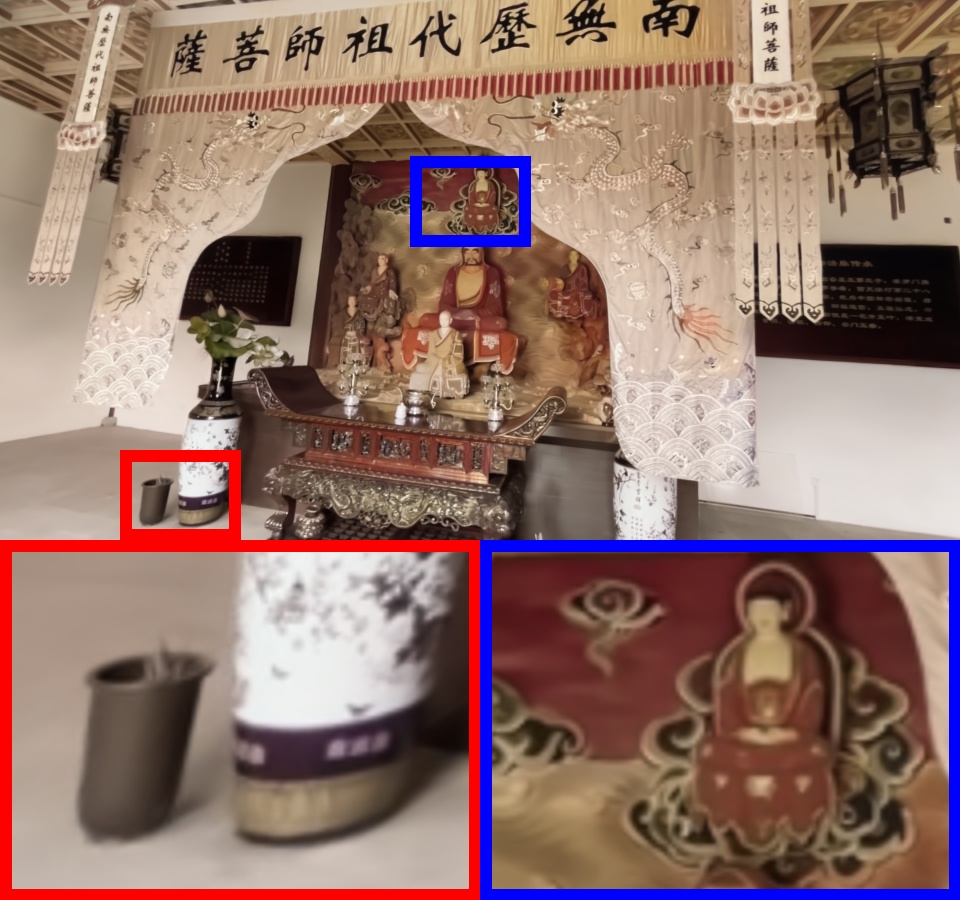} &
        \includegraphics[width=\imgw]{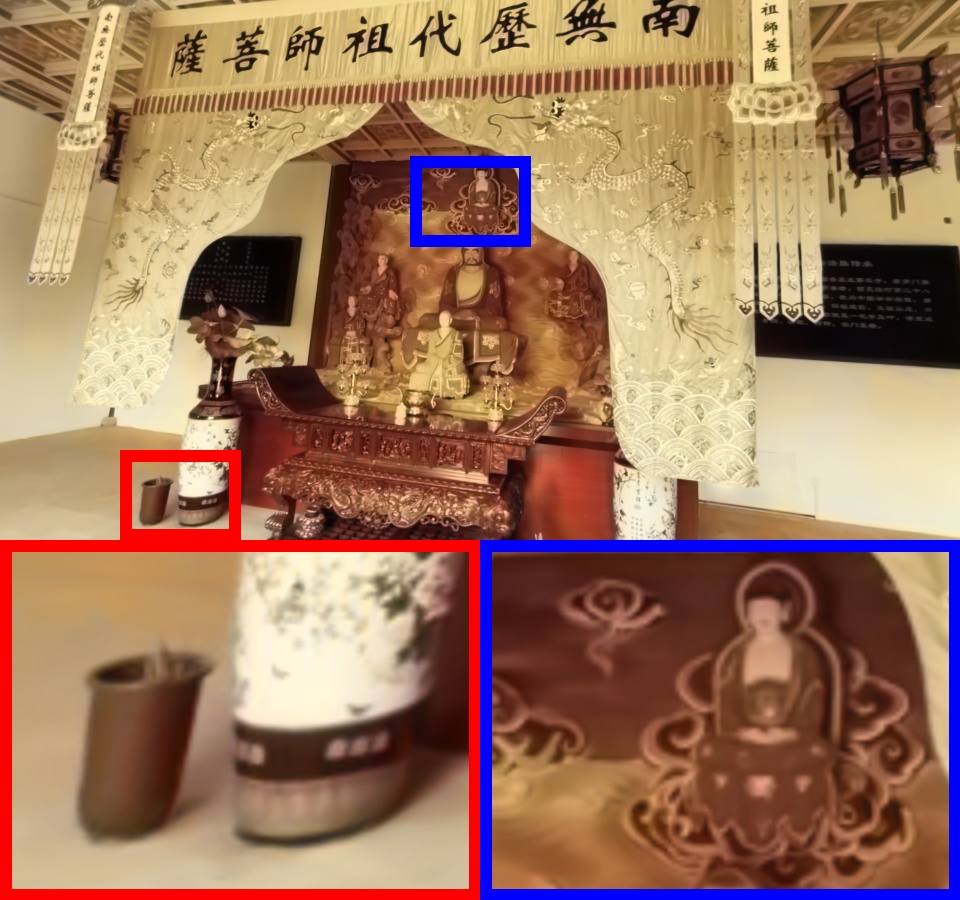} &
        \includegraphics[width=\imgw]{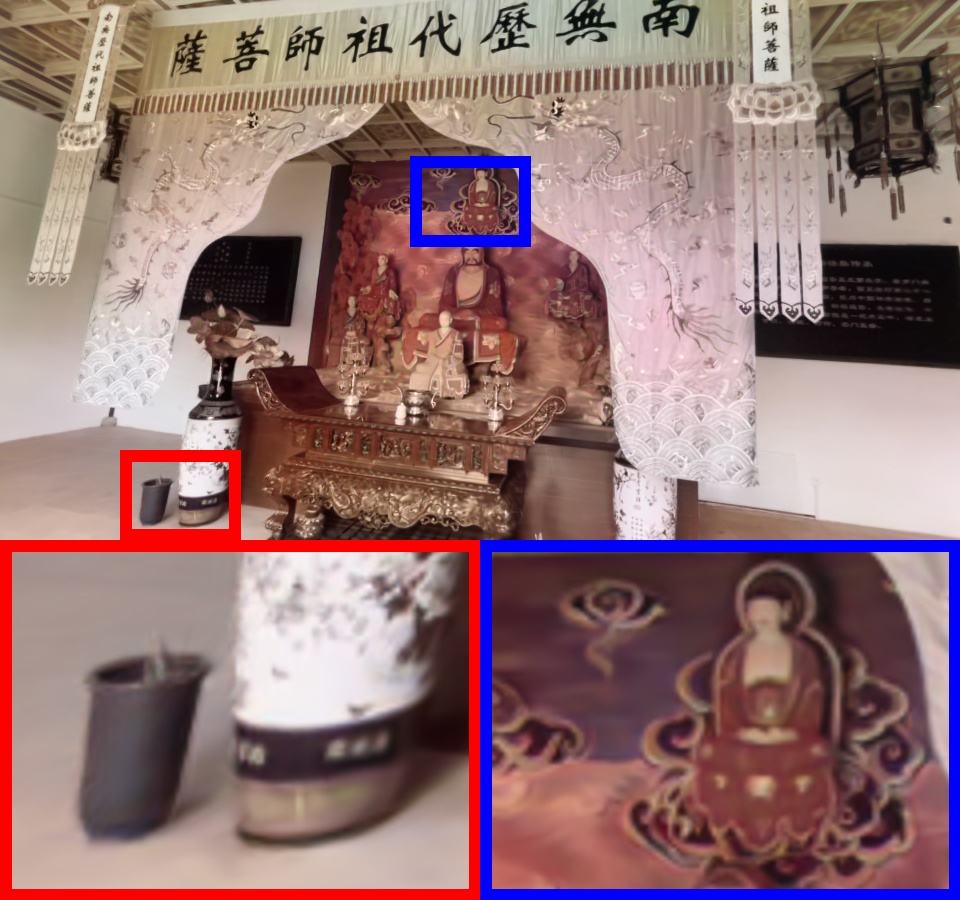} &
        \includegraphics[width=\imgw]{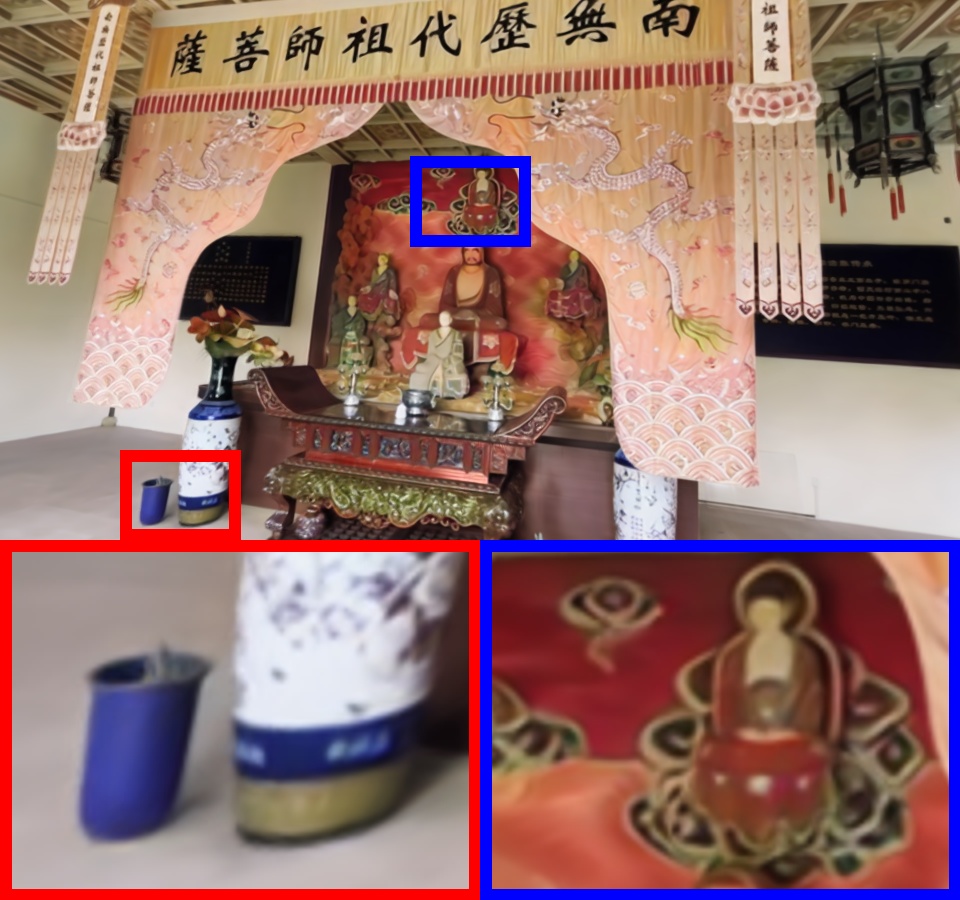} \\
        
        \raisebox{3pt}{\rotatebox{90}{DL3DV-Scene2}} &
        \includegraphics[width=\imgw]{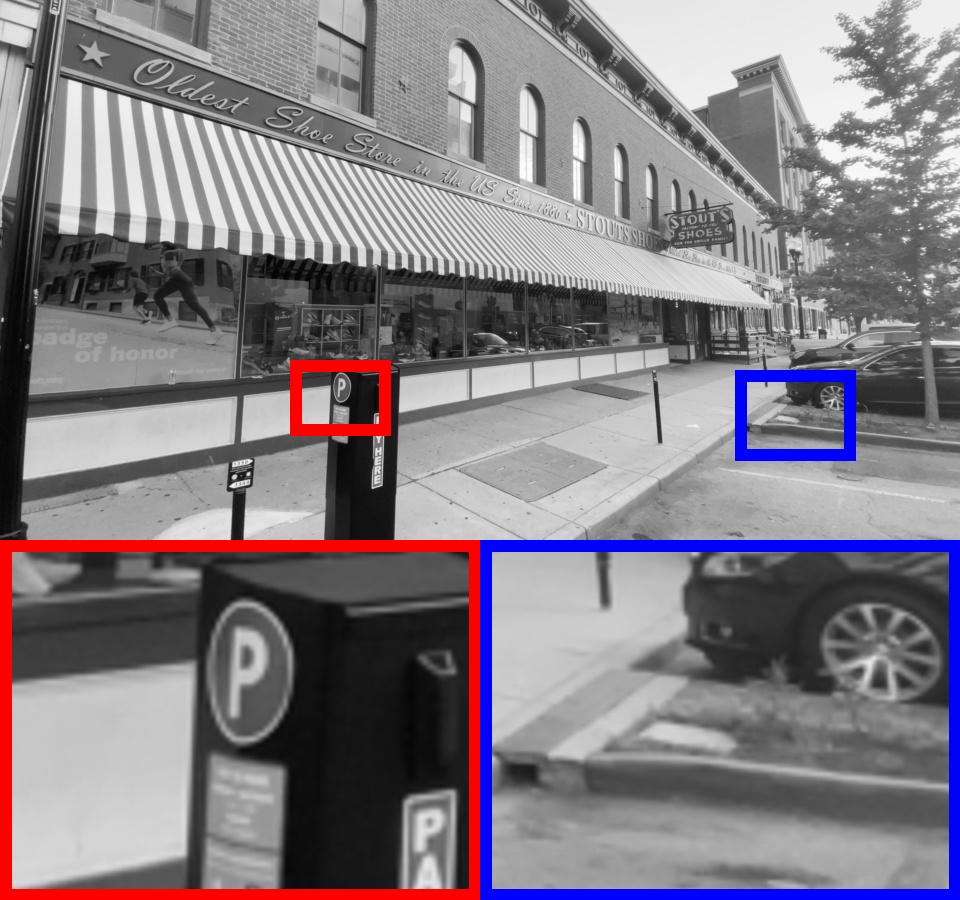} &
        \includegraphics[width=\imgw]{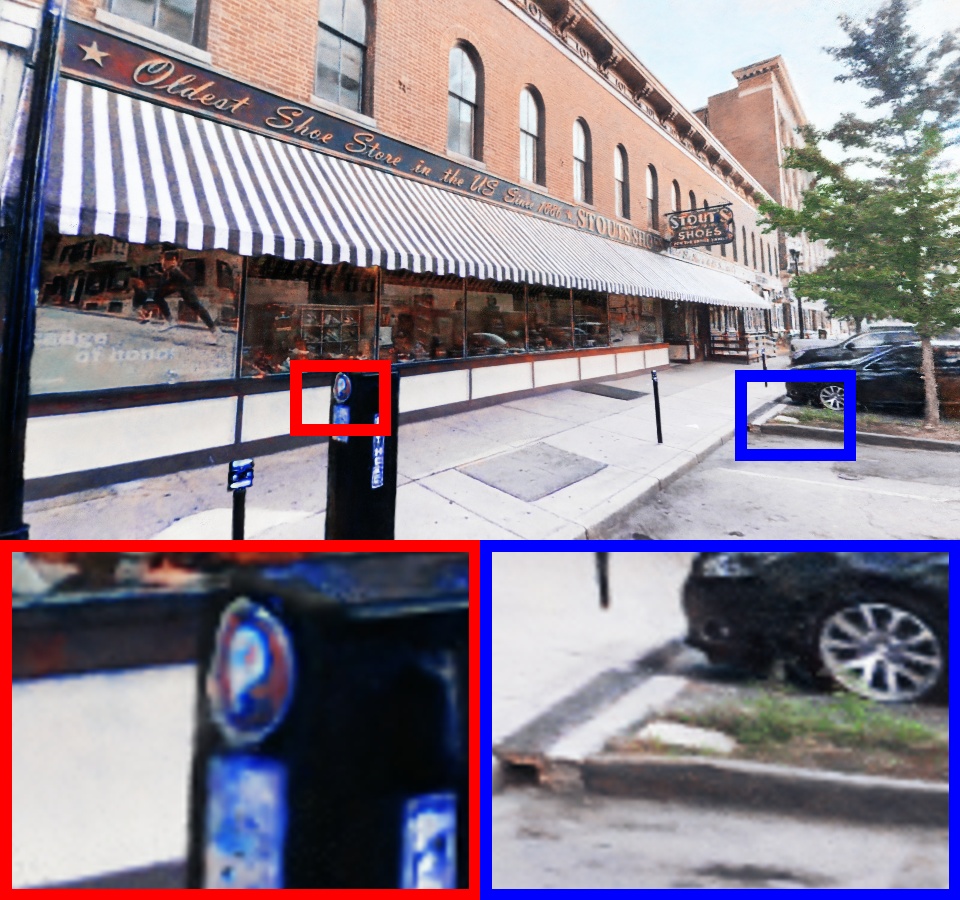} &
        \includegraphics[width=\imgw]{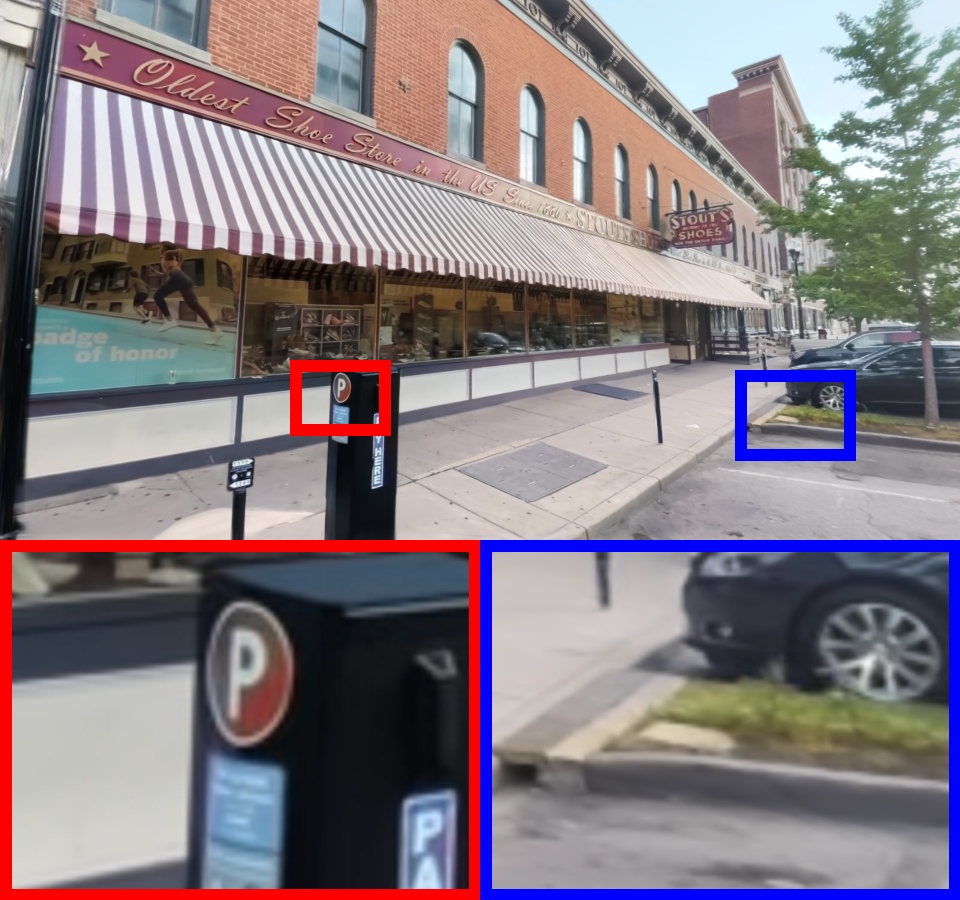} &
        \includegraphics[width=\imgw]{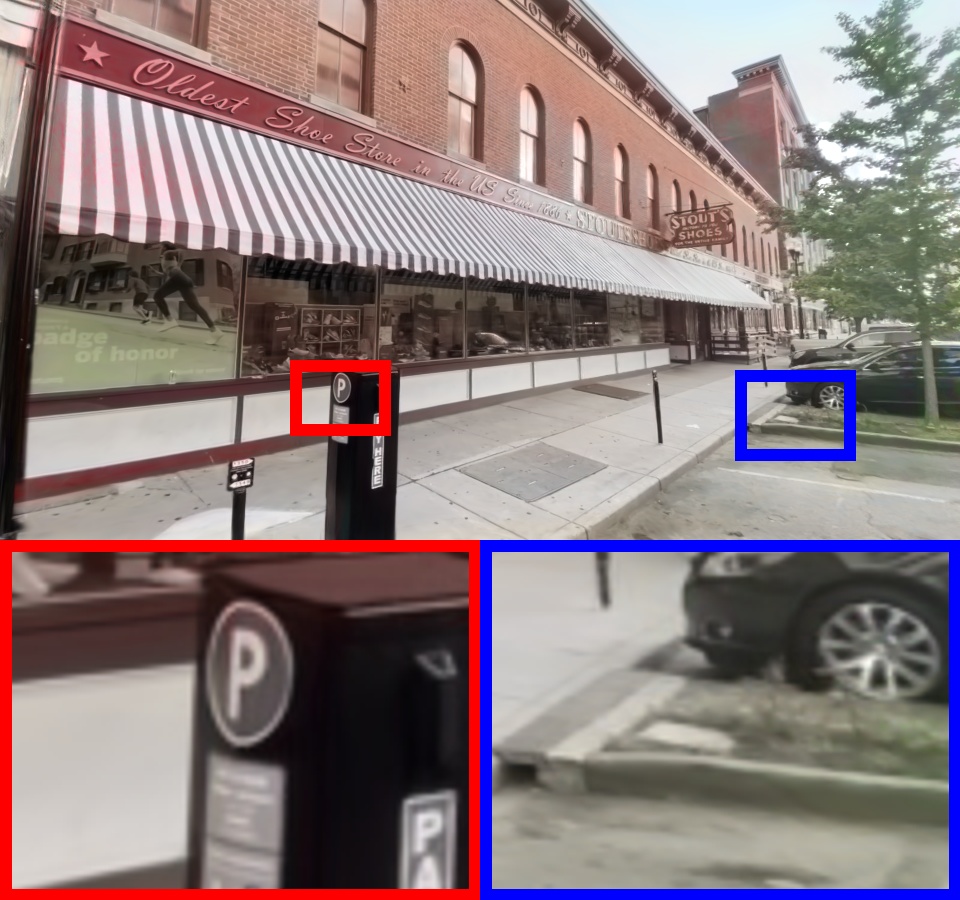} &
        \includegraphics[width=\imgw]{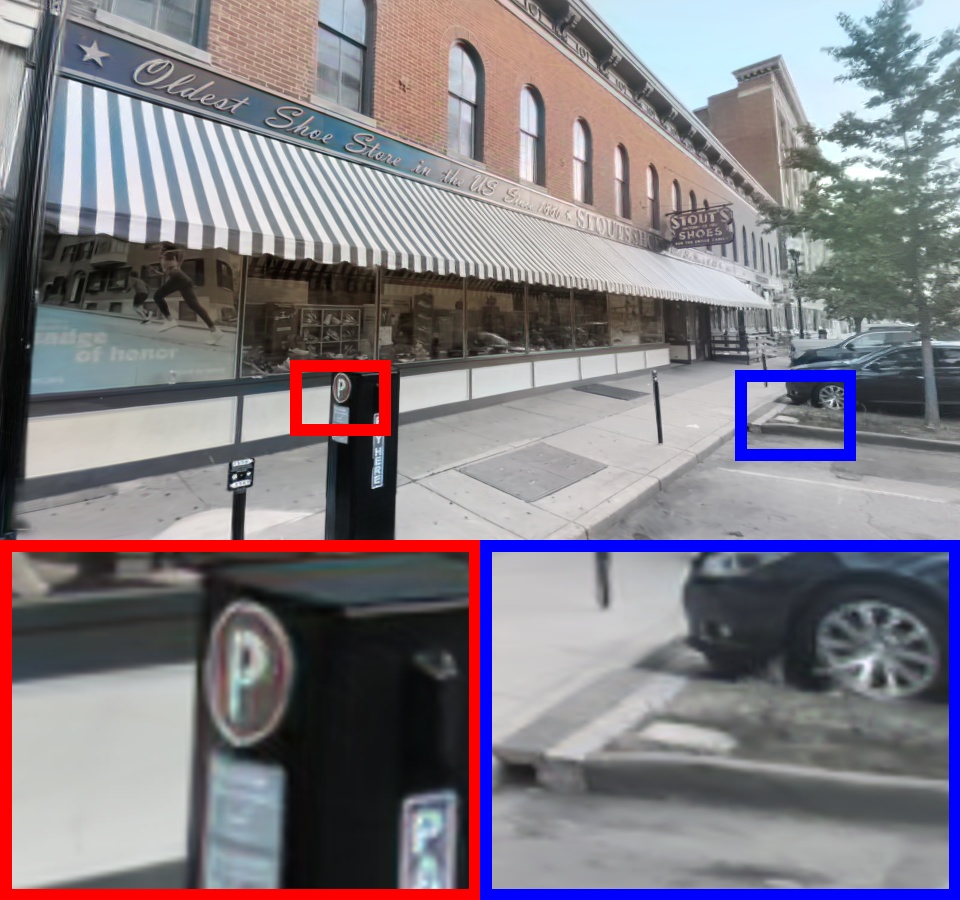} &
        \includegraphics[width=\imgw]{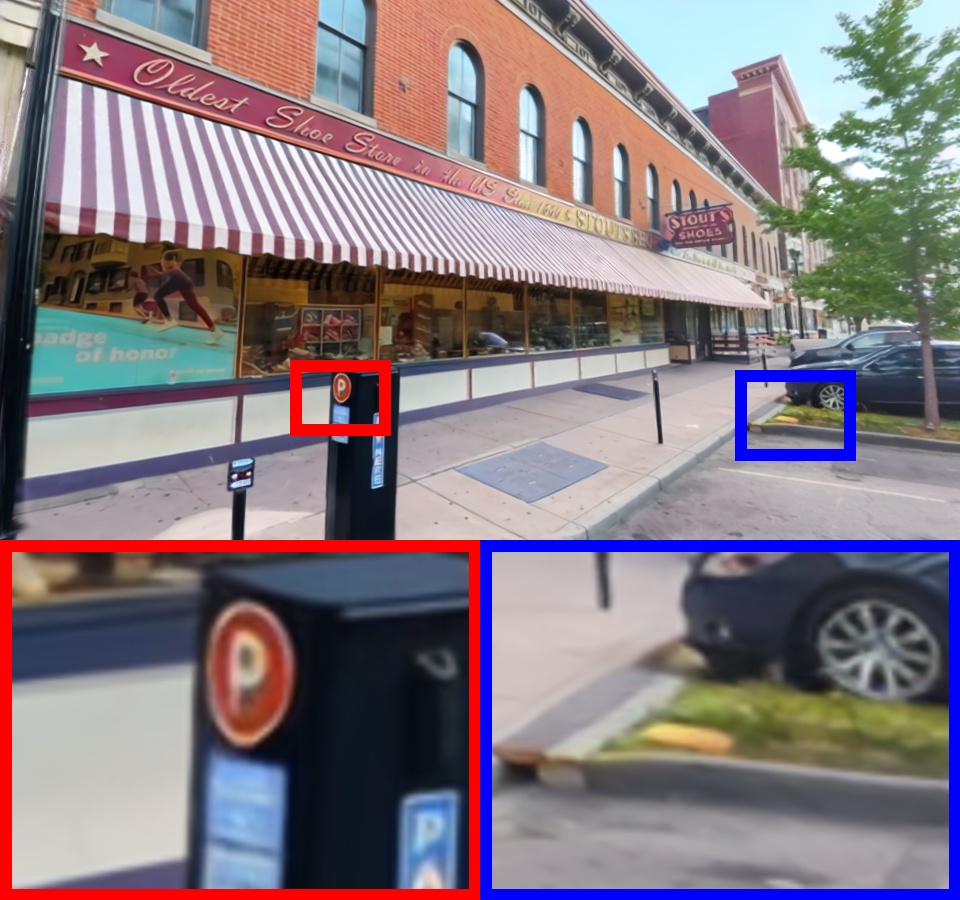} \\        
        
    \end{tabular}
    }
    \caption{\textbf{Qualitative comparison on Tanks and Temples (TnT), Mip-NeRF 360 (360) and DL3DV-10K-Benchmarks (DL3DV) --} Our method successfully reconstructs plausible and diverse colors across all regions, effectively preserving fine-grained details such as the blue label in `Bonsai', the fruits in `Garden', and the street lamp and sign in `Truck'. In contrast, existing works often struggle in these complex scenes, frequently yielding monotonous results and failing to colorize small objects. This is primarily due to the averaging effect during 3D optimization, where inconsistencies across multiple views lead to the loss of distinct color information.
    }
    \label{fig:qualitative}
\end{figure}

\section{Experiments}
\subsection{Experimental setup}

\subsubsection{Implementation details.}
We implement our method using the PyTorch framework, based on the official 3DGS implementation. 
We train the initial single-channel 3DGS model for 30K iterations. 
For the scene decomposition, we empirically set the number of subscenes $K=4$ for unbounded datasets, and  $K=1$ for the forward-facing dataset.
For our multi-view colorization model $\Phi_{MV}$, we fine-tune the pre-trained SD-Turbo model~\cite{sauer2024adversarial} by integrating the reference mixing layer from~\cite{difix3d}. 
We train $\Phi_{MV}$ for 25K iterations using the AdamW optimizer~\cite{loshchilovdecoupled} with a constant learning rate $2 \times 10^{-5}$. 
For our fine-tuning loss $\mathcal{L}_{\text{fine-tune}}$, we set the perceptual weight to $\lambda_{\text{LPIPS}} = 1.0$ and $\lambda_{\text{Gram}} = 0.5$.
For our fine-tuning dataset, we utilize two sources, which are mixed during training:
\begin{itemize}
    \item DL3DV-10K-Benchmark~\cite{ling2024d13dv}: Used to train multi-view referencing. We select one reference view from each scene and set others as grayscale inputs, yielding 4,920 input pairs across 112 scenes.
    \item Flickr8k~\cite{hodosh2013flickr8k}: Used to learn the color distribution for image-to-image translation. We apply random crops at 80-100\% of the original scale, yielding 8,000 input pairs.
\end{itemize}
We use DDColor~\cite{kang2023ddcolor} as our plug-and-play image colorization model $\mathcal{F}$. 
The final 3D color component optimization is run for 7K iterations. 
We provide further details in the supplementary material.

\begin{table}[t]
\centering
\small
\caption{\textbf{Quantitative comparison on 360-degree datasets --} We provide quantitative results across 360-degree datasets. SC and LC denote short- and long-term consistency, respectively, while ``--'' indicates values unavailable in the Color3D~\cite{color3d} paper. \firstc{First}, \secondc{second}, and \thirdc{third} best results are highlighted. Overall, our method achieves the best or second-best results across the majority of benchmarks and metrics.}
\vspace{-5pt}
\label{tab:360}
\resizebox{0.99\textwidth}{!}{%

\begin{tabular}{l|ccccc|ccccc|ccccc}
\toprule
& \multicolumn{5}{c|}{Mip-NeRF360}
& \multicolumn{5}{c|}{TnT}
& \multicolumn{5}{c}{DL3DV} \\
Method
& FID$\downarrow$ & Color$\uparrow$ & nColor$\uparrow$ & SC$\downarrow$ & LC$\downarrow$ 
& FID$\downarrow$ & Color$\uparrow$ & nColor$\uparrow$ & SC$\downarrow$ & LC$\downarrow$ 
& FID$\downarrow$ & Color$\uparrow$ & nColor$\uparrow$ & SC$\downarrow$ & LC$\downarrow$ \\
\midrule
GenN2N 
& 58.17 & 31.82 & 19.45 & 0.015 & \second{0.020} 
& 45.66 & 22.92 & 17.29 & 0.011 & 0.021 
& 44.85 & \second{34.64} & \third{22.48} & 0.010 & 0.014 \\

ColorNeRF-GS 
& 47.86 & \first{41.84} & \third{21.23} & \third{0.012} & 0.023 
& 36.22 & \third{27.01} & \third{20.68} & \third{0.007} & \third{0.019} 
& 45.94 & \third{33.62} & 21.51 & \second{0.005} & \second{0.011} \\

ChromaDistill-GS 
& 52.42 & \second{39.84} & \second{22.16} & 0.014 & 0.027 
& \third{33.88} & \second{27.03} & \second{21.95} & 0.013 & 0.031 
& 38.92 & 33.31 & \second{23.49} & 0.008 & 0.020 \\

ColorMNet 
& \third{45.13} & 27.69 & 16.83 & \first{0.008} & \first{0.013} 
& \second{30.64} & 22.18 & 16.89 & \first{0.006} & \first{0.013} 
& \second{35.54} & 28.39 & 20.64 & \first{0.004} & \first{0.008} \\

Color3D 
& \second{39.03} & 33.36 & -  & \multicolumn{2}{c|}{0.016}
& -  & -  & -  & -  & -  
& \third{37.48} & 32.65 & -  & \multicolumn{2}{c}{0.017} \\

Ours 
& \first{38.42} & \third{37.32} & \first{23.50} & \second{0.011} & \second{0.020} 
& \first{18.02} & \first{29.56} & \first{23.32} & \first{0.006} & \second{0.016} 
& \first{34.92} & \first{39.35} & \first{32.61} & \second{0.005} & \third{0.012} \\

\bottomrule
\end{tabular}
}
\end{table}

\begin{figure}[t]
\centering
\small

\begin{minipage}{0.43\textwidth}
\captionof{table}{\textbf{Quantitative comparison on LLFF}}
    \resizebox{1.0\linewidth}{!}{%
    \begin{tabular}{l|ccccc}
    \toprule
    Method
    & FID$\downarrow$ & Color$\uparrow$ & nColor$\uparrow$ & SC$\downarrow$ & LC$\downarrow$ \\
    \midrule
    GenN2N 
    & 61.91 & 30.71 & 21.78 & 0.012 & 0.015 \\
    
    ColorNeRF 
    & 76.19 & 35.58 & 22.93 & 0.010 & 0.017 \\
    
    ColorNeRF-GS 
    & 68.47 & 34.85 & 20.29 & 0.008 & 0.014 \\
    
    ChromaDistill 
    & 72.07 & 22.29 & 16.99 & \second{0.007} & \first{0.008} \\
    
    ChromaDistill-GS 
    & 86.93 & \first{37.92} & \first{27.75} & 0.015 & 0.026 \\
    
    ColorMNet 
    & \third{58.92} & \second{36.29} & \third{25.24} & \first{0.006} & \second{0.010} \\
    
    Color3D 
    & \first{35.10} & 33.99 & -  & \second{0.007} & - \\
    
    Ours 
    & \second{58.20} & \third{35.59} & \second{26.39} & \second{0.007} & \third{0.012} \\
    \bottomrule
    \end{tabular}
    }
    \label{tab:llff}
\end{minipage}
\hfill
\begin{minipage}{0.55\textwidth}
\centering
\newcommand{\imgw}{0.4\textwidth}
\resizebox{1.0\linewidth}{!}{%
\begin{tabular}{cccc}
    Input & GenN2N & ColorMNet & ChromaDistill \\
    \includegraphics[width=\imgw]{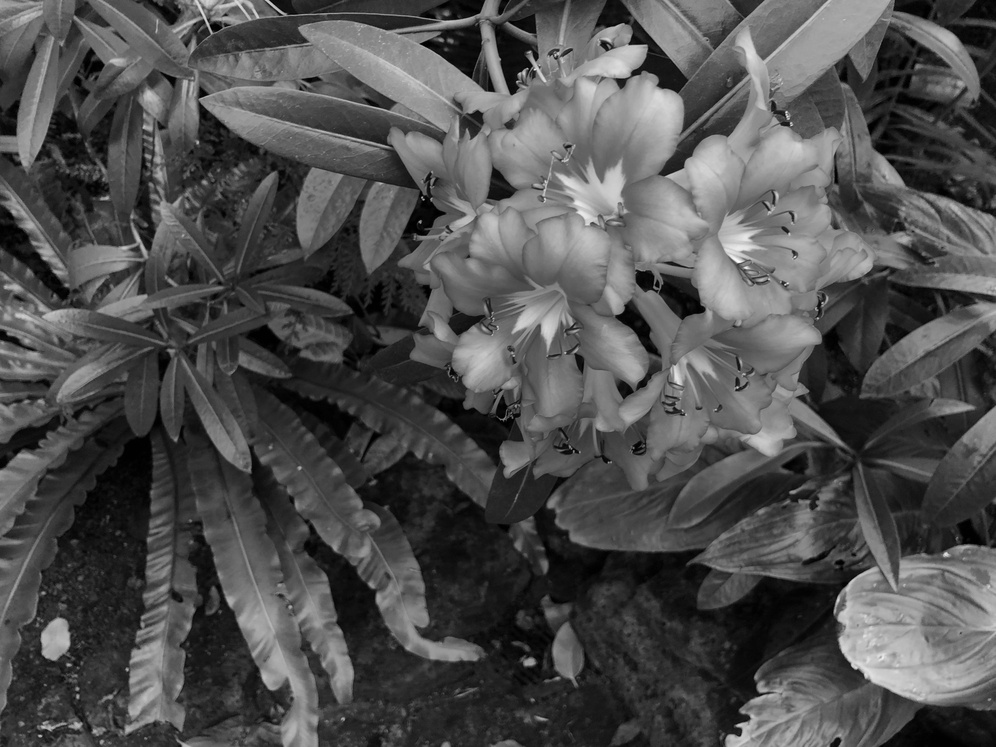} &
    \includegraphics[width=\imgw]{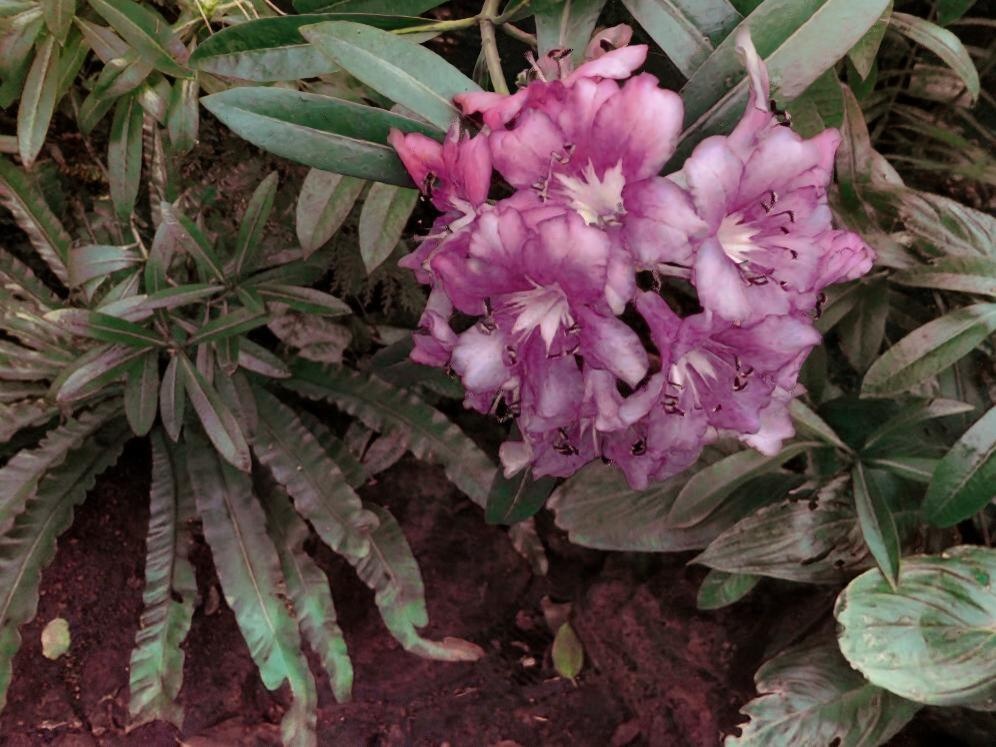} &
    \includegraphics[width=\imgw]{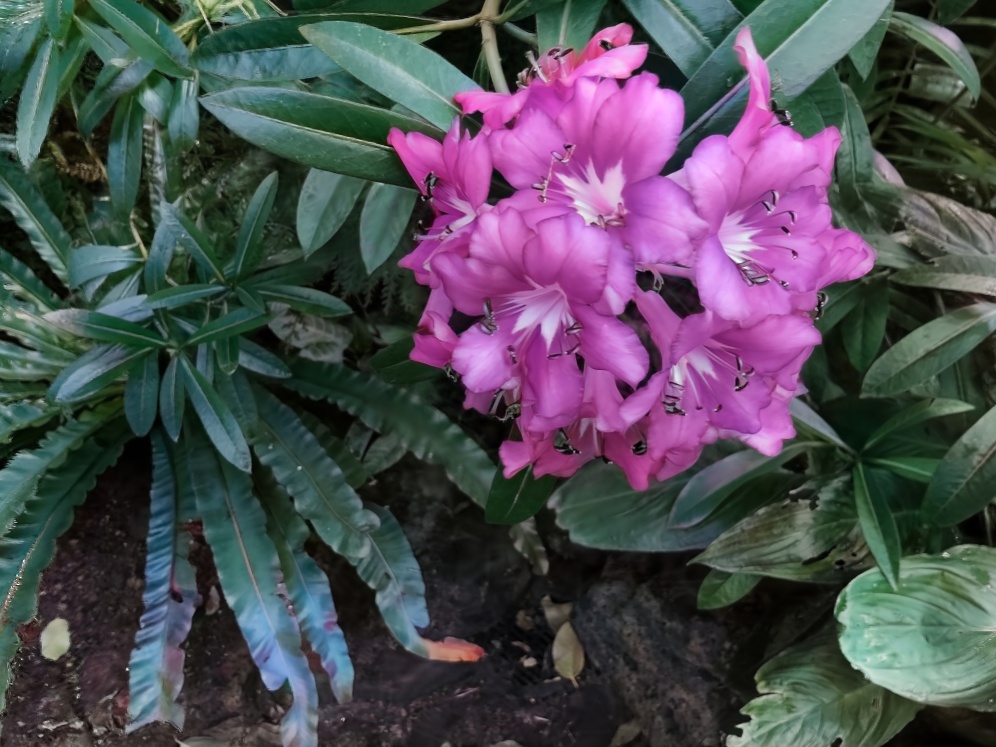} &
    \includegraphics[width=\imgw]{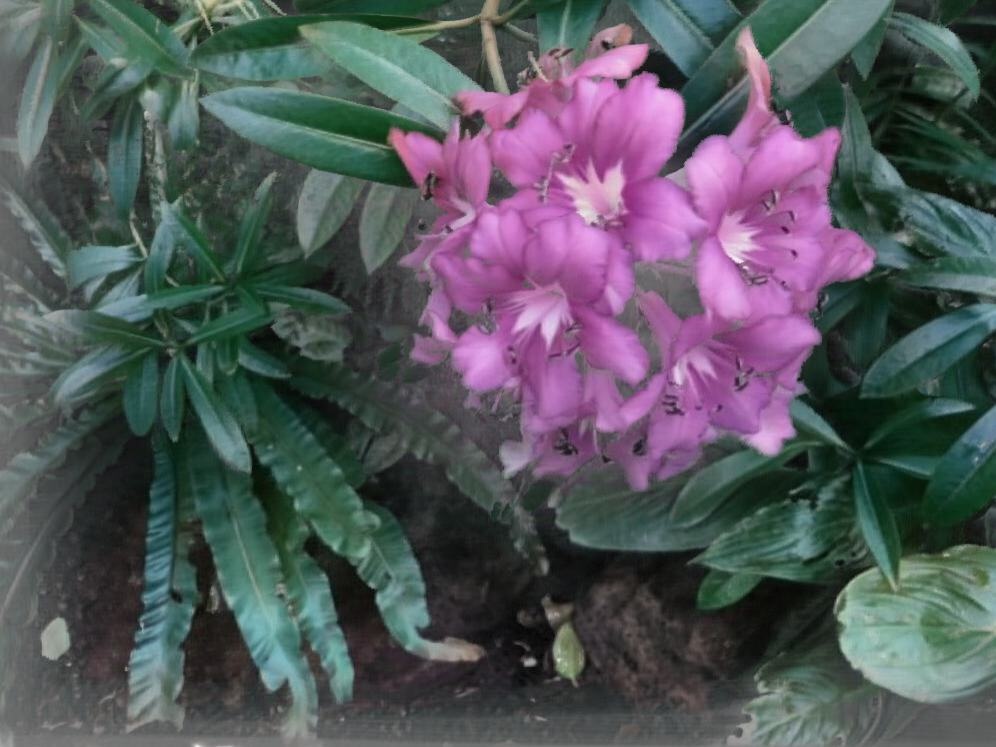} \\

    ChromaDistill-GS & ColorNeRF & ColorNeRF-GS & Ours \\
    \includegraphics[width=\imgw]{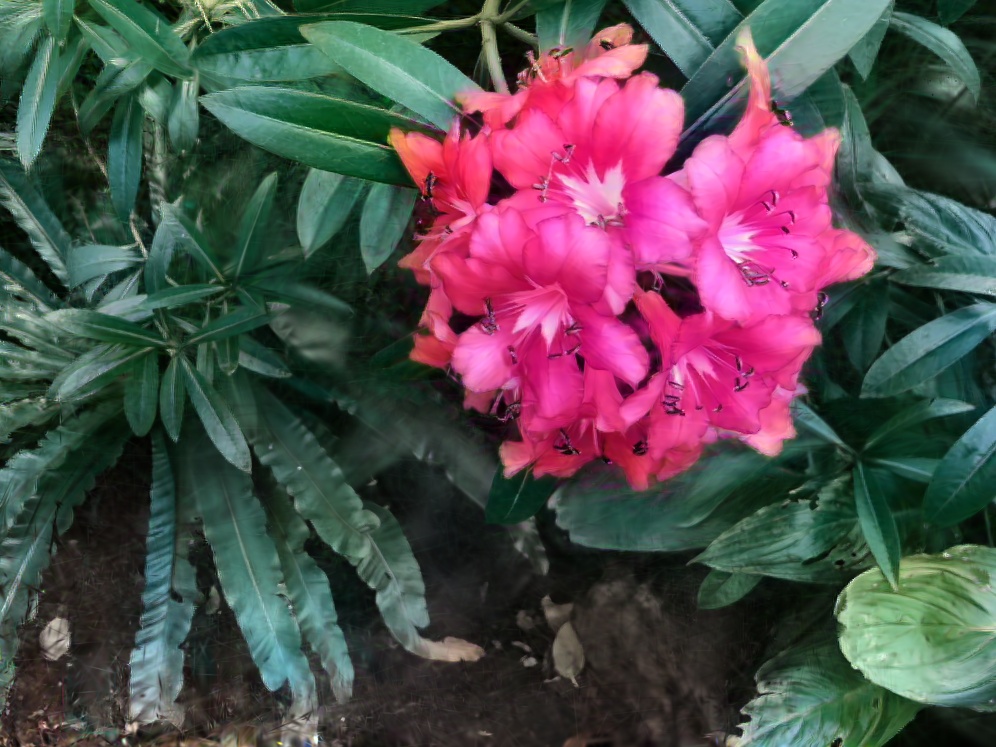} &
    \includegraphics[width=\imgw]{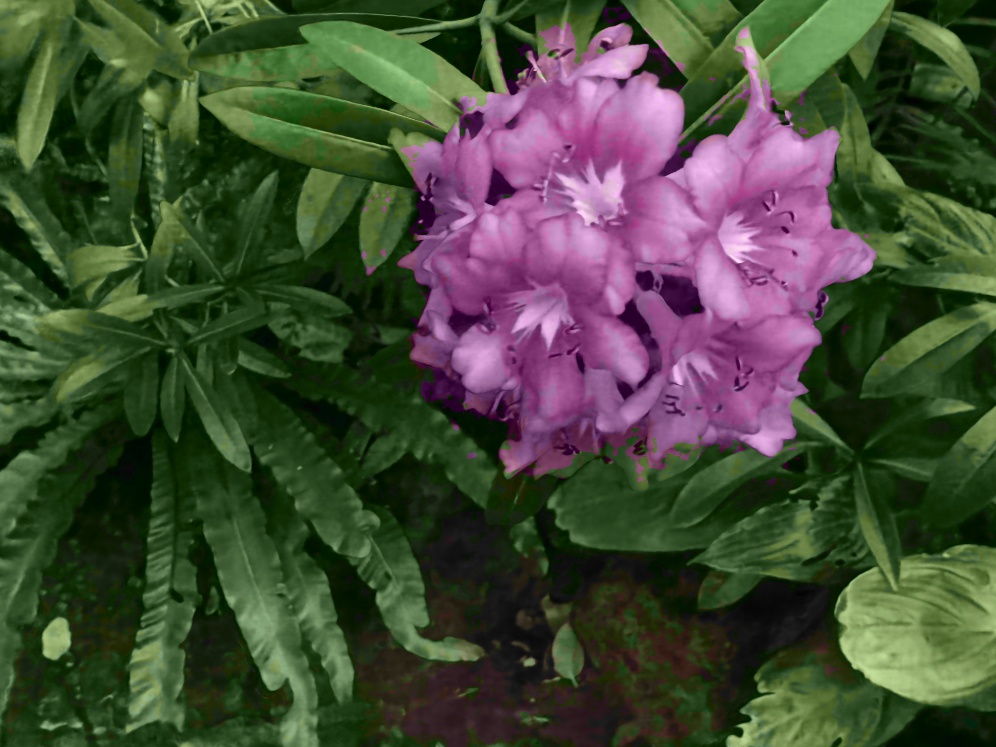} &
    \includegraphics[width=\imgw]{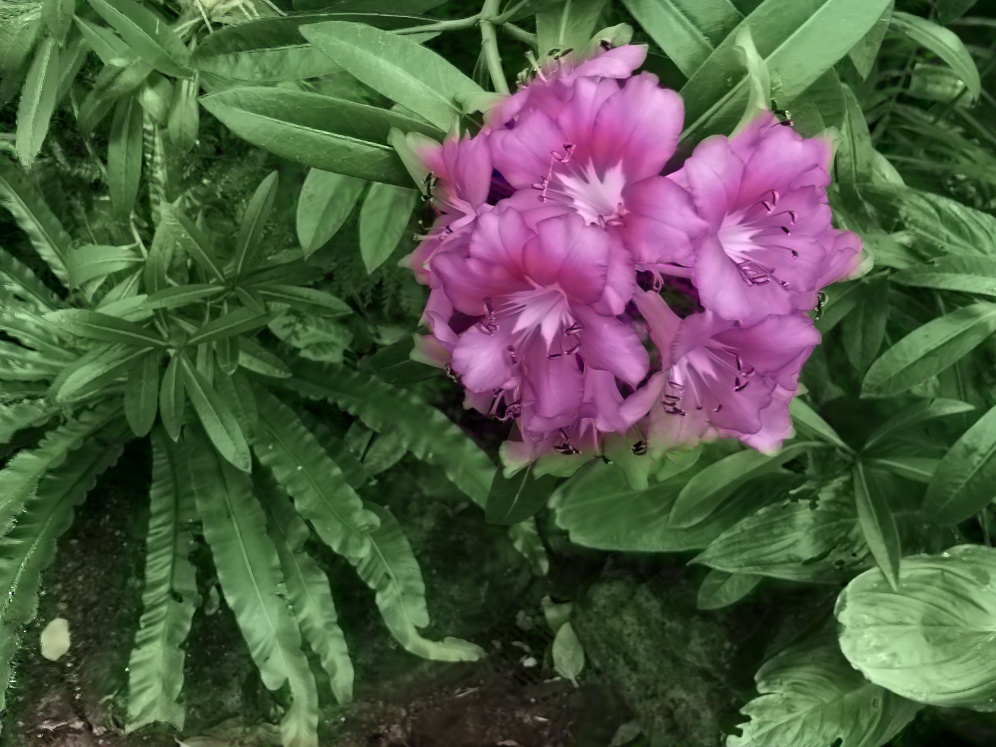} &
    \includegraphics[width=\imgw]{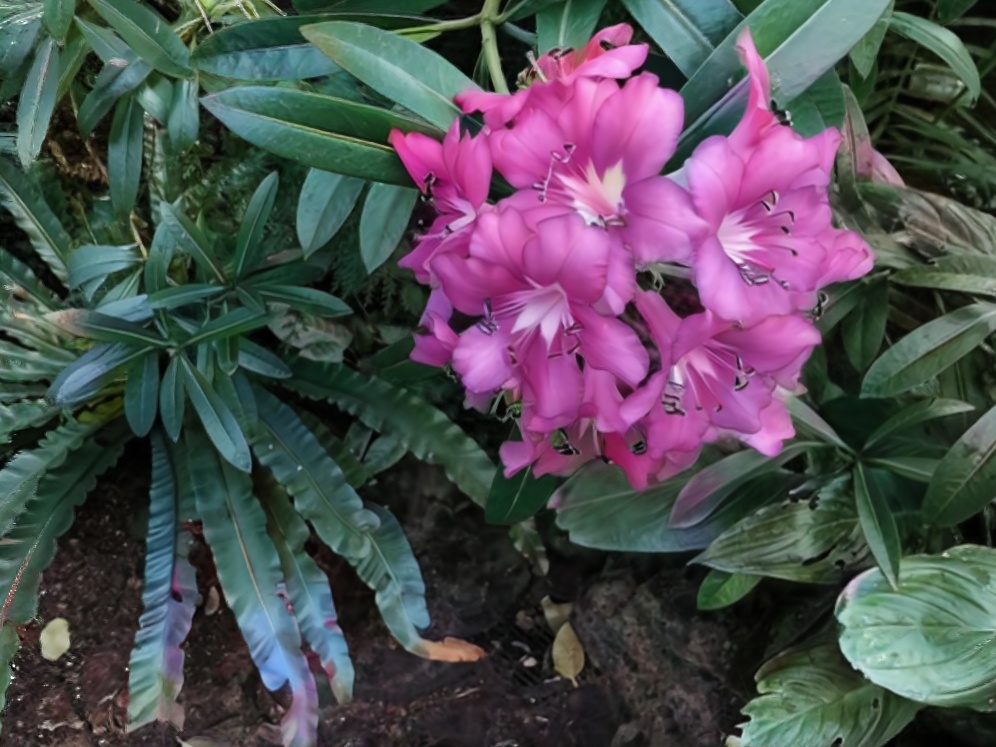} \\
\end{tabular}
}
\vspace{-5pt}
\captionof{figure}{\textbf{Qualitative results on LLFF}}
\label{fig:llff}
\end{minipage}
\end{figure}

\subsubsection{Datasets.}
We evaluate our method on:
\begin{itemize}
    \item LLFF~\cite{mildenhall2019llff}: Forward-facing dataset consisting of 24 scenes. We evaluate our method on the 8 scenes commonly adopted as benchmarks in prior works: \textit{fern}, \textit{flower}, \textit{fortress}, \textit{horns}, \textit{leaves}, \textit{orchids}, \textit{room}, and \textit{trex}.
    \item Mip-NeRF 360~\cite{mipnerf360}: 360-degree dataset consisting of 4 indoor and 3 outdoor scenes. We evaluate on the whole dataset.
    \item Tanks and Temples~\cite{knapitsch2017tanks}: 360-degree dataset consisting of 20 scenes. We evaluate on 4 scenes used in prior works: \textit{Horse}, \textit{M60}, \textit{Train} and \textit{Truck}.
    \item DL3DV-10K-Benchmark~\cite{ling2024d13dv}: A 360-degree dataset containing 140 scenes. For evaluation, we select a subset of 28 scenes, listed in supp. mat.
\end{itemize}

\subsubsection{Baselines.}
We compare our method with ColorNeRF~\cite{colornerf} and ChromaDistill~\cite{chromadistill}, which are primarily designed for 3D colorization.
However, as they struggle to reconstruct geometry in 360-degree scenes, we use our 3DGS-based implementations for robust reconstruction. 
We also evaluate ColorMNet~\cite{colormnet}, since video colorization inherently maintains cross-view consistency, and GenN2N~\cite{genn2n}, a 3D-to-3D translation framework capable of 3D colorization. 
Finally, we directly report results from the concurrent work Color3D~\cite{color3d}, as its code is unavailable.

\subsubsection{Evaluation metrics.}
To evaluate our 3D colorization, we assess color diversity, consistency, and plausibility.
\textbf{1) Colorfulness and normalized Colorfulness.} While Colorfulness~\cite{hasler2003measuring} is a standard metric in colorization tasks, it is insufficient to evaluate whether diverse objects in a complex 3D scene successfully recover their distinct colors.
Such limitations are especially evident in 3D colorization, where the averaging effect across viewpoints often induces a dominant color cast.
To address this, we additionally report \emph{normalized Colorfulness (nColorfulness)}, which calculates Colorfulness after removing the global scene tint. 
Please refer to the supp. mat. for a detailed formulation.
\textbf{2) Consistency.} To evaluate view consistency, we follow existing works~\cite{chromadistill, lai2018learning} and report both short- and long-term consistency metrics.
\textbf{3) Fréchet Inception Distance (FID).} We evaluate the visual plausibility of colorization results using FID score, following~\cite{color3d}.

\begin{figure}[tb]
    \centering
    \newcommand{\imgw}{0.2\textwidth}
    \setlength{\tabcolsep}{1pt}
    
    
    \begin{subfigure}{0.99\linewidth}
    \centering
        \begin{tabular}{@{}cccc@{}}
        \includegraphics[width=\imgw]{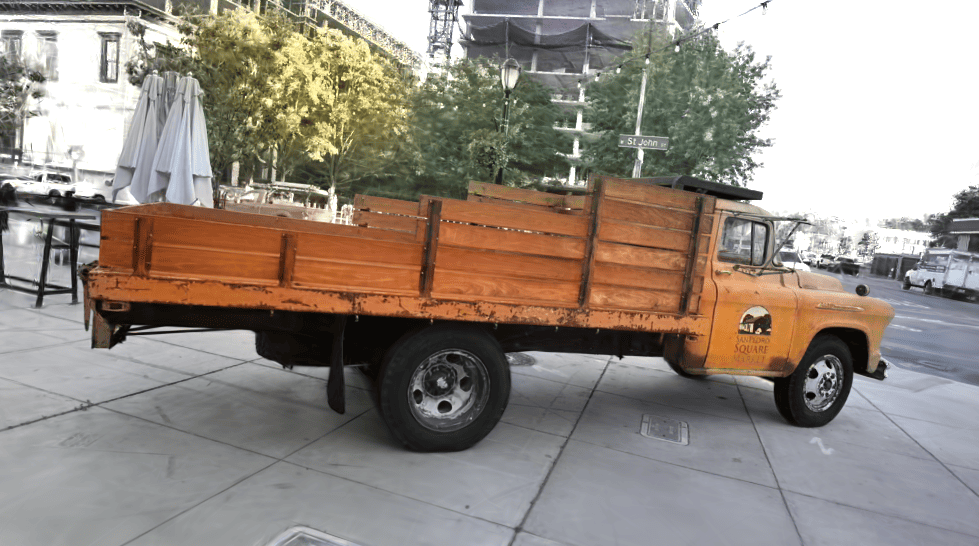} &
        \includegraphics[width=\imgw]{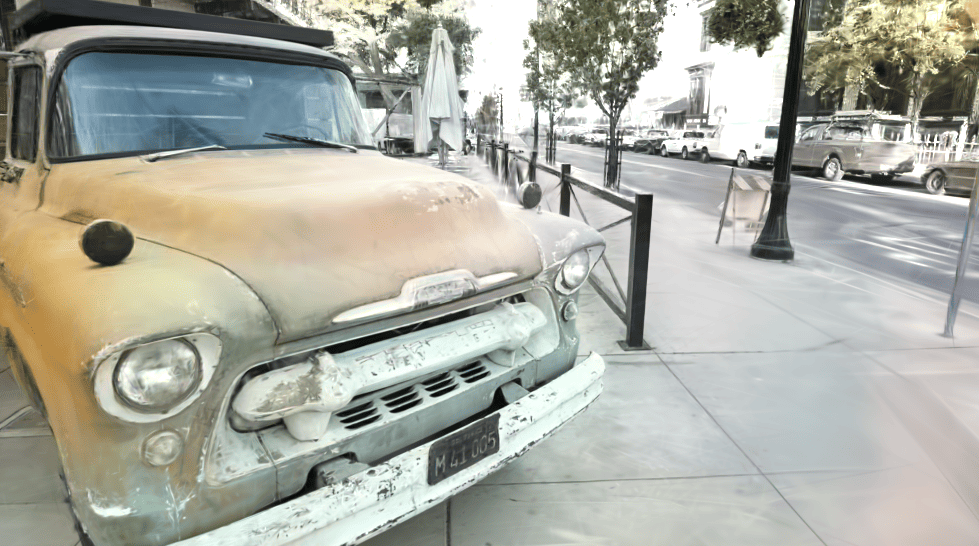} &
        \includegraphics[width=\imgw]{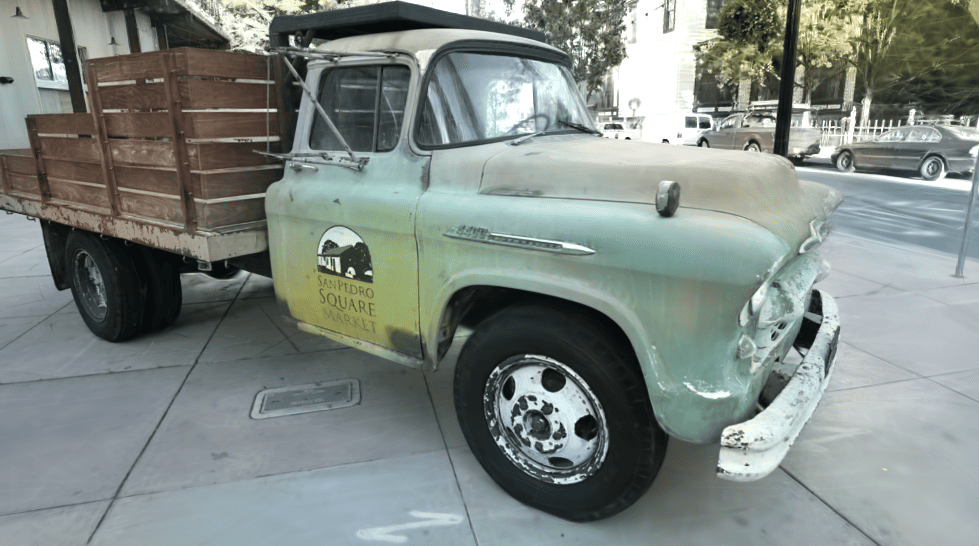} &
        \includegraphics[width=\imgw]{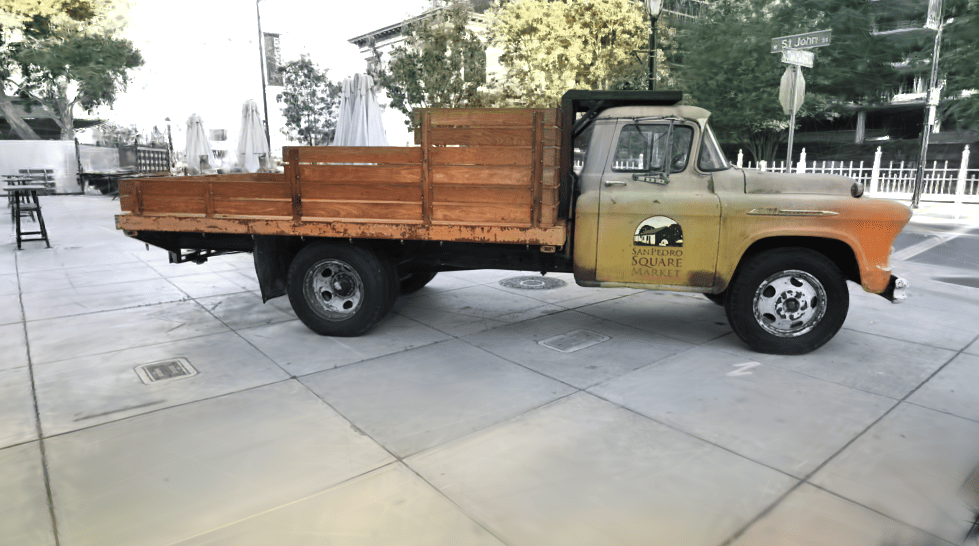} \\
        \end{tabular}
    \vspace{-2pt}
    \caption{ChromaDistill-GS}
    \vspace{2pt}
    \end{subfigure}
    
    \begin{subfigure}{0.99\linewidth}
    \centering
        \begin{tabular}{@{}cccc@{}}
        \includegraphics[width=\imgw]{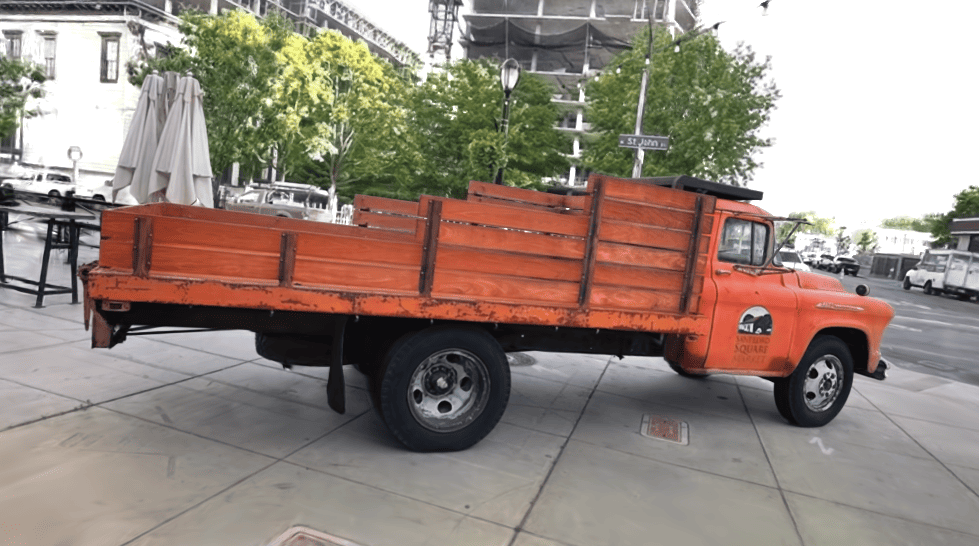} &
        \includegraphics[width=\imgw]{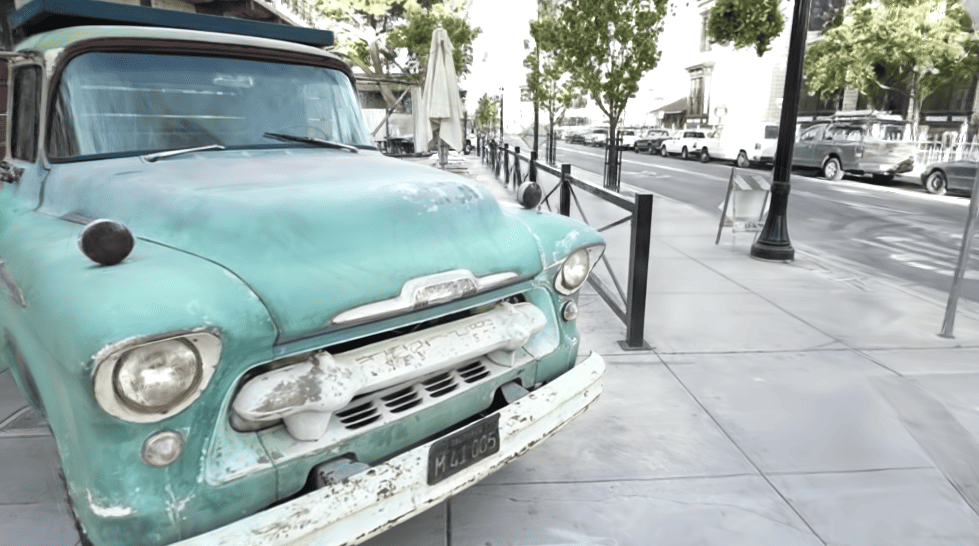} &
        \includegraphics[width=\imgw]{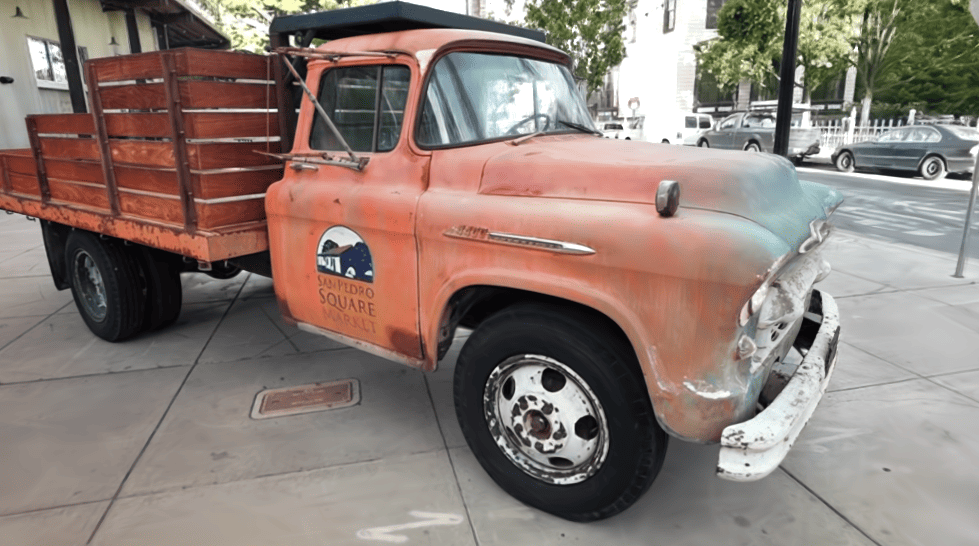} &
        \includegraphics[width=\imgw]{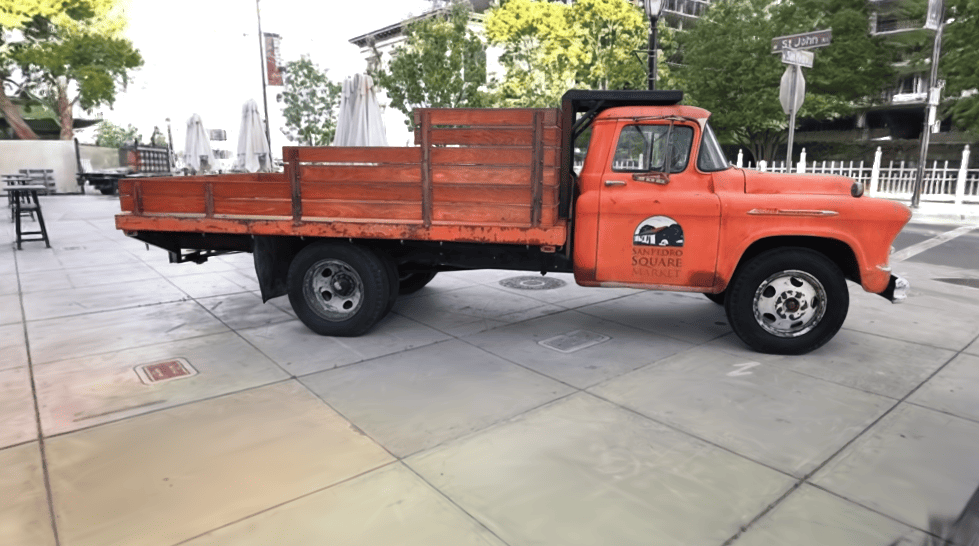} \\
        \end{tabular}
    \vspace{-2pt}
    \caption{w/o Global Calibration}
    \vspace{2pt}
    \end{subfigure}
    
    \begin{subfigure}{0.99\linewidth}
    \centering
        \begin{tabular}{@{}cccc@{}}
        \includegraphics[width=\imgw]{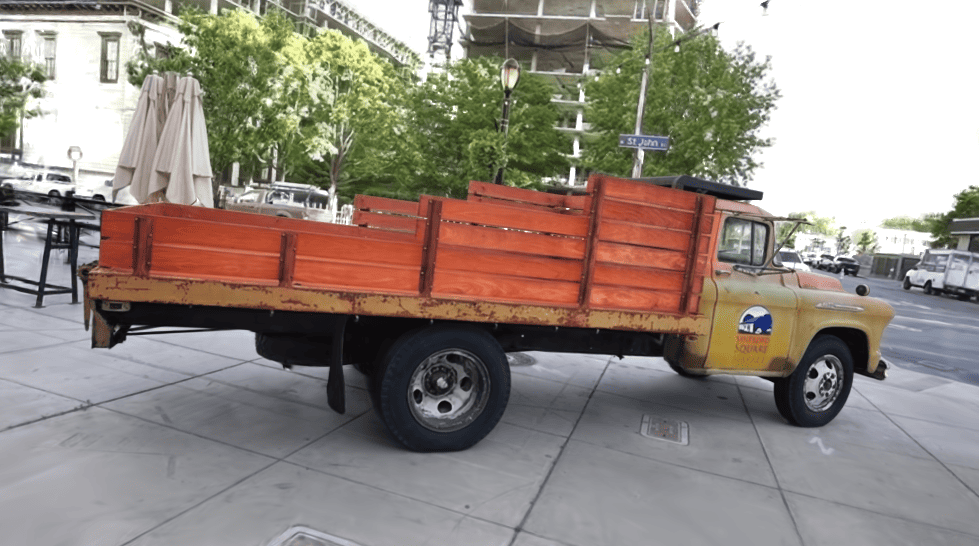} &
        \includegraphics[width=\imgw]{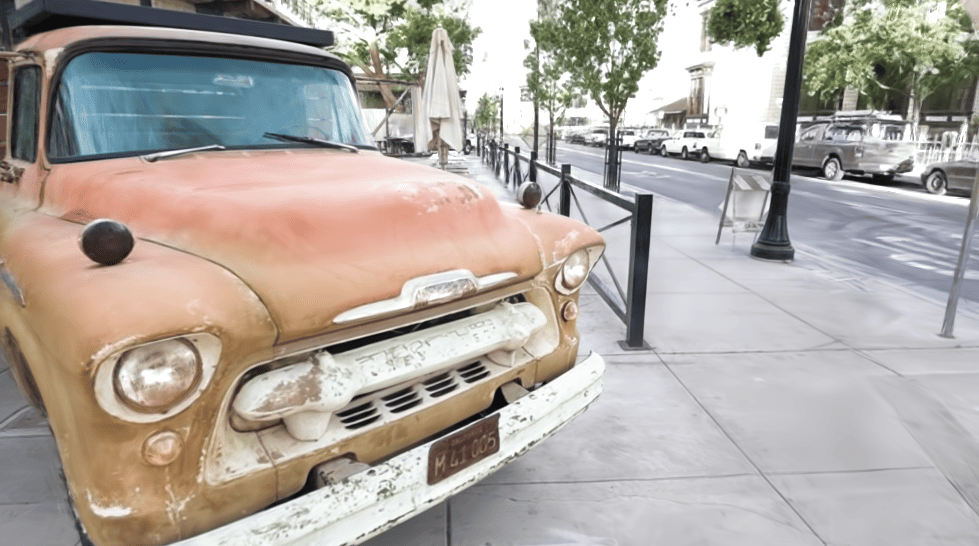} &
        \includegraphics[width=\imgw]{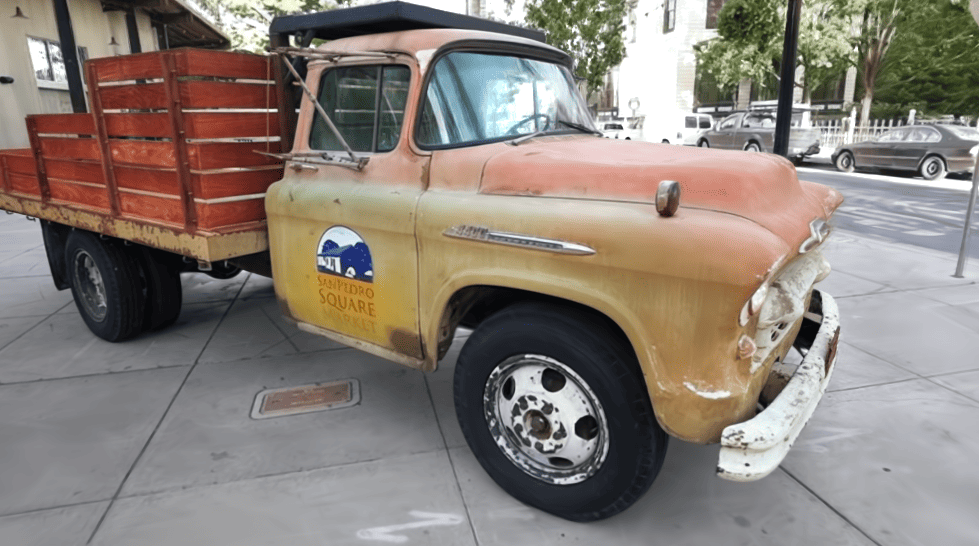} &
        \includegraphics[width=\imgw]{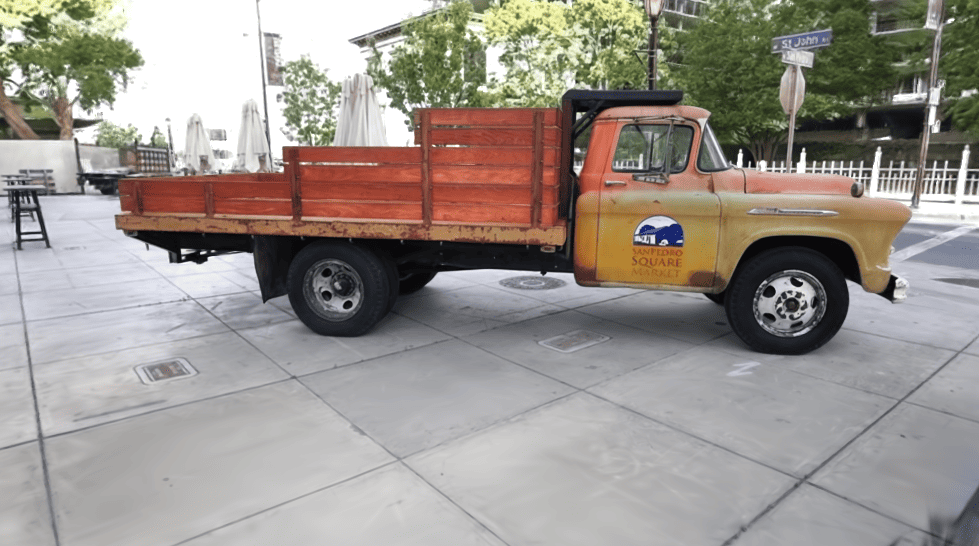} \\
        \end{tabular}
    \vspace{-2pt}
    \caption{Our Method}
    \vspace{2pt}
    \end{subfigure}

    \caption{
    \textbf{Consistency comparison and ablation --} Our global calibration step is essential for mitigating the continuous color shift observed in uncalibrated views, ensuring global consistency. Furthermore, compared to ChromaDistill-GS, our global calibration produces significantly more consistent results, highlighting its effectiveness in maintaining consistent color across views.
    }%

    \label{fig:consistency}
\end{figure}

\subsection{Results}

\subsubsection{Qualitative results.}
\cref{fig:llff} illustrates the results on a forward-facing dataset. 
While most methods perform reasonably well in this setup, ColorNeRF~\cite{colornerf}, which is highly susceptible to the averaging effect, fails to distinguish individual leaf colors. 
However, these baselines encounter significant challenges when applied to 360-degree scenes.
\cref{fig:qualitative} presents qualitative comparisons on 360-degree datasets. 
Overall, our approach successfully colorizes complex scenes with plausible and diverse colors, effectively capturing even the smallest objects. 
In contrast, while the 3D colorization baselines ColorNeRF~\cite{colornerf} and ChromaDistill~\cite{chromadistill} may initially capture colors for small details from 2D image colorizers, the inconsistencies across multiple views often cause these fine-grained features to be averaged out during the 3D optimization process.
For instance, in `DL3DV-Scene1', baselines paint small leaves with the same color as their surroundings, whereas our method correctly recovers their distinct green hue. 
This capability to preserve fine-grained color details is also evident in the blue label within the `Bonsai' scene, the berries in the `Garden' scene, and the street lamp and sign of the `Truck' scene.
Furthermore, while ColorMNet~\cite{colormnet}, a video colorization model, ensures view consistency, its reliance on a single reference view provides insufficient color information for the entire 3D scene, resulting in a lack of color diversity.
Additionally, as GenN2N~\cite{genn2n} is a general-purpose 3D translation framework not specifically tailored for colorization, it tends to introduce subtle chromatic noise and artifacts during the translation process. 
In contrast, our method achieves both superior color diversity and robust 3D consistency.

\begin{table}[tb]
    \centering
    \caption{\textbf{Quantitative ablation on Mip-NeRF 360 dataset --} Our full method achieves the best overall performance across all metrics.}
    \vspace{-5pt}
    \label{tab:ablation}
    \resizebox{0.6\linewidth}{!}{%
        \begin{tabular}{l ccccc}
        \toprule
        Method
        & FID$\downarrow$ & Color$\uparrow$ & nColor$\uparrow$ & SC$\downarrow$ & LC$\downarrow$ \\ 
        \midrule
        w/o $\Phi_\text{MV}$ & 40.99 & \first{37.65} & 22.91 & 0.014 & 0.026 \\
        w/o Flickr8k  & \first{36.44} & 35.47 & \second{22.97} & 0.013 & \third{0.022} \\
        w/o global calibration & \third{39.34} & \third{36.56} & \third{22.95} & \third{0.012} & 0.027 \\
        w/o multi-view referencing & 39.92 & 35.07 & 21.47 & \first{0.011} & \second{0.021} \\
        \midrule
        \textbf{Ours} 
        & \second{38.42} & \second{37.32} & \first{23.50} & \first{0.011} & \first{0.020} \\
        \bottomrule
        \end{tabular}
    }
\end{table}

\begin{figure}[tb]
    \centering
    \newcommand{\imgw}{0.35\textwidth}
    \setlength{\tabcolsep}{1pt}
    \small

    \begin{subfigure}[c]{0.46\textwidth}
    \resizebox{0.99\linewidth}{!}{
    \centering
    \begin{tabular}{cccc}
    & {$K=1$} & {$K=4$} & {$K=10$} \\

    \raisebox{5pt}{\rotatebox{90}{View 1}} &
    \includegraphics[width=\imgw]{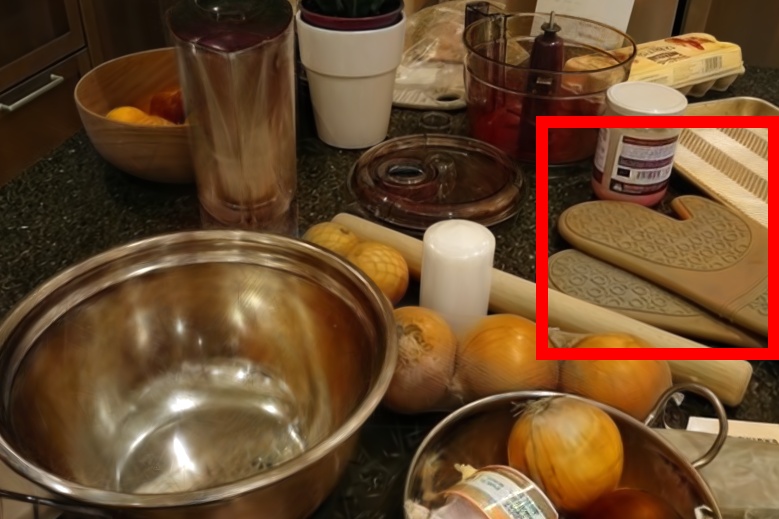} &
    \includegraphics[width=\imgw]{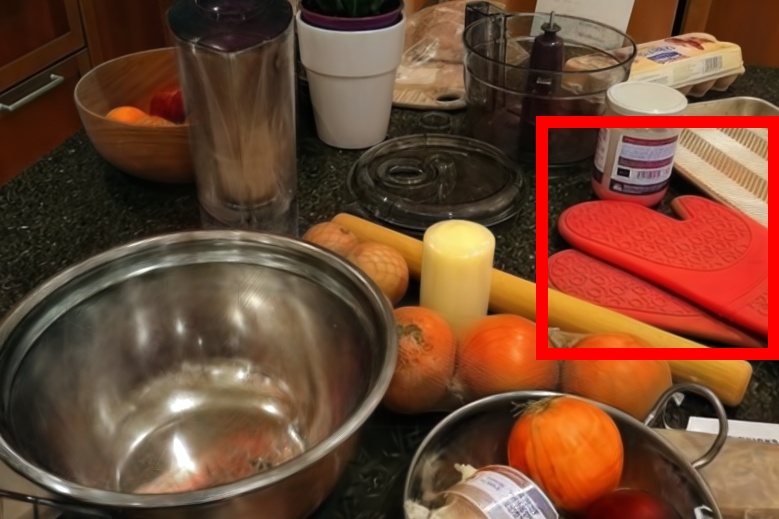} &
    \includegraphics[width=\imgw]{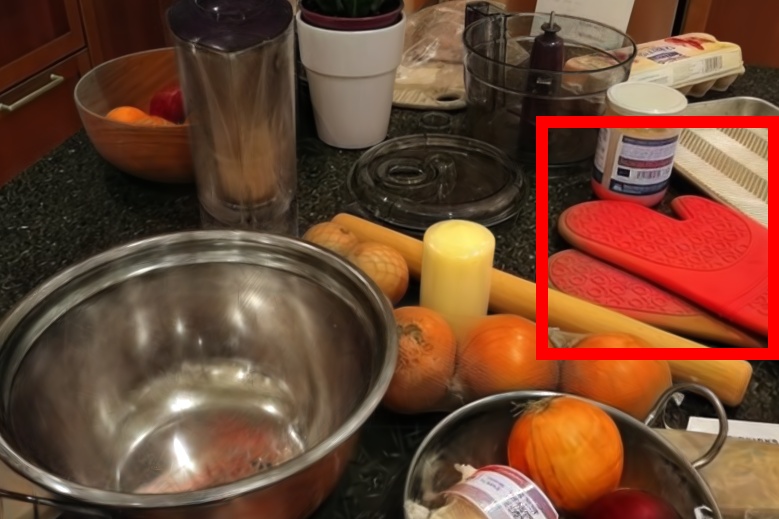} \\
    
    \raisebox{4pt}{\rotatebox{90}{View 2}} &
    \includegraphics[width=\imgw]{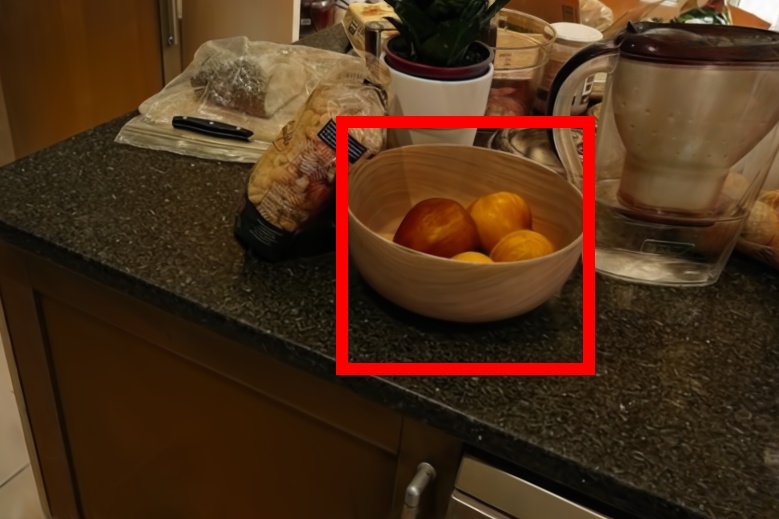} &
    \includegraphics[width=\imgw]{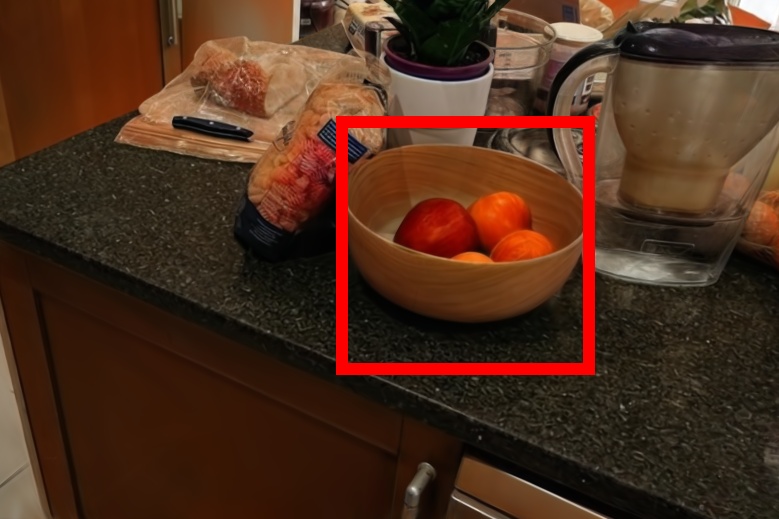} &
    \includegraphics[width=\imgw]{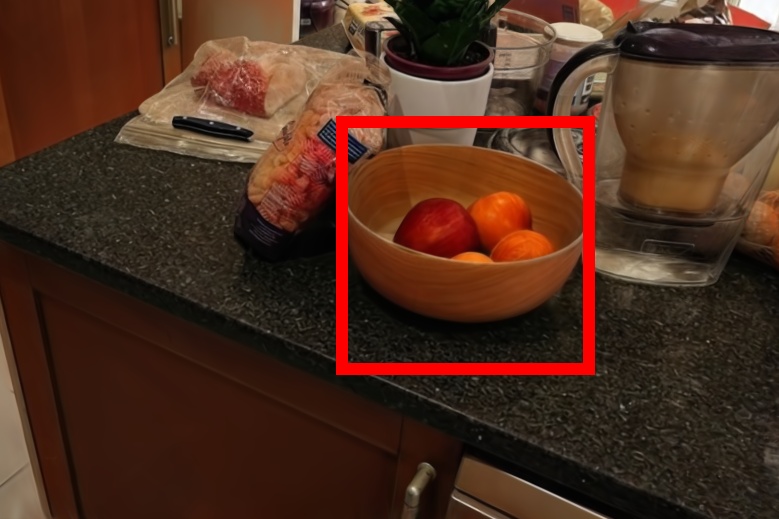} \\

    \end{tabular}
    }
    \end{subfigure}
    \hfill
    \begin{subfigure}[c]{0.53\textwidth}
    \centering
    \includegraphics[width=0.9\textwidth]{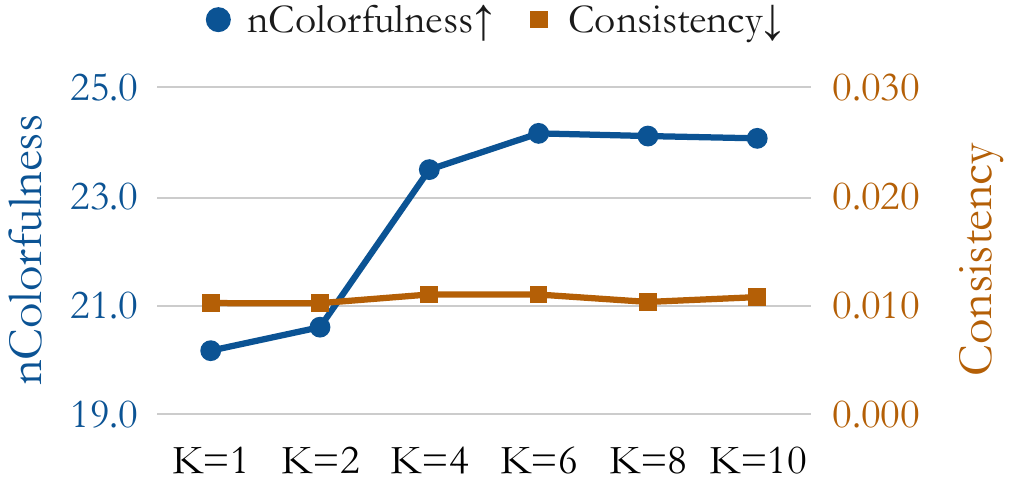}
    \end{subfigure}

    \caption{
    \textbf{Ablation on the number of base views --} Thanks to our global calibration, consistency remains stable regardless of $K$. In addition, available color information expands as $K$ increases, leading to higher nColorfulness. Notably, the color diversity gain is most significant up to $K=4$ and diminishes thereafter. To balance this plateauing performance with the self-attention cost, we select $K=4$ as the optimal value.
    }
    \label{fig:ablation}
\end{figure}

\subsubsection{Quantitative results.}
We provide the quantitative results in \cref{tab:360} and \cref{tab:llff}.
Overall, our method demonstrates superior performance across all five metrics, with particularly significant gains on 360-degree datasets.
Notably, we observe a clear trade-off between consistency and colorfulness among the baseline methods.
For instance, while ColorMNet achieves the highest consistency, its low Colorfulness on 360-degree scenes suggests that such stability stems from monotonous colorization, due to the lack of color information in complex scenes.
In contrast, our method successfully achieves high color diversity while maintaining competitive consistency.
Note that the consistency metric for Color3D is the average of its short- and long-term consistency values.

\subsection{Ablation studies}
In \cref{tab:ablation}, we evaluate the contribution of each component in our framework. 

\noindent\textbf{Multi-view colorization model ($\Phi_\text{mv}$).} We evaluate the impact of our multi-view diffusion model by ablating $\Phi_\text{mv}$. 
When we ablate $\Phi_\text{mv}$ by independently colorizing all training views using a 2D image colorization model, the result suffers from multi-view inconsistency due to inconsistent color guidance.

\noindent\textbf{Multi-view referencing.} Removing multi-view referencing during Local Color Propagation deprives the model of sufficient reference colors to condition on, which significantly degrades the overall colorization quality.

\noindent\textbf{Training data.} While the DL3DV-10K-Benchmark is sufficient for learning multi-view referencing, it lacks the color distribution of various real-world images. 
We thus supplement our training with the Flickr8k dataset to provide rich color diversity. Our ablation study confirms that training on DL3DV alone results in a limited color distribution (see supp. mat. for qualitative results).

\noindent\textbf{Global consistency calibration.} In \cref{fig:consistency}, we ablate global calibration and compare against ChromaDistill-GS~\cite{chromadistill}. Without global calibration, the object exhibits continuous color shifts across the 360-degree scene, leading to unnatural results. In contrast, our method maintains consistent colorization, outperforming both the ablated version and the baseline.

\noindent\textbf{The number of base views.}
In \cref{fig:ablation}, global calibration maintains stable consistency across all $K$. While increasing $K$ improves nColorfulness, the gains diminish after $K=4$. Thus, we set $K=4$ to balance color diversity against self-attention costs.
 
\subsection{Application}
\begin{figure}[tb]
    \centering
    \newcommand{\imgw}{0.2\textwidth}
    \setlength{\tabcolsep}{2pt}
    
    \resizebox{0.99\linewidth}{!}{
    \centering
    \begin{tabular}{cccccc}
    
    \multicolumn{3}{c}{Input}
    & \multicolumn{3}{c}{Colorized} \\

    \cmidrule(lr){1-3}
    \cmidrule(lr){4-6}
    
    \includegraphics[width=\imgw]{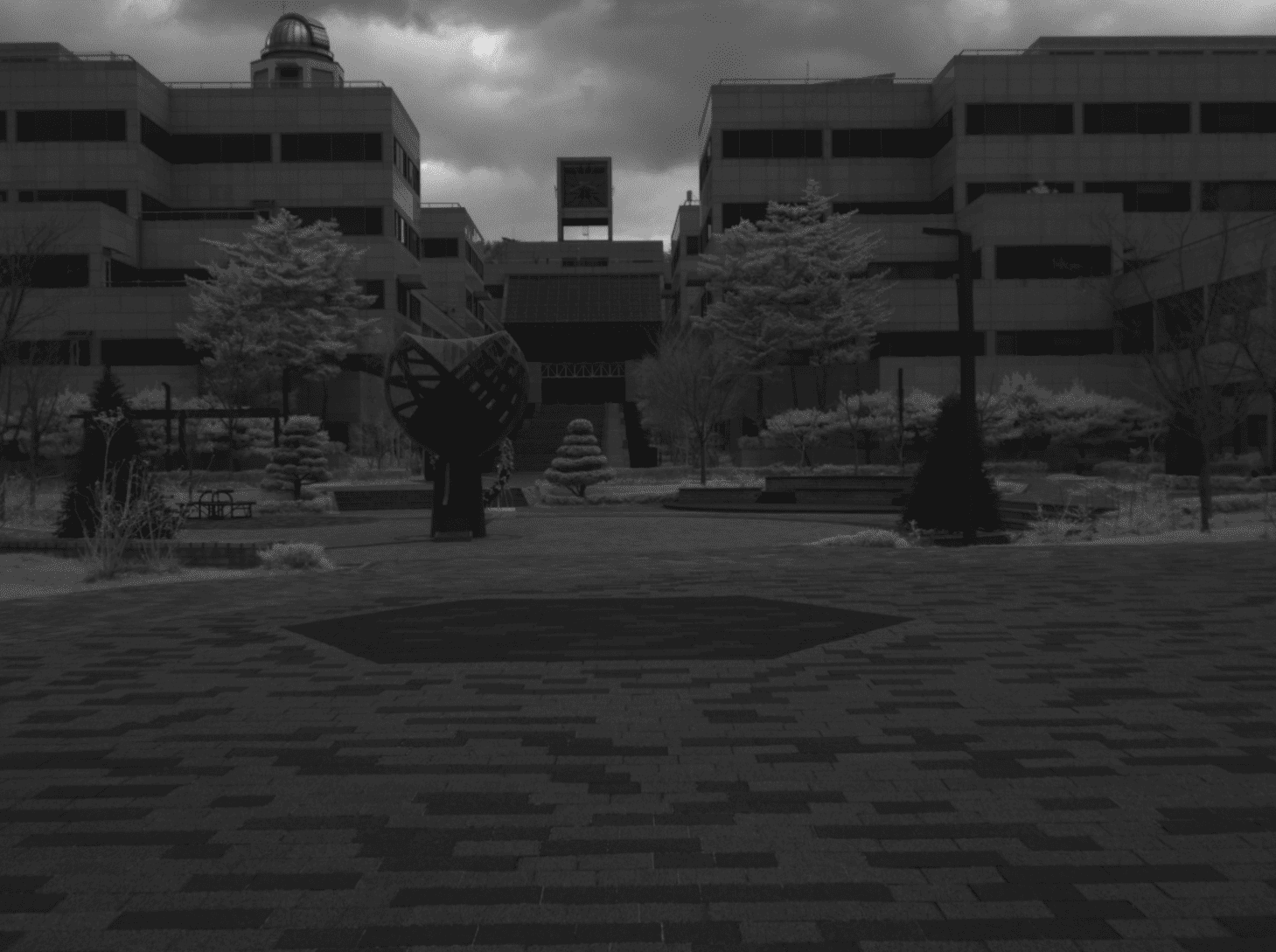} &
    \includegraphics[width=\imgw]{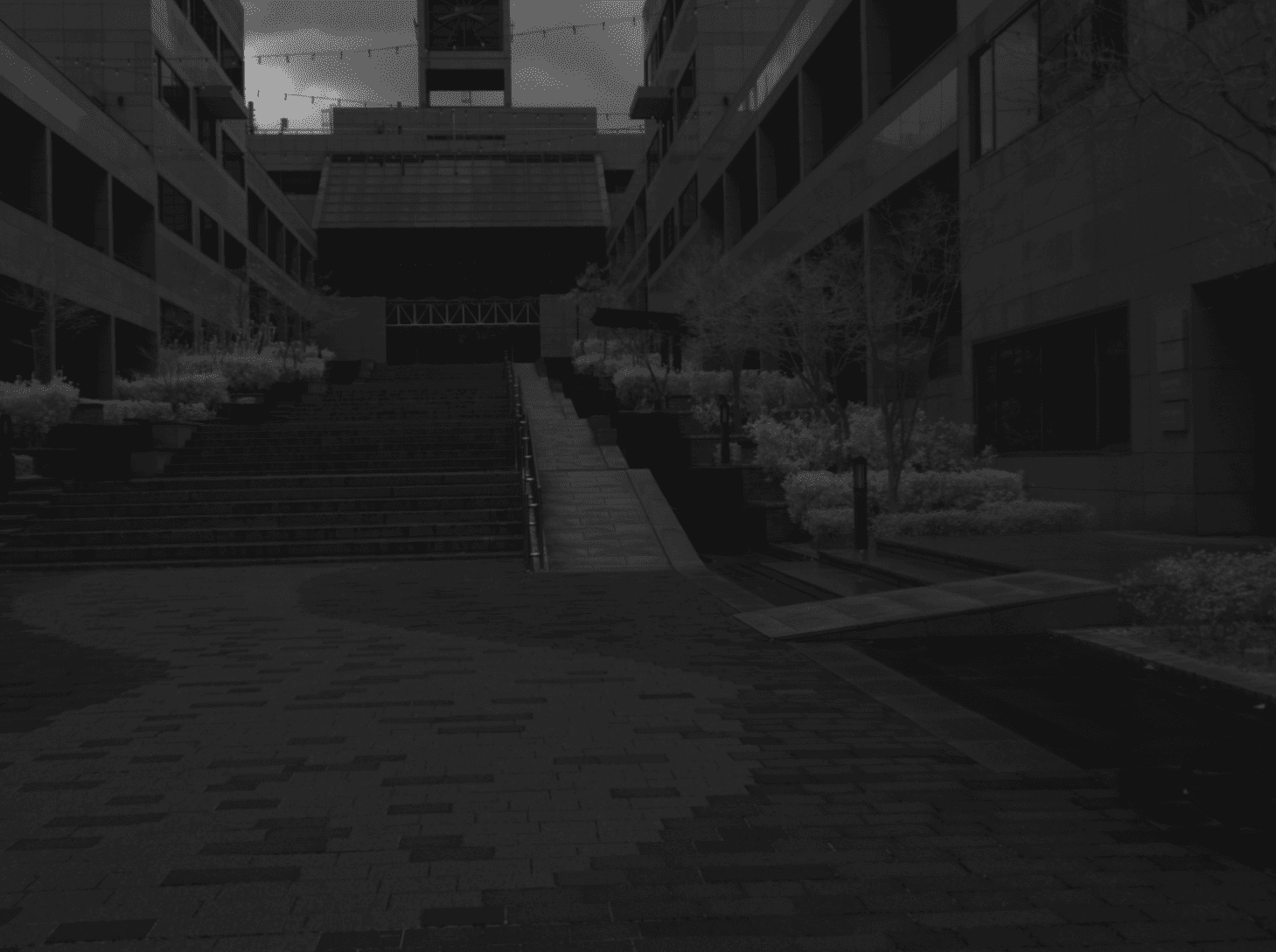} &
    \includegraphics[width=\imgw]{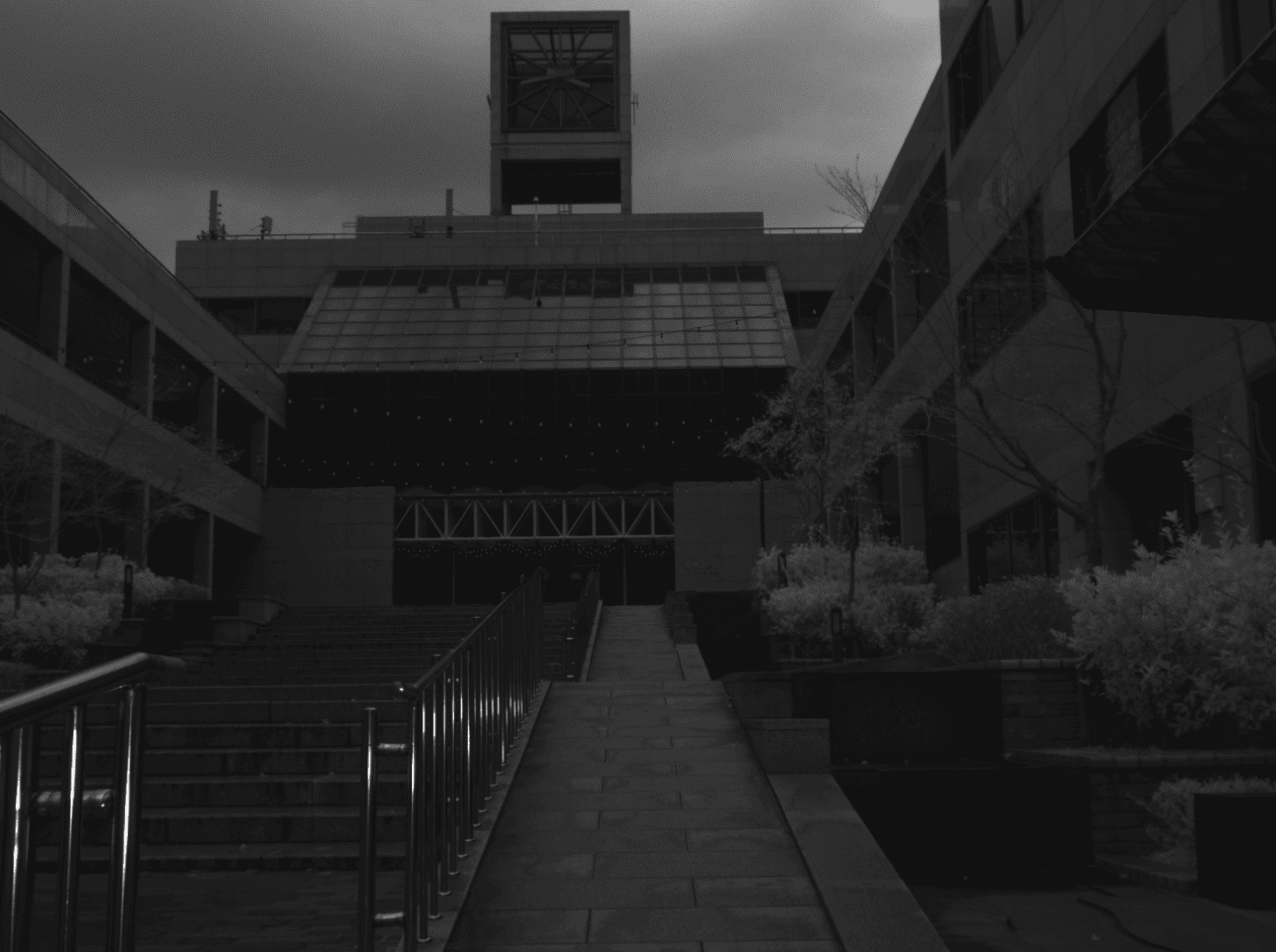} &
    \includegraphics[width=\imgw]{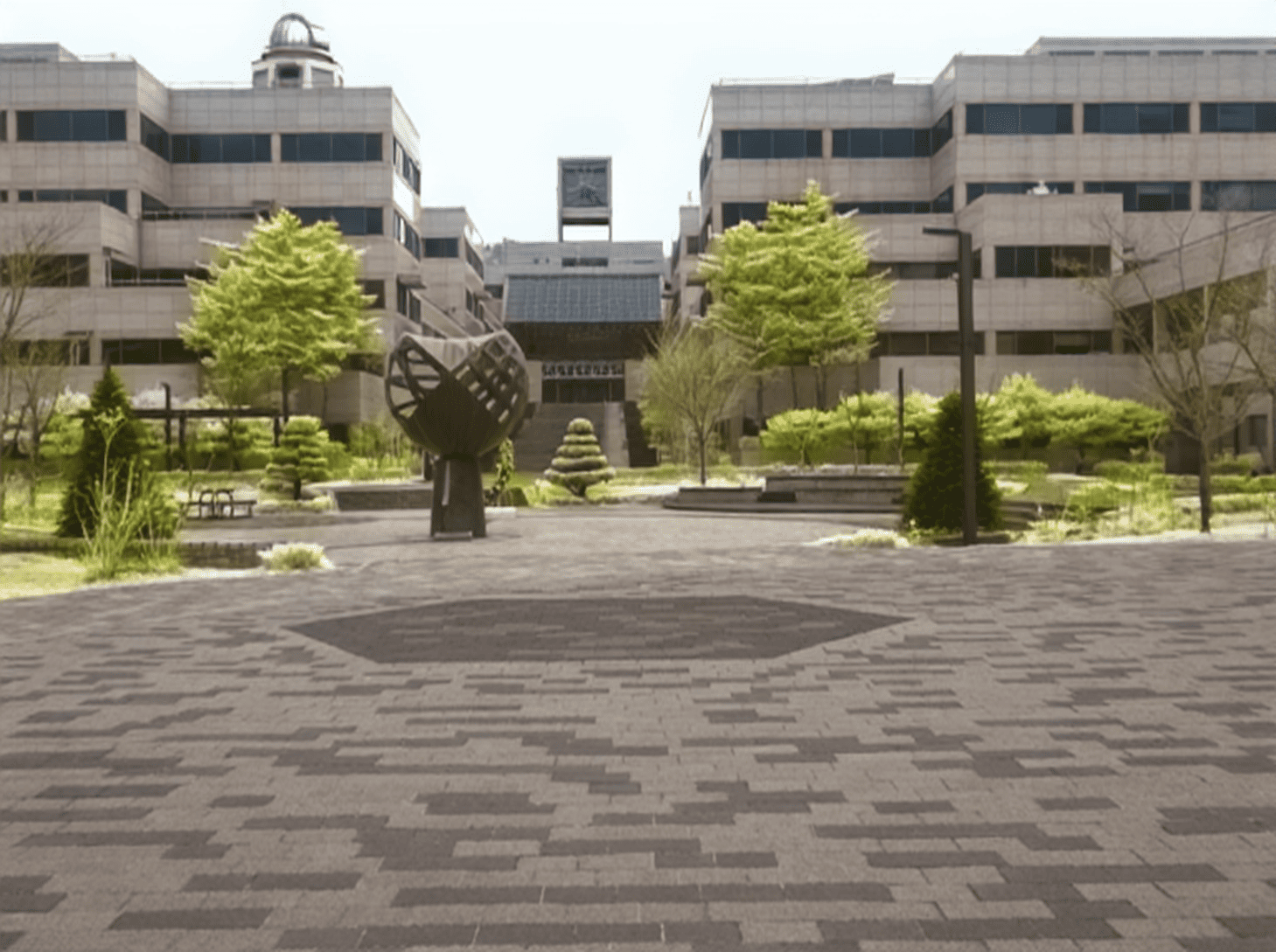} &
    \includegraphics[width=\imgw]{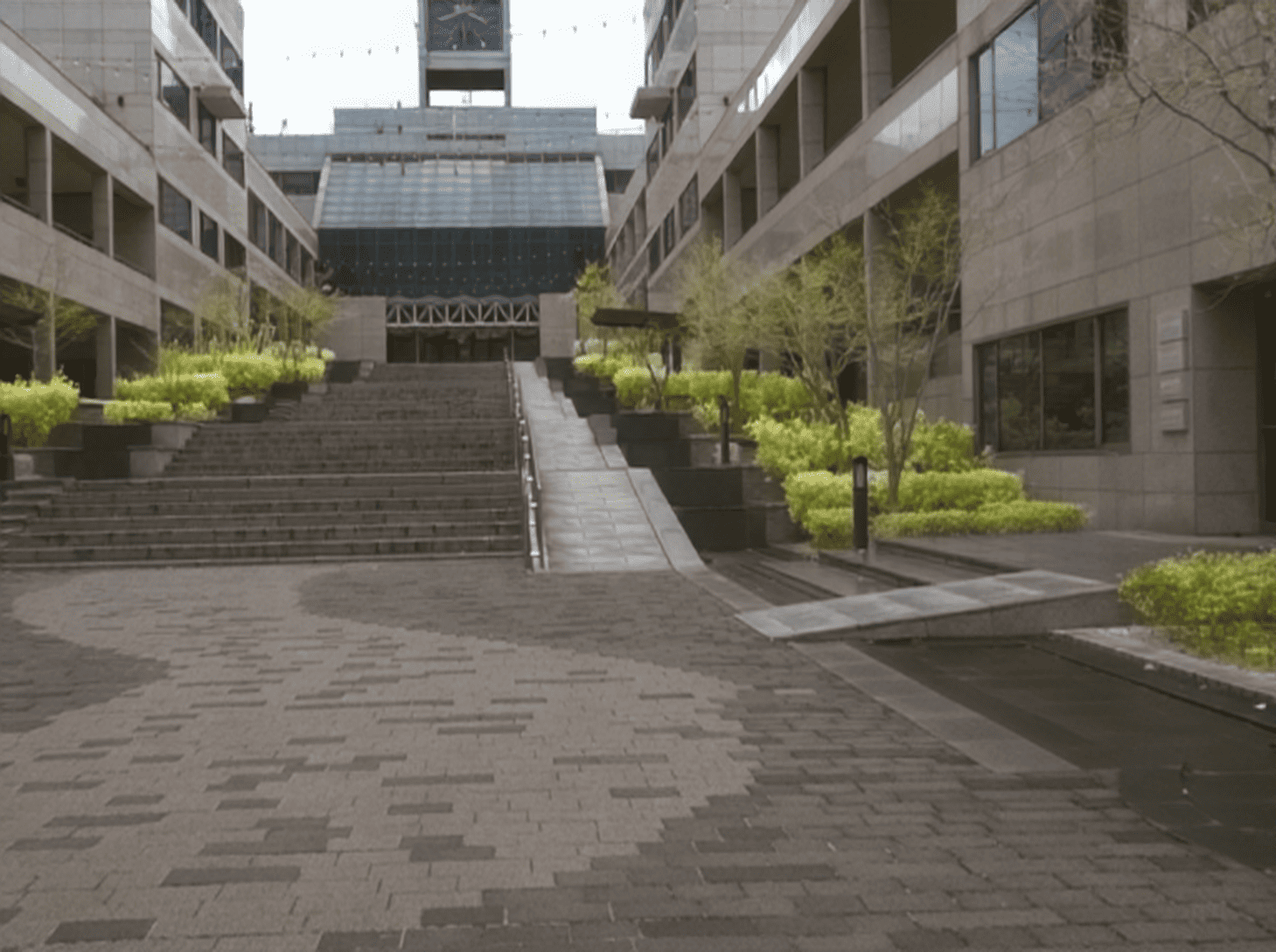} &
    \includegraphics[width=\imgw]{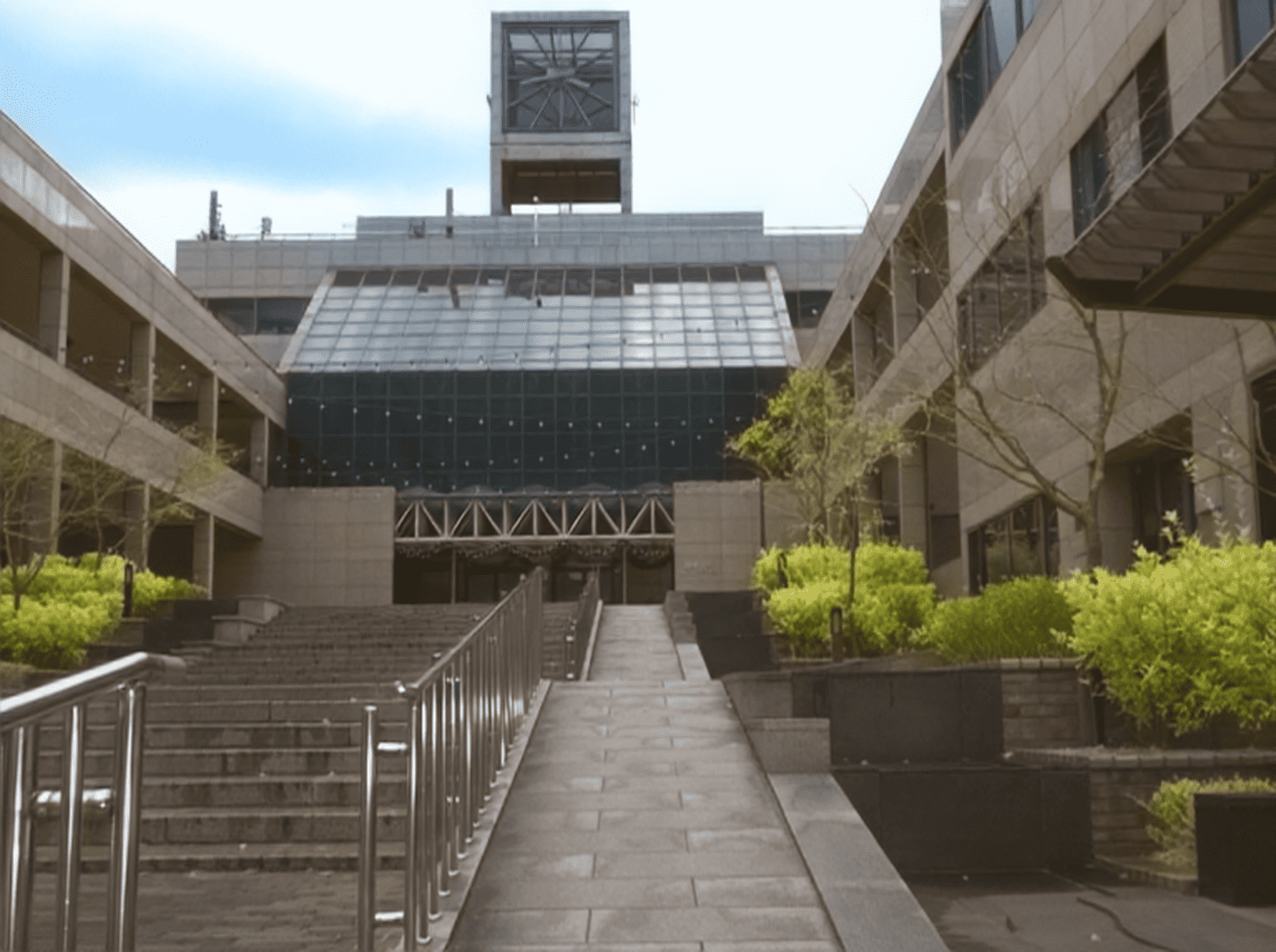} \\

    \end{tabular}
    }
    \vspace{-5pt}
    \caption{
    \textbf{Result on NIR images --} Our method extends to NIR multi-view images, achieving consistent and realistic colorization beyond grayscale images.
    }

    \label{fig:application}
\end{figure}

To demonstrate that our method is applicable not only to grayscale images but also to general single-channel images, we present results of applying our method to NIR multi-view images from the Pixel-aligned RGB-NIR Stereo dataset~\cite{kim2025pixelnir}.
In this case, because the image colorization model cannot colorize NIR images due to the domain gap in the input images, we replace the image colorization model with NanoBanana2~\cite{google2026nanobanana2} to translate the base view NIR images into colorized visible images. 
As shown in \cref{fig:application}, our method successfully achieves plausible and consistent colorization from multi-view single-channel images.
\section{Conclusion}

In this work, we present \ourmethod, a novel approach for colorizing single-channel 3D reconstructions.
We identify that existing methods rely on guidance-averaging to achieve multi-view color consistency, which fails to capture the color diversity of complex 360-degree scenes, leading to monotonous results.
Our method minimizes this averaging process. 
However, this leads to a consistency challenge, so we propose a Local-Global pipeline that partitions the scene and uses a fine-tuned multi-view diffusion model to manage both inter-subscene and intra-subscene consistency. 
Our experiments demonstrate that \ourmethod produces 3D models that are simultaneously consistent and color-diverse, enhancing their versatility for downstream applications like VR/AR.


\bibliographystyle{splncs04}
\bibliography{main}

@String(CVPR  = {IEEE Conf. Comput. Vis. Pattern Recog.})

@String(ICCV  = {Int. Conf. Comput. Vis.})

@String(ECCV  = {Eur. Conf. Comput. Vis.})

@String(NeurIPS = {Adv. Neural Inform. Process. Syst.})

@String(ICML  = {Int. Conf. Mach. Learn.})

@String(ICLR  = {Int. Conf. Learn. Represent.})

@String(CVPRW = {IEEE Conf. Comput. Vis. Pattern Recog. Worksh.})

@String(AAAI  = {AAAI})

@String(ICIP  = {IEEE Int. Conf. Image Process.})

@String(TOG   = {ACM Trans. Graph.})

@String(ACMMM = {ACM Int. Conf. Multimedia})

@String(ARXIV = {arXiv preprint})

@String(WACV = {IEEE/CVF Winter Conference on Applications of Computer Vision})

@String(CVPR  = {CVPR})

@String(ICCV  = {ICCV})

@String(ECCV  = {ECCV})

@String(NeurIPS = {NeurIPS})

@String(ICML  = {ICML})

@String(ICLR  = {ICLR})

@String(CVPRW = {CVPRW})

@String(ICIP  = {ICIP})

@String(TOG   = {ACM TOG})

@String(ACMMM = {ACM MM})

@inproceedings{zhang2016colorful,
  title     = {Colorful Image Colorization},
  author    = {Zhang, Richard and Isola, Phillip and Efros, Alexei A.},
  booktitle = ECCV,
  pages     = {649--666},
  year      = {2016},
  publisher = {Springer},
}

@inproceedings{nerf,
  title={NeRF: Representing Scenes as Neural Radiance Fields for View Synthesis},
  author={Mildenhall, Ben and Srinivasan, Pratul P and Tancik, Matthew and Barron, Jonathan T and Ramamoorthi, Ravi and Ng, Ren},
  booktitle=ECCV,
  year={2020},
}

@article{3dgs,
  title={3D Gaussian Splatting for Real-Time Radiance Field Rendering},
  author={Kerbl, Bernhard and Kopanas, Georgios and Leimk{\"u}hler, Thomas and Drettakis, George},
  journal=TOG,
  volume={42},
  year={2023},
  number = 4
}

@inproceedings{colornerf,
  title={Colorizing monochromatic radiance fields},
  author={Cheng, Yean and Wan, Renjie and Weng, Shuchen and Zhu, Chengxuan and Chang, Yakun and Shi, Boxin},
  booktitle=AAAI,
  volume={38},
  pages={1317--1325},
  year={2024}
}

@inproceedings{chromadistill,
  title={ChromaDistill: Colorizing Monochrome Radiance Fields with Knowledge Distillation},
  author={Dhiman, Ankit and Srinath, R and Sarkar, Srinjay and Boregowda, Lokesh R and Babu, R Venkatesh},
  booktitle=WACV,
  pages={2400--2410},
  year={2025},
  organization={IEEE}
}

@inproceedings{img2img,
  title={Palette: Image-to-image diffusion models},
  author={Saharia, Chitwan and Chan, William and Chang, Huiwen and Lee, Chris and Ho, Jonathan and Salimans, Tim and Fleet, David and Norouzi, Mohammad},
  booktitle={ACM SIGGRAPH 2022 conference proceedings},
  pages={1--10},
  year={2022}
}

@article{img2imgturbo,
  title={One-step image translation with text-to-image models},
  author={Parmar, Gaurav and Park, Taesung and Narasimhan, Srinivasa and Zhu, Jun-Yan},
  journal={arXiv preprint arXiv:2403.12036},
  year={2024}
}

@inproceedings{plenoxels,
  title={Plenoxels: Radiance fields without neural networks},
  author={Fridovich-Keil, Sara and Yu, Alex and Tancik, Matthew and Chen, Qinhong and Recht, Benjamin and Kanazawa, Angjoo},
  booktitle=CVPR,
  pages={5501--5510},
  year={2022}
}

@inproceedings{difix3d,
  title={Difix3d+: Improving 3d reconstructions with single-step diffusion models},
  author={Wu, Jay Zhangjie and Zhang, Yuxuan and Turki, Haithem and Ren, Xuanchi and Gao, Jun and Shou, Mike Zheng and Fidler, Sanja and Gojcic, Zan and Ling, Huan},
  booktitle=CVPR,
  pages={26024--26035},
  year={2025}
}

@inproceedings{lai2018learning,
  title={Learning blind video temporal consistency},
  author={Lai, Wei-Sheng and Huang, Jia-Bin and Wang, Oliver and Shechtman, Eli and Yumer, Ersin and Yang, Ming-Hsuan},
  booktitle=ECCV,
  pages={170--185},
  year={2018}
}

@inproceedings{liu2019pccn,
  title={Pccn: Point cloud colorization network},
  author={Liu, Jitao and Dai, Songmin and Li, Xiaoqiang},
  booktitle=ICIP,
  pages={3716--3720},
  year={2019},
  organization={IEEE}
}

@inproceedings{cao2018point,
  title={Point cloud colorization based on densely annotated 3d shape dataset},
  author={Cao, Xu and Nagao, Katashi},
  booktitle={International Conference on Multimedia Modeling},
  pages={436--446},
  year={2018},
  organization={Springer}
}

@inproceedings{shinohara2021point2color,
  title={Point2color: 3d point cloud colorization using a conditional generative network and differentiable rendering for airborne lidar},
  author={Shinohara, Takayuki and Xiu, Haoyi and Matsuoka, Masashi},
  booktitle=CVPR,
  pages={1062--1071},
  year={2021}
}

@inproceedings{gao2023scene,
  title={Scene-level Point Cloud Colorization with Semantics-and-geometry-aware Networks},
  author={Gao, Rongrong and Xiang, Tian-Zhu and Lei, Chenyang and Park, Jaesik and Chen, Qifeng},
  booktitle={Proceedings-IEEE International Conference on Robotics and Automation},
  volume={2023},
  pages={2818--2824},
  year={2023},
  organization={IEEE}
}

@inproceedings{sauer2024adversarial,
  title={Adversarial diffusion distillation},
  author={Sauer, Axel and Lorenz, Dominik and Blattmann, Andreas and Rombach, Robin},
  booktitle=ECCV,
  pages={87--103},
  year={2024},
  organization={Springer}
}

@misc{sdturbo,
  author = {{Stability AI}},
  title = {SD-Turbo},
  year = {2023},
  url = {https://huggingface.co/stabilityai/sd-turbo},
}

@inproceedings{mipnerf360,
  title={Mip-nerf 360: Unbounded anti-aliased neural radiance fields},
  author={Barron, Jonathan T and Mildenhall, Ben and Verbin, Dor and Srinivasan, Pratul P and Hedman, Peter},
  booktitle=CVPR,
  pages={5470--5479},
  year={2022}
}

@article{knapitsch2017tanks,
  title={Tanks and temples: Benchmarking large-scale scene reconstruction},
  author={Knapitsch, Arno and Park, Jaesik and Zhou, Qian-Yi and Koltun, Vladlen},
  journal=TOG,
  volume={36},
  number={4},
  pages={1--13},
  year={2017},
  publisher={ACM New York, NY, USA}
}

@Article{iizuka2016let,
  author = {Satoshi Iizuka and Edgar Simo-Serra and Hiroshi Ishikawa},
  title = {Let there be Color!: Joint End-to-end Learning of Global and Local Image Priors for Automatic Image Colorization with Simultaneous Classification},
  journal = TOG,
  year = 2016,
  volume = 35,
}

@inproceedings{kumar2021colorization,
  title={Colorization Transformer},
  author={Kumar, Disha and Singh, Gaurav and Lee, Yong Man},
  booktitle=ICCV,
  pages={16361--16370},
  year={2021}
}

@inproceedings{nazeri2018image,
  title={Image Colorization Using Generative Adversarial Networks},
  author={Nazeri, Kamyar and Ng, Eric and Joseph, Fahim and Qureshi, Faisal and Ebrahimi, Mehran},
  booktitle=CVPRW,
  pages={708--715},
  year={2018}
}

@inproceedings{zhang2022bigcolor,
  title={BigColor: Colorization using a Big GAN},
  author={Zhang, Yijun and Ma, Kede and Chen, Yinqiang},
  booktitle=ECCV,
  pages={345--362},
  year={2022}
}

@inproceedings{ho2020ddpm,
  title={Denoising Diffusion Probabilistic Models},
  author={Ho, Jonathan and Jain, Ajay and Abbeel, Pieter},
  booktitle=NeurIPS,
  pages={6840--6851},
  year={2020}
}

@article{luo2022pearlgan,
  title     = {Thermal Infrared Image Colorization for Nighttime Driving Scenes With Top-Down Guided Attention},
  author    = {Luo, Fuya and Li, Yunhan and Zeng, Guang and Peng, Peng and Wang, Gang and Li, Yongjie},
  journal   = {IEEE Transactions on Intelligent Transportation Systems},
  volume    = {23},
  pages     = {15808--15823},
  year      = {2022},
  publisher = {IEEE}
}

@inproceedings{berg2018generating,
  title     = {Generating Visible Spectrum Images from Thermal Infrared},
  author    = {Berg, Amanda and Ahlberg, J{\"o}rgen and Felsberg, Michael},
  booktitle = CVPRW,
  pages     = {1143--1152},
  year      = {2018}
}

@inproceedings{nair2023t2vddpm,
  title     = {T2V-DDPM: Thermal to Visible Face Translation using Denoising Diffusion Probabilistic Models},
  author    = {Nair, Nithin Gopalakrishnan and Patel, Vishal M.},
  booktitle = {Proceedings of the IEEE 17th International Conference on Automatic Face and Gesture Recognition (FG)},
  pages     = {1--7},
  year      = {2023},
  publisher = {IEEE},
  doi       = {10.1109/FG57933.2023.10042661}
}

@inproceedings{cabrelles2009,
  author    = {M. Cabrelles and S. Galcer{\'a} and S. Navarro and J. L. Lerma and T. Akasheh and N. Haddad},
  title     = {Integration of 3D Laser Scanning, Photogrammetry and Thermography to Record Architectural Monuments},
  booktitle = {Proc. 22nd CIPA Symposium},
  publisher = {CIPA},
  address   = {Kyoto, Japan},
  year      = {2009}
}

@inproceedings{iwaszczuk2011,
  author    = {Dominik Iwaszczuk and Lena Hoegner and Uwe Stilla},
  title     = {Matching of 3D Building Models with IR Images for Texture Extraction},
  booktitle = {Proc. Joint Urban Remote Sensing Event (JURSE)},
  publisher = {IEEE},
  address   = {Munich, Germany},
  pages     = {25--28},
  year      = {2011}
}

@article{chen2015reconstruction,
  title     = {3D Reconstruction from IR Thermal Images and Reprojective Evaluations},
  author    = {Chen, Chia-Yen and Yeh, Chia-Hung and Chang, Bao Rong and Pan, Jun-Ming},
  journal   = {Mathematical Problems in Engineering},
  volume    = {2015},
  pages     = {1--8},
  year      = {2015},
  publisher = {Hindawi}
}

@inproceedings{li2021nirpolar,
  author    = {Xiang Li and Jian Zhang and Xue Bai and Qing Wang},
  title     = {Near-Infrared Monocular 3D Reconstruction via Polarisation Imaging},
  booktitle = CVPR,
  publisher = {IEEE},
  pages     = {15616--15630},
  year      = {2021}
}

@inproceedings{ye2024thermalnerf,
  author    = {Tianxiang Ye and Qi Wu and Junyuan Deng and Guoqing Liu and Liu Liu and Songpengcheng Xia and Liang Pang and Wenxian Yu and Ling Pei},
  title     = {Thermal-NeRF: Neural Radiance Fields from an Infrared Camera},
  booktitle = CVPR,
  publisher = {IEEE},
  pages     = {1--10},
  year      = {2024}
}

@article{liu2025thermalgs,
  author    = {Y. Liu and X. Zhang and J. Wang and H. Zhao and F. Xu},
  title     = {Dynamic Thermal 3D Reconstruction with Gaussian Splatting},
  journal   = {Remote Sensing},
  publisher = {MDPI},
  volume    = {17},
  number    = {2},
  pages     = {335},
  year      = {2025},
}

@inproceedings{yu2021pixelnerf,
  title={pixelnerf: Neural radiance fields from one or few images},
  author={Yu, Alex and Ye, Vickie and Tancik, Matthew and Kanazawa, Angjoo},
  booktitle=CVPR,
  pages={4578--4587},
  year={2021}
}

@inproceedings{seo2023mixnerf,
  title={Mixnerf: Modeling a ray with mixture density for novel view synthesis from sparse inputs},
  author={Seo, Seunghyeon and Han, Donghoon and Chang, Yeonjin and Kwak, Nojun},
  booktitle=CVPR,
  pages={20659--20668},
  year={2023}
}

@inproceedings{seo2023flipnerf,
  title={Flipnerf: Flipped reflection rays for few-shot novel view synthesis},
  author={Seo, Seunghyeon and Chang, Yeonjin and Kwak, Nojun},
  booktitle=ICCV,
  pages={22883--22893},
  year={2023}
}

@inproceedings{yang2023freenerf,
  title={Freenerf: Improving few-shot neural rendering with free frequency regularization},
  author={Yang, Jiawei and Pavone, Marco and Wang, Yue},
  booktitle=CVPR,
  pages={8254--8263},
  year={2023}
}

@inproceedings{rematas2021sharf,
  title     = {ShaRF: Shape-conditioned Radiance Fields from a Single View},
  author    = {Rematas, Konstantinos and Martin-Brualla, Ricardo and Ferrari, Vittorio},
  booktitle = ICML,
  pages     = {8948--8958},
  year      = {2021},
  volume    = {139},
  series    = {Proceedings of Machine Learning Research},
  publisher = {PMLR}
}

@inproceedings{chen2021mvsnerf,
  title={Mvsnerf: Fast generalizable radiance field reconstruction from multi-view stereo},
  author={Chen, Anpei and Xu, Zexiang and Zhao, Fuqiang and Zhang, Xiaoshuai and Xiang, Fanbo and Yu, Jingyi and Su, Hao},
  booktitle=ICCV,
  pages={14124--14133},
  year={2021}
}

@inproceedings{gafni2021dynamic,
  title={Dynamic neural radiance fields for monocular 4d facial avatar reconstruction},
  author={Gafni, Guy and Thies, Justus and Zollhofer, Michael and Nie{\ss}ner, Matthias},
  booktitle=CVPR,
  pages={8649--8658},
  year={2021}
}

@article{muller2022instant,
  title={Instant neural graphics primitives with a multiresolution hash encoding},
  author={M{\"u}ller, Thomas and Evans, Alex and Schied, Christoph and Keller, Alexander},
  journal=TOG,
  volume={41},
  number={4},
  pages={1--15},
  year={2022},
  publisher={ACM New York, NY, USA}
}

@inproceedings{yu2021plenoctrees,
  title={Plenoctrees for real-time rendering of neural radiance fields},
  author={Yu, Alex and Li, Ruilong and Tancik, Matthew and Li, Hao and Ng, Ren and Kanazawa, Angjoo},
  booktitle=ICCV,
  pages={5752--5761},
  year={2021}
}

@inproceedings{mildenhall2019llff,
  title={Local Light Field Fusion: Practical View Synthesis with Prescriptive Sampling Guidelines},
  author={Mildenhall, Ben and Srinivasan, Pratul P. and Ortiz-Cayon, Rodrigo and Kalantari, Nima Khademi and Ramamoorthi, Ravi and Ng, Ren and Kar, Abhishek},
  booktitle=CVPR,
  pages={5511--5520},
  year={2019}
}

@article{hodosh2013flickr8k,
  title={Framing image description as a ranking task: Data, models and evaluation metrics},
  author={Hodosh, Micah and Young, Peter and Hockenmaier, Julia},
  journal={Journal of Artificial Intelligence Research},
  volume={47},
  pages={853--899},
  year={2013}
}

@inproceedings{ling2024d13dv,
  title     = {DL3DV-10K: A Large-Scale Scene Dataset for Deep Learning-based 3D Vision},
  author    = {Ling, Lu and Sheng, Yichen and Tu, Zhi and Zhao, Wentian and Xin, Cheng and Wan, Kun and Yu, Lantao and Guo, Qianyu and Yu, Zixun and Lu, Yawen and Li, Xuanmao and Sun, Xingpeng and Ashok, Rohan and Mukherjee, Aniruddha and Kang, Hao and Kong, Xiangrui and Hua, Gang and Zhang, Tianyi and Benes, Bedrich and Bera, Aniket},
  booktitle = CVPR,
  month     = {June},
  year      = {2024},
  pages     = {22160–22169}
}

@inproceedings{kang2023ddcolor,
  author    = {Kang, Xiaoyang and Yang, Tao and Ouyang, Wenqi and Ren, Peiran and Li, Lingzhi and Xie, Xuansong},
  title     = {DDColor: Towards Photo-Realistic Image Colorization via Dual Decoders},
  booktitle = ICCV,
  year      = {2023},
  pages     = {328–338}
}

@misc{google2026nanobanana2,
  author = {{Google DeepMind}},
  title  = {Gemini 3.1 Flash Image (Nano Banana 2) model card},
  url    = {https://deepmind.google/models/model-cards/gemini-3-1-flash-image/},
  year   = {2026},
  note   = {accessed: 2026-03-05}
}

@inproceedings{singh2023robotnerf,
  title={Robot-NeRF: A Neural Radiance Field for Robotic Manipulation},
  author={Singh, Rohit and Kamat, Shubham and Sushkov, Oleg and Fox, Dieter and Lee, Youngwoon},
  booktitle={Proceedings of the IEEE International Conference on Robotics and Automation (ICRA)},
  pages={11223--11230},
  year={2023}
}

@inproceedings{hu2022efficientnerf,
  title={EfficientNeRF: Efficient Neural Radiance Fields},
  author={Hu, Tao and Liu, Shu and Chen, Yilun and Shen, Tiancheng and Jia, Jiaya},
  booktitle=CVPR,
  pages={12902--12911},
  year={2022}
}

@inproceedings{cai2024saxnerf,
  title     = {Structure-Aware Sparse-View X-ray 3D Reconstruction},
  author    = {Cai, Yuanhao and Wang, Jiahao and Yuille, Alan and Zhou, Zongwei and Wang, Angtian},
  booktitle = CVPR,
  pages     = {13552--13561},
  year      = {2024}
}

@inproceedings{lin2024gaussianflow,
  title     = {Gaussian-Flow: 4D Reconstruction with Dynamic 3D Gaussian Particle},
  author    = {Lin, Youtian and Dai, Zuozhuo and Zhu, Siyu and Yao, Yao},
  booktitle = CVPR,
  pages     = {21136--21145},
  year      = {2024}
}

@inproceedings{cai2024xgaussian,
  title     = {Radiative Gaussian Splatting for Efficient X-ray Novel View Synthesis},
  author    = {Cai, Yuanhao and Liang, Yixun and Wang, Jiahao and Wang, Angtian and Zhang, Yulun and Yang, Xiaokang and Zhou, Zongwei and Yuille, Alan},
  booktitle = ECCV,
  pages     = {111--127},
  year      = {2024}
}

@article{gaussctrl2024,
author = {Wu, Jing and Bian, Jia-Wang and Li, Xinghui and Wang, Guangrun and Reid, Ian and Torr, Philip and Prisacariu, Victor},
title = {GaussCtrl: Multi-View Consistent Text-Driven 3D Gaussian Splatting Editing},
journal = ECCV,
year = {2024},
}

@inproceedings{yuan2022nerfediting,
  title     = {NeRF-Editing: Geometry Editing of Neural Radiance Fields},
  author    = {Yuan, Yifan and He, Xudong and Ye, Xiangyu and Liu, Mingming and Zhang, Jing and Tao, Dacheng},
  booktitle = CVPR,
  pages     = {18353--18362},
  year      = {2022}
}

@article{zhang2024text2nerf,
  title     = {Text2NeRF: Text-Driven 3D Scene Generation With Neural Radiance Fields},
  author    = {Zhang, Jingbo and Li, Xiaoyu and Wan, Ziyu and Wang, Can and Liao, Jing},
  journal   = {IEEE Transactions on Visualization and Computer Graphics},
  year      = {2024}
}

@article{poole2022dreamfusion,
  title     = {DreamFusion: Text-to-3D using 2D Diffusion},
  author    = {Poole, Ben and Jain, Ajay and Barron, Jonathan T. and Mildenhall, Ben},
  journal   = {CoRR},
  volume    = {abs/2209.14988},
  year      = {2022},
}

@inproceedings{loshchilovdecoupled,
  title={Decoupled Weight Decay Regularization},
  author={Loshchilov, Ilya and Hutter, Frank},
  booktitle=ICLR,
  year = {2019}
}

@article{hasler2003measuring,
  title     = {Measuring colourfulness in natural images},
  author    = {Hasler, David and S{\"u}sstrunk, Sabine},
  journal   = {Proceedings of SPIE--The International Society for Optical Engineering, Human Vision and Electronic Imaging VIII},
  volume    = {5007},
  pages     = {87--95},
  year      = {2003},
  publisher = {SPIE}
}

@inproceedings{yan2024multi,
  title={Multi-scale 3d gaussian splatting for anti-aliased rendering},
  author={Yan, Zhiwen and Low, Weng Fei and Chen, Yu and Lee, Gim Hee},
  booktitle=CVPR,
  pages={20923--20931},
  year={2024}
}

@inproceedings{yu2024mip,
  title={Mip-splatting: Alias-free 3d gaussian splatting},
  author={Yu, Zehao and Chen, Anpei and Huang, Binbin and Sattler, Torsten and Geiger, Andreas},
  booktitle=CVPR,
  pages={19447--19456},
  year={2024}
}

@inproceedings{liang2024analytic,
  title={Analytic-splatting: Anti-aliased 3d gaussian splatting via analytic integration},
  author={Liang, Zhihao and Zhang, Qi and Hu, Wenbo and Zhu, Lei and Feng, Ying and Jia, Kui},
  booktitle=ECCV,
  pages={281--297},
  year={2024},
  organization={Springer}
}

@inproceedings{tang2024dreamgaussian,
  title     = {DreamGaussian: Generative Gaussian Splatting for Efficient 3D Content Creation},
  author    = {Tang, Jiaxiang and Ren, Jiawei and Zhou, Hang and Liu, Ziwei and Zeng, Gang},
  booktitle = ICLR,
  year      = {2024}
}

@inproceedings{yuan2024gavatar,
  title={Gavatar: Animatable 3d gaussian avatars with implicit mesh learning},
  author={Yuan, Ye and Li, Xueting and Huang, Yangyi and De Mello, Shalini and Nagano, Koki and Kautz, Jan and Iqbal, Umar},
  booktitle=CVPR,
  pages={896--905},
  year={2024}
}

@inproceedings{zou2024triplane,
  title={Triplane meets gaussian splatting: Fast and generalizable single-view 3d reconstruction with transformers},
  author={Zou, Zi-Xin and Yu, Zhipeng and Guo, Yuan-Chen and Li, Yangguang and Liang, Ding and Cao, Yan-Pei and Zhang, Song-Hai},
  booktitle=CVPR,
  pages={10324--10335},
  year={2024}
}

@inproceedings{lin2025diffsplat,
  title={DiffSplat: Repurposing Image Diffusion Models for Scalable 3D Gaussian Splat Generation},
  author={Lin, Chenguo and Pan, Panwang and Yang, Bangbang and Li, Zeming and Mu, Yadong},
  booktitle=ICLR,
  year={2025}
}

@inproceedings{yang2024deformable,
  title={Deformable 3d gaussians for high-fidelity monocular dynamic scene reconstruction},
  author={Yang, Ziyi and Gao, Xinyu and Zhou, Wen and Jiao, Shaohui and Zhang, Yuqing and Jin, Xiaogang},
  booktitle=CVPR,
  pages={20331--20341},
  year={2024}
}

@inproceedings{wu20244d,
  title={4D gaussian splatting for real-time dynamic scene rendering},
  author={Wu, Guanjun and Yi, Taoran and Fang, Jiemin and Xie, Lingxi and Zhang, Xiaopeng and Wei, Wei and Liu, Wenyu and Tian, Qi and Wang, Xinggang},
  booktitle=CVPR,
  pages={20310--20320},
  year={2024}
}

@inproceedings{yan20244d,
  title={4D gaussian splatting with scale-aware residual field and adaptive optimization for real-time rendering of temporally complex dynamic scenes},
  author={Yan, Jinbo and Peng, Rui and Tang, Luyang and Wang, Ronggang},
  booktitle=ACMMM,
  pages={7871--7880},
  year={2024}
}

@inproceedings{teed2020raft,
  title={Raft: Recurrent all-pairs field transforms for optical flow},
  author={Teed, Zachary and Deng, Jia},
  booktitle=ECCV,
  pages={402--419},
  year={2020},
  organization={Springer}
}

@inproceedings{colormnet,
  title={ColorMNet: A memory-based deep spatial-temporal feature propagation network for video colorization},
  author={Yang, Yixin and Dong, Jiangxin and Tang, Jinhui and Pan, Jinshan},
  booktitle=ECCV,
  pages={336--352},
  year={2024},
  organization={Springer}
}

@inproceedings{genn2n,
    author    = {Liu, Xiangyue and Xue, Han and Luo, Kunming and Tan, Ping and Yi, Li},
    title     = {GenN2N: Generative NeRF2NeRF Translation},
    booktitle = CVPR,
    month     = {June},
    year      = {2024},
    pages     = {5105-5114}
}

@article{color3d,
  title={Color3D: Controllable and Consistent 3D Colorization with Personalized Colorizer},
  author={Wan, Yecong and Shao, Mingwen and Wu, Renlong and Zuo, Wangmeng},
  journal={arXiv preprint arXiv:2510.10152},
  year={2025}
}

@inproceedings{kim2025pixelnir,
  author    = {Kim, Jinnyeong and Baek, Seung-Hwan},
  title     = {Pixel-aligned {RGB-NIR} Stereo Imaging and Dataset for Robot Vision},
  booktitle = {CVPR},
  year      = {2025},
  pages     = {11482-11492}
}

\clearpage
\setcounter{page}{1}
\setcounter{figure}{0}
\setcounter{table}{0}

\renewcommand{\thefigure}{S.\arabic{figure}}
\renewcommand{\thetable}{S.\arabic{table}}
\appendix

\begin{center}
    \Large
    \textbf{\\LoGoColor: Local-Global 3D Colorization\\ for 360\textdegree~Scenes} \\
    \vspace{0.5em}
    {\large Supplementary Material}
    \vspace{1.0em}
\end{center}

\noindent
In this appendix, we provide additional implementation details, a comprehensive analysis, and additional experiments. The supplementary material is organized as follows:
\begin{itemize}[label=\textbullet, itemsep=3pt]
    \item \cref{sec:details}: Implementation and Dataset Details
        \begin{itemize}[label=$\circ$, itemsep=3pt, topsep=1pt]
        \item \cref{sec:other_details}: Implementation details
        \item \cref{sec:scene_list}: DL3DV-10K evaluation scene list
        \end{itemize}

    \item \cref{sec:metric}: Discussion on Evaluation Metrics
        \begin{itemize}[label=$\circ$, itemsep=3pt, topsep=1pt]
        \item \cref{sec:ncolorfulness}: Normalized Colorfulness
        \item \cref{sec:consistency}: Consistency
        \end{itemize}

    \item \cref{sec:supp_exp}: Experiments
    \begin{itemize}[label=$\circ$, itemsep=3pt, topsep=1pt]
        \item \cref{sec:base_views}: Example base views
        \item \cref{sec:supp_qual}: Extended qualitative comparisons
        \item \cref{sec:supp_ablation_qual}: Extended qualitative ablation
        \item \cref{sec:supp_application}: Additional application results
    \end{itemize}

    \item \cref{sec:limitation}: Limitations
    \item \cref{sec:video}: \textbf{Video Results}
\end{itemize}

\section{Implementation and Dataset Details}
\label{sec:details}

\subsection{Implementation details}
\label{sec:other_details}

We calculate short- and long-term consistency by first rendering the scene using the training view cameras and sorting the resulting frames by sequence order. 
We set the temporal delta frame to 1 for short-term consistency and 5 for long-term consistency. 
Following \cite{chromadistill}, we utilize the RAFT~\cite{teed2020raft} optical flow method; the optical flow is computed on the grayscale version of the rendered images to minimize the impact of color inconsistencies. 
Consistency metrics are then calculated in the LAB space, using only the AB channels.

For the sake of transparency, we include the implementation code for our 3DGS-based versions of ColorNeRF and ChromaDistill.

\subsection{DL3DV-10K evaluation scene list}
\label{sec:scene_list}
The names of the scenes used for evaluation are listed below.
\begin{itemize}
    {\small
    \item \texttt{0569e83fdc248a51fc0ab082ce5e2baff15755c53c207f545e6d02d91f01d166}
    \item \texttt{0bfdd020cf475b9c68e4b469d1d1a2d0cad303eefe8b78fb2307855afdaac8be}
    \item \texttt{1ba74c22670ad047981441581d00f26f4a148d1010bcac7468c615adf5fa4d5d}
    \item \texttt{1da888bdedfc9629c0fa9f82cf3f5d96f8103baee0ff64d9311aea1224a9f2ae}
    \item \texttt{26fd23358fa11fff0fb3180ef0b65591b486e20dcf753ce4a7aae49a37e370c7}
    \item \texttt{341b4ff3dfd3d377d7167bd81f443bedafbff003bf04881b99760fc0aeb69510}
    \item \texttt{35317e621976e87f0c143e66fc61fb8cddb4ff134304da7a00e32ac1983105b4}
    \item \texttt{3b16a10ec9b4ab71580958b634485a979ffd6df0d368dbbf6fc1c5ffacf46b7a}
    \item \texttt{3bb894d1933f3081134ad2d40e54de5f0636bd8b502b0a8561873bb63b0dce85}
    \item \texttt{54bf355ca7e08ed1bc86f5772e564ac0f92981ca25dab24d86b694e915fc4c43}
    \item \texttt{599ca3e04cae3ec83affc426af7d0d7ab36eb91cd8e539edbc13070a4d455792}
    \item \texttt{6e11e7f4fea305c7c4658d2c1f8df29e6f299330860cf48ffbf1c5ff8b96c0a8}
    \item \texttt{71b2dc8a2aa553da09b8b94b9f0d5e8abcca307def74d26301616ee238464d46}
    \item \texttt{90cb7ef95384138c2370f13a9ae1698fb1b5bdd68e8b3d01f8e53d38933a4b92}
    \item \texttt{9e9a89ae6fed06d6e2f4749b4b0059f35ca97f848cedc4a14345999e746f7884}
    \item \texttt{a62c330f5403e2e41a82a74c4e865b705c5706843b992fae2fe2e538b122d984}
    \item \texttt{adf35184a12d4cfa3f4248b87aa5adb4f39f179df460d6d76136e13d37299a2a}
    \item \texttt{ba55c875d20c34ee85ffc72264c4d77710852e5fb7d9ce4b9c26a8442850e98f}
    \item \texttt{c076929db6501cf7ebe386c70e6d77ea3af844a745e794f2ec17c981c465a69b}
    \item \texttt{c37109a55effe0000f8e40652ca935376e75bcb2a0b56de8eabd20a26e2a0f68}
    \item \texttt{d3812aad538261e7f73c75762ff55f23b468bcc76f376d52ac86ca6cf3c44b4b}
    \item \texttt{d9b6376623741313bf6da6bf4cdb9828be614a2ce9390ceb3f31cd535d661a75}
    \item \texttt{e78f8cebd2bd93d960bfaeac18fac0bb2524f15c44288903cd20b73e599e8a81}
    \item \texttt{e9360e7a89bee835dc847cf8796093e634b759ff582558788dcfe8326f6e8901}
    \item \texttt{eb4cf52988f805e6fce11d1b239fa9de32eb157364cff06ebac0aa50e0a46567}
    \item \texttt{ec1e44d4dc0f8fa77610866495f9297a7f82158c43e1777668b84fd4b736c7bc}
    \item \texttt{ef59aac437132bfc1dd45a7e1f8e4800978e7bb28bf98c4428d26fb3e1da3e90}
    \item \texttt{fb3b73f1d3fe9d192f21f55f5100fd258887aef345f778e0a64fc0587930a6f9}
    }
\end{itemize}

\section{Discussion on Evaluation Metrics}
\label{sec:metric}

\begin{figure}[t]
    \centering

    \includegraphics[width=0.7\textwidth]{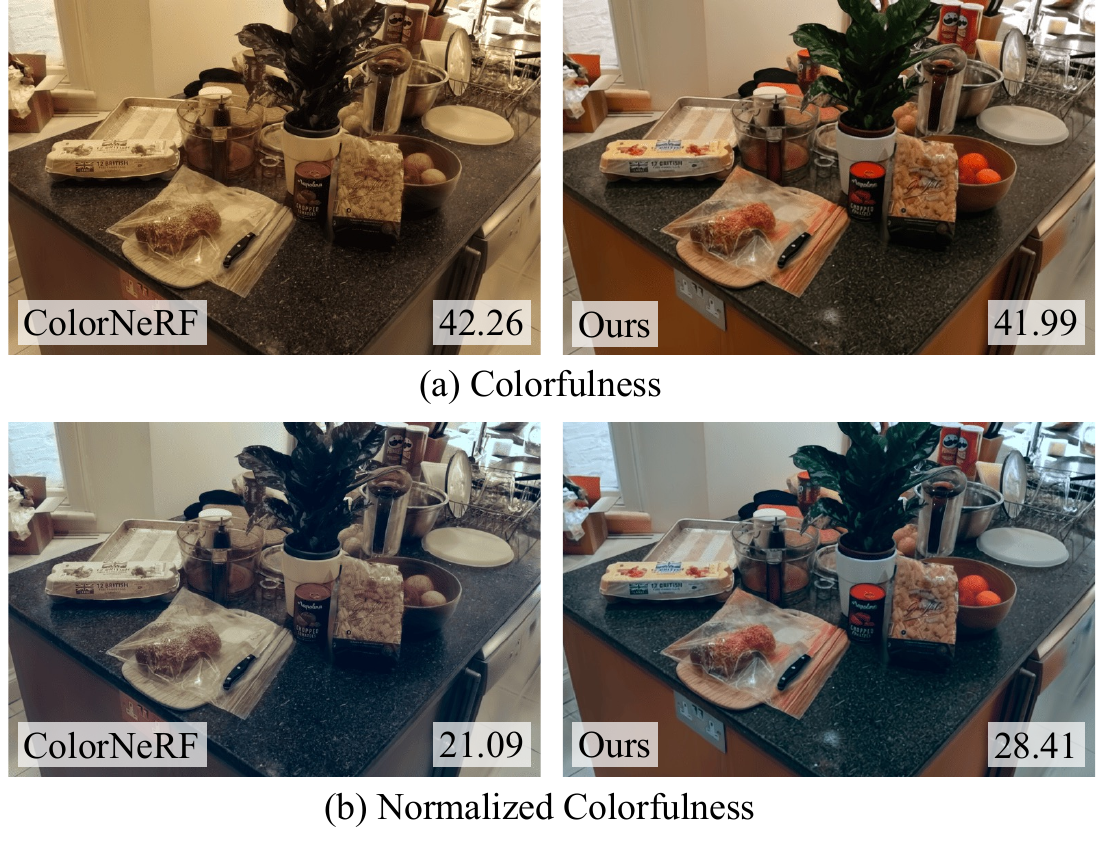} \\
    \vspace{-5pt}
    \caption{
    \textbf{Colorfulness --} While the yellowish-cast image of ColorNeRF records a higher score on the Colorfulness metric in (a), it obtains a low nColorfulness in (b).
    }%

    \label{fig:ncolorfulness}
\end{figure}

\subsection{Normalized Colorfulness}
\label{sec:ncolorfulness}
Colorfulness~\cite{hasler2003measuring} is a standard metric in colorization tasks.
However, 3D colorization introduces additional challenges compared to image colorization, such as maintaining cross-view color consistency.
Common failure modes in 3D colorization include: 1) ``averaging out'' colors across various objects instead of applying distinct hues, or 2) colorizing with inconsistent colors. 
In the former case, the standard Colorfulness metric is limited in assessing color diversity, as it may yield high scores due to a dominant global color tint rather than a diverse distribution of hues, as shown in \cref{fig:ncolorfulness}(a).
To address this, we propose Normalized Colorfulness (nColorfulness), which focuses specifically on color diversity by removing the global chromatic bias.

Specifically, given an image $I$, we first convert it to the LAB color space, where $L$ represents lightness and $a, b$ represent chromaticity layers. 
We then normalize the chromatic channels by subtracting their respective means:
\begin{equation}
    a' = a - \mu_a, \quad b' = b - \mu_b,
\end{equation}
where $\mu_a$ and $\mu_b$ are the mean values of the $a$ and $b$ channels across all pixels. 
The image is then converted back to the RGB space, denoted as $I_\text{norm}$. 
Finally, we compute the Colorfulness score on $I_\text{norm}$ using the standard formula from \cite{hasler2003measuring}:
\begin{equation}
\text{nColorfulness} = \sqrt{\sigma_{rg}^2 + \sigma_{yb}^2} + 0.3\sqrt{\mu_{rg}^2 + \mu_{yb}^2},
\end{equation}
where $rg = R - G$ and $yb = 0.5 (R + G) - B$. 
As illustrated in \cref{fig:ncolorfulness}(b), this normalization effectively eliminates the global tint, ensuring that higher scores are awarded to scenes with diverse color distributions across different objects. 

\subsection{Consistency}
\label{sec:consistency}
While warping-based consistency metrics are effective to evaluate view pairs with small disparities, their application to expansive 360-degree scenes involves practical limitations. 
These metrics depend on the precision of optical flow estimation, which can become increasingly complex as the distance between views increases. 
Consequently, quantitative metrics may not fully reflect long-term consistency across extended camera trajectories.
Given these considerations, we believe video results offer a more comprehensive and intuitive representation of the color consistency achieved by our method. 
We therefore encourage readers to refer to videos on our project page for a comprehensive visual evaluation.

\section{Experiments}
\label{sec:supp_exp}

\subsection{Example base views}
\label{sec:base_views}
\begin{figure}[t]
    \centering
    \newcommand{\imgw}{0.24\textwidth}
    \setlength{\tabcolsep}{1pt}
    
    \begin{subfigure}{0.99\linewidth}
    \centering
        \begin{tabular}{cccc}
        \includegraphics[width=\imgw]{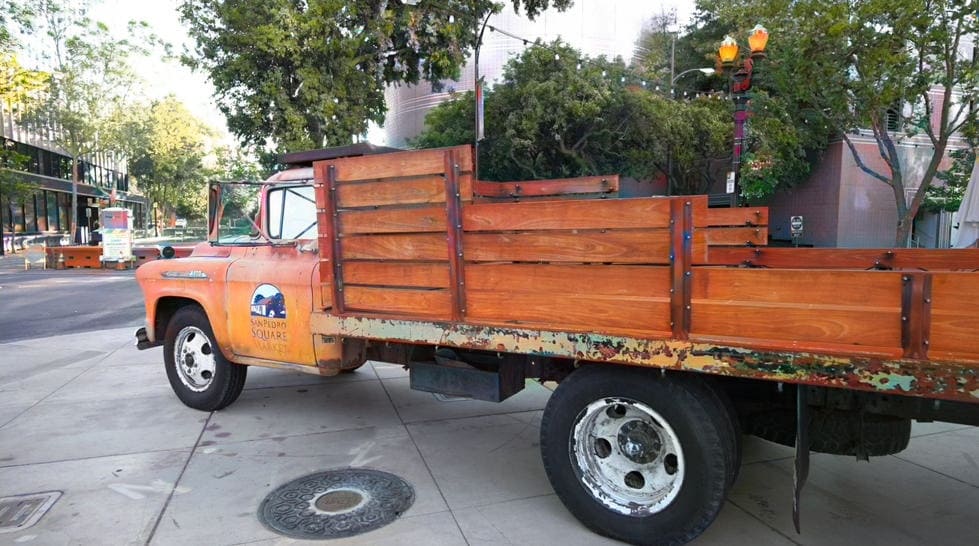} &
        \includegraphics[width=\imgw]{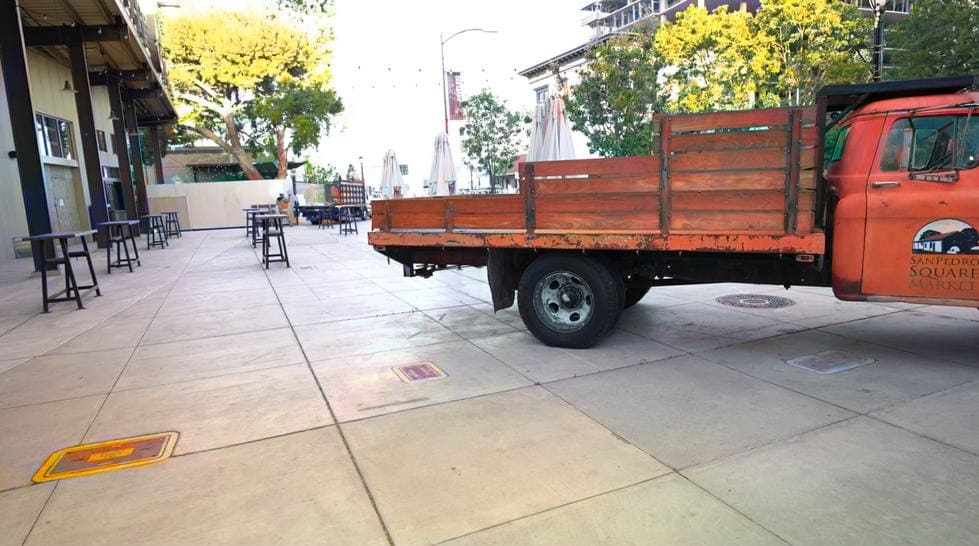} &
        \includegraphics[width=\imgw]{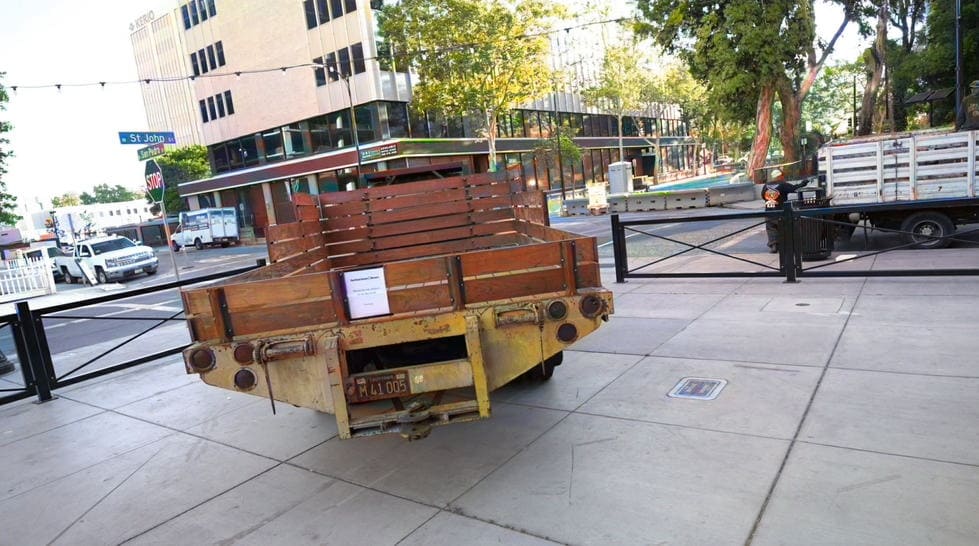} &
        \includegraphics[width=\imgw]{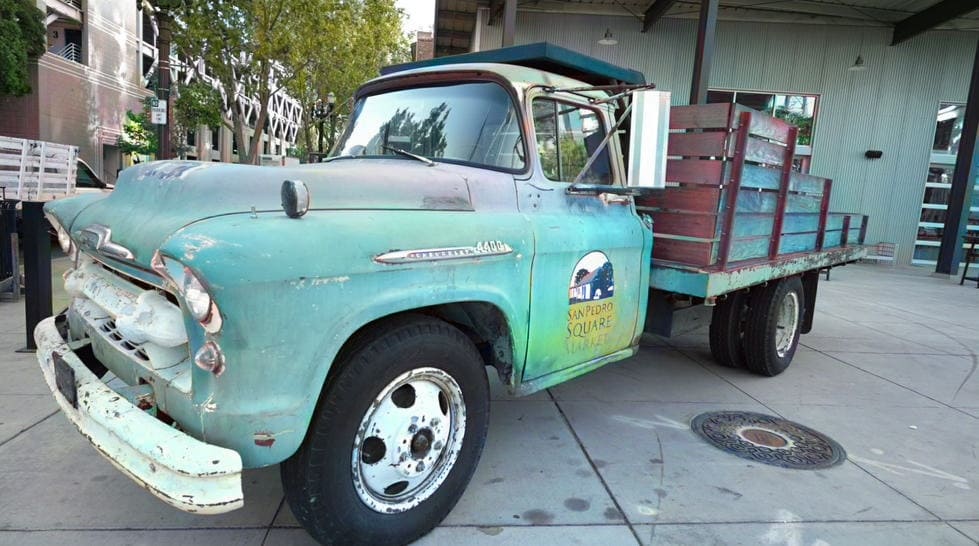} \\
        \end{tabular}
    \vspace{-2pt}
    \caption{Output of $\mathcal{F}$}
    \vspace{2pt}
    \end{subfigure}
    
    \begin{subfigure}{0.99\linewidth}
    \centering
        \begin{tabular}{cccc}
        \includegraphics[width=\imgw]{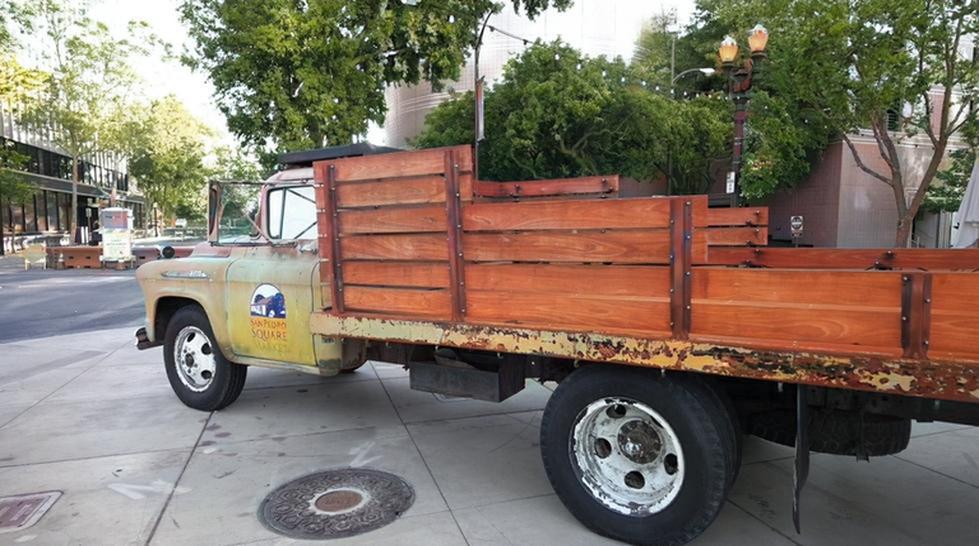} &
        \includegraphics[width=\imgw]{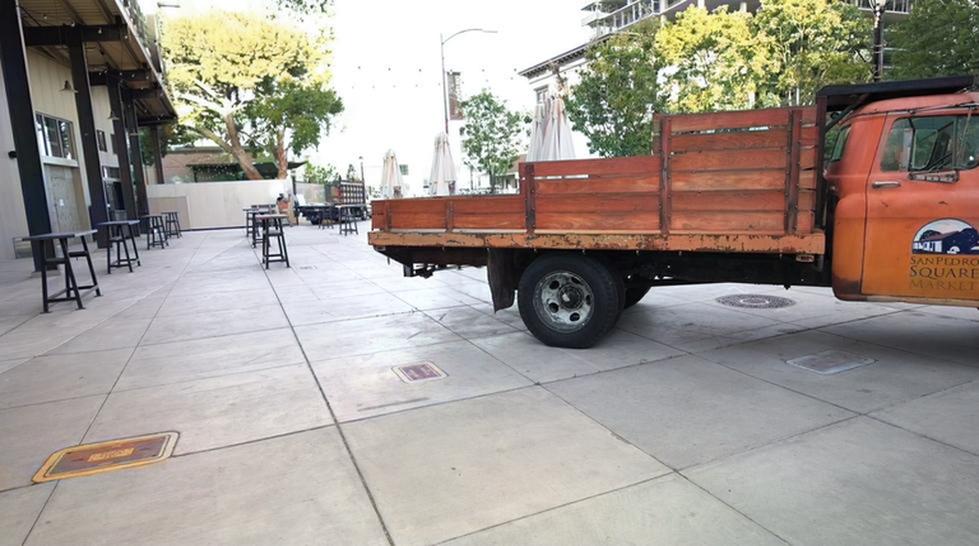} &
        \includegraphics[width=\imgw]{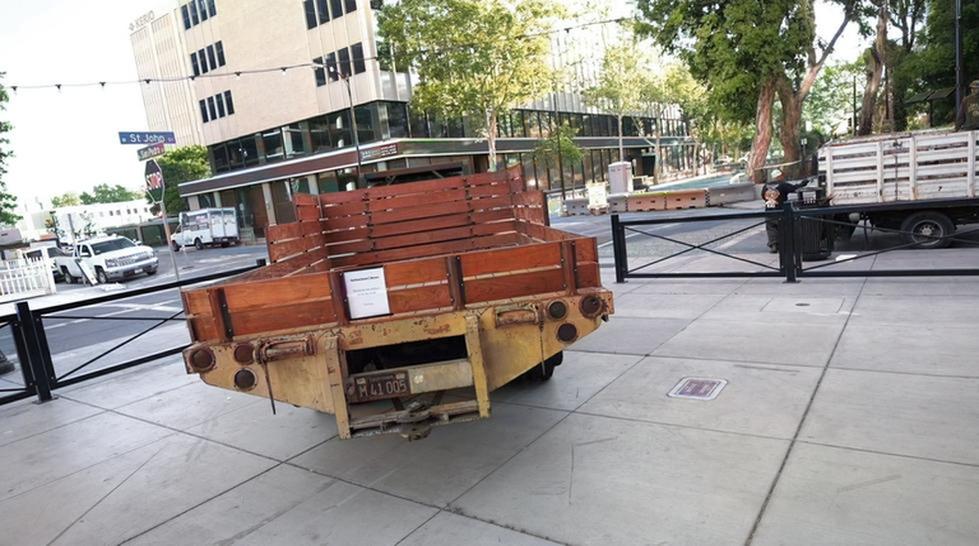} &
        \includegraphics[width=\imgw]{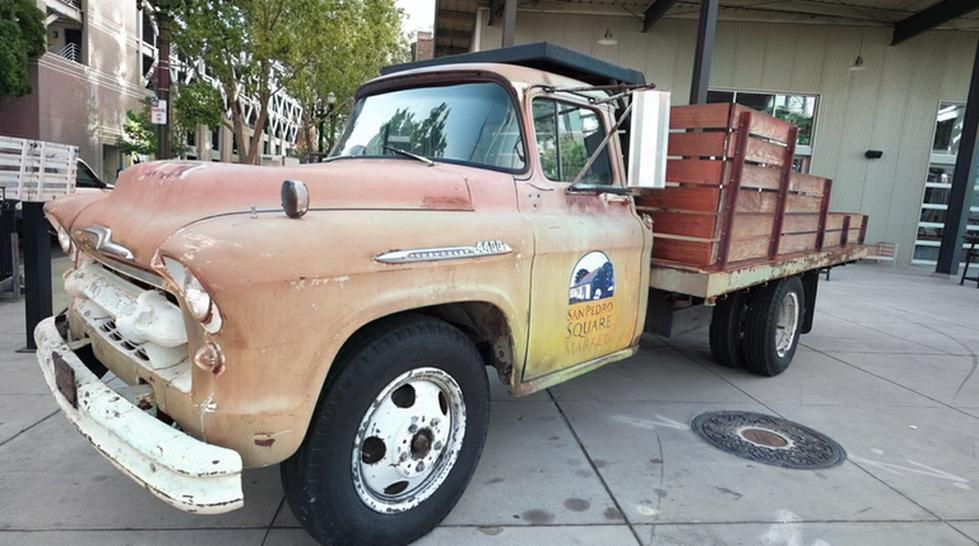} \\
        \end{tabular}
    \vspace{-2pt}
    \caption{After global calibration}
    \vspace{2pt}
    \end{subfigure}

    \caption{
    \textbf{Example base views --} Base views are selected to attain maximum coverage, while global calibration is applied to resolve color inconsistencies across the subscenes.
    }
    \label{fig:baseview}
\end{figure}
We present example base views before and after global calibration in \cref{fig:baseview}.
As shown in (a), our subscene decomposition ensures that the selected base views observe distinct parts of the scene, achieving near-maximum coverage with minimal overlap. 
However, despite this strategic decomposition, the independent application of a 2D colorization model results in chromatic inconsistencies across base views due to its view-agnostic nature. 
In contrast, our global calibration process effectively aligns these disparate colorizations into a set of globally consistent base views, as illustrated in (b).

\begin{figure}[p]
    \centering
    \newcommand{\imgw}{0.2\textwidth}
    \newcommand{\imgh}{0.13\textwidth}
    
    \setlength{\tabcolsep}{0.5pt}
    \resizebox{0.99\linewidth}{!}{
    \begin{tabular}{@{}ccccccc@{}}
        & {Input Image} & {GenN2N} & {ColorMNet} & {ColorNeRF} & {ChromaDistill} & {Our Method} \\
        
        \raisebox{9pt}{\rotatebox{90}{TnT-Horse}} &
        \includegraphics[width=\imgw]{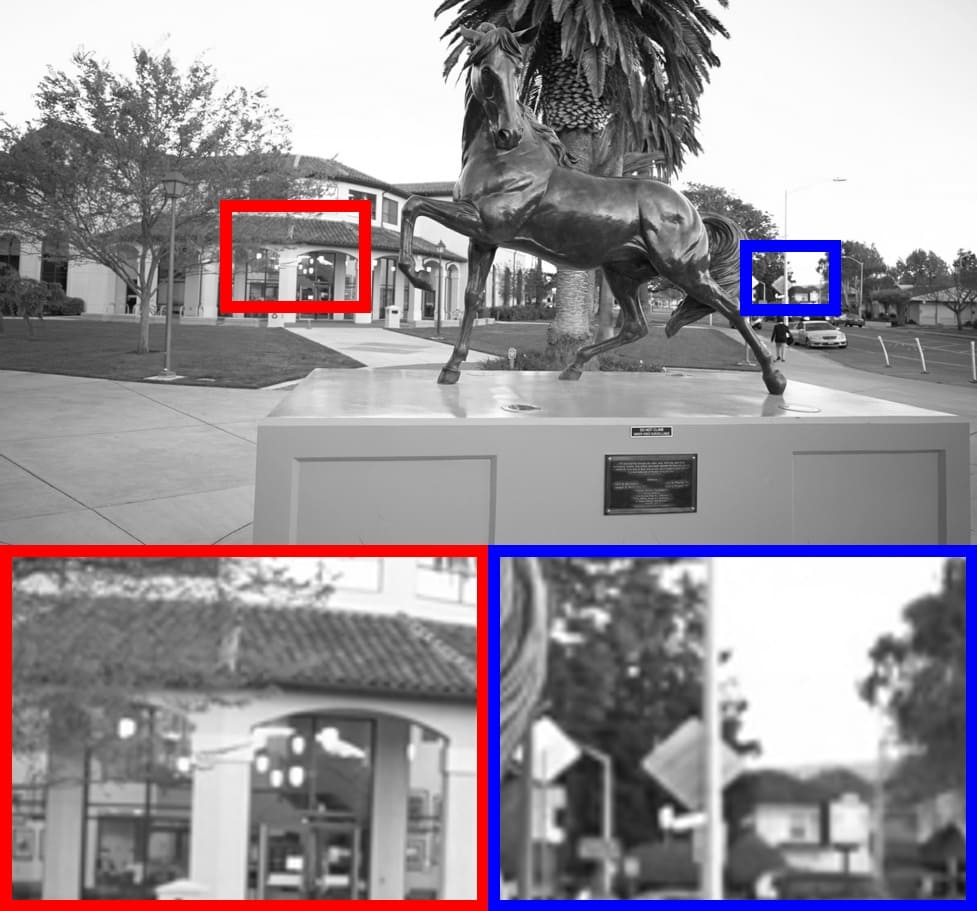} &
        \includegraphics[width=\imgw]{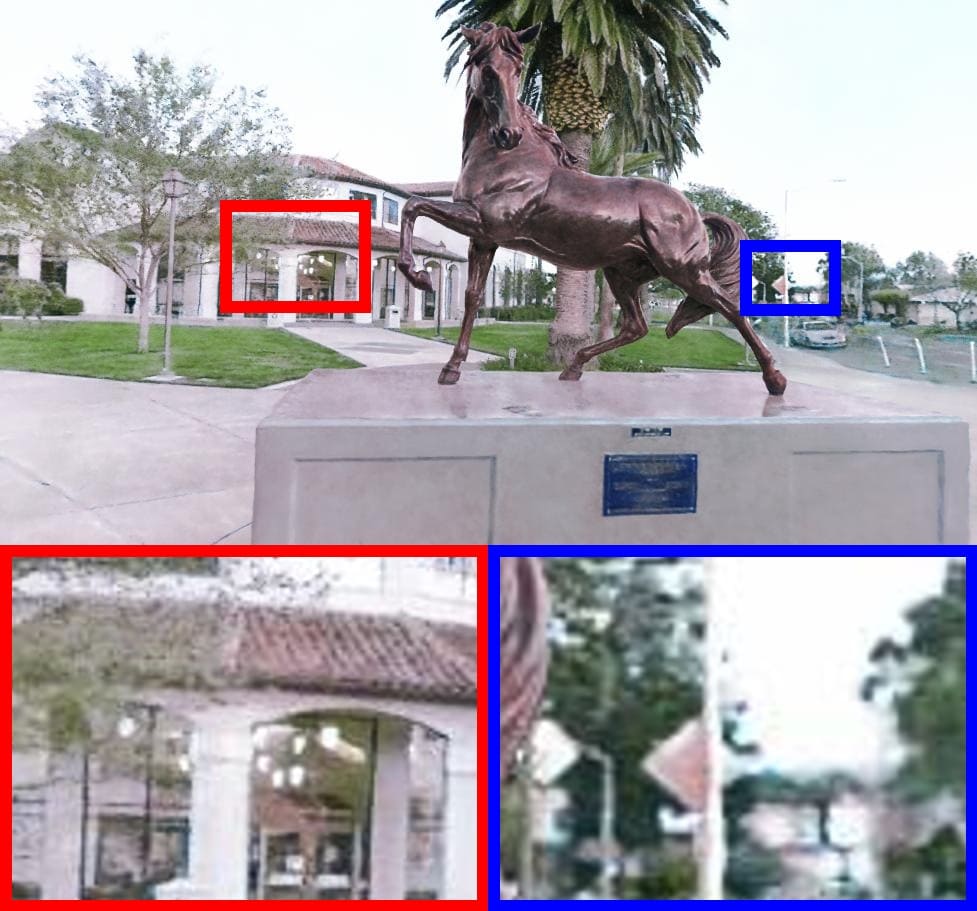} &
        \includegraphics[width=\imgw]{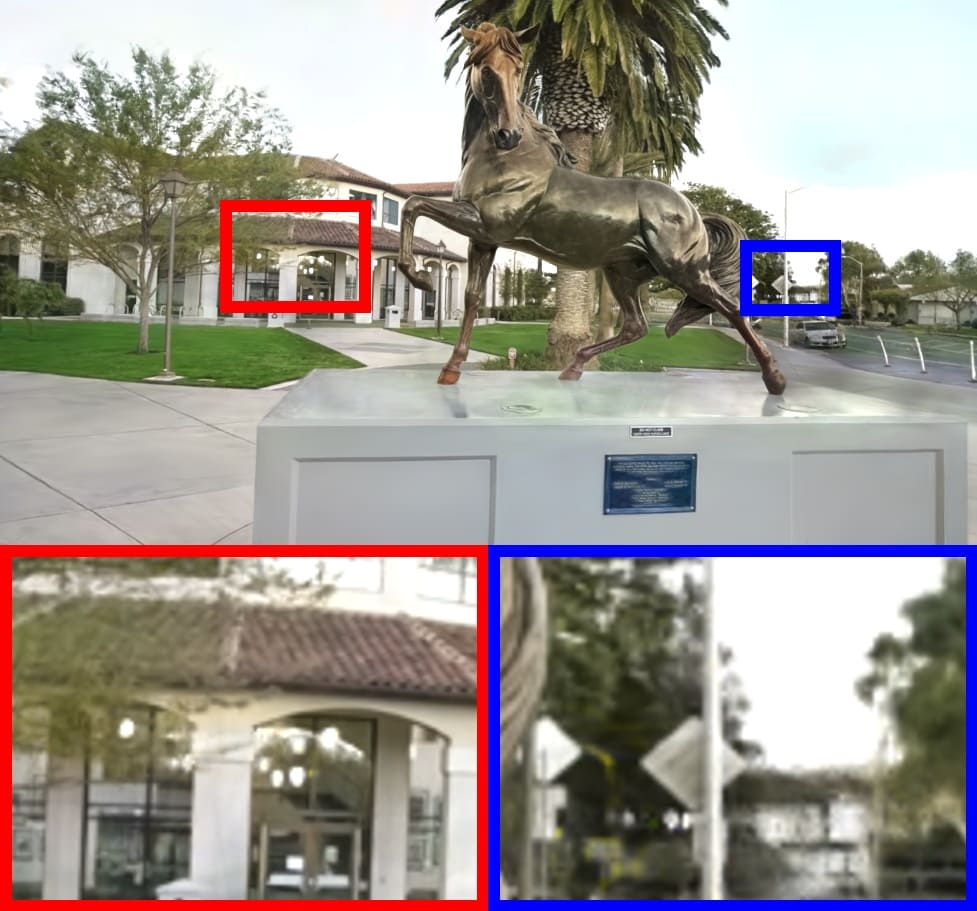} &
        \includegraphics[width=\imgw]{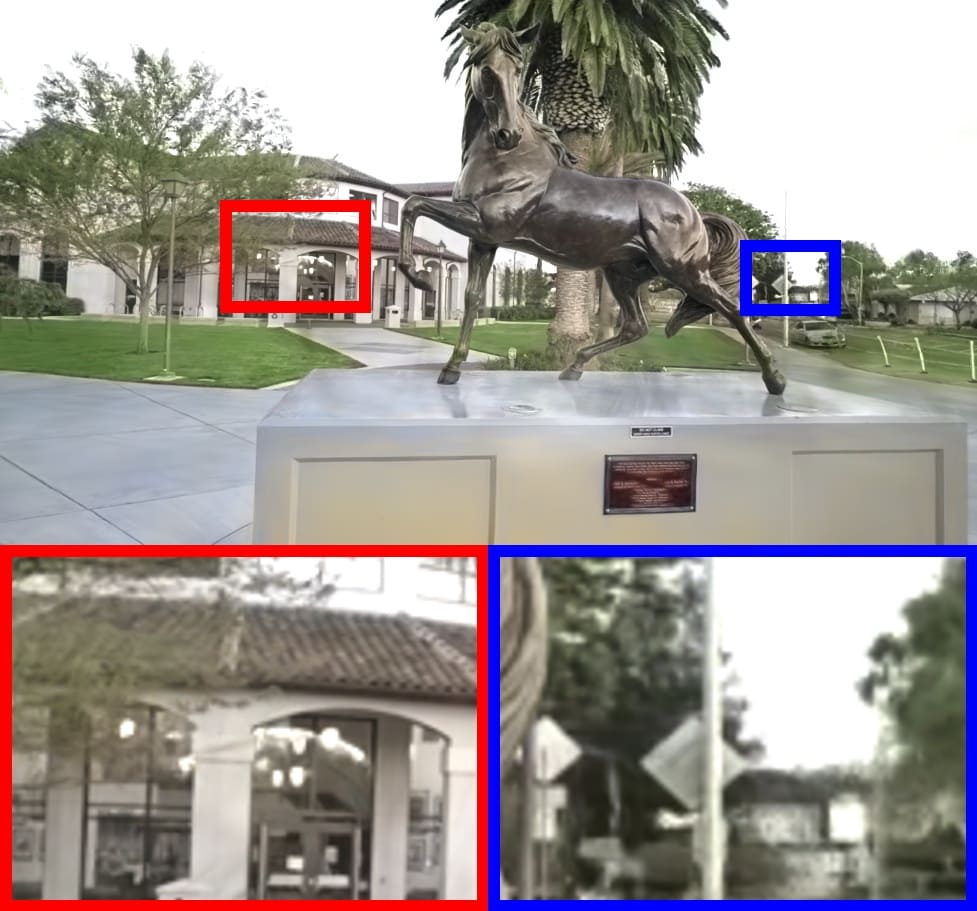} &
        \includegraphics[width=\imgw]{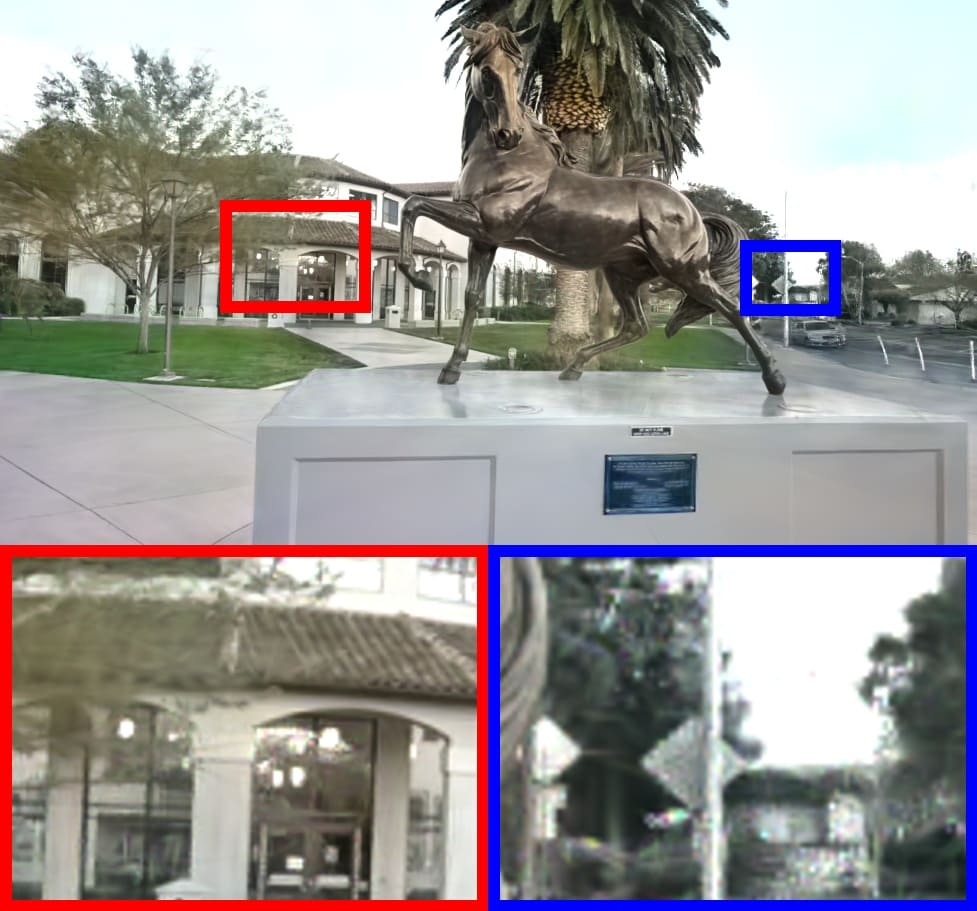} &
        \includegraphics[width=\imgw]{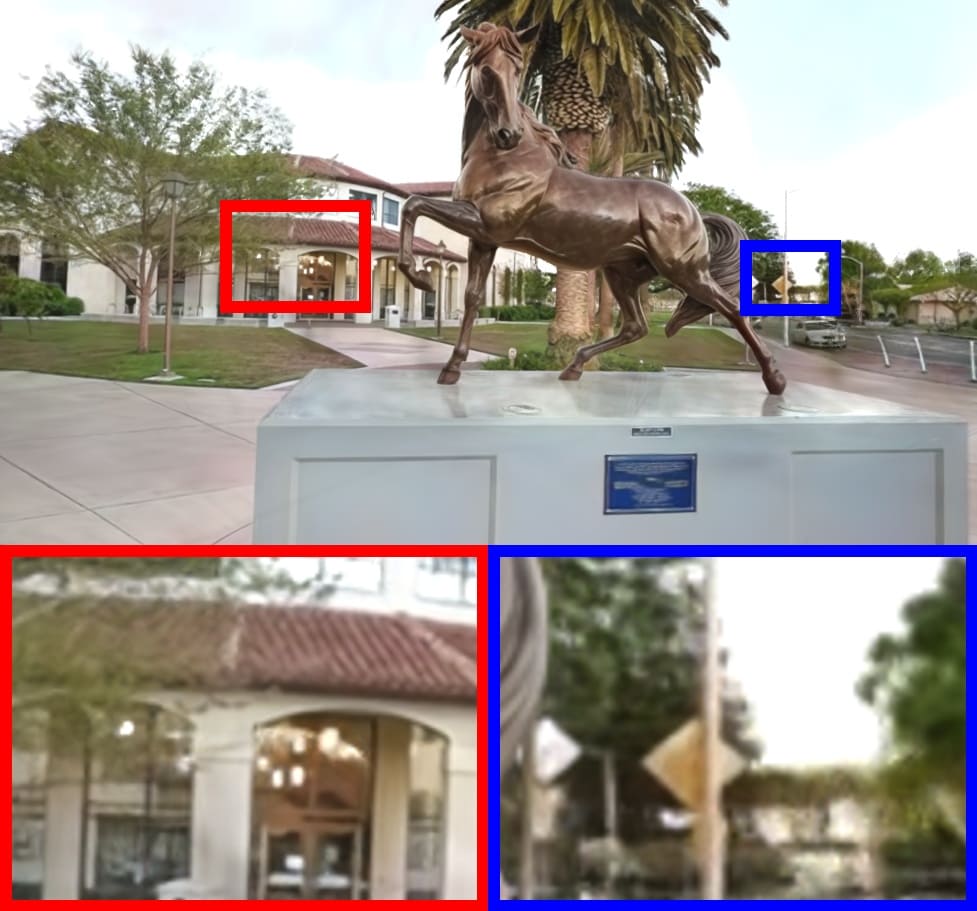} \\
        
        \raisebox{13pt}{\rotatebox{90}{360-Kitchen}} &
        \includegraphics[width=\imgw]{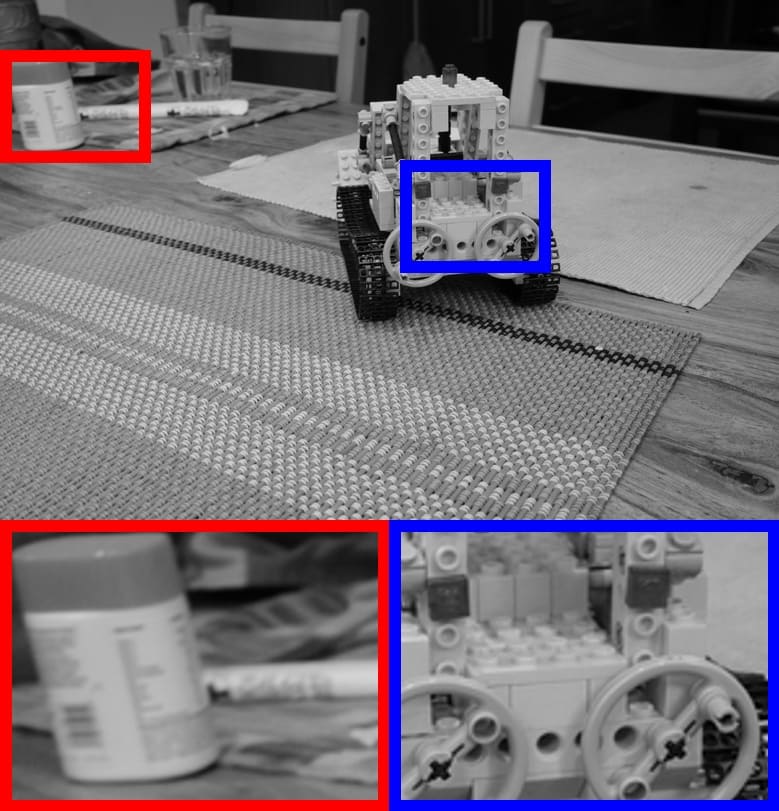} &
        \includegraphics[width=\imgw]{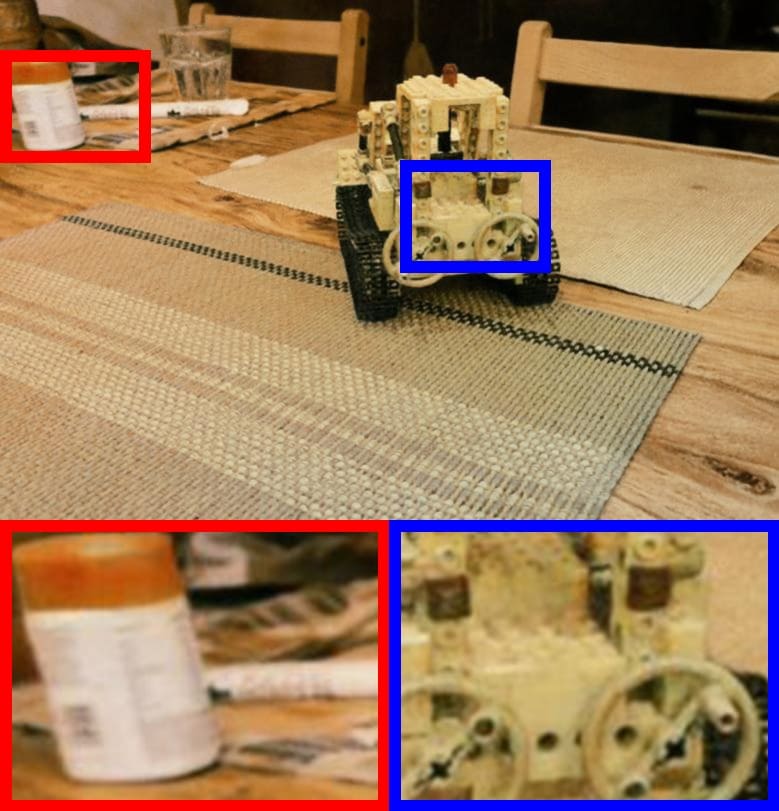} &
        \includegraphics[width=\imgw]{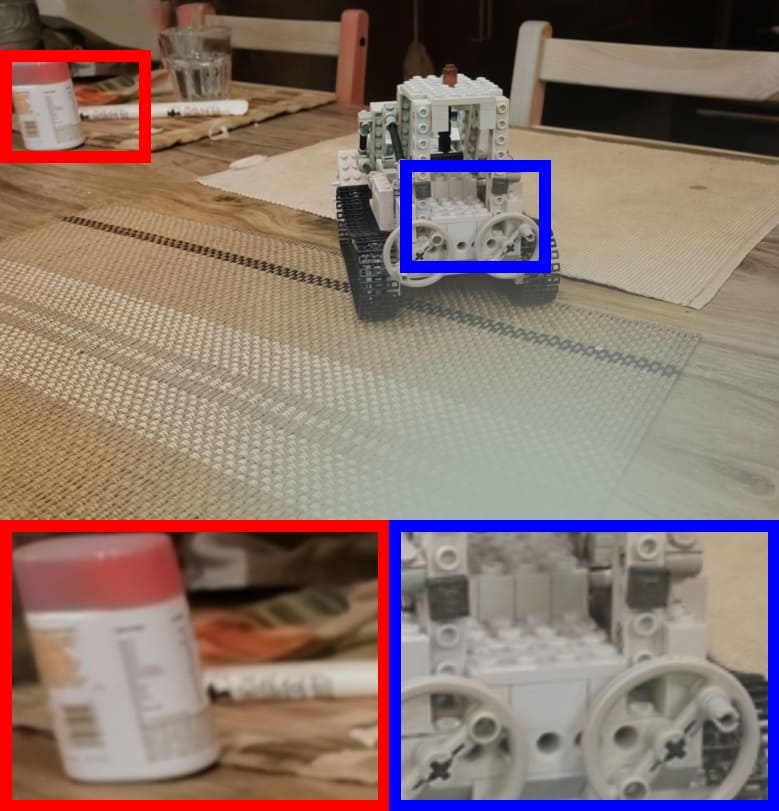} &
        \includegraphics[width=\imgw]{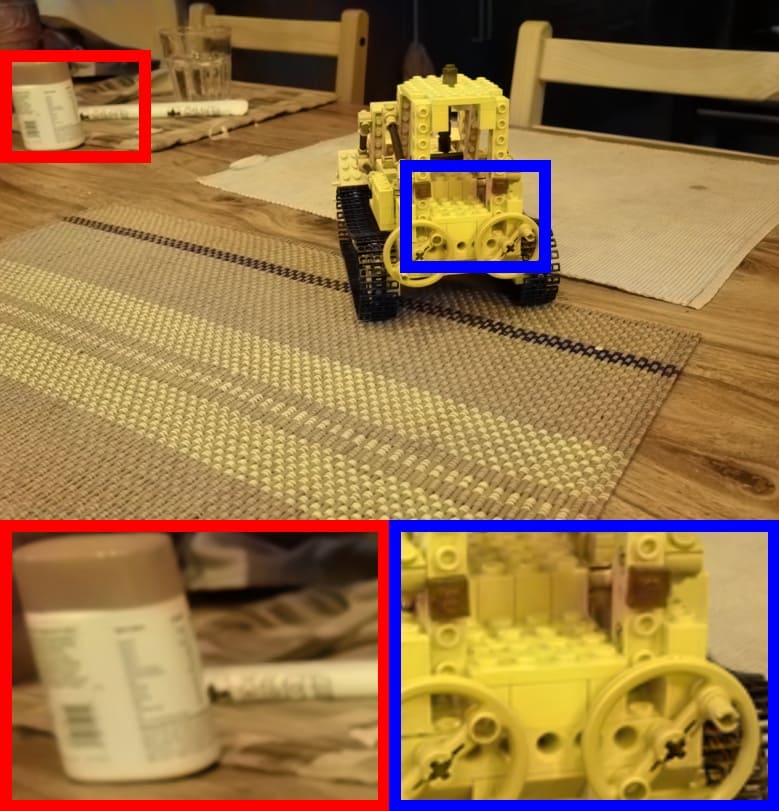} &
        \includegraphics[width=\imgw]{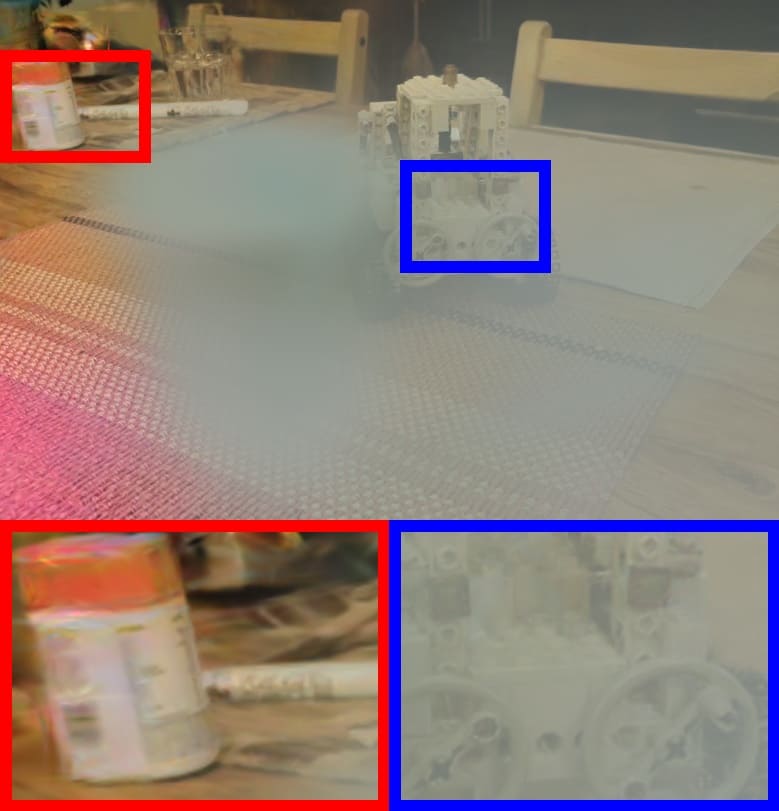} &
        \includegraphics[width=\imgw]{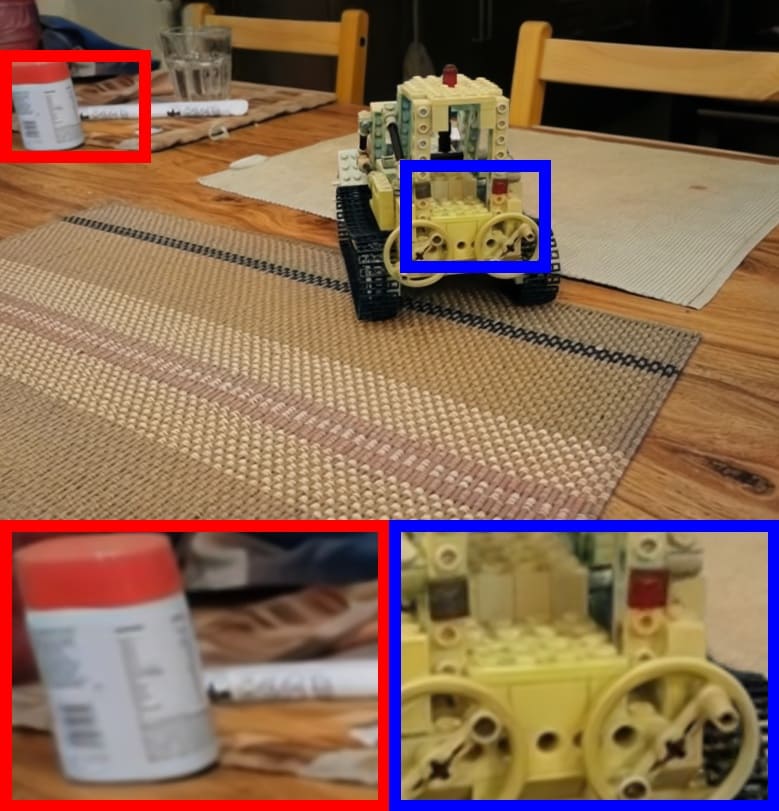} \\
        
        \raisebox{16pt}{\rotatebox{90}{360-Room}} &
        \includegraphics[width=\imgw]{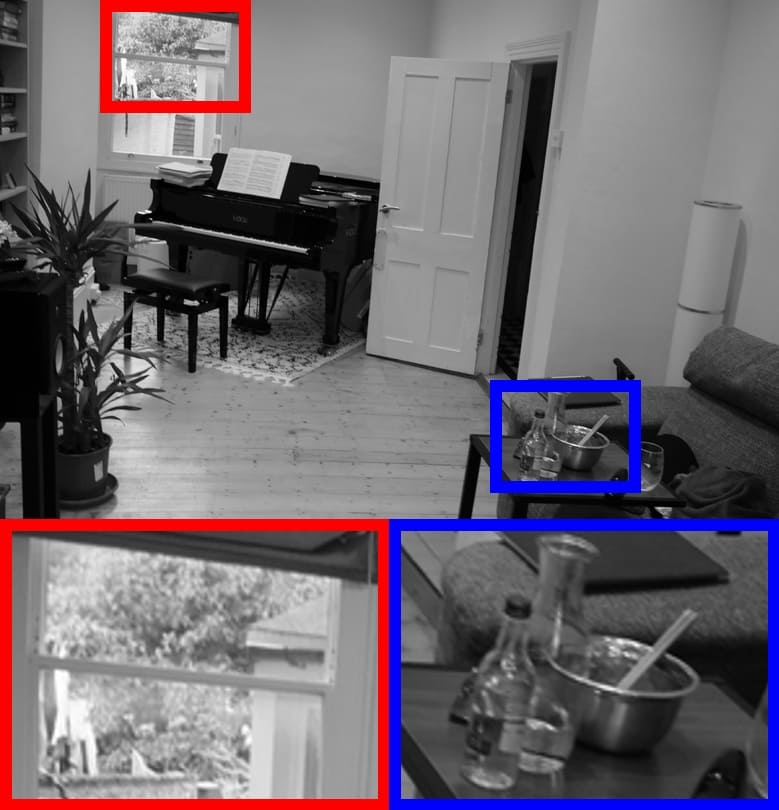} &
        \includegraphics[width=\imgw]{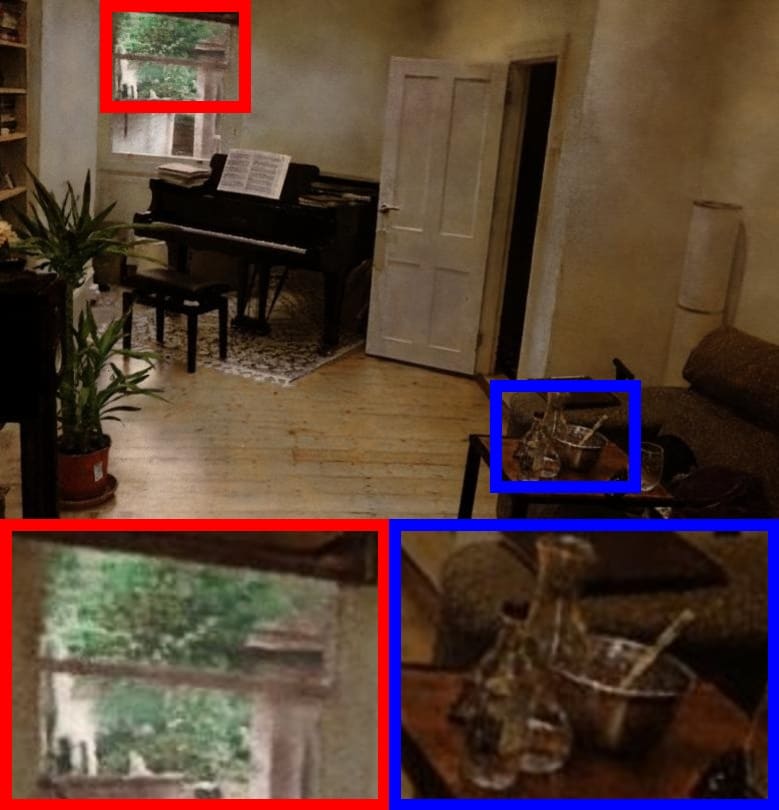} &
        \includegraphics[width=\imgw]{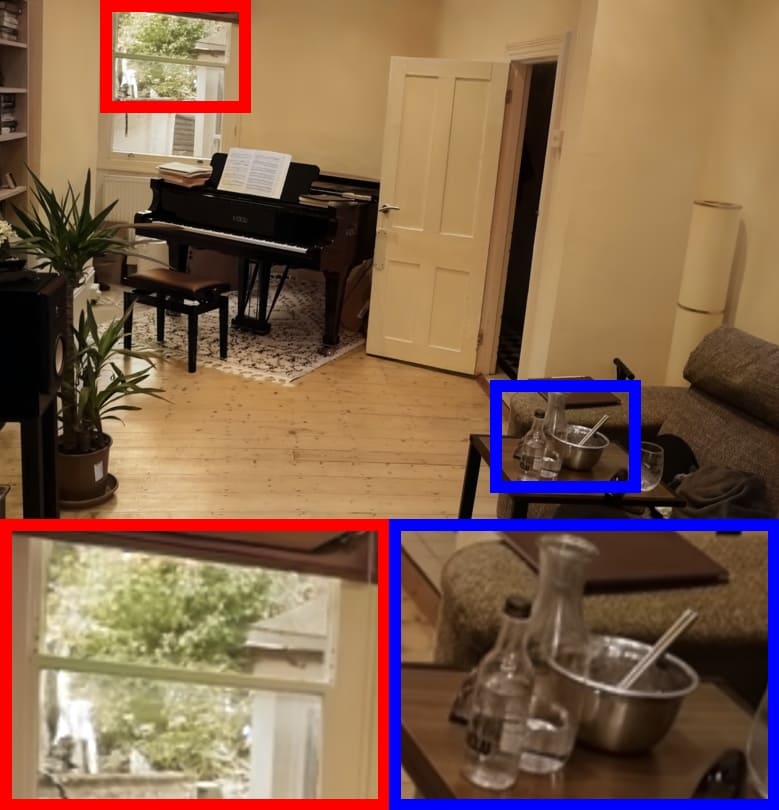} &
        \includegraphics[width=\imgw]{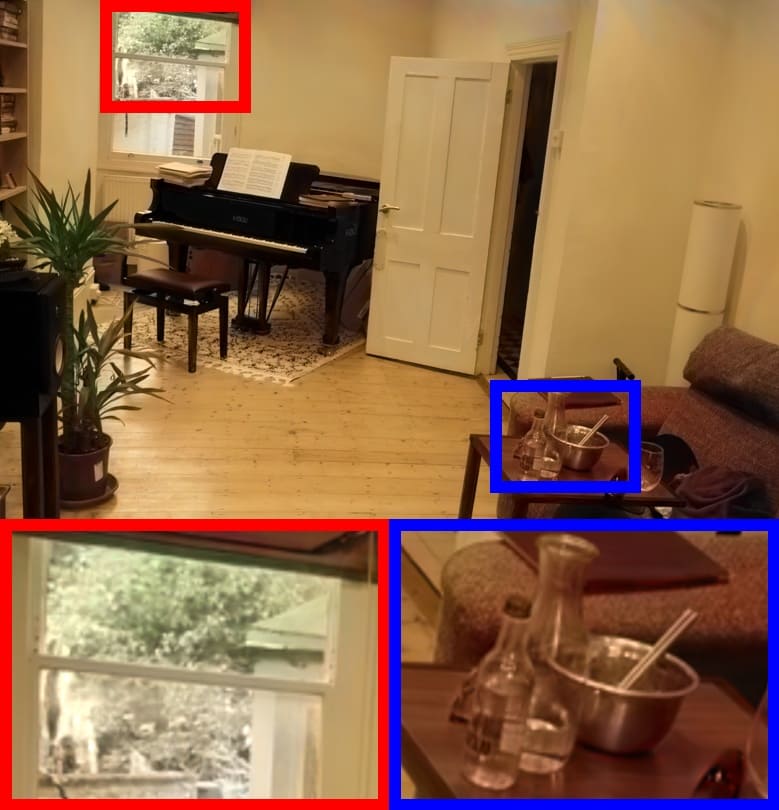} &
        \includegraphics[width=\imgw]{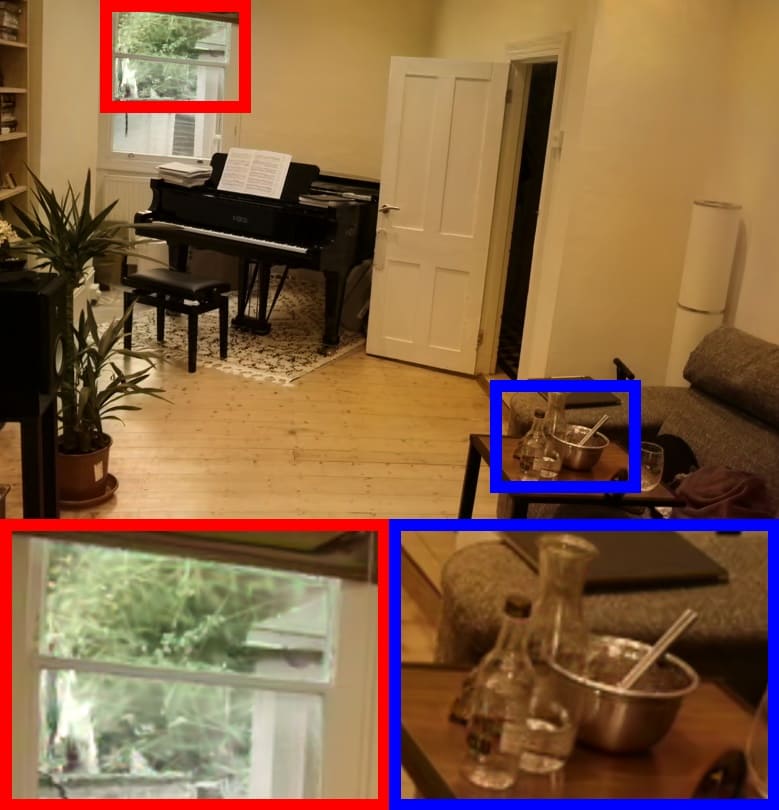} &
        \includegraphics[width=\imgw]{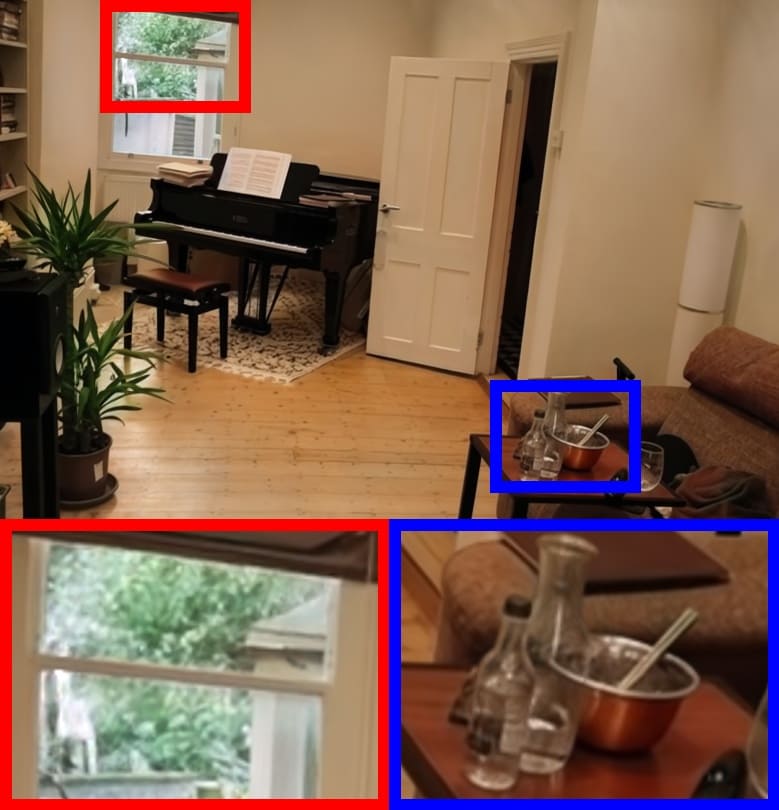} \\

        \raisebox{2pt}{\rotatebox{90}{DL3DV-Scene3}} &
        \includegraphics[width=\imgw]{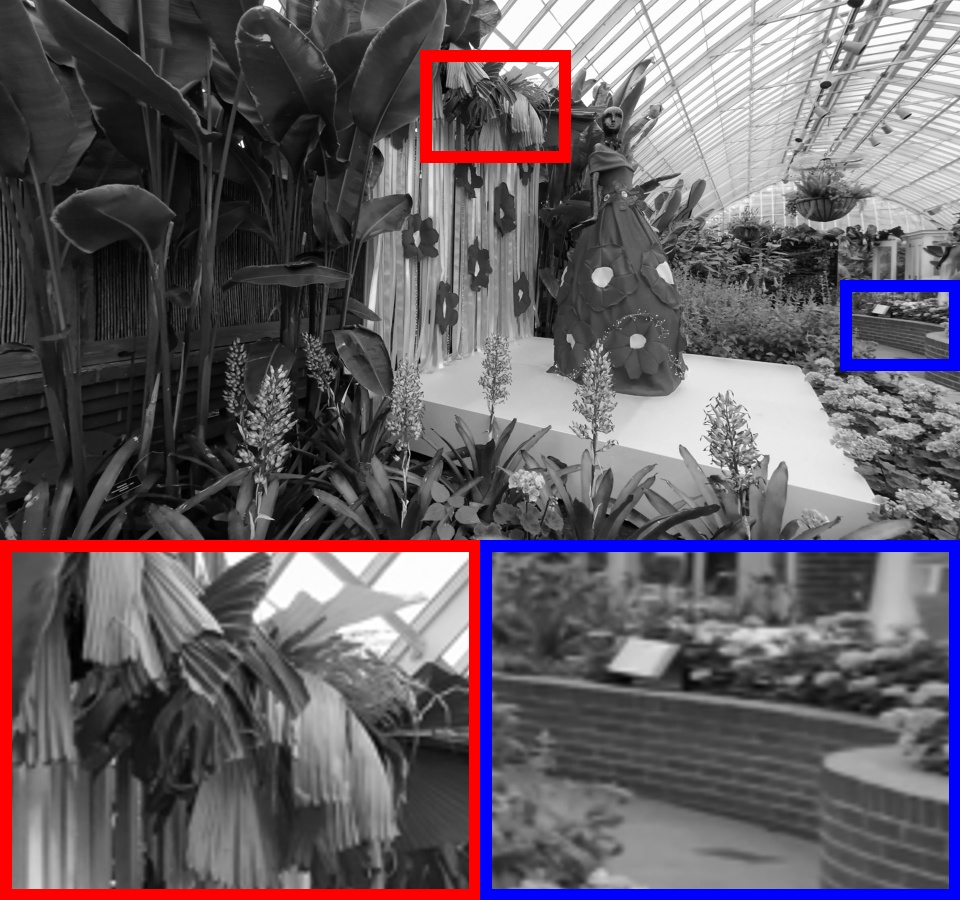} &
        \includegraphics[width=\imgw]{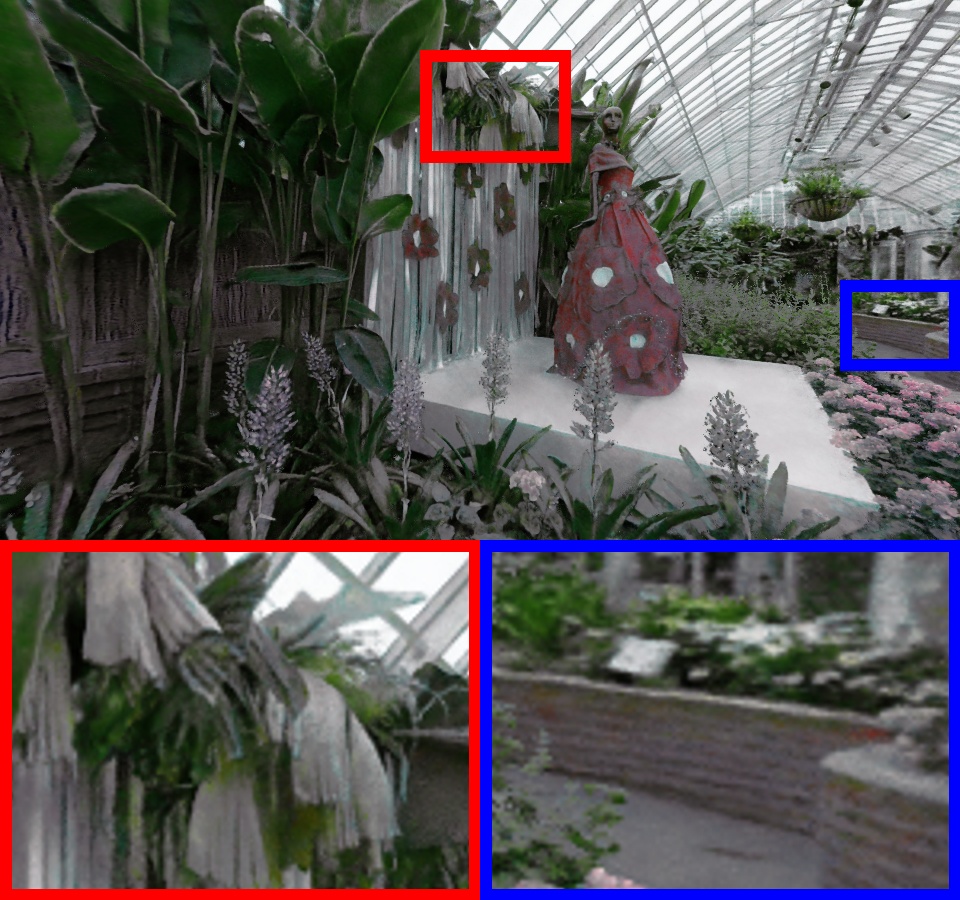} &
        \includegraphics[width=\imgw]{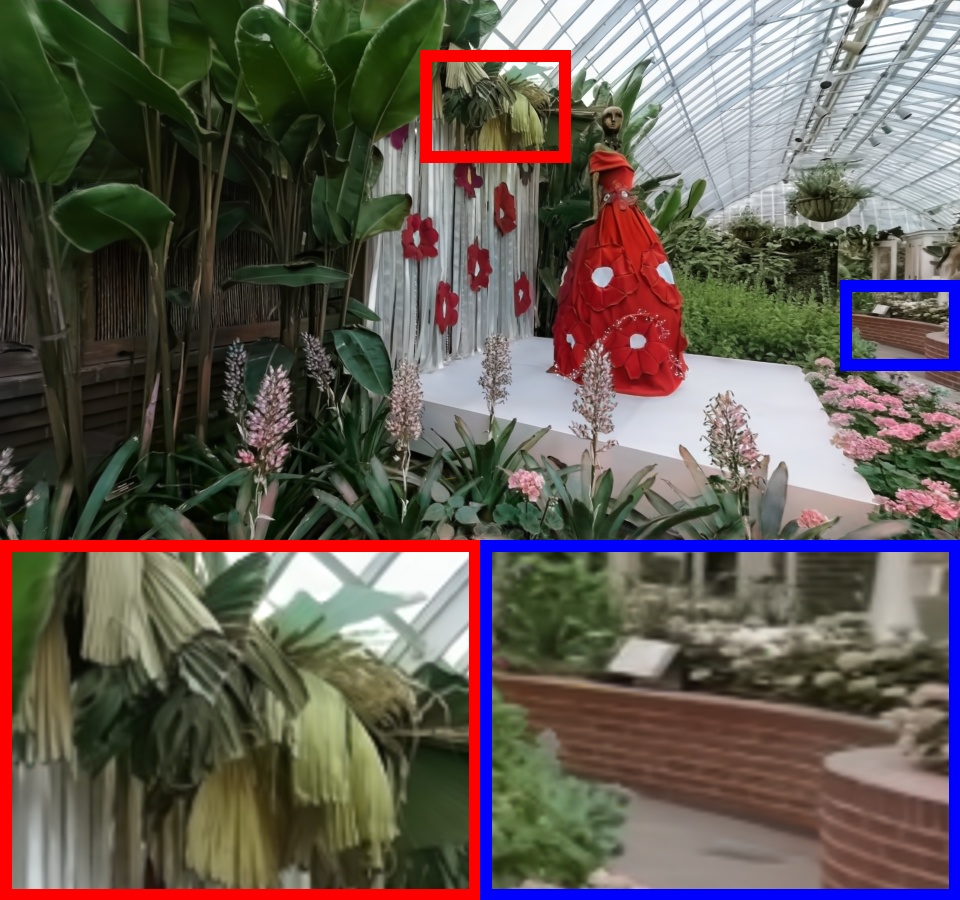} &
        \includegraphics[width=\imgw]{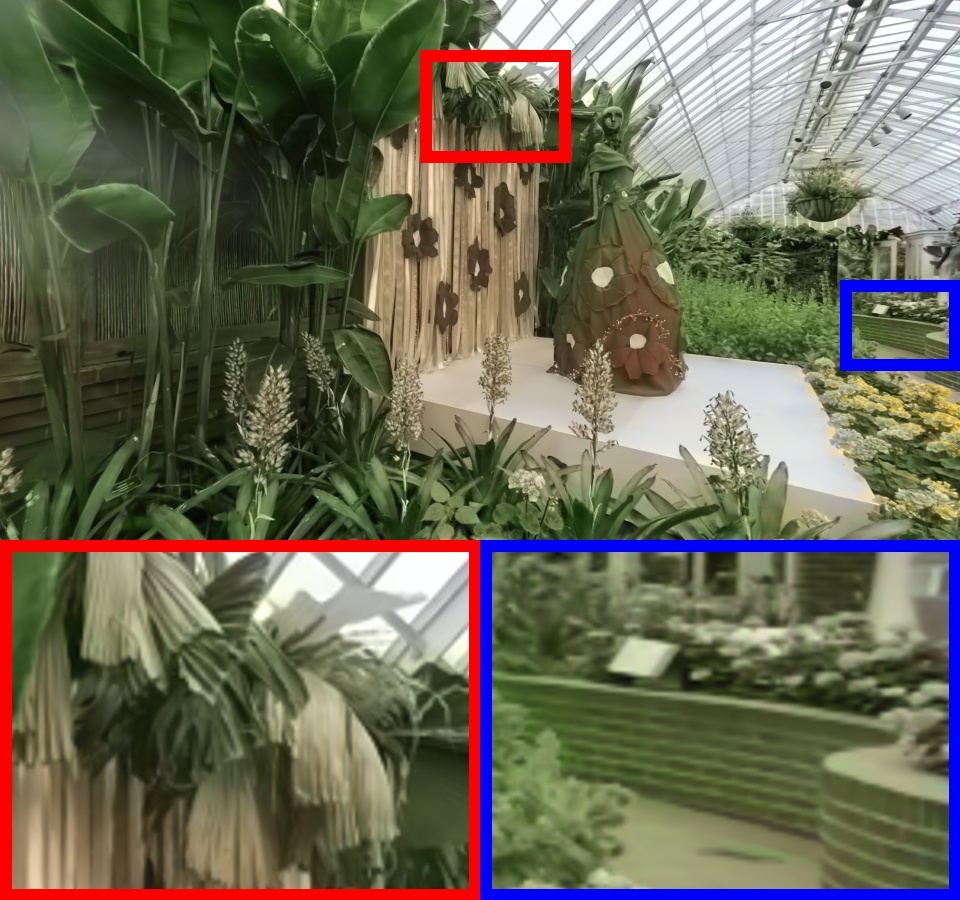} &
        \includegraphics[width=\imgw]{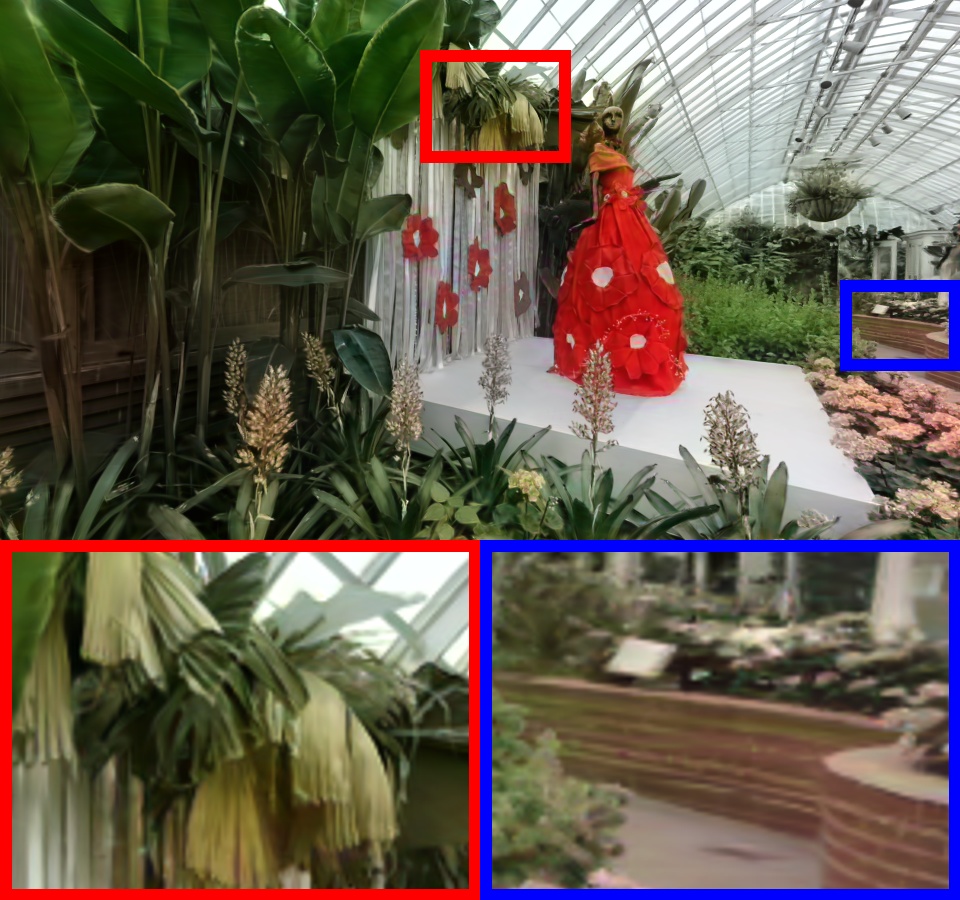} &
        \includegraphics[width=\imgw]{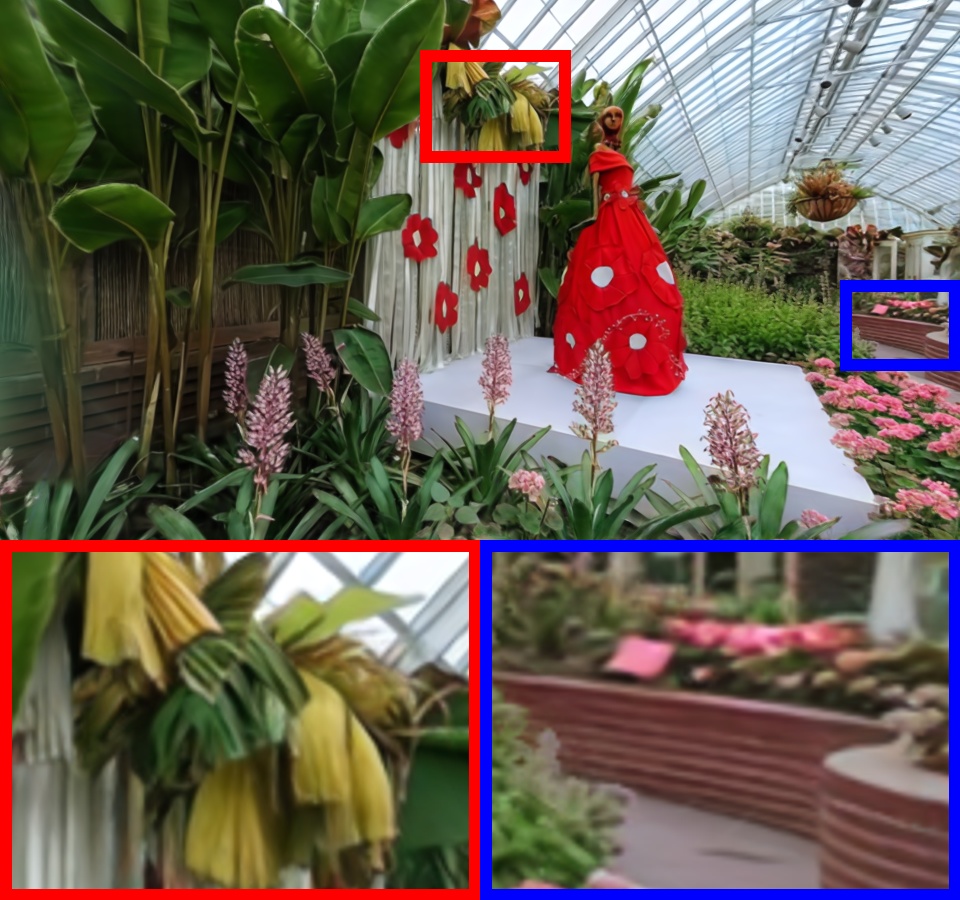} \\
        
        \raisebox{2pt}{\rotatebox{90}{DL3DV-Scene4}} &
        \includegraphics[width=\imgw]{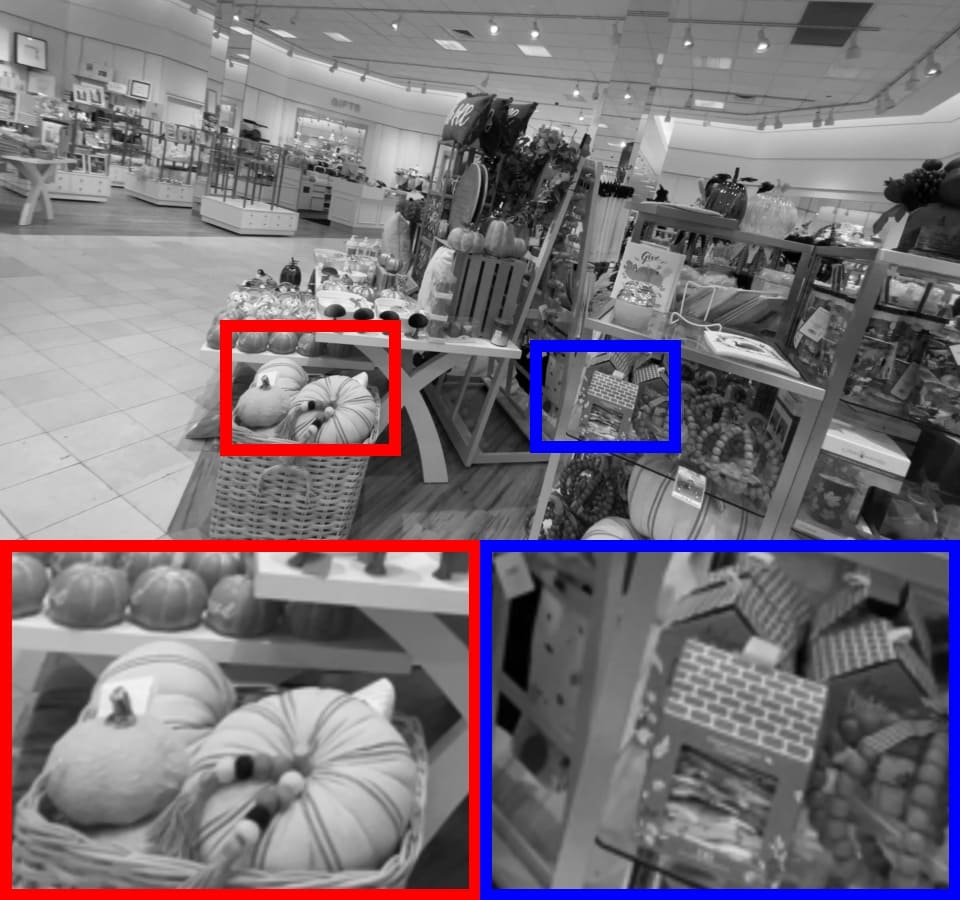} &
        \includegraphics[width=\imgw]{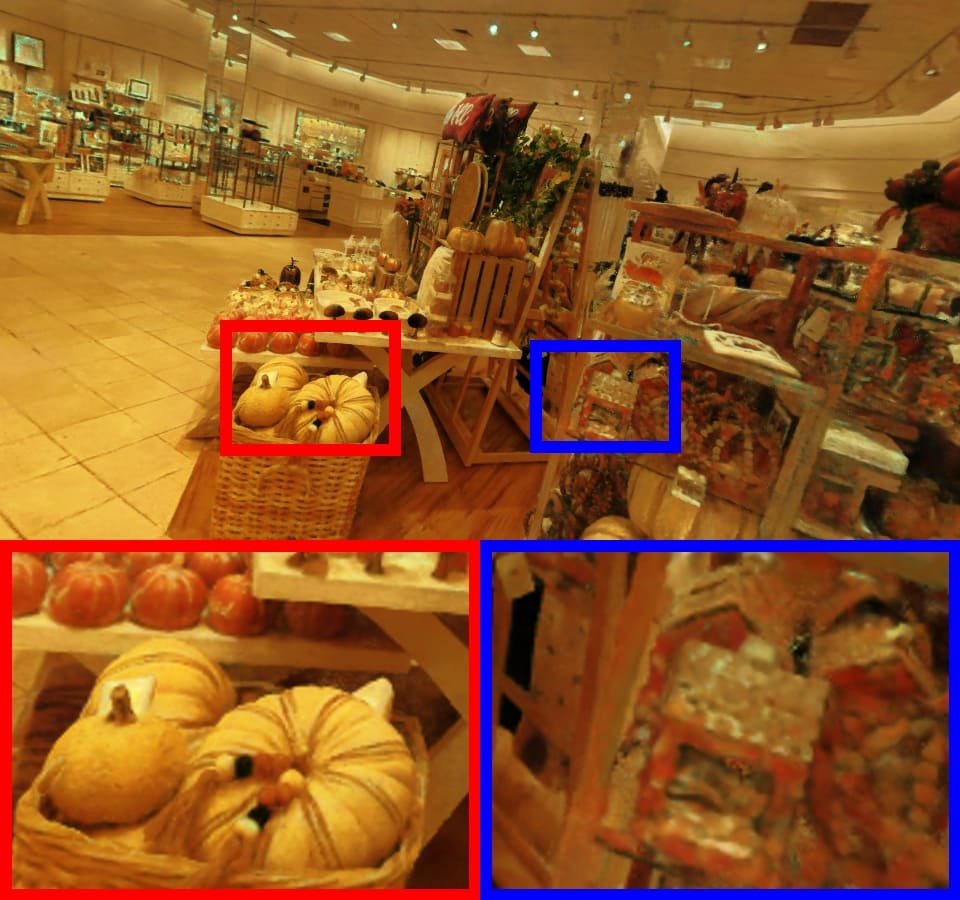} &
        \includegraphics[width=\imgw]{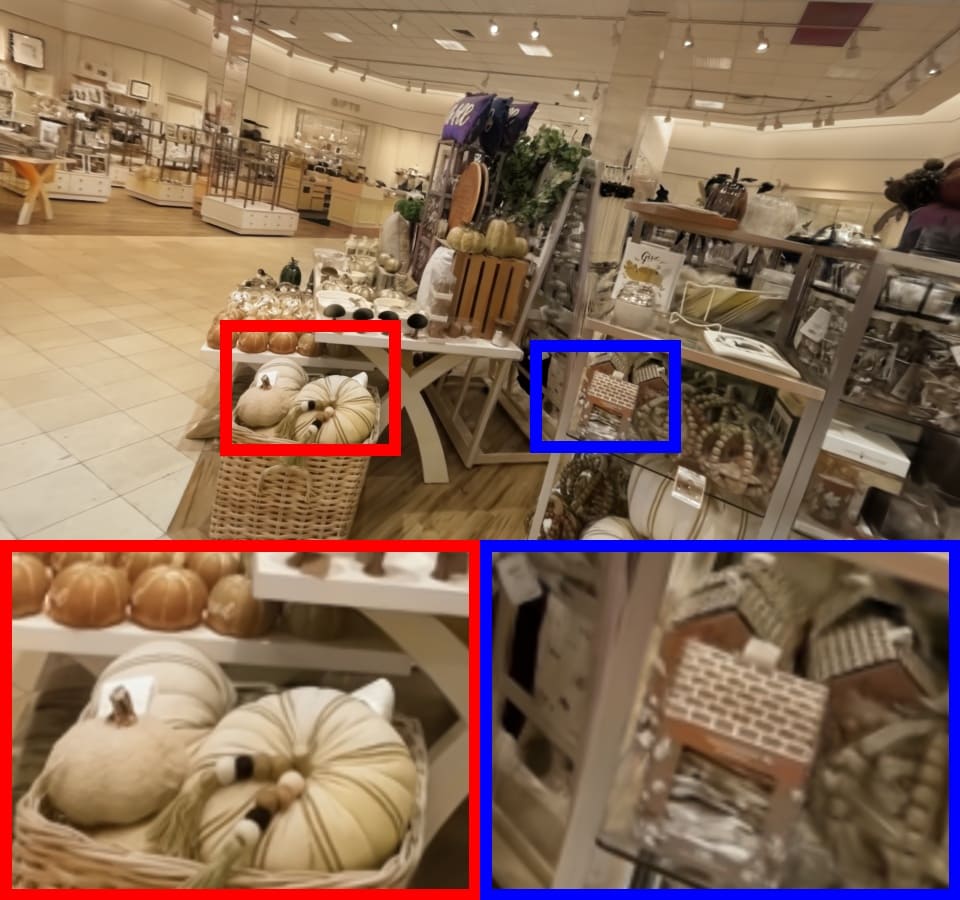} &
        \includegraphics[width=\imgw]{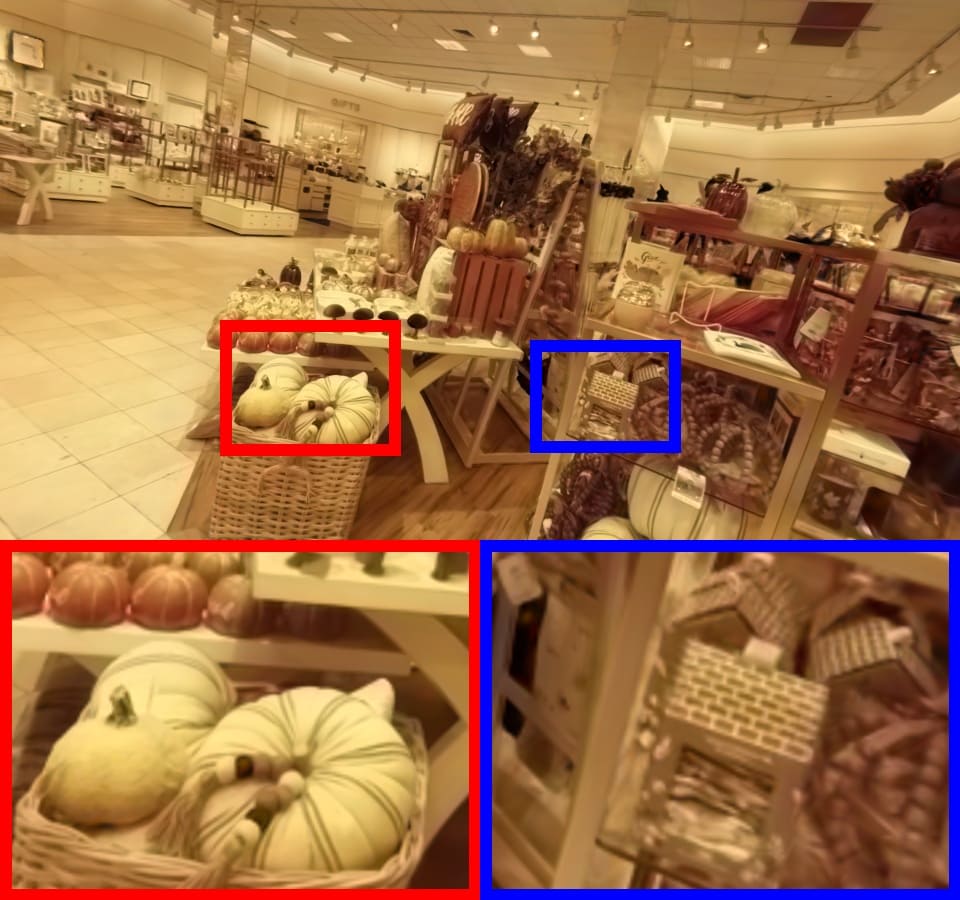} &
        \includegraphics[width=\imgw]{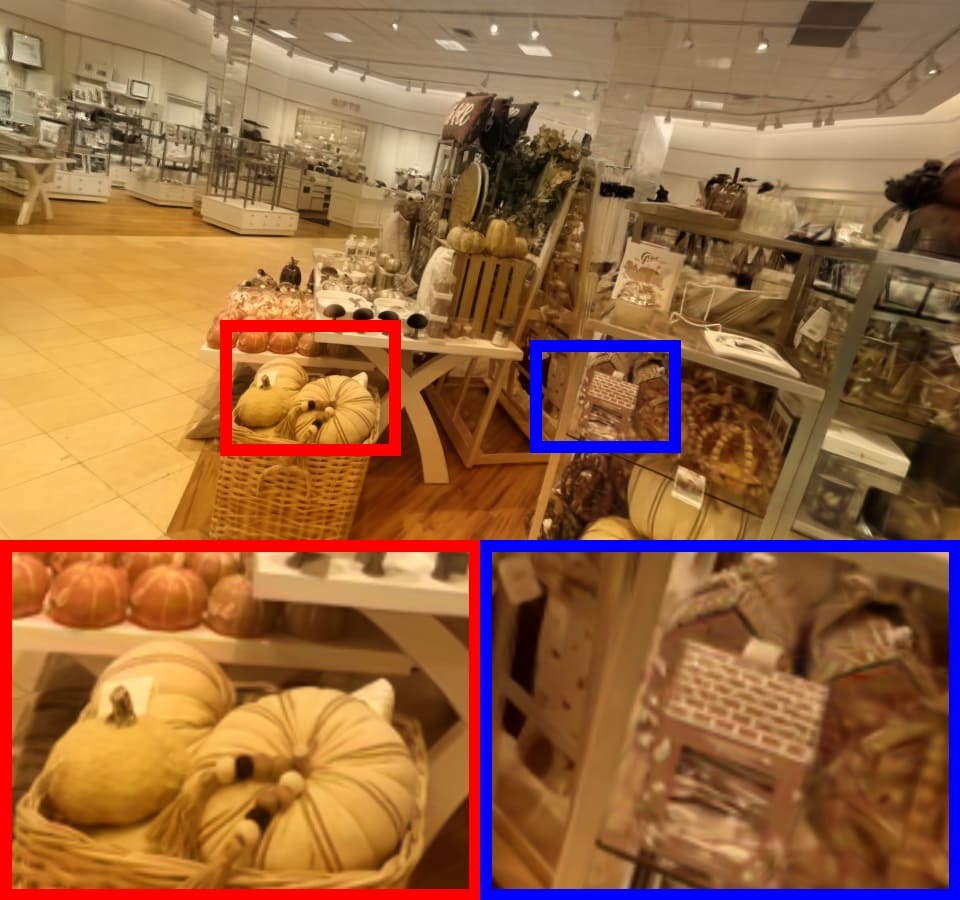} &
        \includegraphics[width=\imgw]{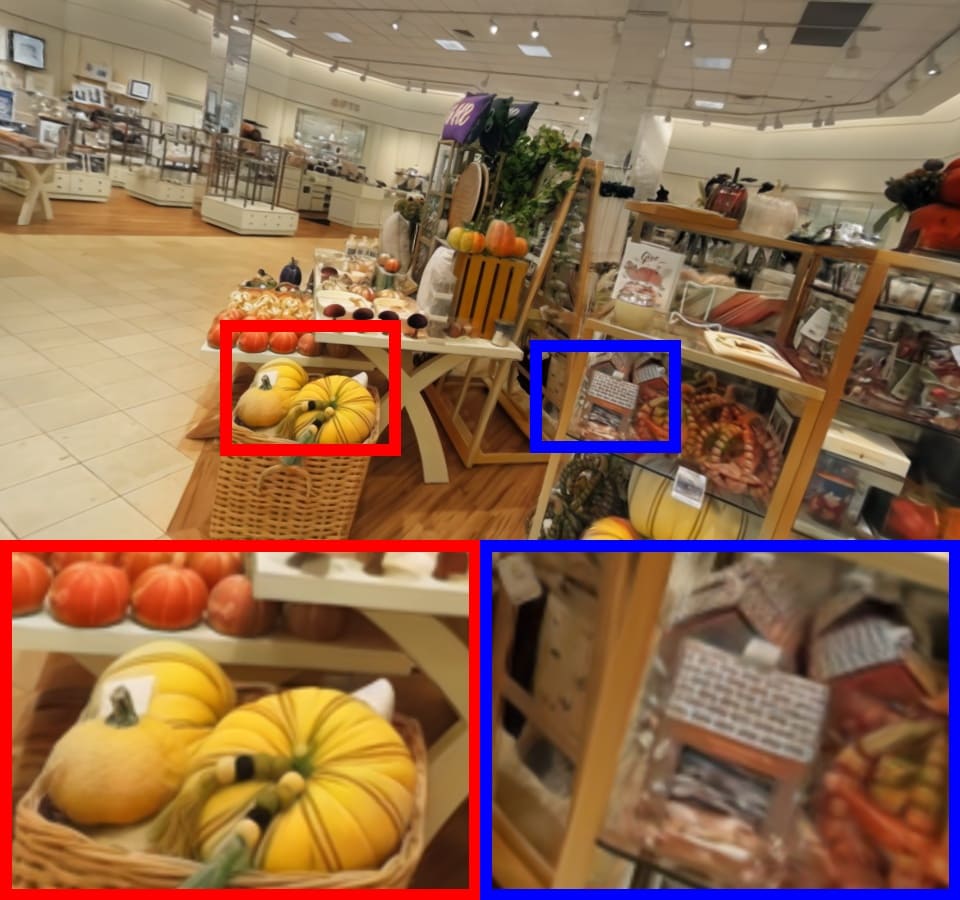} \\

        \raisebox{2pt}{\rotatebox{90}{DL3DV-Scene5}} &
        \includegraphics[width=\imgw]{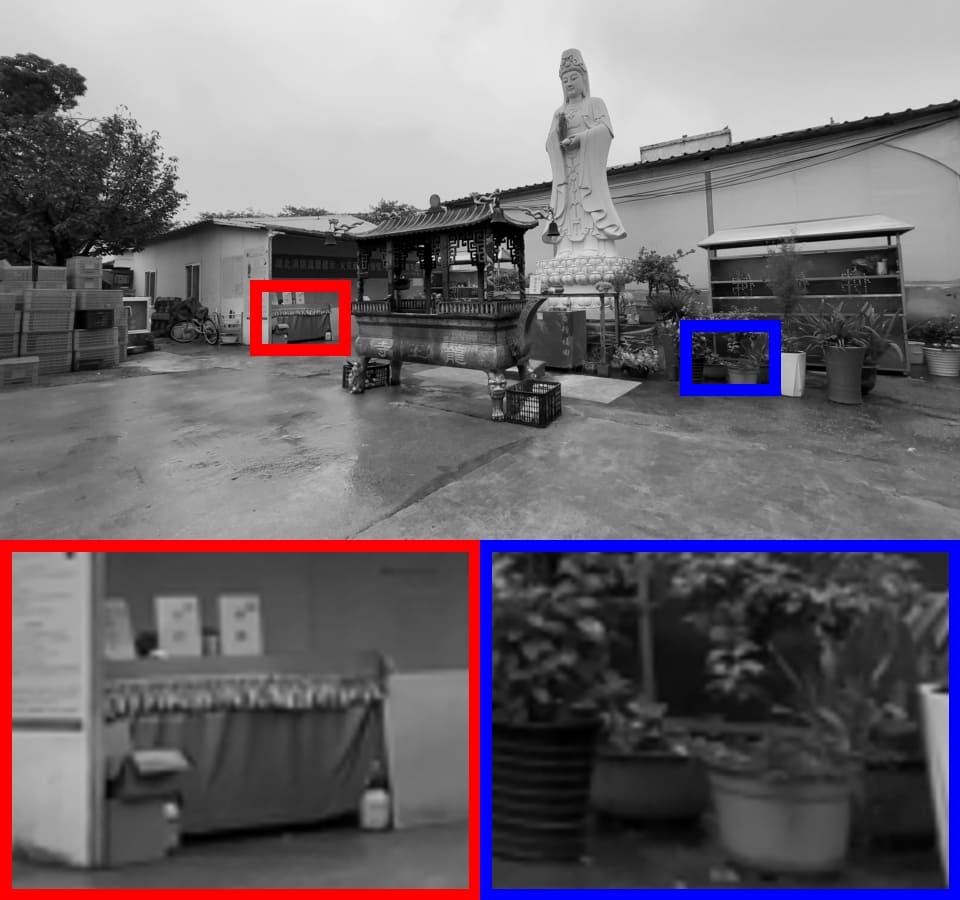} &
        \includegraphics[width=\imgw]{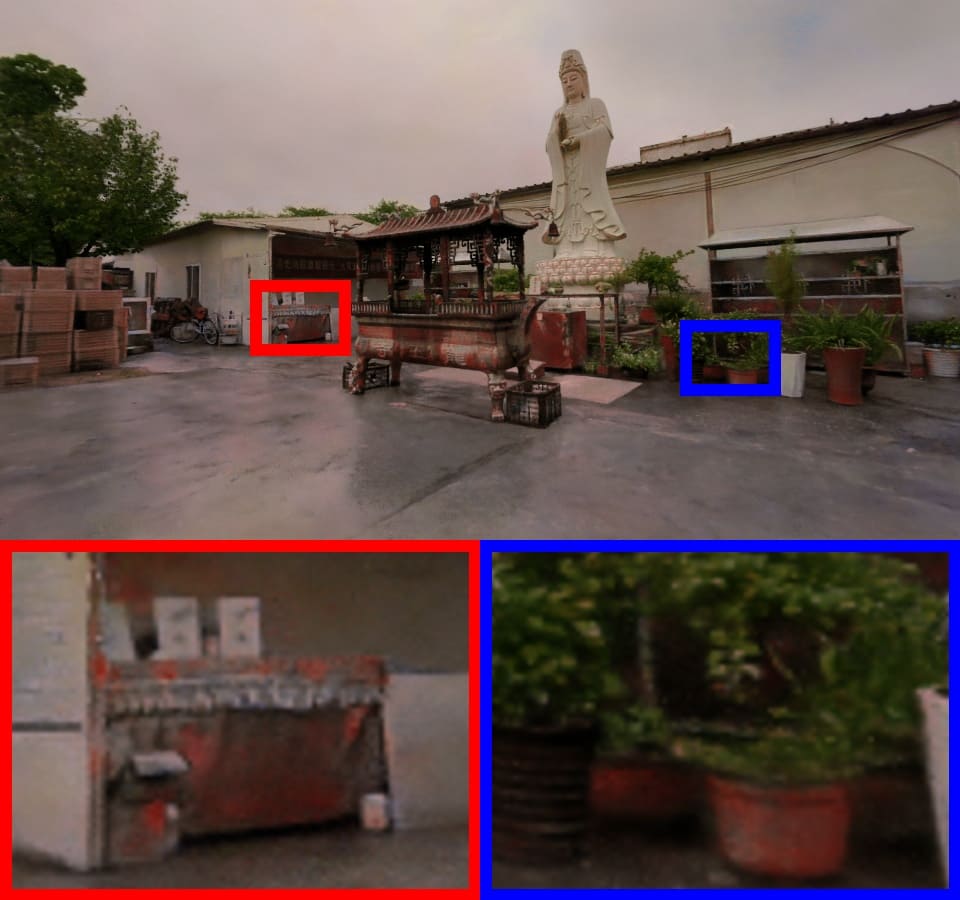} &
        \includegraphics[width=\imgw]{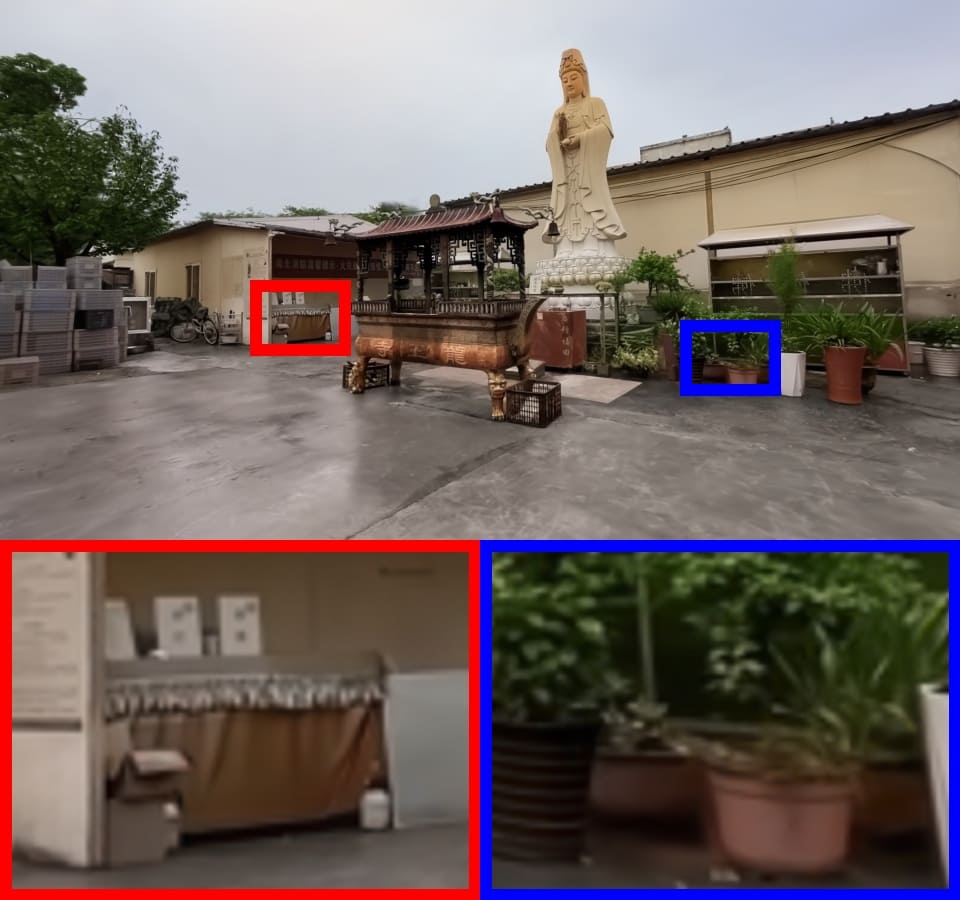} &
        \includegraphics[width=\imgw]{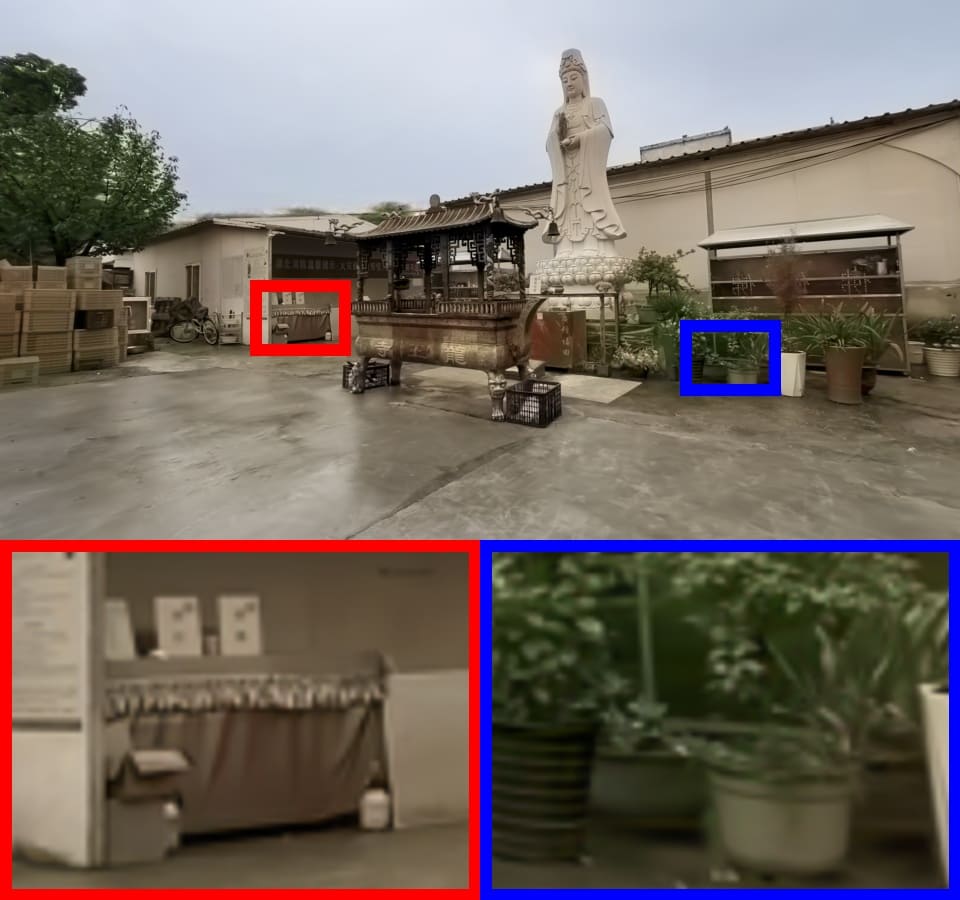} &
        \includegraphics[width=\imgw]{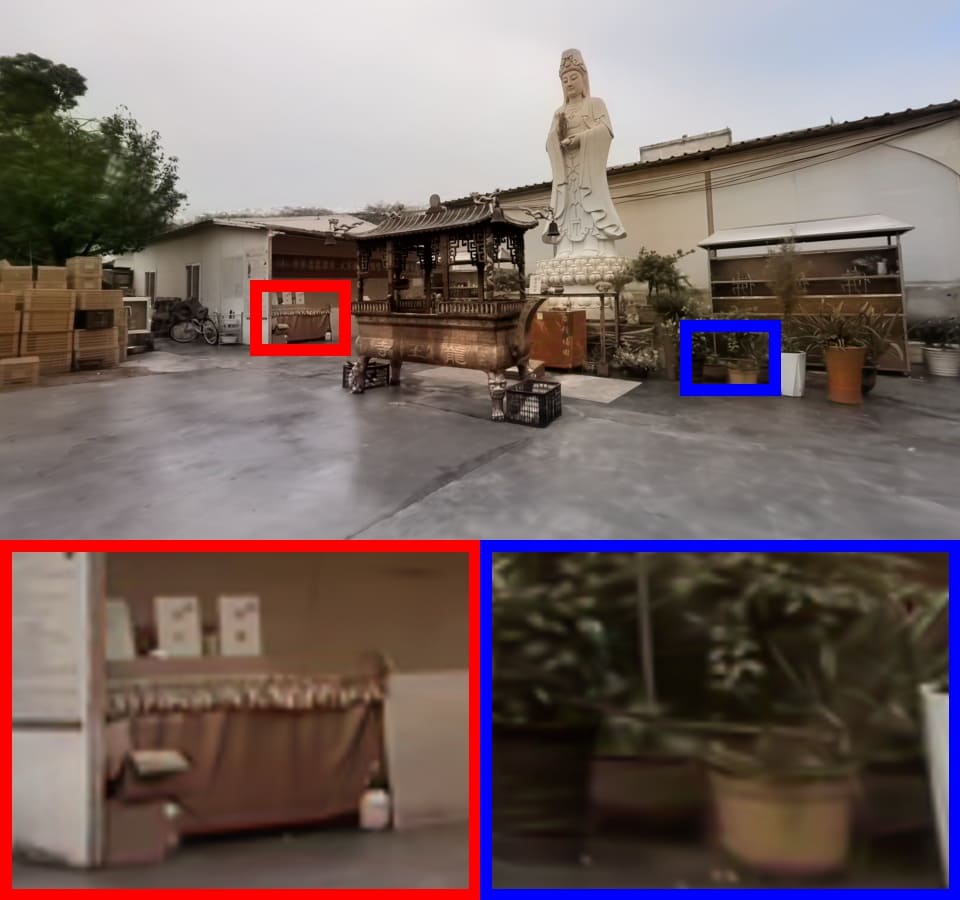} &
        \includegraphics[width=\imgw]{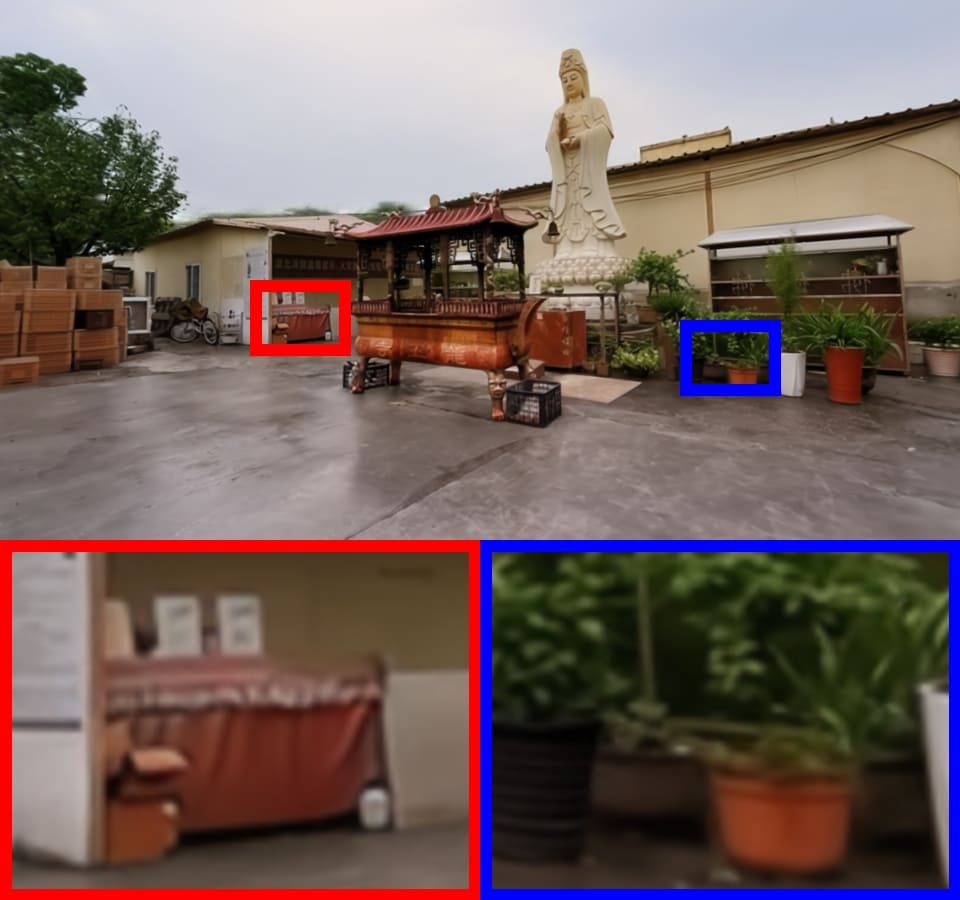} \\
        
        \raisebox{2pt}{\rotatebox{90}{DL3DV-Scene6}} &
        \includegraphics[width=\imgw]{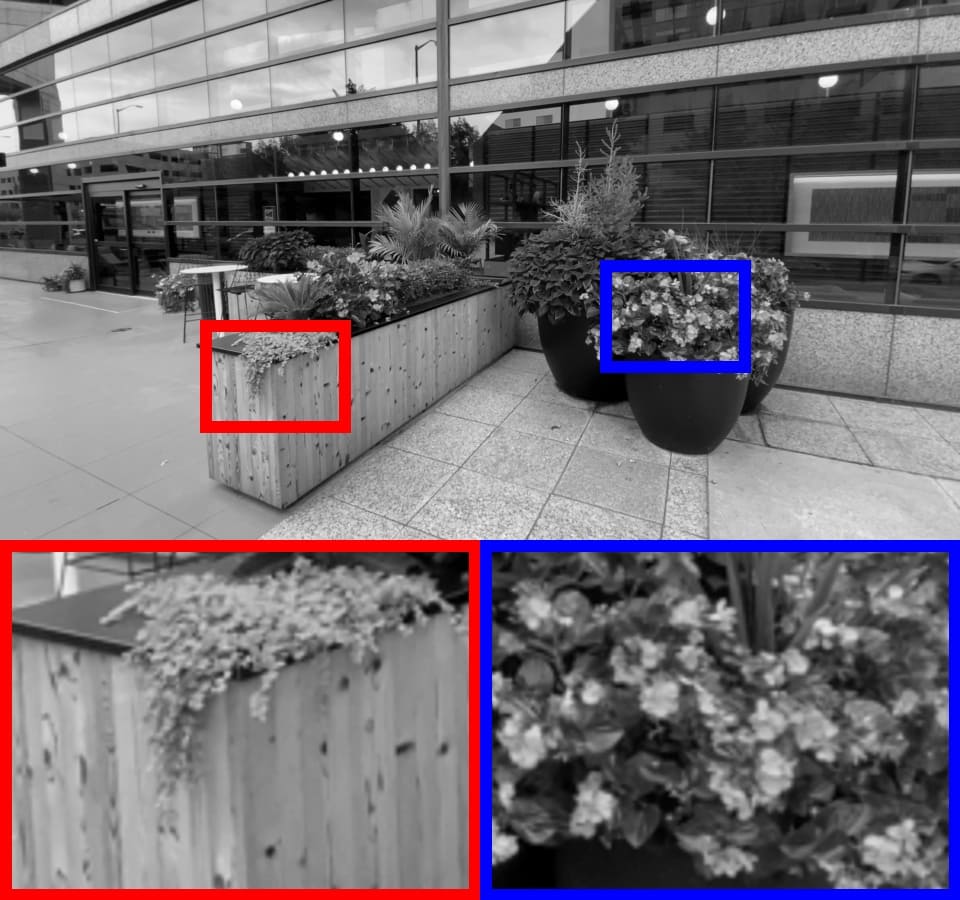} &
        \includegraphics[width=\imgw]{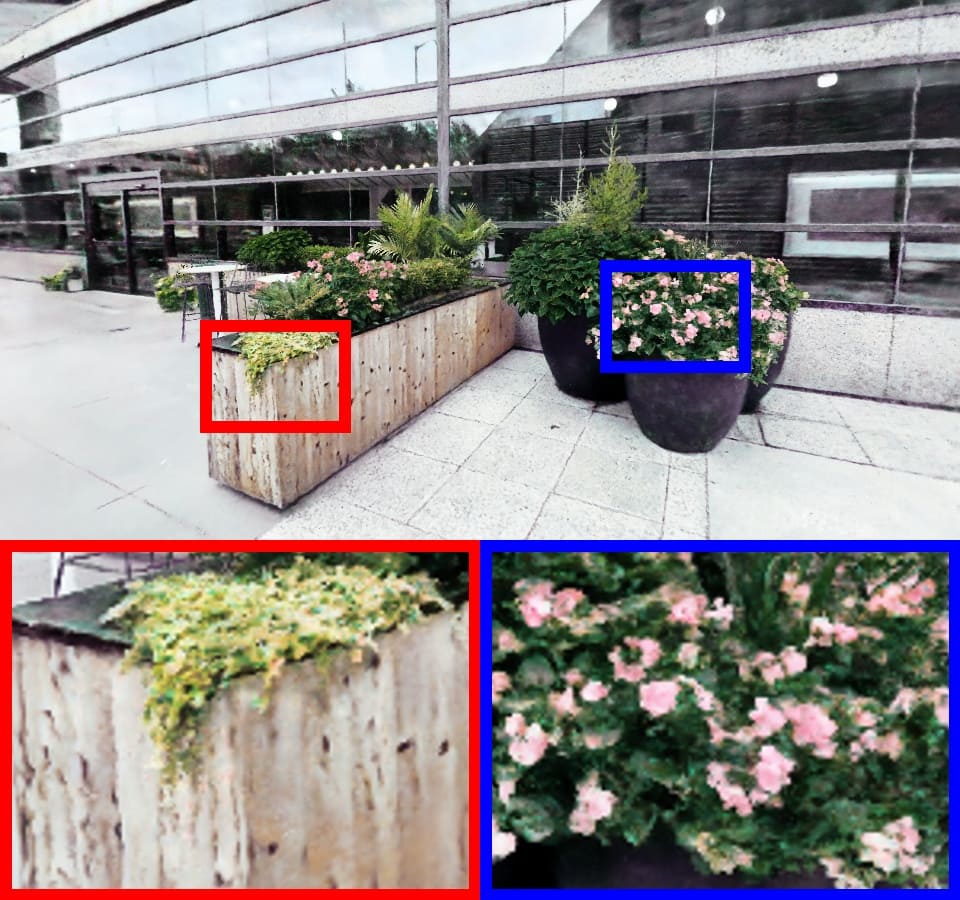} &
        \includegraphics[width=\imgw]{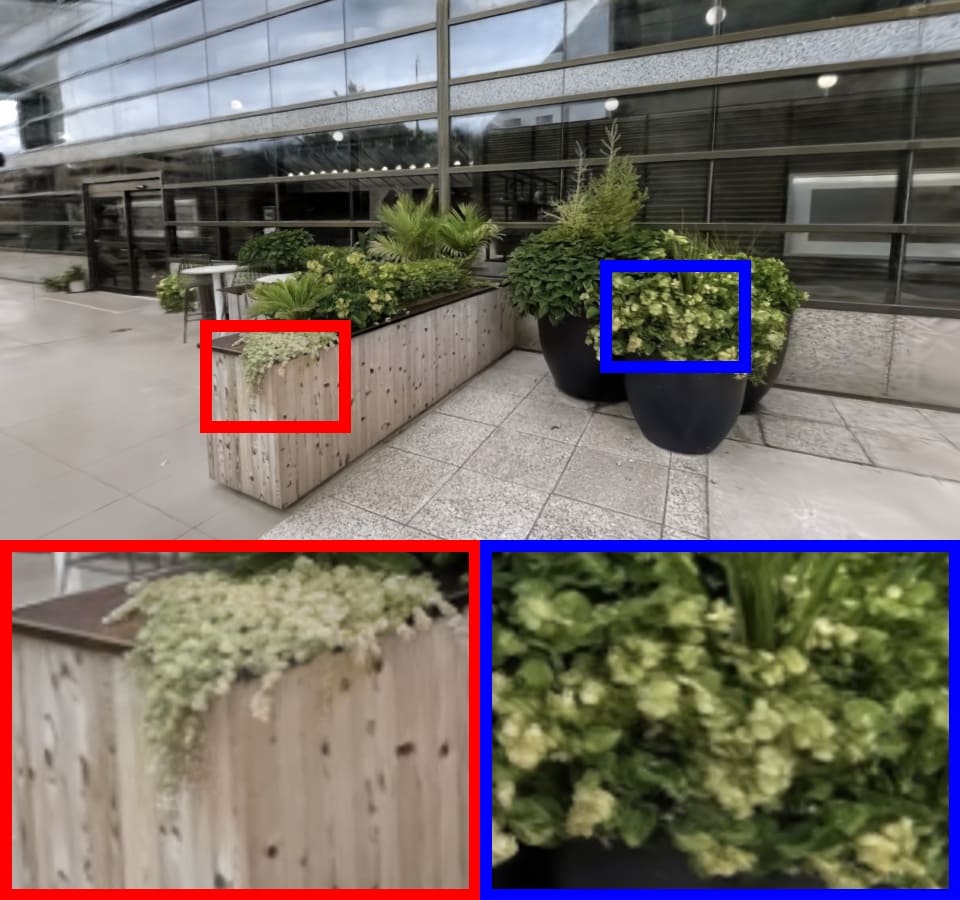} &
        \includegraphics[width=\imgw]{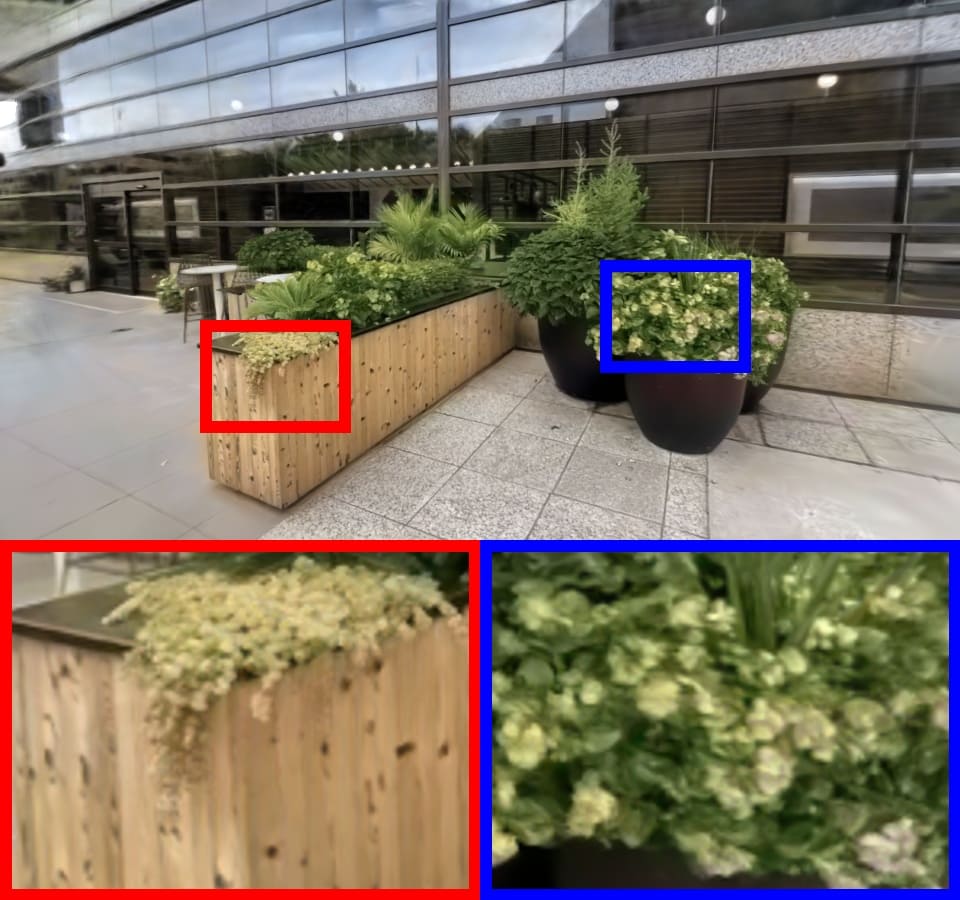} &
        \includegraphics[width=\imgw]{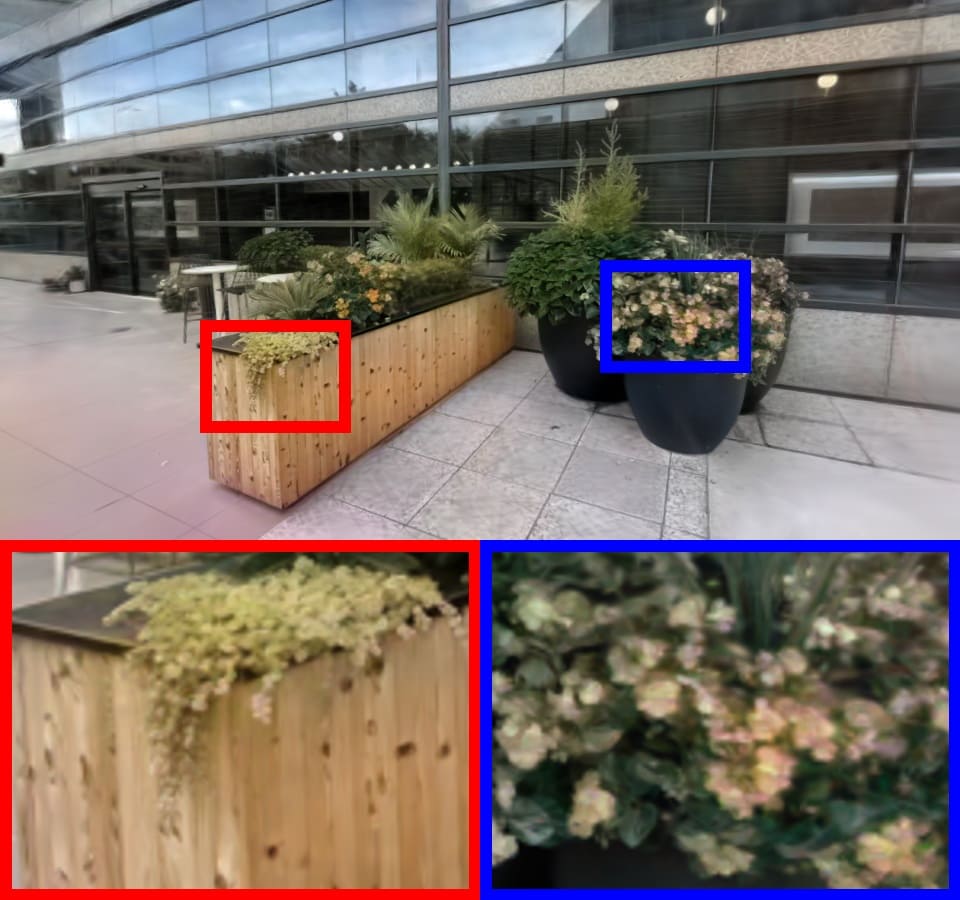} &
        \includegraphics[width=\imgw]{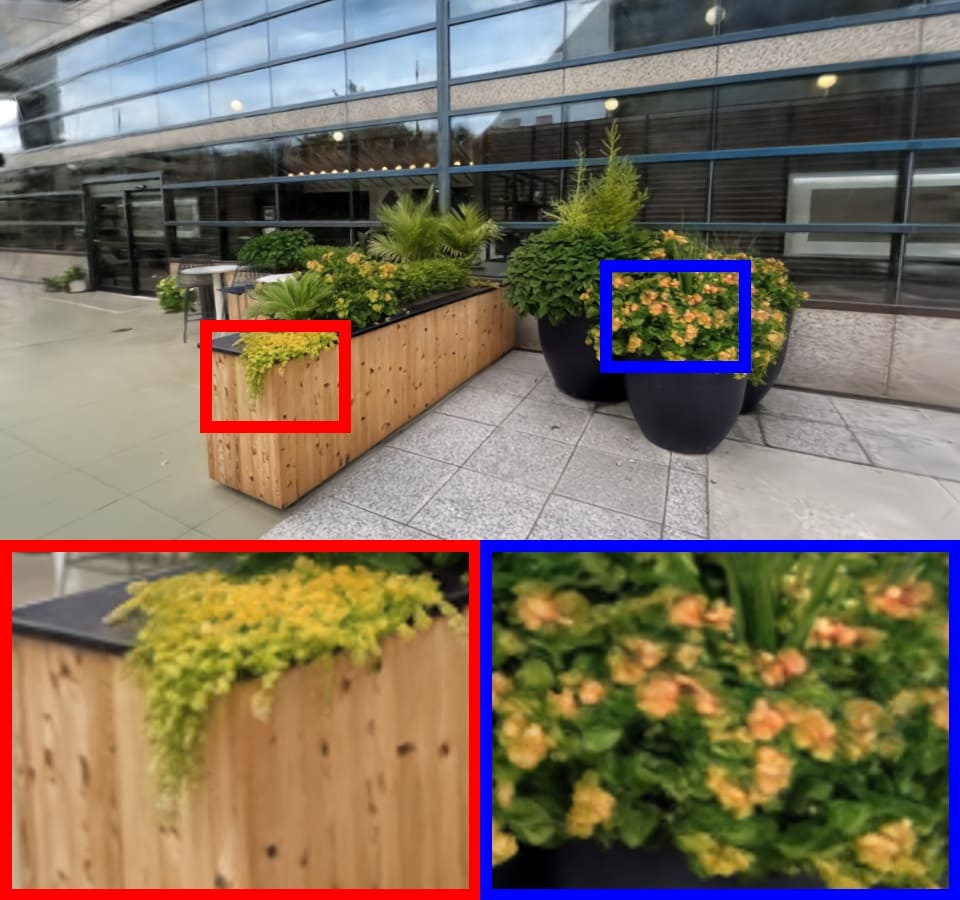} \\   
        
        \raisebox{2pt}{\rotatebox{90}{DL3DV-Scene7}} &
        \includegraphics[width=\imgw]{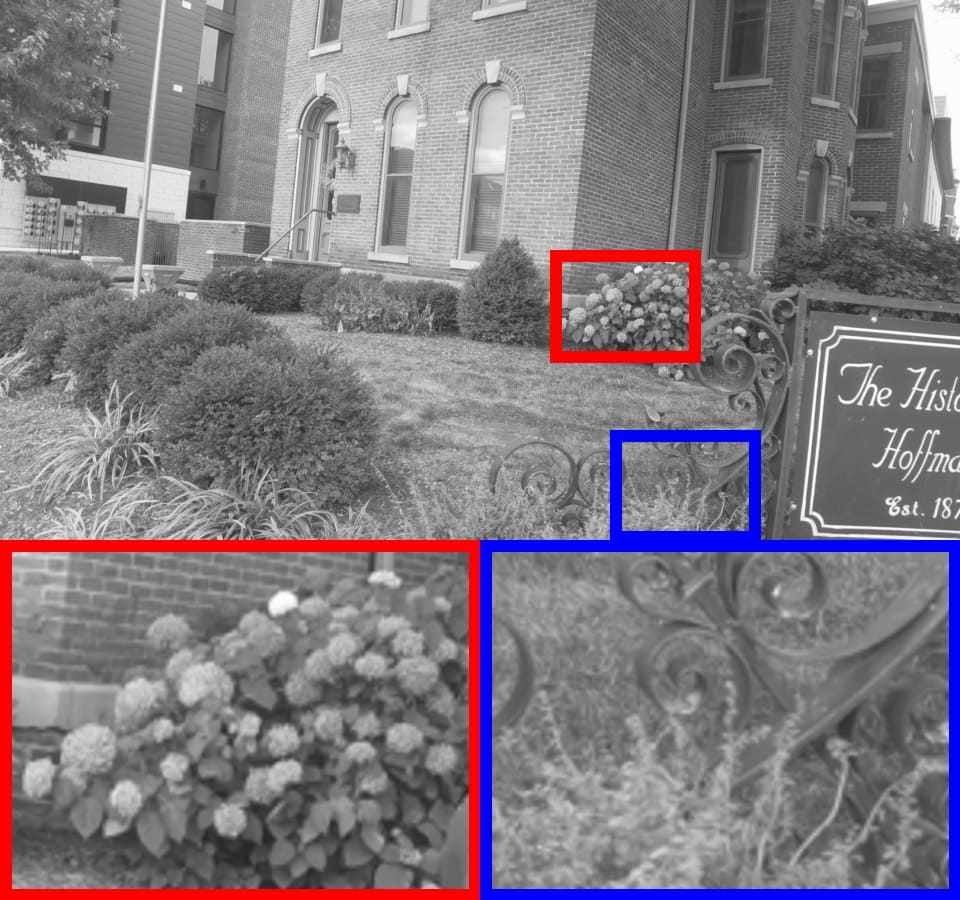} &
        \includegraphics[width=\imgw]{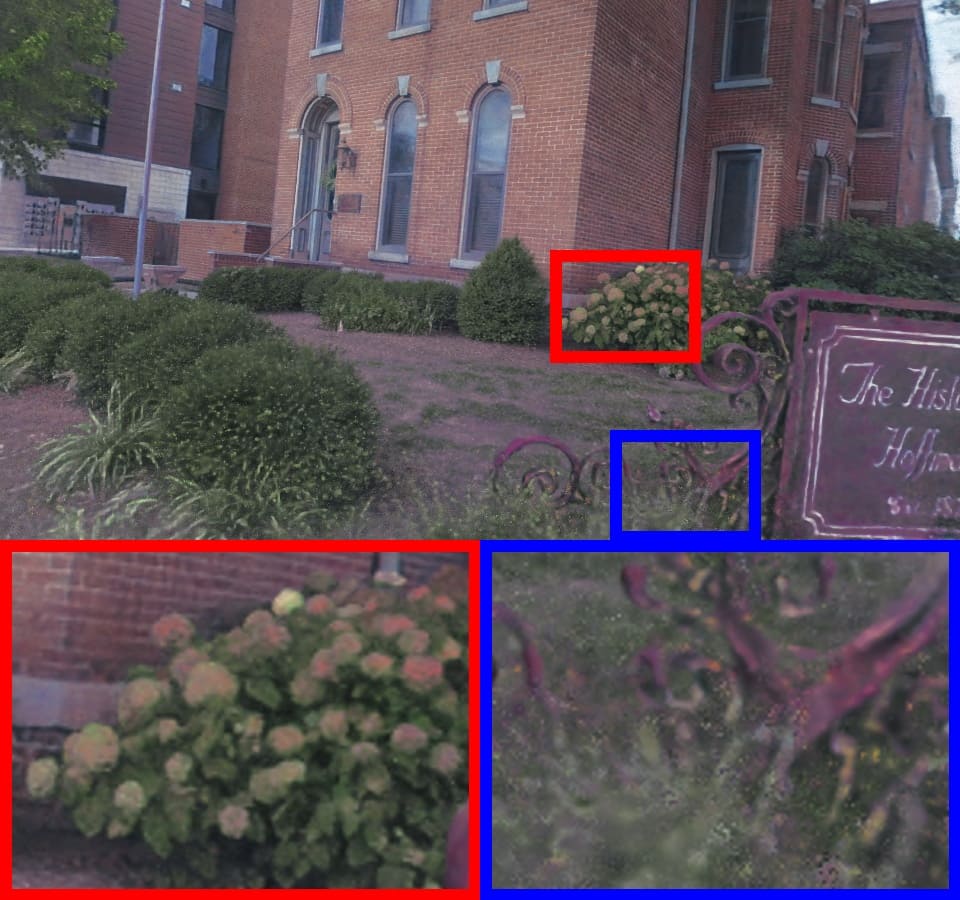} &
        \includegraphics[width=\imgw]{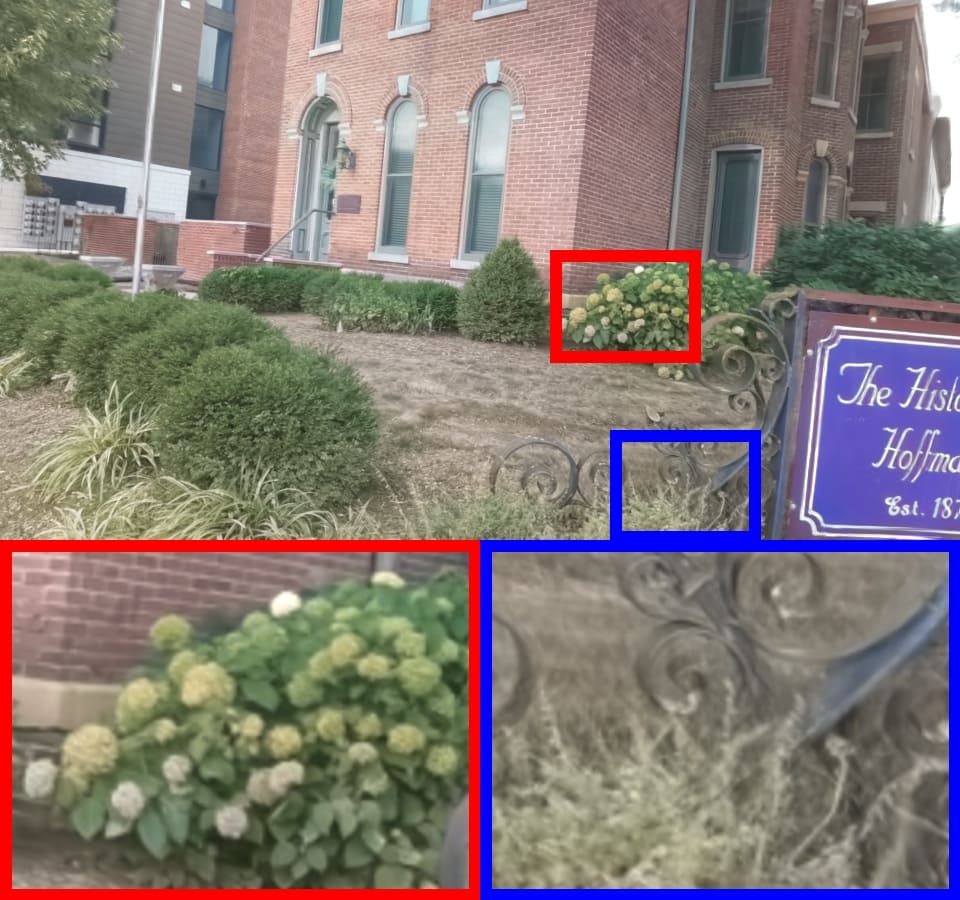} &
        \includegraphics[width=\imgw]{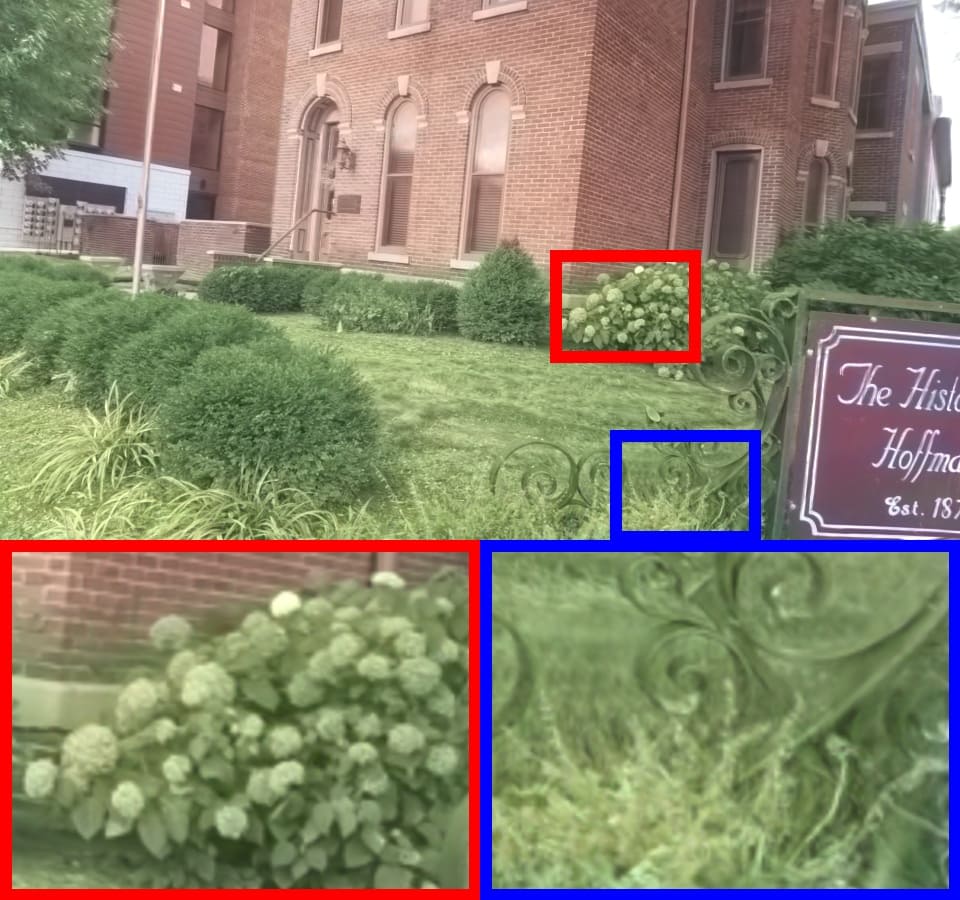} &
        \includegraphics[width=\imgw]{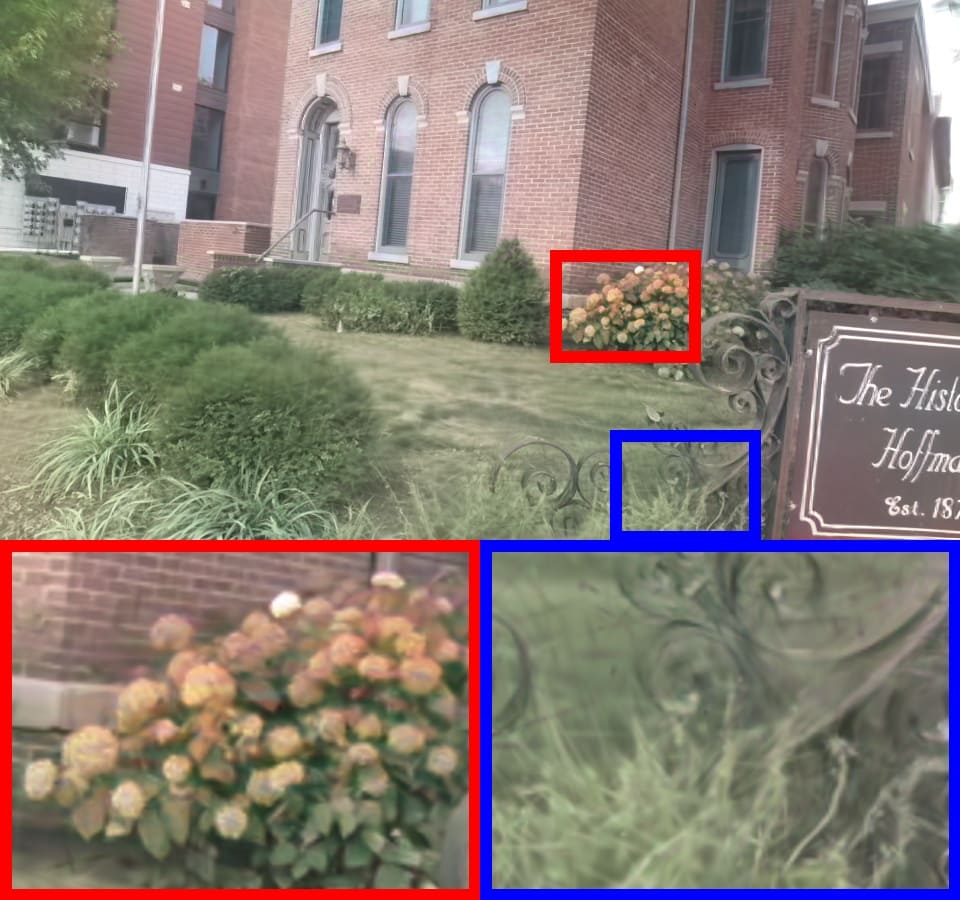} &
        \includegraphics[width=\imgw]{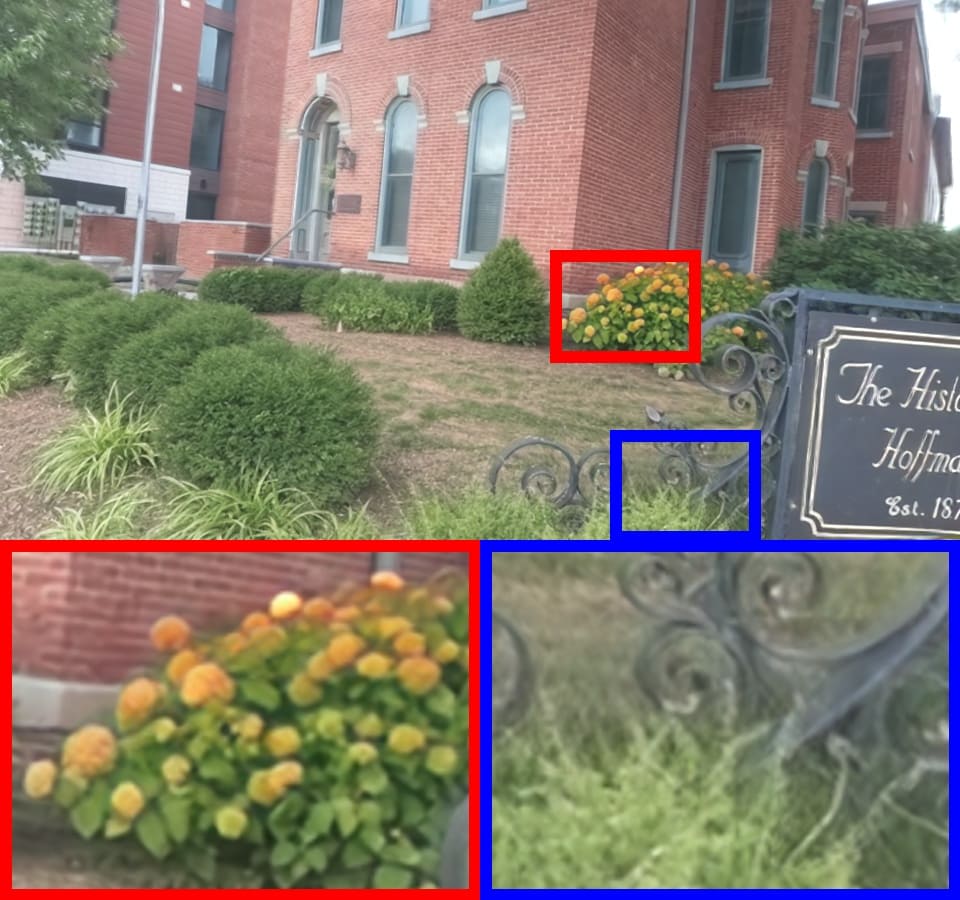} \\   
        
    \end{tabular}
    }
    \caption{\textbf{Additional qualitative comparison on Tanks and Temples (TnT), Mip-NeRF 360 (360) and DL3DV-10K-Benchmarks (DL3DV) --} We successfully achieve high-fidelity colorization across all parts of the scene, including intricate and fine-grained details.
    }
    \label{fig:supp_qualitative}
\end{figure}

\subsection{Extended qualitative comparisons}
\label{sec:supp_qual}
To complement the results in the main paper, we provide additional qualitative comparisons in \cref{fig:supp_qualitative}. 
Our method consistently achieves high-fidelity colorization across various scenes, successfully capturing fine-grained details even within complex geometries.

\subsection{Extended qualitative ablation}
\label{sec:supp_ablation_qual}
\begin{figure}[tb]
    \centering
    \newcommand{\imgw}{0.25\textwidth}
    \setlength{\tabcolsep}{1.5pt}
    
    \centering

    \begin{tabular}{ccc}

    \includegraphics[width=\imgw]{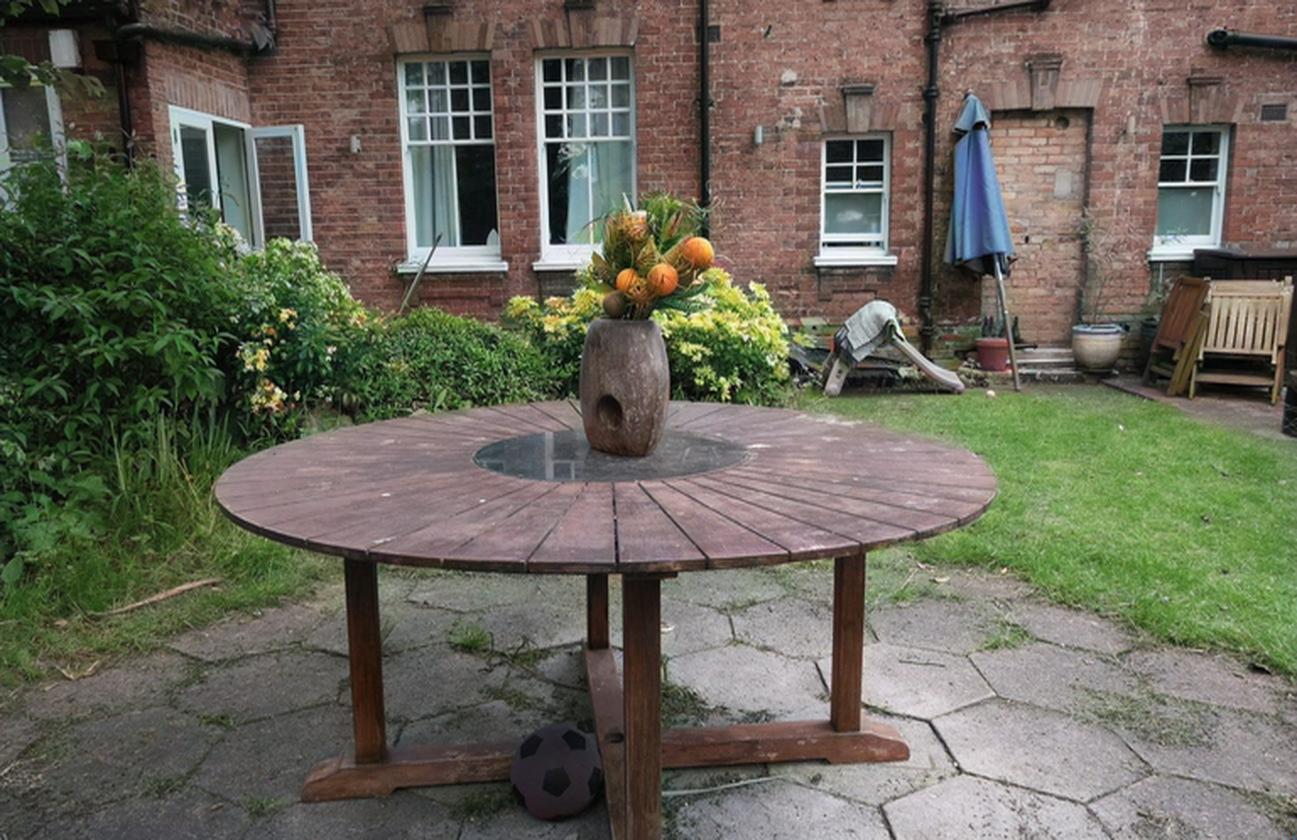} &
    \includegraphics[width=\imgw]{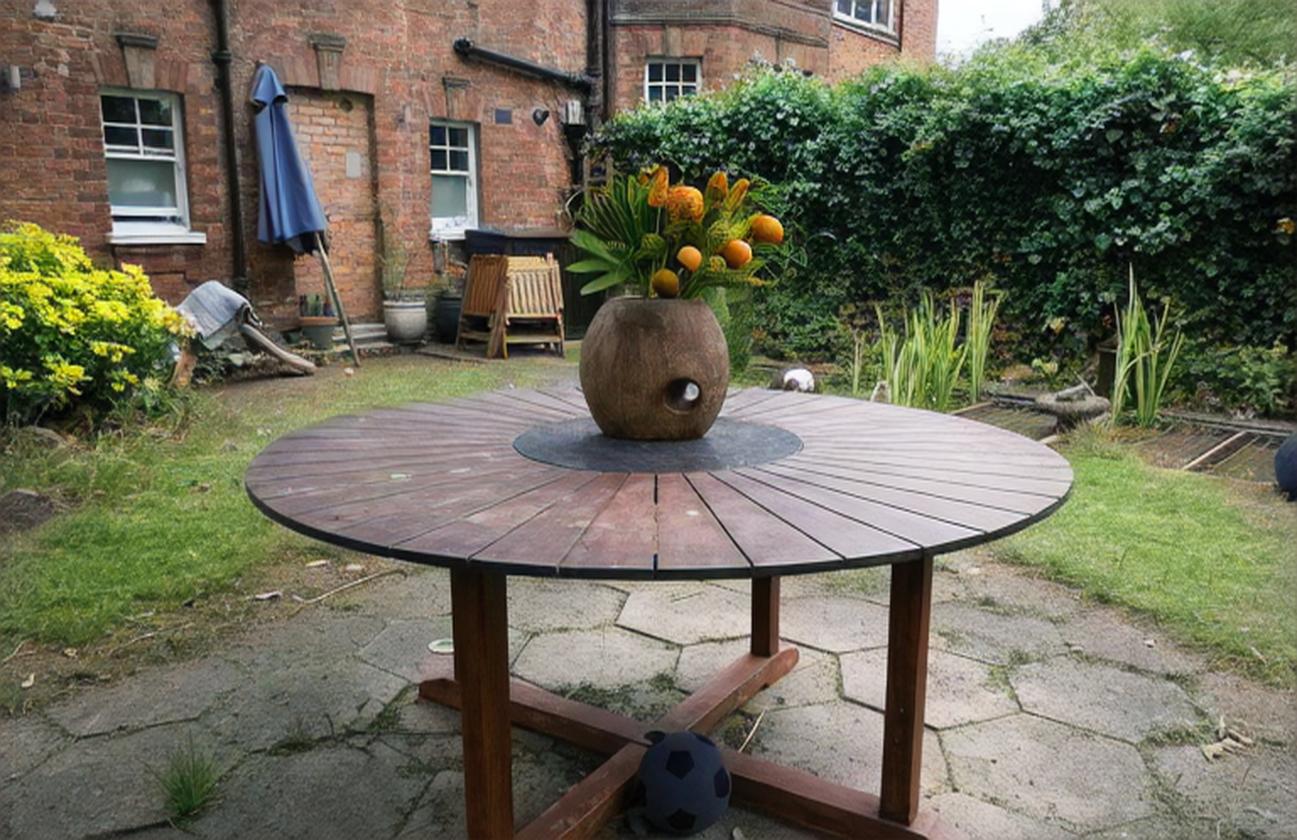} &
    \includegraphics[width=\imgw]{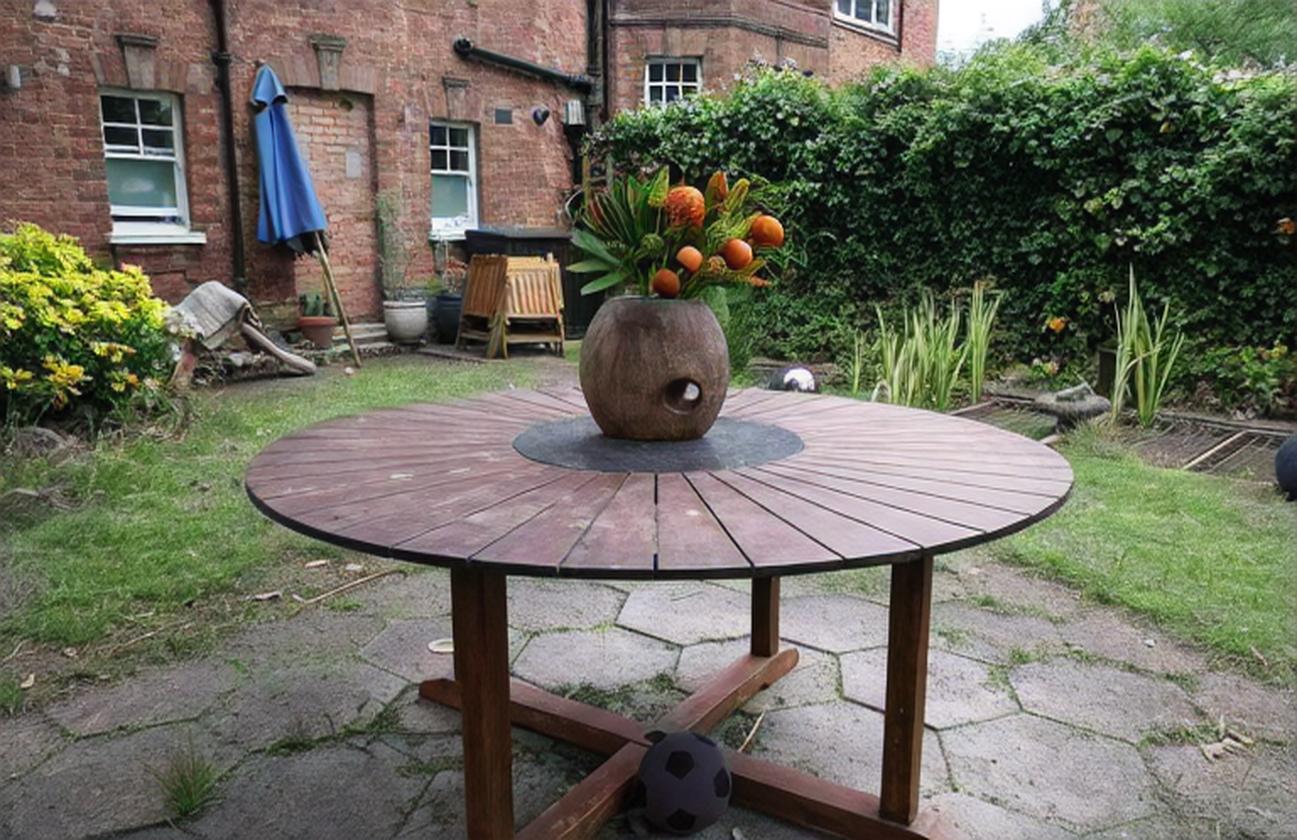} \\
    
    \includegraphics[width=\imgw]{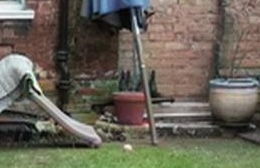} &
    \includegraphics[width=\imgw]{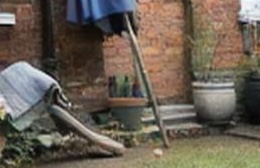} &
    \includegraphics[width=\imgw]{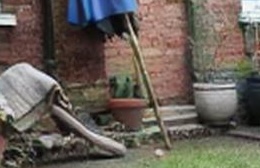} \\
    
    \includegraphics[width=\imgw]{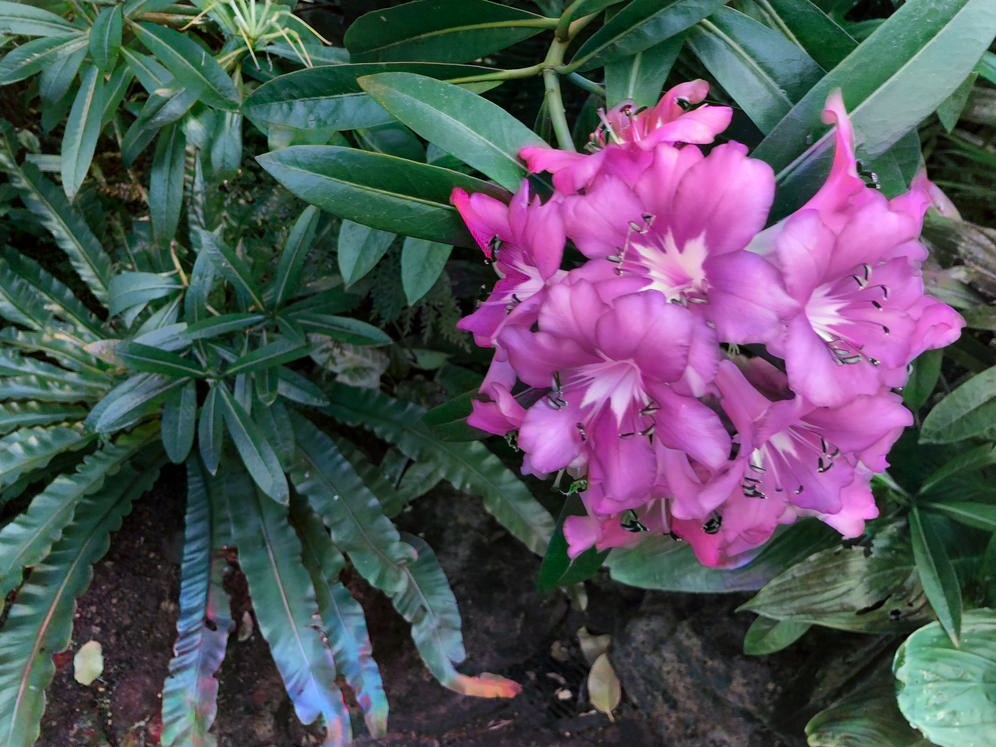} &    
    \includegraphics[width=\imgw]{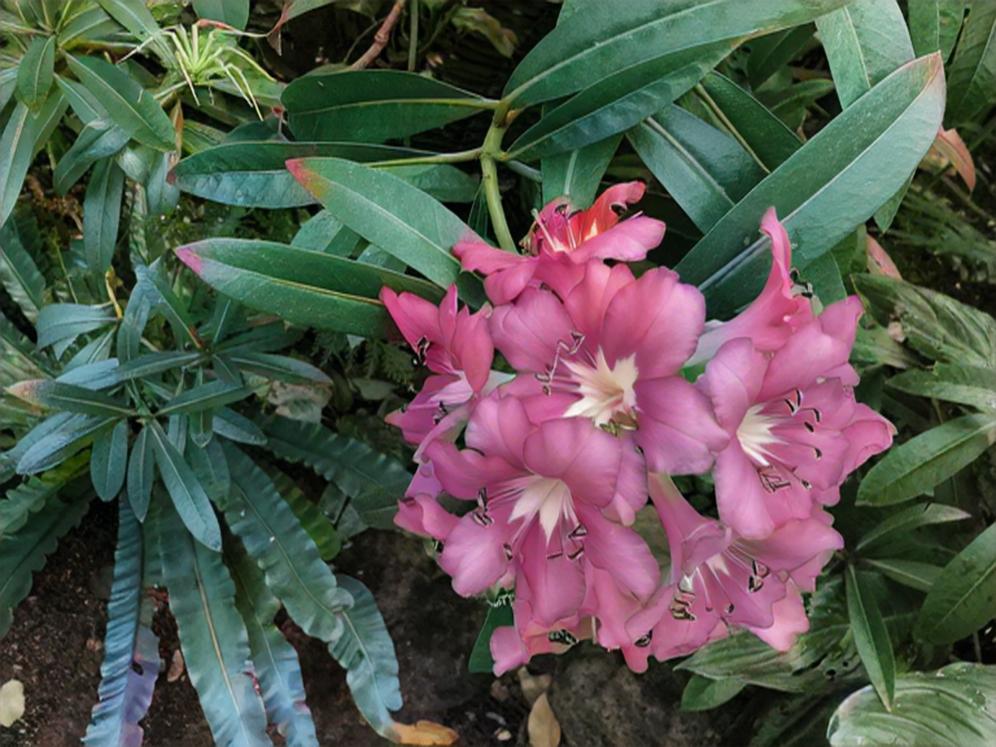} &
    \includegraphics[width=\imgw]{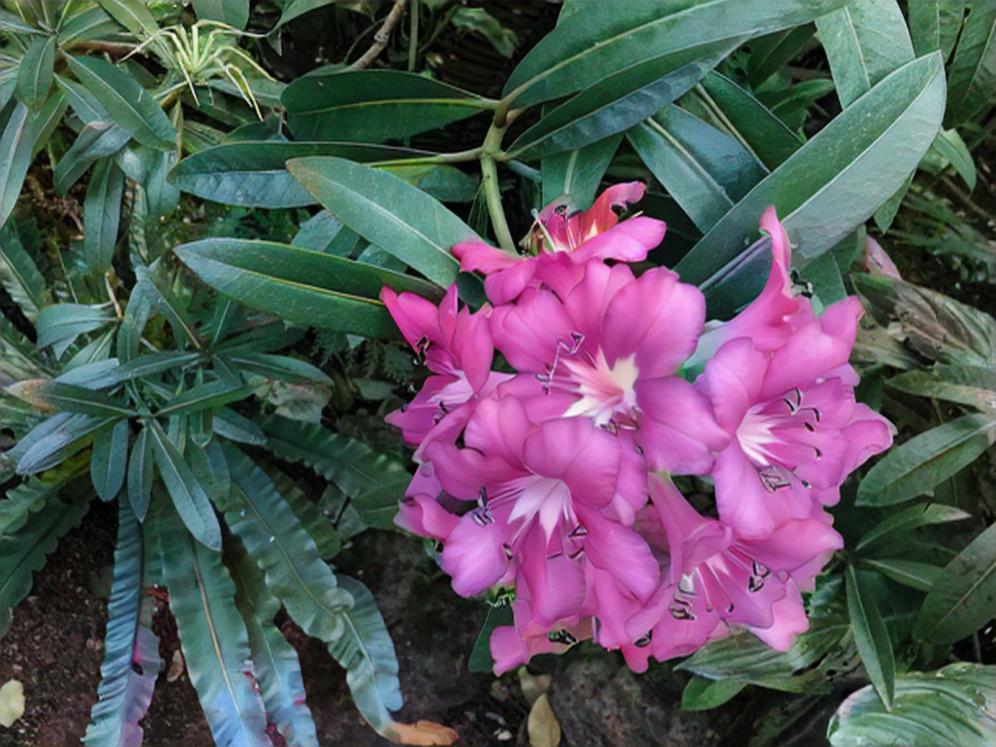} \\
    Base view & w/o Flickr8k & Our method \\

    \end{tabular}
    \caption{
    \textbf{Ablation on the training data --} Note the subtle color variations. While the DL3DV-10K-Benchmark dataset is sufficient for training the multi-view referencing capability to the model, it fails to capture the full spectrum of real-world color distribution necessary for plausible colorization. Our method, with the Flickr8k dataset, successfully captures realistic color distribution.
    }%

    \label{fig:ablation_flickr}
\end{figure}

\subsubsection{Training data.}
While the DL3DV-10K-Benchmark~\cite{ling2024d13dv} provides high-quality multi-view data essential for referencing capability, its training split--consisting of 112 scenes after excluding 28 for evaluation--cannot adequately represent the full spectrum of real-world color distribution.
Since diffusion model performance relies heavily on its learned color space, we supplement our fine-tuning with the Flickr8k~\cite{hodosh2013flickr8k} dataset to provide this required color diversity. 
The ablation study validating this choice is presented in \cref{fig:ablation_flickr}.
While the global visual impact may be subtle, incorporating Flickr8k significantly improves its ability to transfer color accurately from the reference base view as confirmed by the results in \cref{tab:ablation}.
This improvement is best observed in zoomed-in views in the second row.

\begin{figure}[t]
    \centering
    \newcommand{\imgw}{0.2\textwidth}
    
    \resizebox{0.99\linewidth}{!}{
    \centering
    \begin{tabular}{ccccccc}
    
    & \multicolumn{3}{c}{Input}
    & \multicolumn{3}{c}{Colorized} \\

    \cmidrule(lr){2-4}
    \cmidrule(lr){5-7}

    \raisebox{13pt}{\rotatebox{90}{Scene 1}} &
    \includegraphics[width=\imgw]{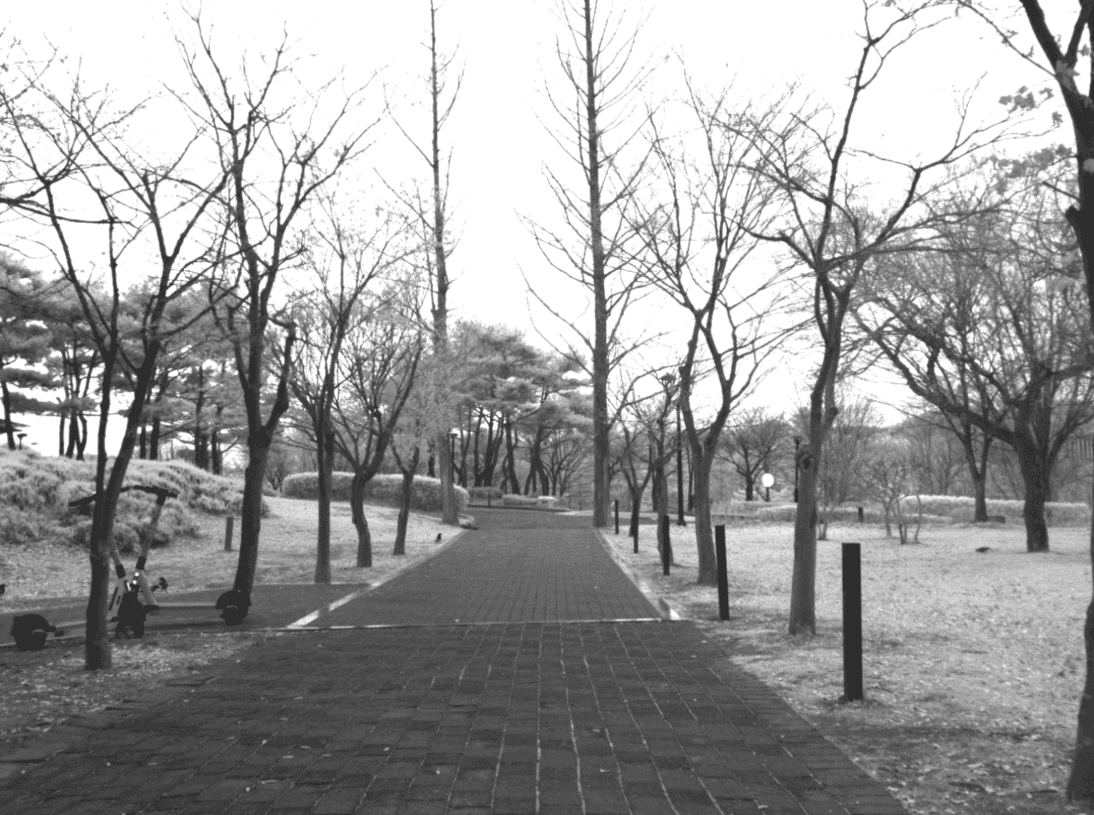} &
    \includegraphics[width=\imgw]{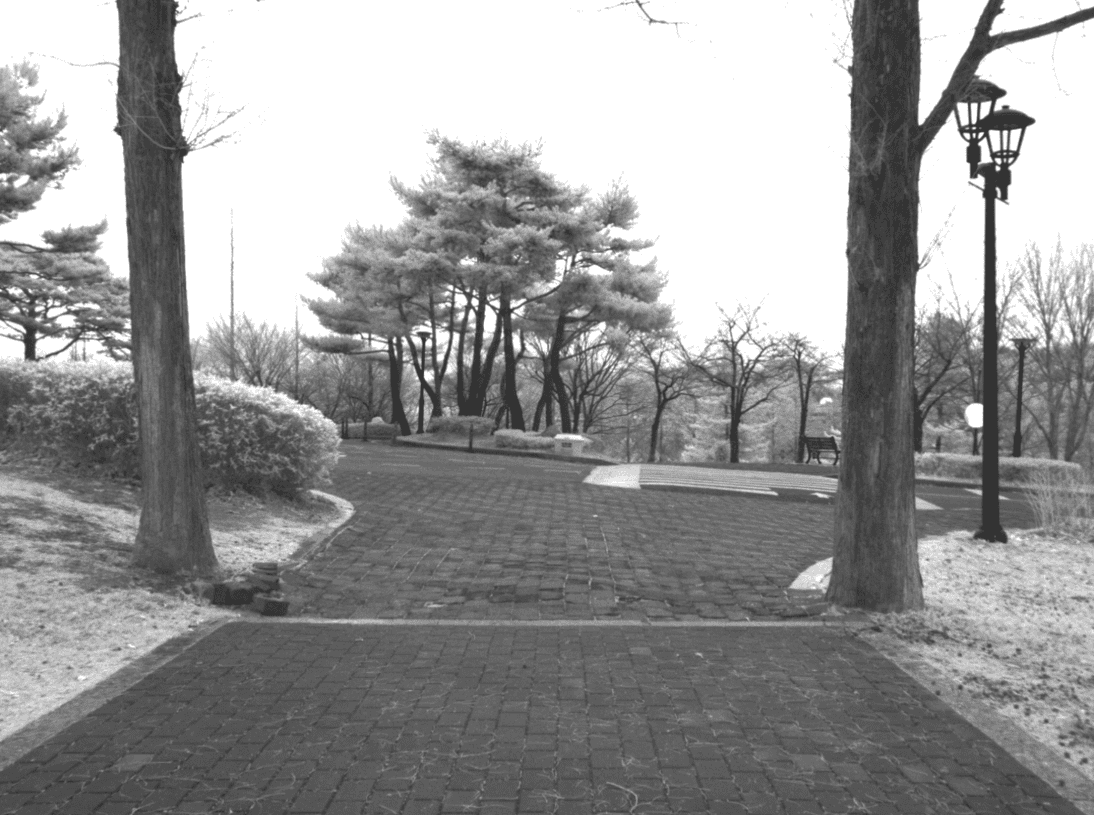} &
    \includegraphics[width=\imgw]{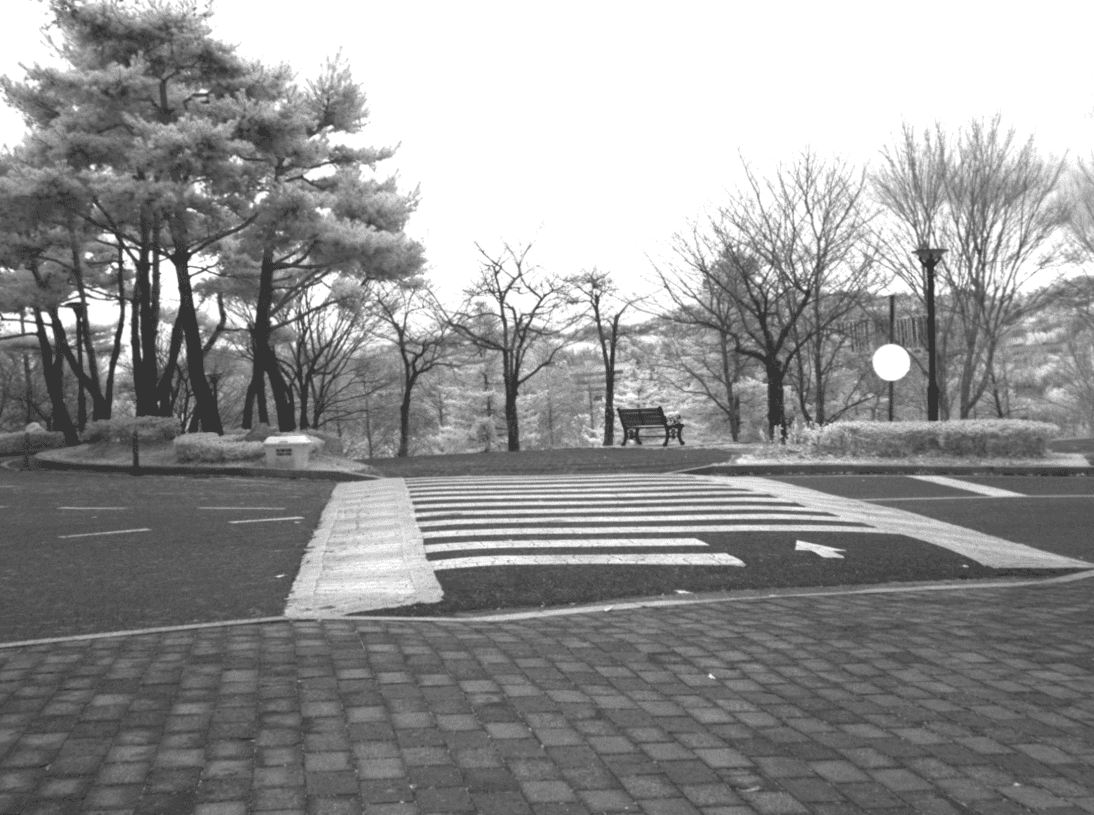} &
    \includegraphics[width=\imgw]{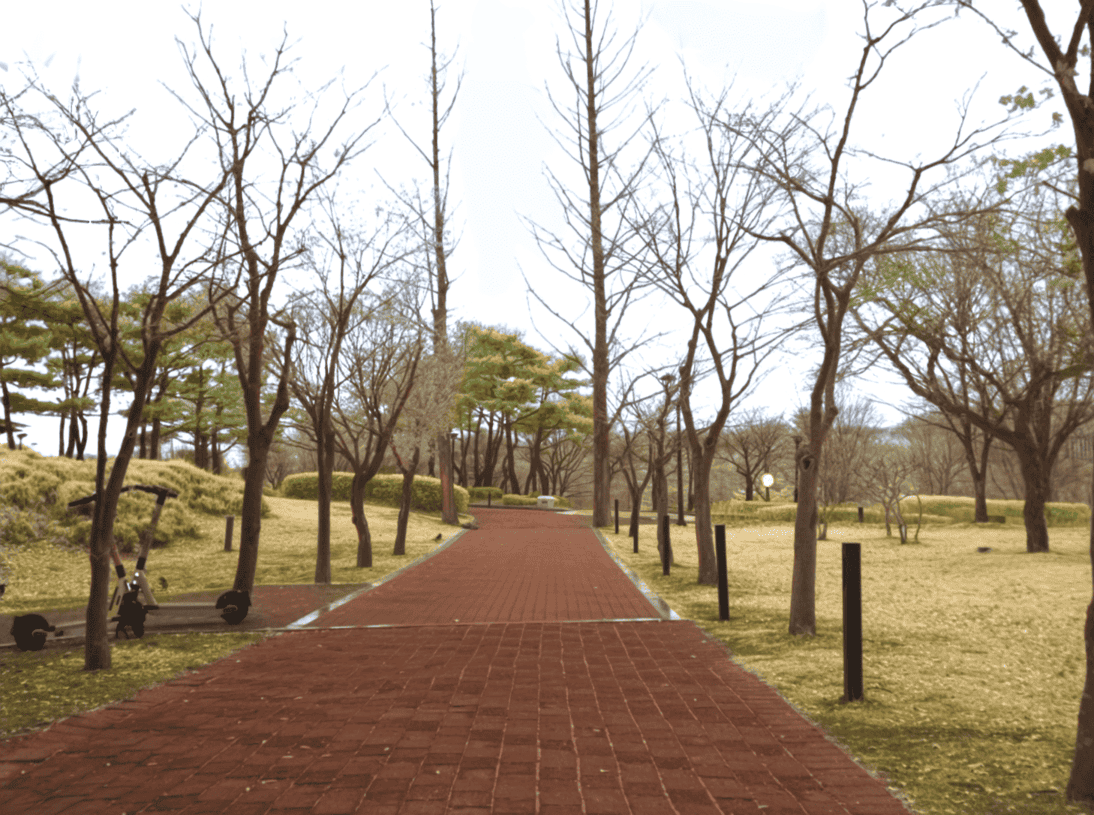} &
    \includegraphics[width=\imgw]{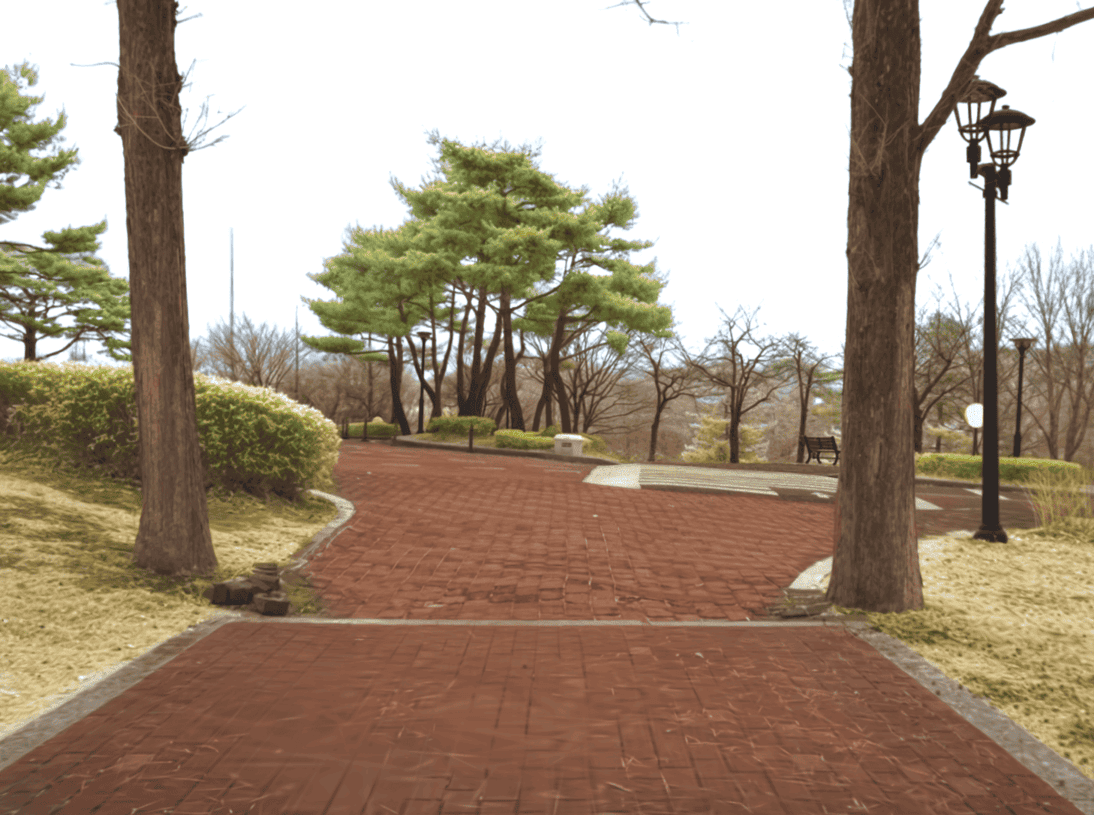} &
    \includegraphics[width=\imgw]{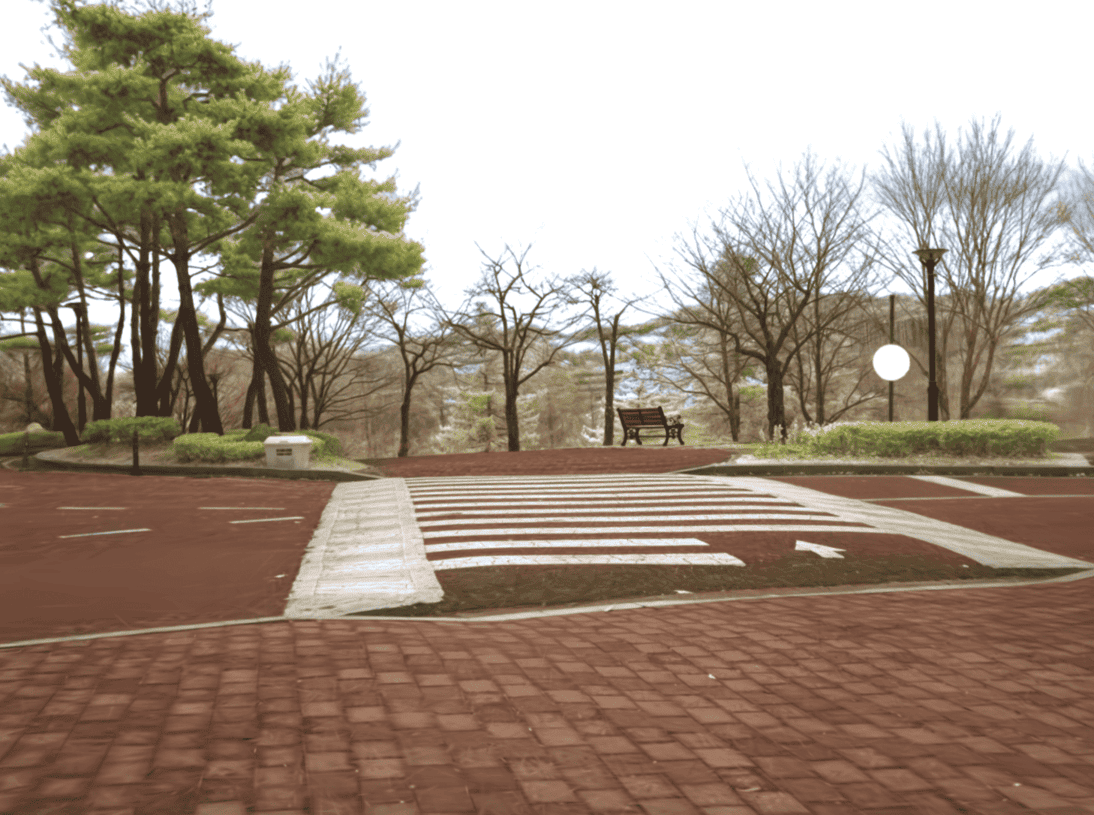} \\

    \raisebox{13pt}{\rotatebox{90}{Scene 2}} &
    \includegraphics[width=\imgw]{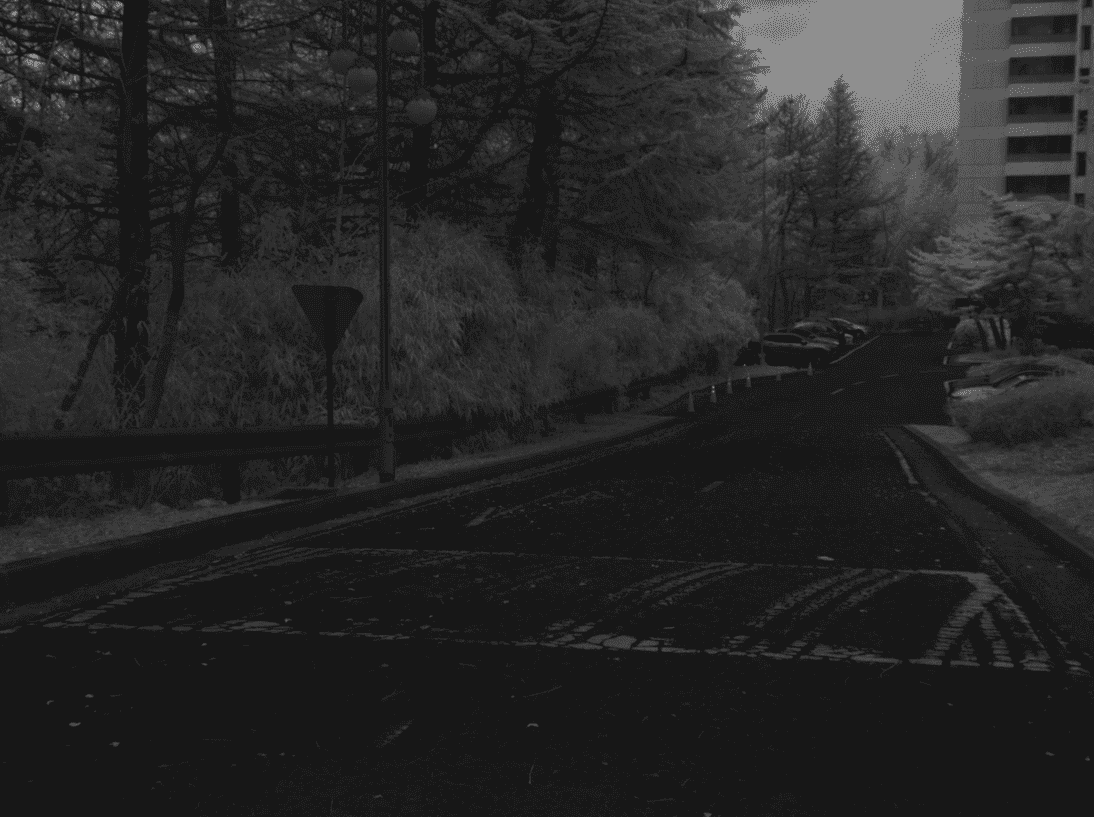} &
    \includegraphics[width=\imgw]{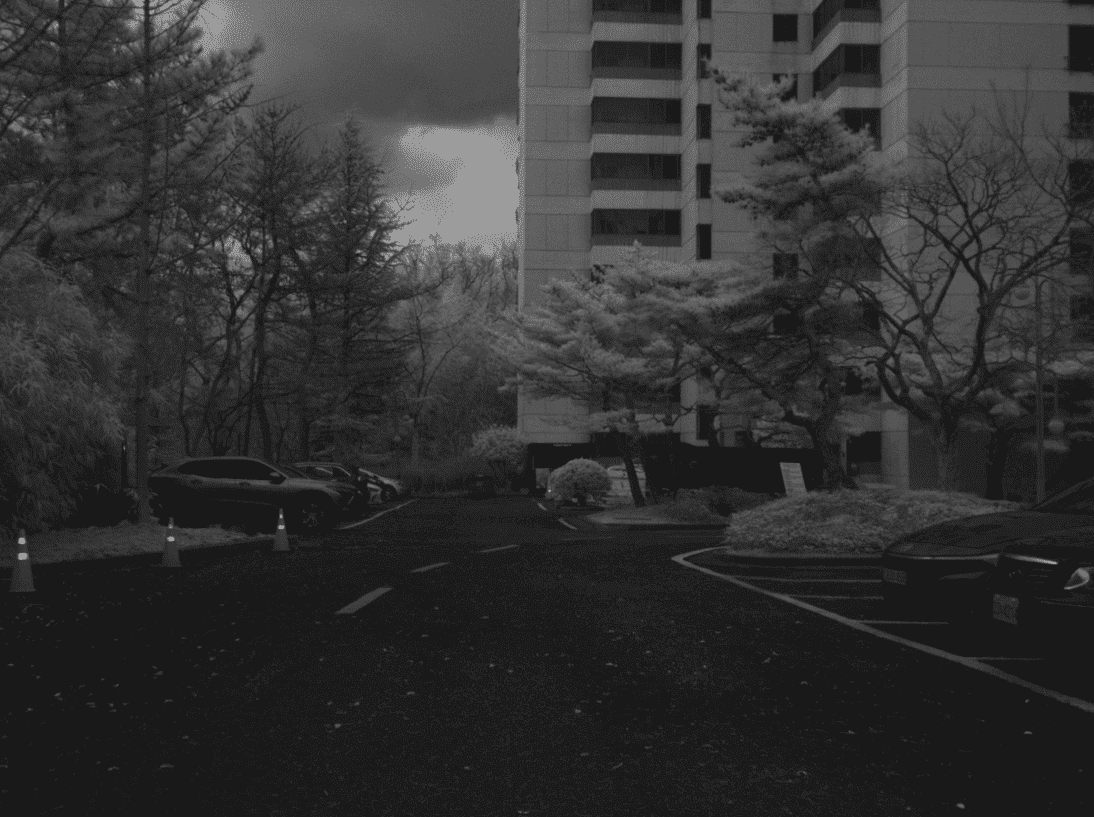} &
    \includegraphics[width=\imgw]{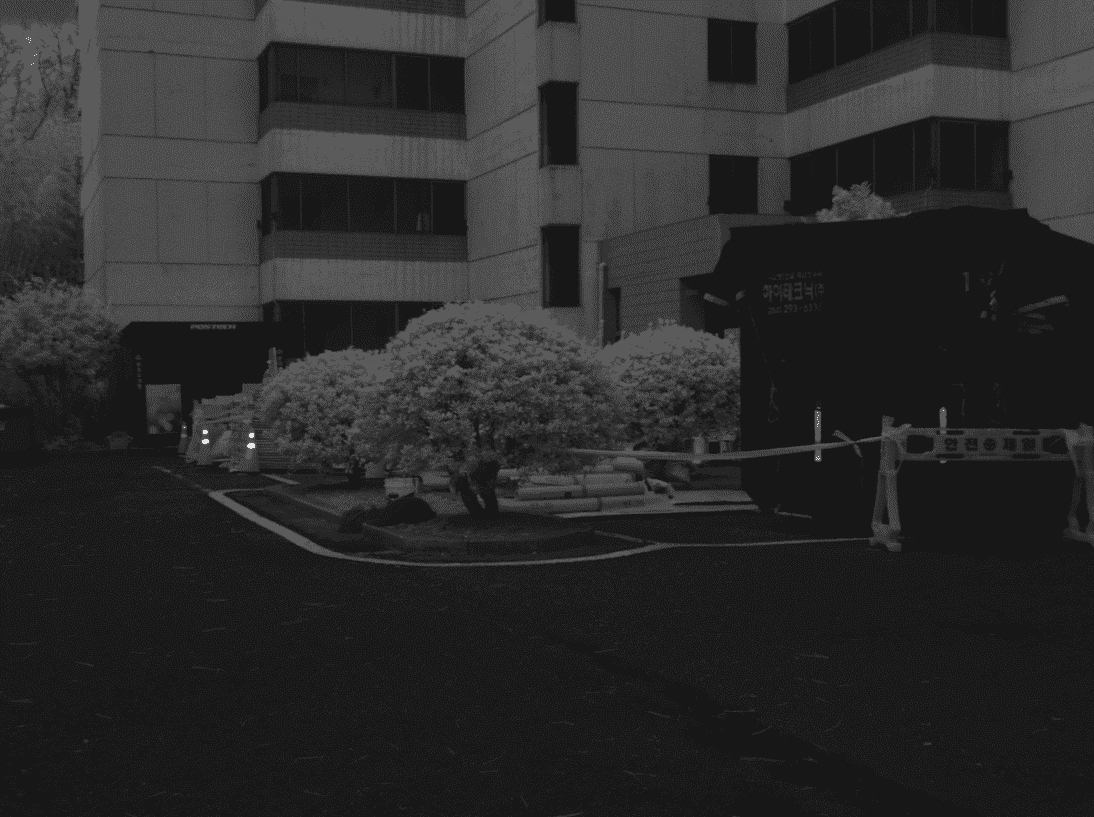} &
    \includegraphics[width=\imgw]{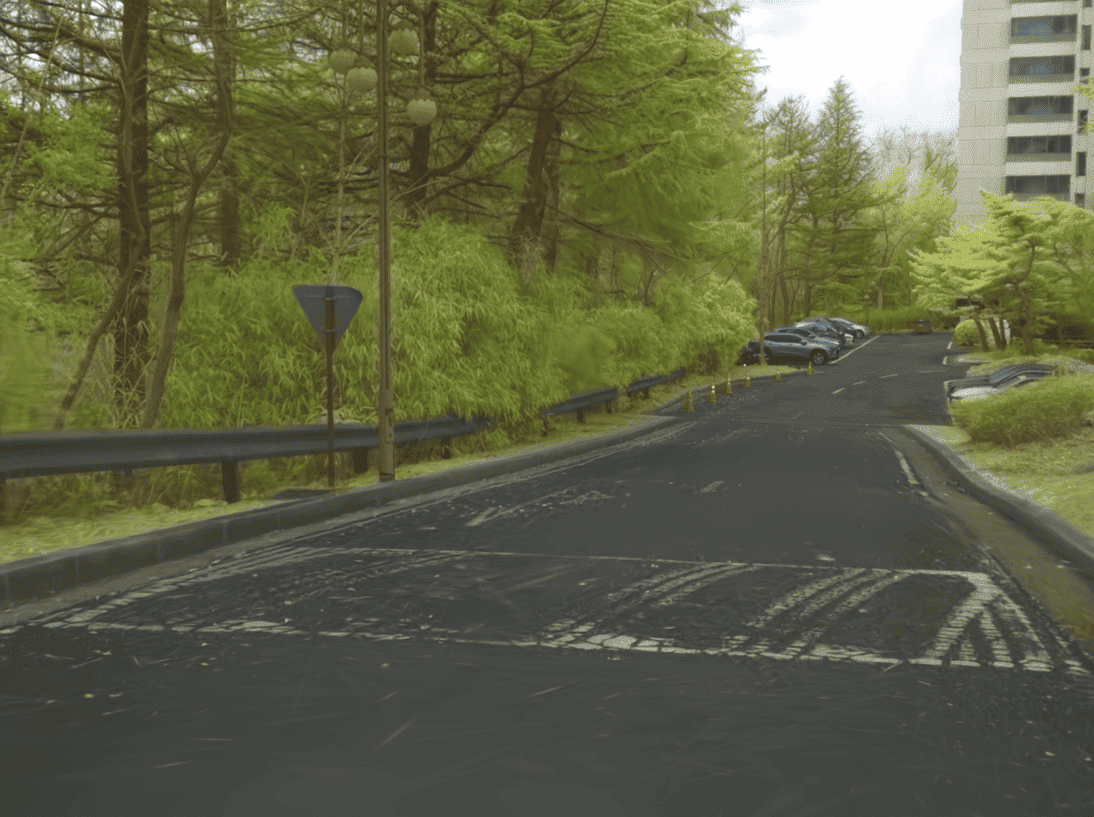} &
    \includegraphics[width=\imgw]{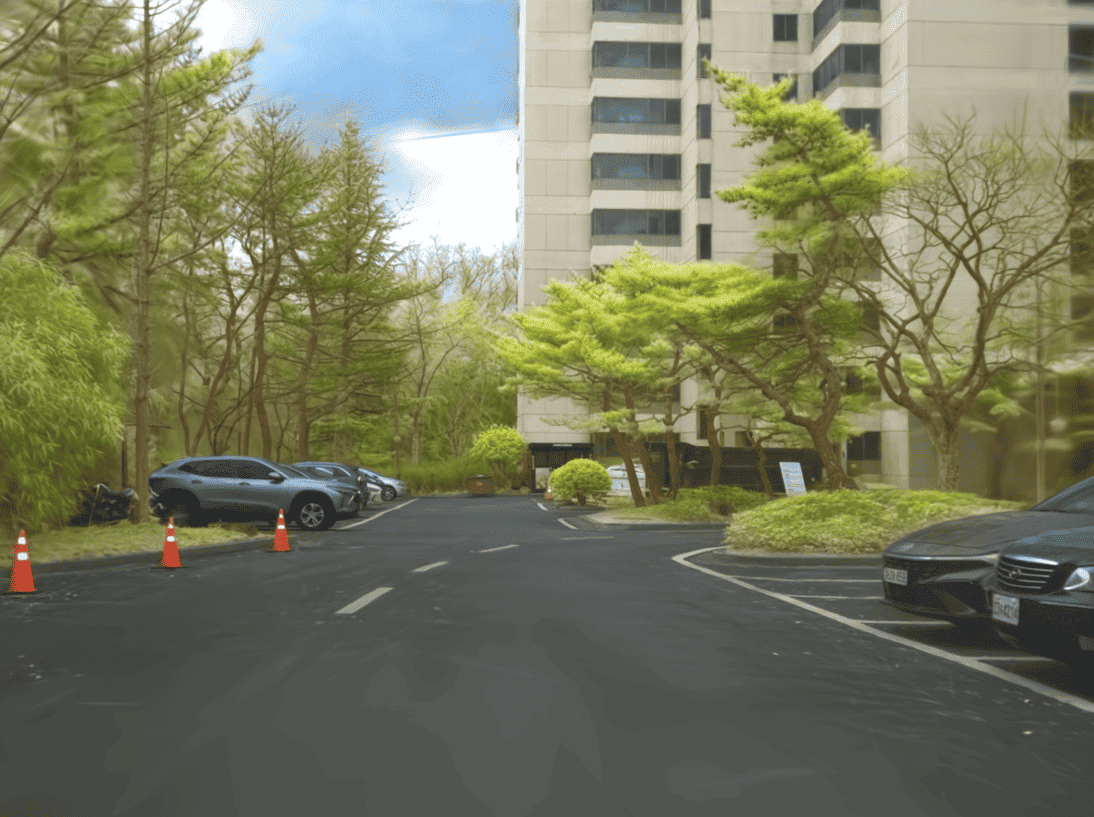} &
    \includegraphics[width=\imgw]{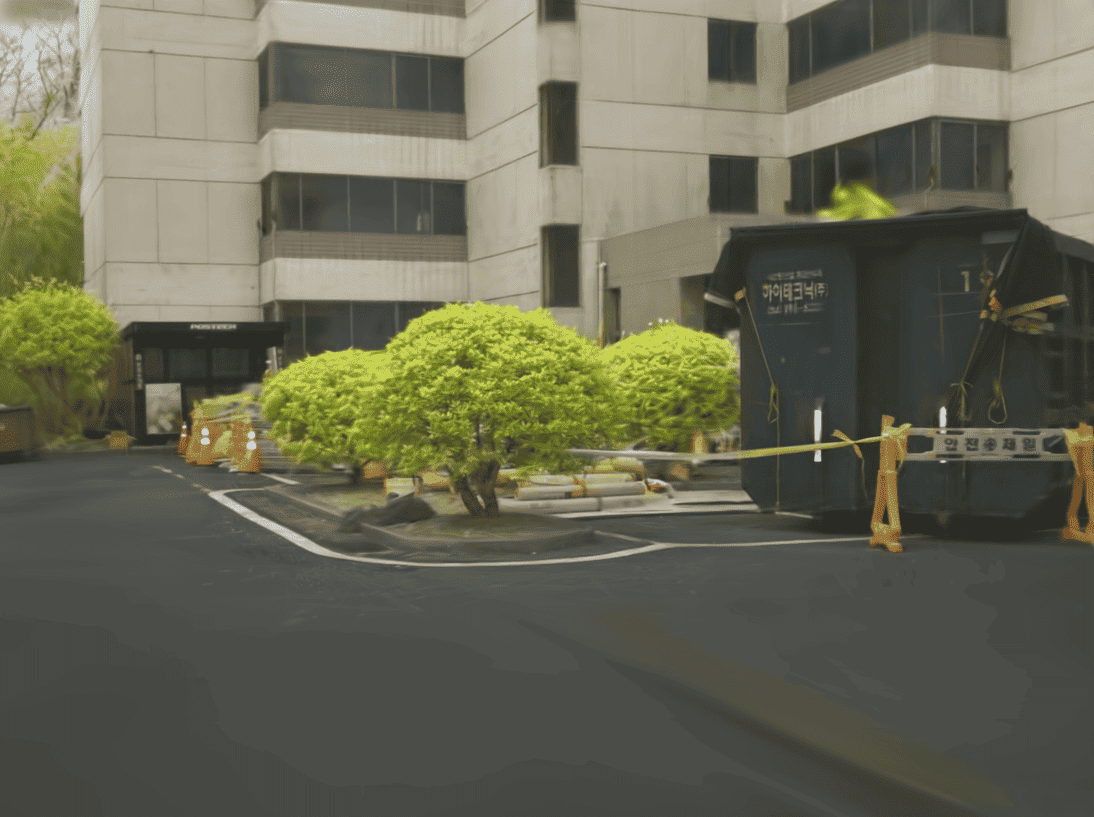} \\

    \raisebox{13pt}{\rotatebox{90}{Scene 3}} &
    \includegraphics[width=\imgw]{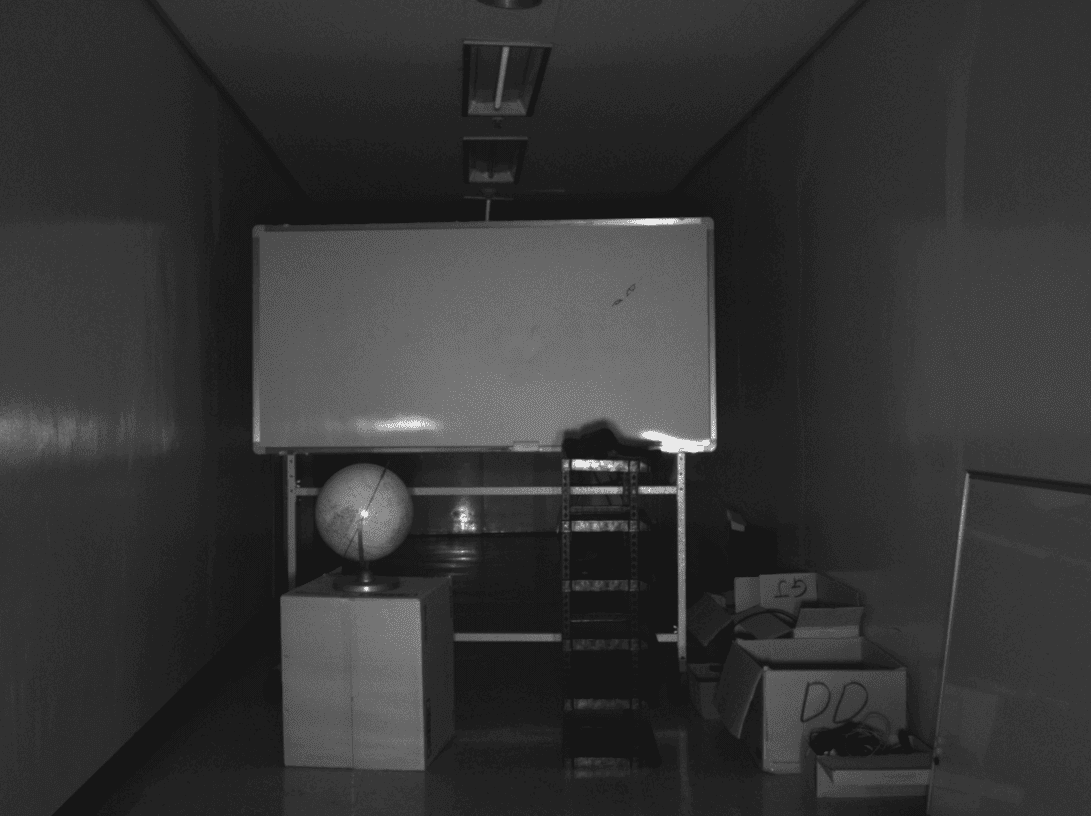} &
    \includegraphics[width=\imgw]{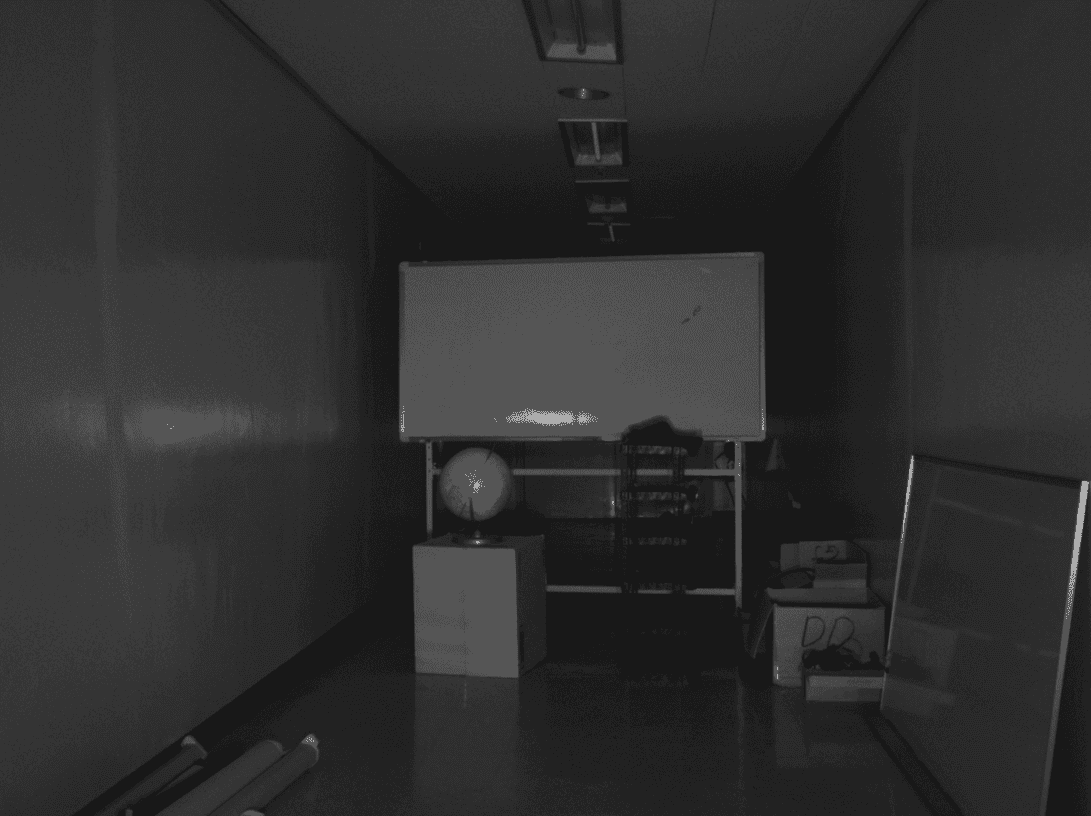} &
    \includegraphics[width=\imgw]{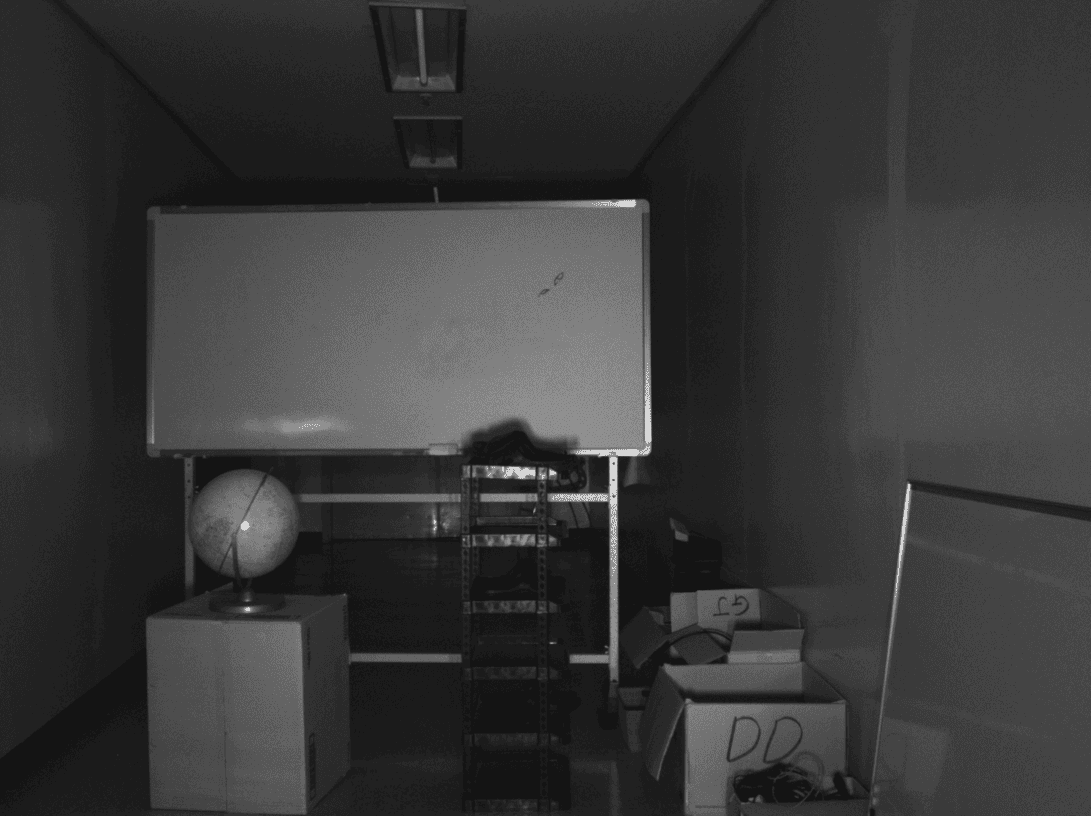} &
    \includegraphics[width=\imgw]{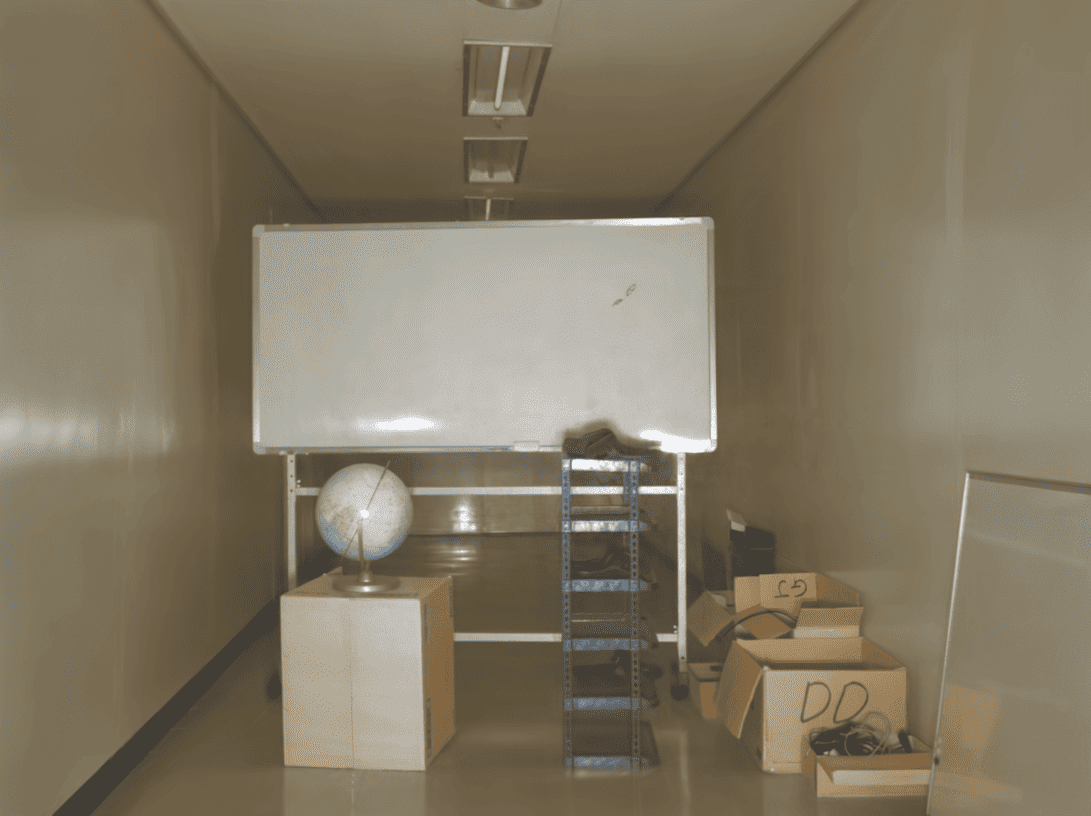} &
    \includegraphics[width=\imgw]{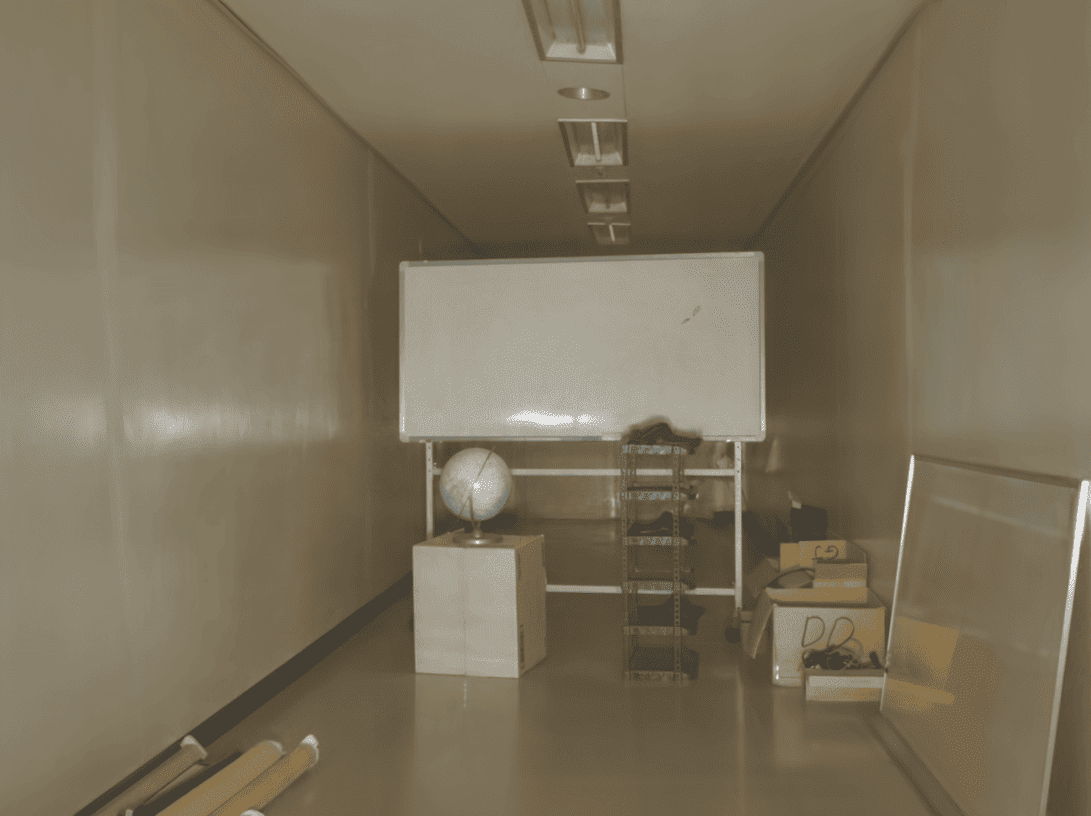} &
    \includegraphics[width=\imgw]{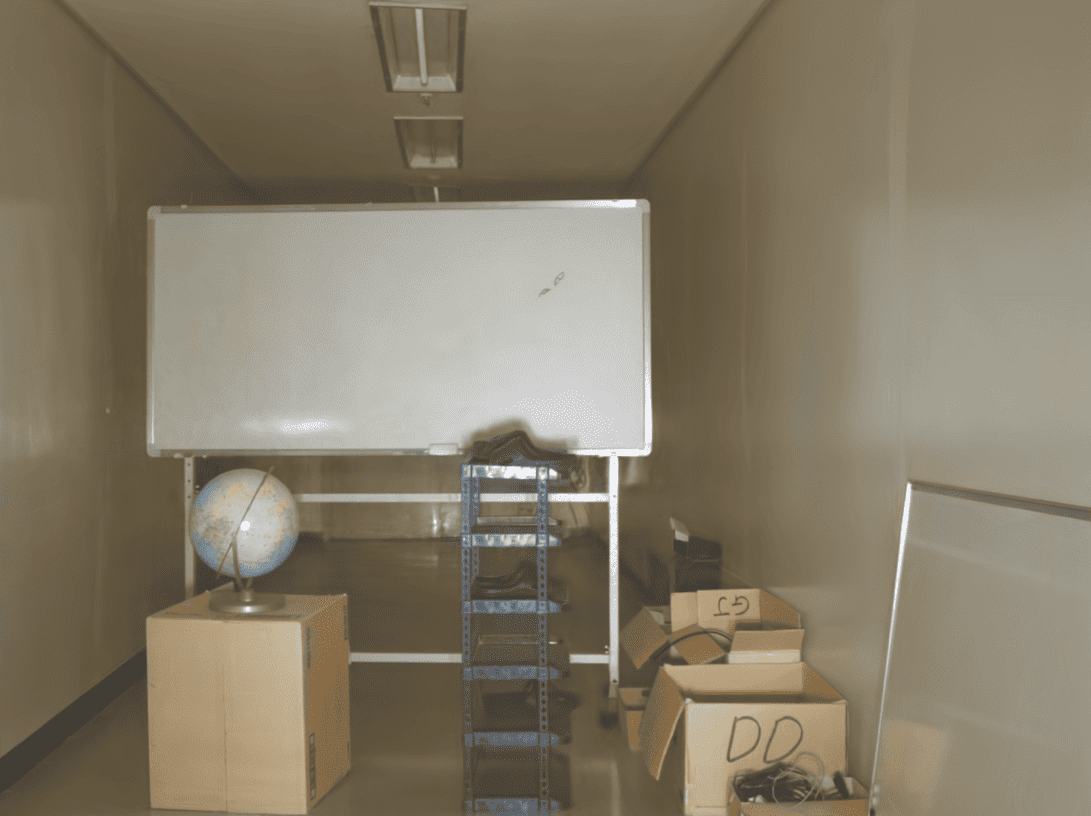} \\

    \raisebox{13pt}{\rotatebox{90}{Scene 4}} &
    \includegraphics[width=\imgw]{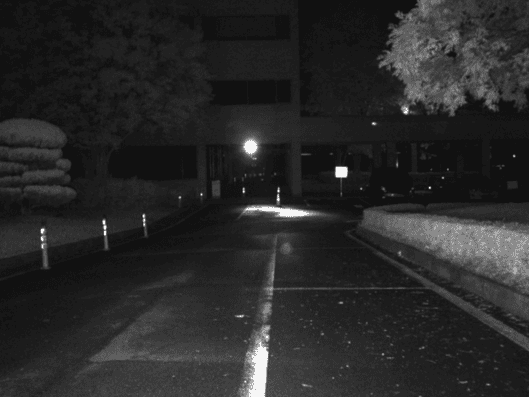} &
    \includegraphics[width=\imgw]{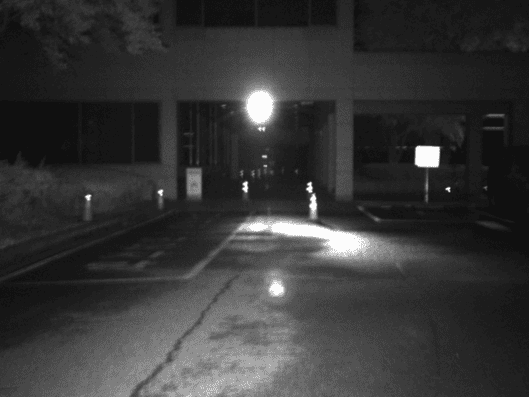} &
    \includegraphics[width=\imgw]{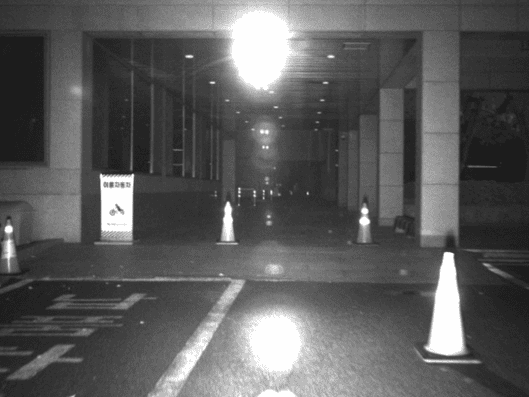} &
    \includegraphics[width=\imgw]{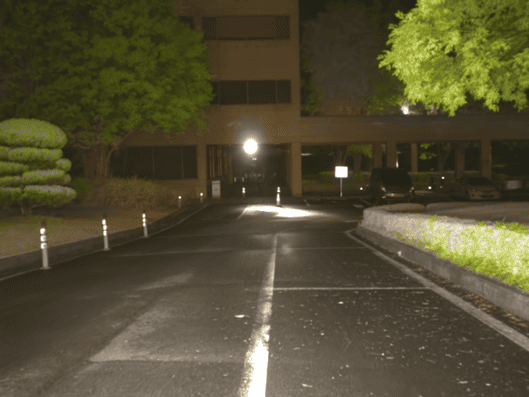} &
    \includegraphics[width=\imgw]{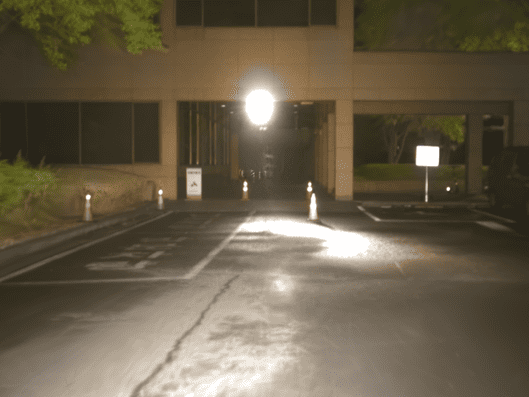} &
    \includegraphics[width=\imgw]{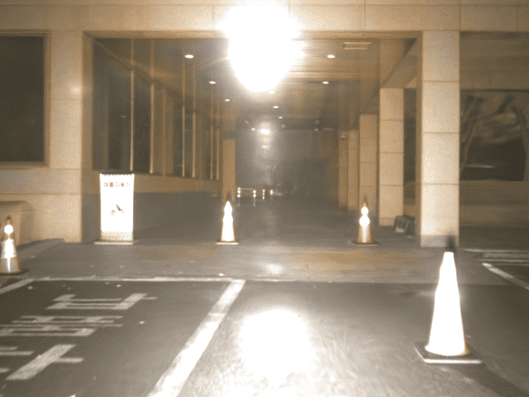} \\
    
    \end{tabular}
    }
    \caption{
    \textbf{Additional results on NIR images --} The results on NIR multi-view inputs further illustrate the robustness of our method across diverse indoor and outdoor lighting conditions.
    }

    \label{fig:supp_application}
\end{figure}

\subsection{Additional application results}
\label{sec:supp_application}
In \cref{fig:supp_application}, we present additional colorization results for 3D reconstructions from single-channel NIR multi-view images, using only the NIR images from the Pixel-aligned RGB-NIR Stereo dataset~\cite{kim2025pixelnir}.
According to the dataset description, `Scene 1' and `Scene 2' correspond to well-lit outdoor settings, `Scene 3' is a complex indoor environment, and `Scene 4' is a dark outdoor scene.
Our method not only produces plausible colorizations for various NIR-based 3D scenes but also demonstrates its utility in challenging nighttime scenarios, as exemplified by `Scene 4'.
These results indicate that our approach successfully colorizes NIR multi-view inputs captured in environments where standard visible-light imaging is impractical or unfeasible.

\section{Limitations}
\label{sec:limitation}
Our method relies on the capabilities of the image colorization model; consequently, any artifacts or implausible colors generated by the 2D model are inevitably propagated to our 3D result.
In addition, although we minimize the guidance averaging process, we incorporate a single averaging operation during global calibration. While crucial for establishing global consistency, this step may reduce the resulting color diversity in some cases.

\section{Video Results}
\label{sec:video}
For more qualitative results, please see the videos on our project page \url{https://yeonjin-chang.github.io/LoGoColor/}. 
These videos clearly show that, unlike other methods where the overall scene color and tone fluctuate with the viewpoint, our approach maintains robust color consistency.
We note that these videos are rendered entirely using the 3DGS-based implementation.

\end{document}